\documentclass{article}

\usepackage{microtype}
\usepackage{graphicx}
\usepackage{subcaption}
\usepackage{booktabs}
\usepackage{hyperref}
\usepackage{multirow}

\usepackage[accepted]{icml2026}

\usepackage{amsmath}
\usepackage{amssymb}
\usepackage{mathtools}
\usepackage{amsthm}
\usepackage{bbm}
\usepackage{colortbl}
\usepackage{xcolor}
\usepackage{listings}
\usepackage{tabularx}
\usepackage{booktabs}
\usepackage{enumitem}      
\usepackage[table]{xcolor} 

\usepackage{newunicodechar}
\newunicodechar{π}{\ensuremath{\pi}}
\newunicodechar{ε}{\ensuremath{\epsilon}}
\newunicodechar{α}{\ensuremath{\alpha}}
\newunicodechar{β}{\ensuremath{\beta}}
\newunicodechar{γ}{\ensuremath{\gamma}}
\newunicodechar{δ}{\ensuremath{\delta}}
\newunicodechar{θ}{\ensuremath{\theta}}
\newunicodechar{λ}{\ensuremath{\lambda}}
\newunicodechar{μ}{\ensuremath{\mu}}
\newunicodechar{σ}{\ensuremath{\sigma}}
\newunicodechar{τ}{\ensuremath{\tau}}
\newunicodechar{ω}{\ensuremath{\omega}}
\newunicodechar{Ω}{\ensuremath{\Omega}}
\newunicodechar{≤}{\ensuremath{\leq}}
\newunicodechar{≥}{\ensuremath{\geq}}
\newunicodechar{≠}{\ensuremath{\neq}}
\newunicodechar{≈}{\ensuremath{\approx}}
\newunicodechar{→}{\ensuremath{\rightarrow}}
\newunicodechar{∈}{\ensuremath{\in}}
\newunicodechar{∉}{\ensuremath{\notin}}
\newunicodechar{⊂}{\ensuremath{\subset}}
\newunicodechar{⊆}{\ensuremath{\subseteq}}
\newunicodechar{∪}{\ensuremath{\cup}}
\newunicodechar{∩}{\ensuremath{\cap}}
\newunicodechar{∞}{\ensuremath{\infty}}
\newunicodechar{∇}{\ensuremath{\nabla}}
\newunicodechar{∑}{\ensuremath{\sum}}
\newunicodechar{∏}{\ensuremath{\prod}}
\newunicodechar{∫}{\ensuremath{\int}}
\newunicodechar{∂}{\ensuremath{\partial}}
\newunicodechar{±}{\ensuremath{\pm}}
\newunicodechar{÷}{\ensuremath{\div}}
\newunicodechar{×}{\ensuremath{\times}}
\newunicodechar{—}{---}
\newunicodechar{–}{--}

\usepackage[capitalize,noabbrev]{cleveref}

\theoremstyle{plain}

\theoremstyle{definition}

\theoremstyle{remark}

\newcommand{\benchname}{\textsc{QEDBench}}

\icmltitlerunning{\benchname: Quantifying the Alignment Gap in
  Automated Evaluation of University-Level Mathematical Proofs}

\begin{document}

\twocolumn[
  \icmltitle{\benchname: Quantifying the Alignment Gap in
  Automated Evaluation of University-Level Mathematical Proofs}

  \icmlsetsymbol{equal}{*}
  \begin{icmlauthorlist}
  \icmlauthor{Santiago Gonzalez}{equal,yale}
  \icmlauthor{Alireza Amiri Bavandpour}{equal,yale}
  \icmlauthor{Peter Ye}{equal,yale}
  \icmlauthor{Edward Zhang}{equal,yale}
  \icmlauthor{Ruslans Aleksejevs}{berk}
  \icmlauthor{Todor Antić}{charles}
  \icmlauthor{Polina Baron}{uchicago}
  \icmlauthor{Sujeet Bhalerao}{uiuc}
  \icmlauthor{Shubhrajit Bhattacharya}{uchicago}
  \icmlauthor{Zachary Burton}{mit}
  \icmlauthor{John Byrne}{udel}
  \icmlauthor{Hyungjun Choi}{princeton}
  \icmlauthor{Nujhat Ahmed Disha}{mit}
  \icmlauthor{Koppány István Encz}{idsia}
  \icmlauthor{Yuchen Fang}{berk}
  \icmlauthor{Robert Joseph George}{caltech}
  \icmlauthor{Ebrahim Ghorbani}{tuhamburg}
  \icmlauthor{Alan Goldfarb}{berk}
  \icmlauthor{Jing Guo}{regensburg}
  \icmlauthor{Meghal Gupta}{berk}
  \icmlauthor{Stefano Huber}{idsia}
  \icmlauthor{Annika Kanckos}{helsinki}
  \icmlauthor{Minjung Kang}{illinois}
  \icmlauthor{Hyun Jong Kim}{western}
  \icmlauthor{Dino Lorenzini}{uga}
  \icmlauthor{Levi Lorenzo}{fortlewis}
  \icmlauthor{Tianyi Mao}{helm}
  \icmlauthor{Giovanni Marzenta}{ubc}
  \icmlauthor{Ariane M. Masuda}{cunytt}
  \icmlauthor{Lukas Mauth}{koeln}
  \icmlauthor{Ana Mickovic}{amsterdam}
  \icmlauthor{Andrés Miniguano-Trujillo}{kcl}
  \icmlauthor{Antoine Moulin}{pompeu}
  \icmlauthor{Wenqi Ni}{edinburgh}
  \icmlauthor{Tomos Parry}{bilkent}
  \icmlauthor{Kevin Ren}{princeton}
  \icmlauthor{Hossein Roodbarani}{york}
  \icmlauthor{Mathieu Rundström}{waterloo}
  \icmlauthor{Manjil Saikia}{ahmedabad}
  \icmlauthor{Detchat Samart}{burapha}
  \icmlauthor{Rebecca Steiner}{gutenberg}
  \icmlauthor{Connor Stewart}{cunyhunter}
  \icmlauthor{Dhara Thakkar}{nagoya}
  \icmlauthor{Jeffrey Tse}{oxford}
  \icmlauthor{Vasiliki Velona}{}
  \icmlauthor{Yunhai Xiang}{western}
  \icmlauthor{Sibel Yalçın}{uludag}
  \icmlauthor{Jun Yan}{oxford}
  \icmlauthor{Ji Zeng}{renyi}
  \icmlauthor{Arman Cohan}{yale}
  \icmlauthor{Quanquan C. Liu}{yale}
  \end{icmlauthorlist}

  \icmlaffiliation{yale}{Department of Computer Science, Yale University, New Haven, CT, USA}
  \icmlaffiliation{renyi}{Alfréd Rényi Institute of Mathematics}
  \icmlaffiliation{udel}{University of Delaware}
  \icmlaffiliation{mit}{MIT}
  \icmlaffiliation{western}{University of Western Ontario}
  \icmlaffiliation{koeln}{University of Cologne}
  \icmlaffiliation{cunyhunter}{CUNY Hunter College}
  \icmlaffiliation{berk}{UC Berkeley}
  \icmlaffiliation{cunytt}{New York City College of Technology (CUNY)}
  \icmlaffiliation{waterloo}{University of Waterloo}
  \icmlaffiliation{kcl}{King's College London}
  \icmlaffiliation{princeton}{Princeton University}
  \icmlaffiliation{helm}{Helm.ai}
  \icmlaffiliation{burapha}{Burapha University, Thailand}
  \icmlaffiliation{ahmedabad}{Ahmedabad University, India}
  \icmlaffiliation{illinois}{Illinois Institute of Technology}
  \icmlaffiliation{uludag}{Bursa Uludağ University}
  \icmlaffiliation{amsterdam}{University of Amsterdam}
  \icmlaffiliation{idsia}{USI, IDSIA}
  \icmlaffiliation{oxford}{University of Oxford}

  \icmlaffiliation{gutenberg}{Johannes Gutenberg University Mainz}
  \icmlaffiliation{york}{York University}
  \icmlaffiliation{charles}{Charles University}
  \icmlaffiliation{uiuc}{University of Illinois Urbana-Champaign}
  \icmlaffiliation{uchicago}{The University of Chicago}
  \icmlaffiliation{caltech}{California Institute of Technology}
  \icmlaffiliation{tuhamburg}{Technische Universität Hamburg}
  \icmlaffiliation{regensburg}{Universität Regensburg}
  \icmlaffiliation{helsinki}{University of Helsinki}
  \icmlaffiliation{uga}{University of Georgia}
  \icmlaffiliation{fortlewis}{Fort Lewis College}
  \icmlaffiliation{ubc}{University of British Columbia}
  \icmlaffiliation{edinburgh}{University of Edinburgh}
  \icmlaffiliation{bilkent}{Bilkent University}
  \icmlaffiliation{pompeu}{Universitat Pompeu Fabra}
  \icmlaffiliation{nagoya}{Nagoya University}
  \icmlcorrespondingauthor{Quanquan C. Liu}{quanquan.liu@yale.edu}

  \icmlkeywords{Machine Learning, Math Benchmark, LLM Evaluation, Alignment}

  \vskip 0.3in
]

\printAffiliationsAndNotice{\icmlEqualContribution}

\begin{abstract}
As Large Language Models (LLMs) saturate elementary benchmarks, the research frontier has shifted from generation 
to the reliability of automated evaluation. We demonstrate that standard ``LLM-as-a-Judge'' protocols suffer from 
a systematic \textbf{Alignment Gap} when applied to upper-undergraduate to early graduate level mathematics. 
To quantify this, we introduce \textbf{\benchname}, the first large-scale dual-rubric alignment benchmark to 
systematically measure alignment with human experts on university-level math proofs by contrasting 
\emph{course-specific rubrics} against \emph{expert common knowledge criteria}. By deploying a dual-evaluation 
matrix ($7$ judges $\times$ $5$ solvers) against 1,000+ hours of human evaluation, we reveal that certain 
frontier evaluators like Claude Opus 4.5, DeepSeek-V3, Qwen 2.5 Max, and Llama 4 Maverick exhibit significant positive 
bias (up to $+0.18, +0.20, +0.30, +0.36$ mean score inflation, respectively). Furthermore, we uncover a critical 
reasoning gap in the discrete domain: while \textbf{Gemini 3.0 Pro} achieves state-of-the-art performance (0.91 average
human evaluation score), other reasoning models like \textbf{GPT-5 Pro} and \textbf{Claude Sonnet 4.5} see their performance 
significantly degrade in discrete domains. Specifically, their average human evaluation scores drop to 0.72 and 0.63 
in Discrete Math, and to 0.74 and 0.50 in Graph Theory. In addition to these research results, 
we also release \benchname{} as a public benchmark 
for evaluating and improving AI judges. Our benchmark is publicly published at 
this \href{https://github.com/qqliu/Yale-QEDBench}{Github link}~\cite{qedbench}.
\end{abstract}

\section{Introduction}
\label{sec:intro}


As Large Language Models (LLMs) achieve saturation on elementary mathematical benchmarks \cite{cobbe2021gsm8k, hendrycks2021math_dataset}, the research frontier has shifted from generation to the reliability of automated evaluation. While models can now solve high-school competition problems (approaching gold-medal performance in specific domains), evaluating intricate \textit{proofs} at the upper-undergraduate to early graduate level remains an open alignment challenge. Existing benchmarks face a fundamental trade-off between verifiability and scalability. Auto-formalization frameworks (e.g., Lean~\citep{demoura2015lean}) provide correctness guarantees but incur high annotation costs that limit dataset scale. Conversely, standard `LLM-as-a-Judge' protocols are scalable but correlate poorly with expert judgment. This reliability gap makes it difficult to distinguish between valid logical reasoning and hallucinated pseudo-precision. Indeed, recent investigations reveal that relying strictly on numerical or short-answer evaluation masks profound reasoning shortcomings. Advanced reasoning models frequently employ heuristic shortcuts or hallucinated premises to arrive at correct final answers, highlighting the urgent need for rigorous proof-based diagnostics \citep{guo2025litmus}. Theoretical frameworks corroborate this discrepancy; for instance, \citet{zhang2024dagmath} mapped intermediate mathematical reasoning into directed acyclic graphs, revealing a substantial gap between final-answer accuracy and rule-consistent logical derivation.

This paper introduces \textbf{\benchname}, a rigorous evaluation framework designed not just to test model capabilities in proof-based mathematics at the upper undergraduate and early graduate level, but to audit the LLM judges themselves. Unlike prior works that implicitly tie evaluation quality to model performance, we explicitly decouple these two axes to provide a systematic audit of university proof generation. Our study systematically evaluates proof synthesis and judging across ten core upper-undergraduate/early-graduate disciplines: Analysis, Complex Analysis, Abstract Algebra, Discrete Math, Probability Theory, ODEs, Number Theory, Combinatorics, Algorithms, and Graph Theory.

\textbf{Methodology Overview.} Our experimental design follows a three-stage pipeline to establish a \emph{Human Expert Ground Truth}:

\textit{1. Solution Generation.} First, to ensure our dataset captures the capabilities of current state-of-the-art systems, we generated solutions for our problem set using five frontier models: \texttt{o3-deep-research}, \texttt{GPT-5 Pro}, \texttt{Claude Sonnet 4.5}, \texttt{Gemini 3.0 Pro}, and \texttt{DeepSeek-Prover-V2}. We selected these specific models as they represented the most recent versions of the frontier models available at the time the solution generation and subsequent human evaluation process began (thus pre-dating the release of models like GPT-5.2 Pro and Claude Sonnet 4.6). This diversity ensures our evaluation is not over-fitted to the idiosyncrasies of a single model family. For each problem, we set model-specific token limits (up to 16,384 output tokens) and a timeout window of up to 2,000 seconds for streaming APIs; if a model failed to produce a solution (e.g., due to an API timeout or connection error), we re-attempted generation up to three times. If the model still failed to return a valid proof after all retries, we assigned a default score of $0$, reflecting the model's inability to produce output under real-world deployment conditions.

\textit{2. Expert Annotation \& Rubric Engineering.} Raw model outputs are insufficient for benchmarking without verified ground truth. We employed a team of PhD-level experts to evaluate these solutions on a granular scale $[0, 0.25, 0.5, 0.75, 0.9, 1.0]$. To isolate the impact of rubric specificity on automated grading, we generated two distinct types of rubrics using \texttt{GPT-5.2 Pro} (verified by \texttt{Gemini 3.0 Pro}):
\begin{itemize}
    \item \textbf{Course-Specific Rubric:} Simulating a standard undergraduate grading rubric, focused on textbook definitions for the relevant course.
    \item \textbf{Expert-Domain Rubric:} Rubrics designed for a standard of fundamental correctness assuming expert common knowledge, where proofs are evaluated on logical soundness and implicit steps are permitted if trivial to a domain specialist, mirroring graduate-level evaluation.
\end{itemize}
Crucially, our human experts refined these rubrics iteratively to match their own evaluation criteria for the corresponding problems, establishing a verified standard for correctness.

\textit{3. Judge Calibration.} Finally, we turned to the primary focus of this work: the reliability of automated evaluators. We deployed \emph{seven} judge models (including \texttt{GPT-5.2 Pro}, \texttt{Claude Opus 4.5}, and \texttt{Gemini 3.0 Pro}) to grade the generated proofs against both rubrics. By comparing these $7 \times 5$ evaluation matrices against the human ground truth, we can quantify the alignment gap by isolating systemic judge bias from model-specific noise.

We address three fundamental questions regarding the trustworthiness of LLM evaluators for upper undergraduate to early graduate-level proof writing and correctness:
\begin{enumerate}
    \item \textbf{The Evaluator Reliability Question:} Do off-the-shelf LLMs perform genuine logical verification in abstract domains that align with expert human evaluators?
    \item \textbf{The Reasoning Frontier:} When evaluated against a strict human ground truth, how do frontier models perform on complex proofs requiring constructive reasoning beyond retrieval?
    \item \textbf{The Knowledge Gap:} Do models adhere to the pedagogical constraints of undergraduate definitions, or do they ensure correctness by retrieving advanced, out-of-scope theorems?
\end{enumerate}

\subsection{Key Contributions}
\begin{itemize}
    \item \textbf{Quantifying the Alignment Gap:} We demonstrate that standard LLM judges exhibit a systematic positive bias compared to human experts. Our dual-evaluation matrix reveals that frontier evaluators like \texttt{Claude Opus 4.5} inflate scores by up to \textbf{+0.18} (mean score delta) in abstract domains. Similarly, \texttt{Llama 4 Maverick} acts as a grade inflator, achieving a 90.2\% pass rate against a 67.7\% human baseline.
    \item \textbf{The Discrete-Continuous Divide:} Using our calibrated evaluation, we uncover a critical disparity in reasoning architectures. While \textbf{Gemini 3.0 Pro} achieves state-of-the-art overall performance across most categories (0.91 raw score), other frontier reasoning models like \textbf{Claude Sonnet 4.5} see their performance significantly degrade, achieving 100\% pass rate in ODEs but dropping to \textbf{27.3\%} in Combinatorics (\cref{fig:pass-rate}).
    \item \textbf{A Rigorous Calibration Standard:} \benchname{} provides the first statistically robust standard for automated university-level mathematical proof evaluation, grounded in over 1,000+ hours of hand-evaluation by PhD-level experts. We publicly release the full dataset, the dual-rubric system, and the evaluation logs to enable the community to benchmark future AI judges against a verified human ground truth.
\end{itemize}

\noindent\textbf{Conflict of Interest Disclosure.}
One author is affiliated with Helm.ai. This work does not evaluate models developed by Helm.ai; to the best of our knowledge, the authors have no financial conflicts of interest with the providers of the evaluated systems.




\section{Related Work}

\textbf{Benchmarking the Reasoning Frontier.}
The evaluation of mathematical capability has evolved from foundational arithmetic \cite{cobbe2021gsm8k, hendrycks2021math_dataset} to domains requiring sophisticated abstraction. Recent efforts like \textit{FrontierMath} \cite{frontiermath2024} and \textit{IMProofBench} \cite{schmitt2025improofbench} have pushed the difficulty frontier to expert and research levels. However, a trade-off exists between difficulty and evaluation fidelity: \textit{FrontierMath} relies on closed-form answers to ensure verifying correctness is tractable, while \textit{IMProofBench} offers a smaller-scale (approx.~39 problems) look at research generation. Similarly, \textit{LiveBench} \cite{livebench2024} addresses contamination via novel problem updates but evaluates proofs via ``cloze-style'' reordering rather than \textit{open-ended proof generation} where
only the problem is provided to the LLM without any other information. Other works have explored specific reasoning dimensions, such as temporal dependencies \citep{fatemi2024test}, prompt-induced chain-of-thought \citep{sessler2024benchmarking}, and the ability to generate constructive mathematical objects \citep{balunovic2025mathconstruct}. Concurrently, U-MATH \citep{chernyshev2025umath} has expanded evaluation to the university level, introducing a meta-evaluation subset ($\mu$-MATH) specifically to assess judge inconsistencies. More recently, \citet{zheng2026unmasking} introduced a process-aware benchmark demonstrating that LLMs degrade significantly in structural reasoning scenarios requiring ``multi-constraint coordination.'' Initiatives like the Rosetta Stone project \citep{ho2025rosetta} attempt to unify these heterogeneous metrics into a single rating. \textsc{QedBench} occupies a distinct middle ground: we target the upper undergraduate/early graduate curriculum (harder than high school competitions but more verifiable than novel research) prioritizing the semantic quality of full-text proofs over binary answer matching.

\textbf{Dynamics of Large Reasoning Models (LRMs).}
The emergence of ``Large Reasoning Models'' (e.g., o3, Gemini Thinking) has prompted deeper analysis into the nature of model thought. \citet{bhoopchand2025illusion} analyze these models through the lens of algorithmic complexity, suggesting that current improvements may stem from pattern matching in controlled environments rather than generalized reasoning. Concurrently, \textit{ReasonBENCH} \cite{liu2025reasonbench} highlights the ``instability'' of these reasoning traces, quantifying high variance in logic despite correct final answers. Our work extends this scrutiny to the domain of higher mathematics, investigating whether the ``reasoning stability'' observed in algorithmic tasks holds up when models attempt rigorous theorem proving.

\textbf{The Challenge of Automated Evaluation.}
As generation capabilities saturate, the bottleneck shifts to evaluation. The ``LLM-as-a-Judge'' paradigm \cite{zheng2023judging} has shown promise in competition mathematics. Pioneering efforts like \textit{ProofBench} (also referred to as \textit{Reliable Fine-Grained Evaluation}) \cite{ma2025reliable} and \textit{RefGrader} \cite{mahdavi2025refgrader} demonstrate that fine-tuned evaluators or agentic workflows can achieve high correlation with human judges on Olympiad-style problems (IMO/USAMO). However, recent human-evaluation studies on Olympiad proofs \citep{dekoninck2025openproof, petrov2025prooforbluff, mahdavi2025brains} highlight that LLM evaluators frequently overestimate proof quality and reward superficial plausibility over rigorous logic. Furthermore, recent audits reveal that automated judges rely heavily on stylistic heuristics rather than deep mathematical understanding \citep{mahdavi2025scaling}, exhibiting a severe bias toward the authoritative ``writing style'' of the generator model \citep{stephan2025calculation}. Building on these findings, we identify a critical \textit{Alignment Gap} when evaluation methods scale to the upper-undergraduate/early-graduate level. Unlike \citet{ma2025reliable}, who fine-tune specific judges on competition data, we evaluate the inherent alignment of frontier foundation models. We find that in the absence of rigid competition rubrics, models like Claude Opus 4.5 misalign with human evaluators, sometimes rewarding persuasive language that lacks logical soundness. \benchname{} thus 
serves as a counter-weight to competition-centric benchmarks, revealing that \textit{solving} 
proficiency does not guarantee \textit{grading} reliability in abstract mathematical domains.

\section{\benchname{} Methods}\label{sec:methods}

The \benchname{} benchmark comprises 272 expert-curated, proof-based mathematical problems spanning upper-undergraduate and early-graduate curricula. The domain coverage includes Analysis, Complex Analysis, Abstract Algebra, Discrete Math, Probability Theory, ODEs, Number Theory, Combinatorics, Algorithms, and Graph Theory. Because each problem is solved by five frontier models, \benchname{} contains more than 1,300 full generated proofs with expert scores, public qualitative annotations, and multi-rubric LLM-judge evaluations. Our methodology rests on three pillars: adversarial data curation, a tiered-rubric expert evaluation, and a cross-model evaluation matrix.

\subsection{Problem Curation and Anti-Leakage}
To mitigate the risk of data contamination and memorization, we implemented a rigorous three-stage curation pipeline. 

\begin{itemize}
    \item \textbf{Source Diversity:} Problems were sourced from authoritative graduate texts (e.g., \citealp{beals2004analysis, dummit2003abstract}) and advanced qualifying exams (e.g., Kent State) to ensure high mathematical quality.
    \item \textbf{Syntactic Rewording:} We manually rephrased all problem statements into \textit{ab initio} equivalents. This process alters linguistic patterns while preserving the underlying logical structure, neutralizing simple surface-level memorization.
    \item \textbf{Adversarial Audit:} We subjected every candidate problem to an automated deep-search audit. We utilized \texttt{o3-deep-research} to scan the open web for existing solutions to our rephrased prompts.
\end{itemize}

The exact adversarial audit prompt is provided in \cref{app:search_prompt}. We instructed the agent to strictly distinguish between \textit{equivalent} solutions (contamination) and \textit{analogous} problems (acceptable domain overlap).

Of the 272 problems, we obtained high-confidence solution availability annotations for 214: the audit confirmed online solutions for 88 instances, while 126 returned no matches. The remaining 58 problems were excluded due to ambiguous classification; notably, all 19 Graph Theory problems fell into this ambiguous category and were thus excluded from this search (explaining their absence in downstream contamination plots). In~\cref{sec:results}, we analyze performance variance between these two subsets to quantify the impact of potential contamination.

\subsection{Expert Evaluator Selection}
We recruited 48 expert evaluators, all of whom either have taken PhD-level coursework, are current PhD candidates, or are holders of a PhD in mathematics or theoretical computer science. To ensure high-fidelity grading, we matched evaluators to problems strictly within their publication domains. The human-evaluations resulted in a combined 1,000+ hours of work.

Evaluators graded solutions using a granular tiered Expert Rubric, which is adapted to the specifics of each problem
based on the tiered standard rubric described in~\cref{app:tiered_rubric}. 
Unlike binary pass/fail metrics, this scale $[0, 0.25, 0.5, 0.75, 0.9, 1.0]$ is designed to distinguish 
between fundamental logical failures (0.25) and expository oversights (0.9). Evaluators provided a numerical score, a justification, and a marked-up solution identifying specific error locations.
Each problem was assigned to two independent experts whenever possible; disagreements were adjudicated by a third expert. Graders were explicitly instructed to award full credit for any logically complete proof, including proofs that followed a valid path different from the reference solution.

\subsection{The Evaluation Alignment Gap: Dual-Rubric Strategy}
To decouple ``superficial plausibility'' from ``logical rigor,'' we quantify the evaluation \textbf{Alignment Gap}: the discrepancy between an LLM judge's score and the human expert ground truth. 
We employ a \textbf{Dual-Rubric Strategy} to isolate the source of this gap:
\begin{enumerate}
    \item \textbf{Course-Specific Rubric:} A standard pedagogical rubric focusing on definition compliance and step-by-step derivations, simulating a university-level grading environment.
    \item \textbf{Expert Rubric:} A high-granularity schema calibrated to research standards. This rubric penalizes ``hidden'' logical fallacies and circular reasoning that typically escape surface-level verification.
\end{enumerate}

Initial rubrics were synthesized by \texttt{GPT-5.2 Pro} and verified by \texttt{Gemini 3.0 Pro}. Crucially, human experts then iteratively refined these drafts to align them with their own evaluation and grading standards. 
Full examples illustrating the difference between expert and course-specific rubrics appear in \cref{app:example_rubrics}. Humans evaluated the proofs using the Expert Rubric, while LLM judges evaluated the proofs using both rubrics.

\subsection{Evaluation Framework}
We utilize a $7 \times 5$ \textbf{Evaluator-Solver Matrix} to marginalize individual model bias. We gathered \LaTeX{} solutions from 5 frontier solver models and evaluated them using 7 state-of-the-art judge models (including GPT, Claude, Gemini, Grok, Qwen, DeepSeek, and Llama model families) against both rubrics. All automated evaluations were run at temperature $T=0.0$ for deterministic reproducibility. This fully crossed factorial design allows us to statistically isolate \textit{Judge Bias} from true \textit{Solver Skill}; full evaluator prompts and the complete evaluator-solver pass-rate matrix appear in \cref{app:evaluator_prompts,fig:llm-pass-rate-matrix}.

\section{Results}
\label{sec:results}

In this section, we detail the findings that address the three key questions outlined in \cref{sec:intro}. First, we analyze the performance of frontier models on complex proofs requiring constructive reasoning, using expert human evaluations across ten mathematical disciplines. Second, we quantify the alignment gap between automated judges and human experts. Finally, we investigate the ``knowledge gap,'' assessing whether models adhere to pedagogical constraints or rely on advanced, out-of-scope machinery.

\subsection{Frontier Model Performance}

We evaluated five frontier models: \texttt{Gemini 3.0 Pro}, \texttt{GPT-5 Pro}, \texttt{o3-deep-research}, \texttt{Claude Sonnet 4.5}, and \texttt{DeepSeek-Prover-V2}. These represented the state-of-the-art reasoning architectures at the time of the human evaluation.

We assess competency through two primary metrics:
\begin{enumerate}
    \item \textbf{Average Score:} The mean score (0.0--1.0) assigned by expert human judges, capturing partial credit for conceptual understanding.
    \item \textbf{Pass Rate:} The percentage of problems receiving a score of $\ge 0.9$, indicating a proof that is logically complete with only minor expository gaps.
\end{enumerate}

\begin{figure}[t]
    \centering
    \includegraphics[width=\linewidth]{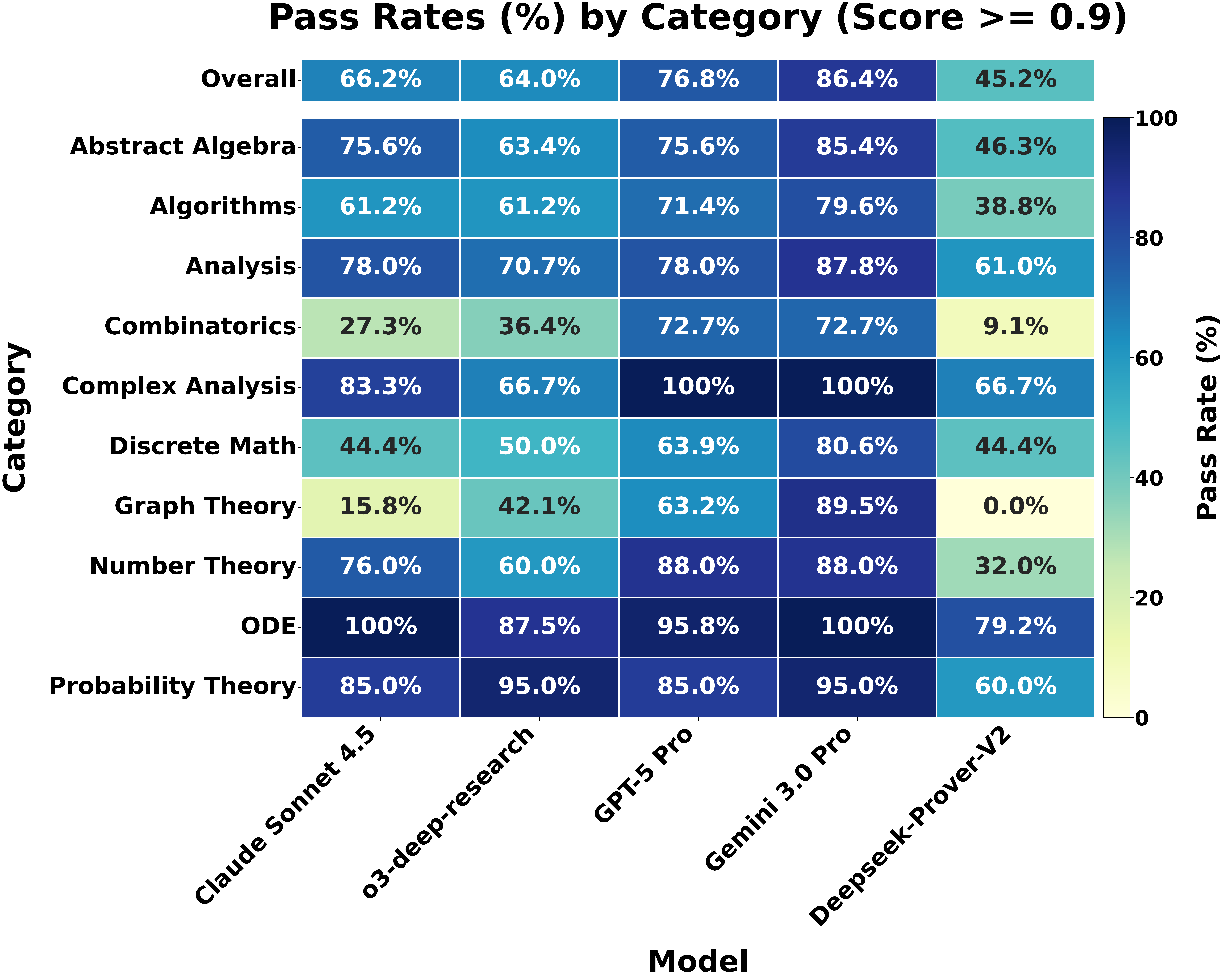}
    \caption{\textbf{Pass Rates by Discipline.} The pass rates (score $\ge 0.9$) of frontier models across various mathematical disciplines. We observe that while models achieve high reliability in calculation-heavy domains like ODEs and Probability, performance drops significantly in structure-heavy domains such as Combinatorics and Graph Theory.}
    \label{fig:pass-rate}
\end{figure}

\textbf{Overall Rankings.}
As summarized in \cref{fig:pass-rate}, \textbf{Gemini 3.0 Pro} achieves state-of-the-art performance with an 86.4\% overall pass rate and an average score of $0.91$, followed by \textbf{GPT-5 Pro} at 76.8\% and an average score of $0.84$.

\begin{figure}[t]
    \centering
    \includegraphics[width=\linewidth]{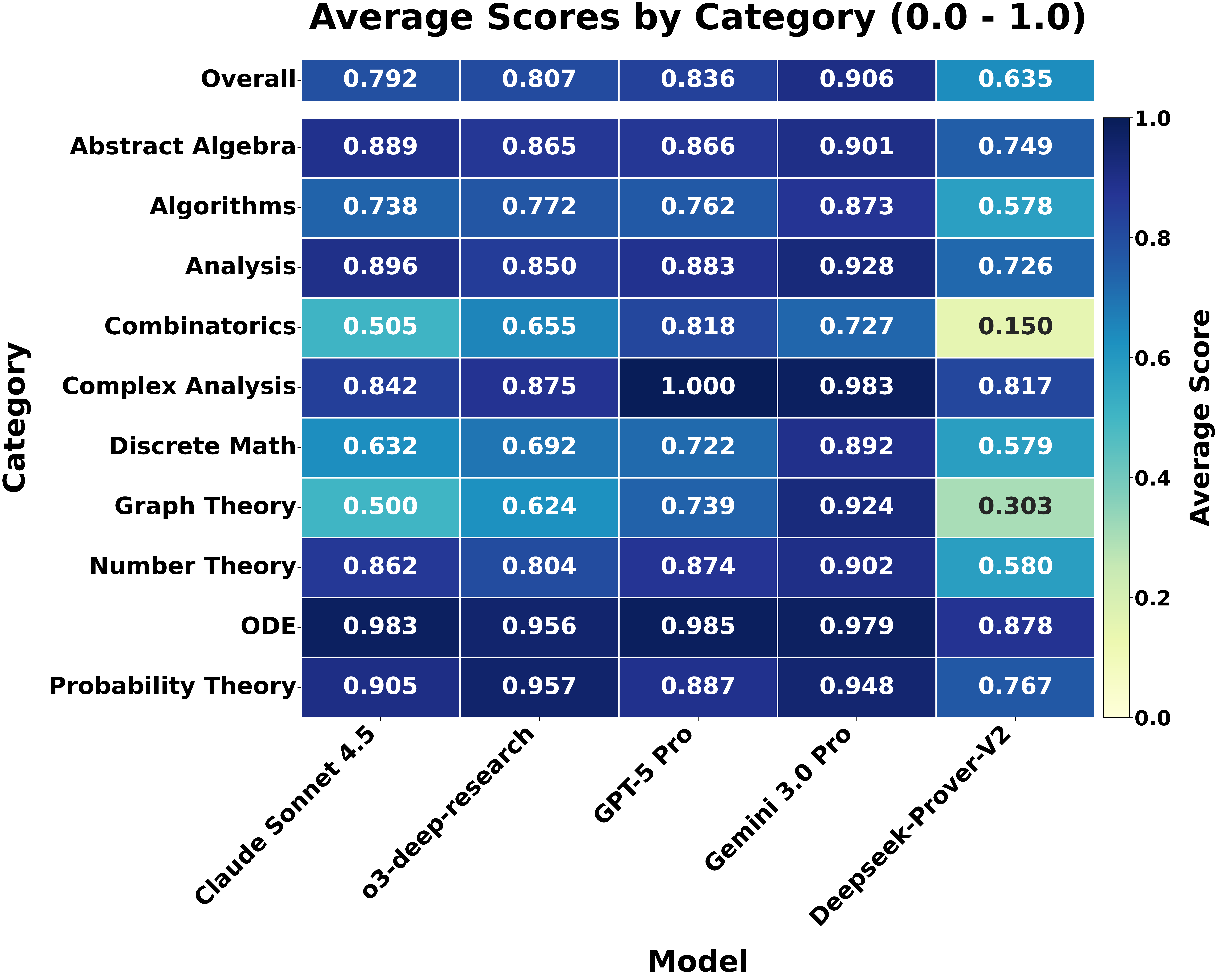}
    \caption{\textbf{Average Score Distribution.} The average evaluation scores (0.0--1.0) assigned by expert judges. Unlike the binary pass rate, this metric accounts for partial credit, revealing that models often demonstrate strong conceptual understanding (high partial scores) in domains like Analysis even when failing to produce fully rigorous proofs.}
    \label{fig:average-score}
\end{figure}

\subsubsection{Key Observations and Analysis}
\label{sec:observations}

Our expert human evaluation reveals systemic divergences in how frontier models approach university-level mathematical reasoning.
We identify three critical phenomena that define the current state of automated proof generation.

\paragraph{1. The ``Template vs. Construction'' Divide.}
Performance does not degrade uniformly; rather, it correlates with the necessity of constructive search (\cref{fig:pass-rate}).
\begin{itemize}
    \item \textbf{Algorithmic Saturation:} In domains like \textit{ODEs} and \textit{Complex Analysis}, where proofs often follow a procedural ``recipe'' (e.g., applying the Residue Theorem or separating variables), frontier models achieve near-saturation. \texttt{Gemini 3.0 Pro} and \texttt{Claude Sonnet 4.5} both achieved \textbf{100.0\%} pass rates in ODEs.
    \item \textbf{Combinatorial Weaknesses:} In contrast, domains requiring the construction of novel objects (e.g., bijections or counter-examples) show high failure rates. In \textit{Combinatorics}, \texttt{DeepSeek-Prover-V2} dropped to a \textbf{9.1\%} pass rate, and \texttt{Claude Sonnet 4.5} achieved only \textbf{27.3\%}. This suggests that while models excel at \textit{retrieving} theorem templates, they struggle to \textit{search} finite state spaces for constructive proofs. This structural deficit is heavily corroborated by recent domain-specific evaluations. The \textit{MathConstruct} benchmark \citep{balunovic2025mathconstruct} demonstrates that while LLMs easily saturate continuous numerical benchmarks, they systematically fail at tasks requiring the explicit construction of novel mathematical objects (e.g., bijections, graphs) satisfying strict constraints. Similarly, \citet{zheng2026unmasking} found that LLMs successfully navigate template-based arithmetic but degrade significantly in structural reasoning scenarios requiring constructive logical synthesis, perfectly mirroring our observed discrete-continuous divide.
\end{itemize}

\paragraph{2. Variance in Discrete Reasoning.}
Contrary to the hypothesis that models are generally ``good'' or ``bad'' at discrete math, we observe large differences in performance between related sub-fields.
\begin{itemize}
    \item While \texttt{Gemini 3.0 Pro}'s weakest area is Combinatorics (72.7\% pass rate), it achieved a state-of-the-art \textbf{89.5\%} pass rate in \textit{Graph Theory}, significantly outperforming \texttt{Claude Sonnet 4.5} (15.8\%) and \texttt{o3-deep-research} (42.1\%). This suggests that rather than a pure reasoning collapse, performance significantly degrades in exhaustive search domains.
    \item This contrasting performance within discrete mathematics suggests that ``reasoning'' is not yet a generalized capability. Instead, model performance may be highly sensitive to the density of domain-specific training data (e.g., the prevalence of graph-theoretic proofs in the pre-training corpus versus combinatorial arguments).
\end{itemize}

\paragraph{3. The ``Partial Credit'' Trap.}
A comparison of Pass Rates (\cref{fig:pass-rate}) vs. Average Scores (\cref{fig:average-score}) reveals that binary success metrics mask significant nuances in model capability.
\begin{itemize}
    \item \textbf{The Almost Correct Solutions:} In domains like \textit{Analysis}, models frequently achieve high average scores (e.g., $\approx 0.88$) despite lower strict pass rates. This discrepancy indicates a high volume of proofs scoring in the $0.75$ band: solutions that are structurally sound and ``plausible''
    but contain a few subtle, fatal logical gaps (e.g., conflating uniform vs. pointwise convergence).
    \item \textbf{Implications for Training:} This prevalence of ``near-miss'' proofs poses a critical challenge for future reinforcement learning. Models have learned the \textit{form} of graduate-level reasoning but often fail on the precise logical implications required for full rigor. A reward model trained only on binary correctness would miss these signals of partial understanding, potentially reinforcing the generation of ``hallucinated rigor'' rather than correcting the underlying logic.
\end{itemize}

\subsection{Evaluation Alignment: Human Experts and LLMs}

To validate our automated evaluation pipeline, we compared seven evaluator models against a human ground truth baseline derived from our expert evaluations. We assessed evaluators on three axes: \textit{Strictness}, \textit{Domain Bias}, and \textit{Agreement}.

\begin{figure}[t]
    \centering
    \includegraphics[width=\linewidth]{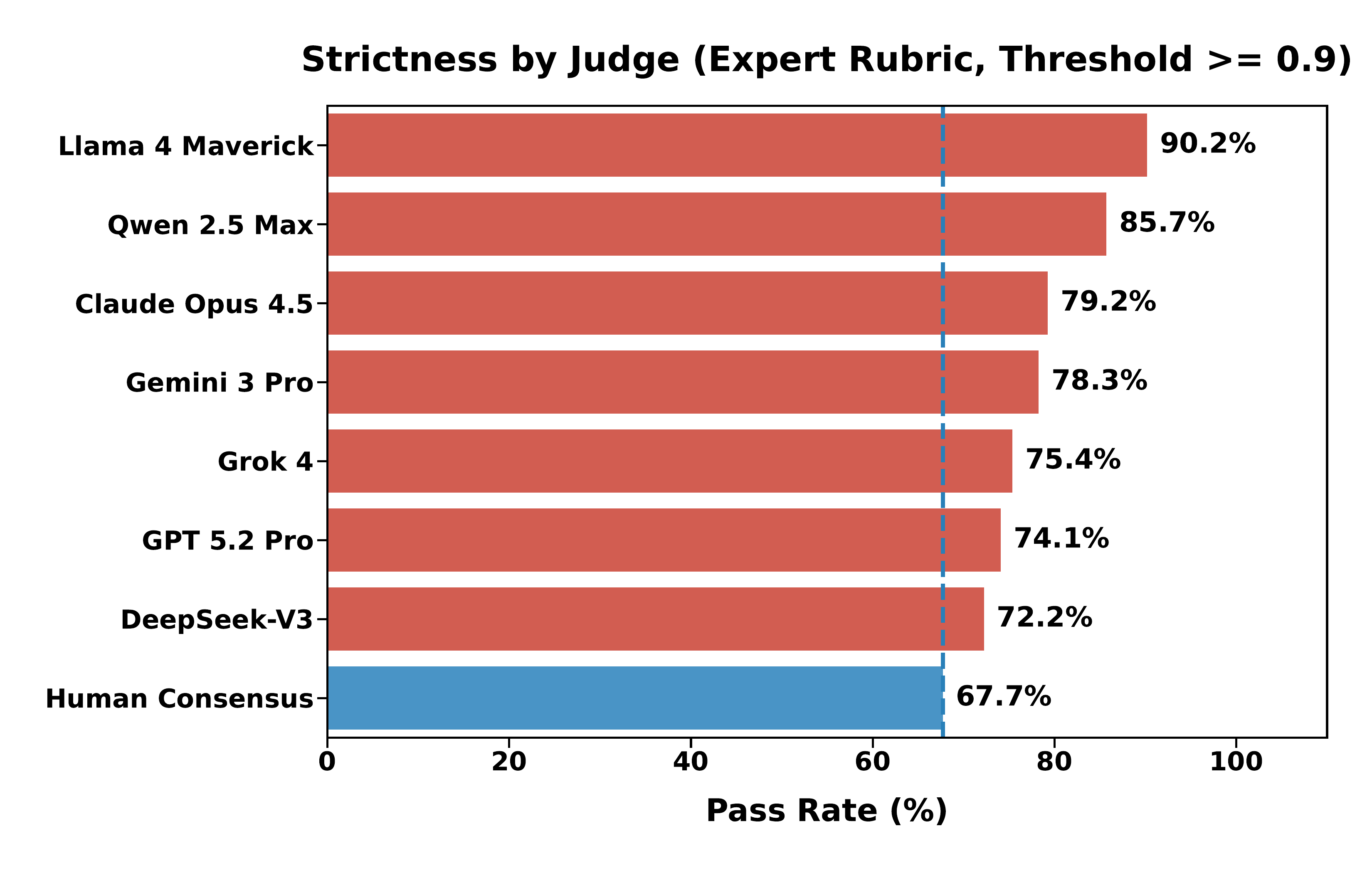}
    \caption{\textbf{Evaluator Strictness.} Comparison of pass rates ($\text{score} \ge 0.9$) across judges. While \texttt{DeepSeek-V3} (72.2\%) aligns most closely with the Human Consensus (67.7\%), \texttt{Llama 4 Maverick} (90.2\%) exhibits significant grade inflation.}
    \label{fig:strictness}
\end{figure}

\paragraph{1. The Strictness Gap.}
As shown in Figure \ref{fig:strictness}, LLM judges vary wildly in their leniency. 
Human experts established a baseline pass rate of \textbf{67.7\%} across all problems.
\begin{itemize}
    \item \textbf{Grade Inflation:} \texttt{Llama 4 Maverick} acted as a significant grade inflator, passing \textbf{90.2\%} of solutions. Qualitative review suggests it frequently failed to distinguish between subtle logical gaps and correct reasoning, approving proofs based on surface-level coherence.
    \item \textbf{Best Alignment:} \texttt{DeepSeek-V3} achieved the closest alignment (72.2\%) to the human baseline pass rate, followed by \texttt{GPT-5.2 Pro} (74.1\%).
\end{itemize}
(For a complementary analysis of strictness using Average Scores rather than Pass Rates, see \cref{app:evaluator_strictness_avg}).

\begin{figure}[t]
    \centering
    \includegraphics[width=\linewidth]{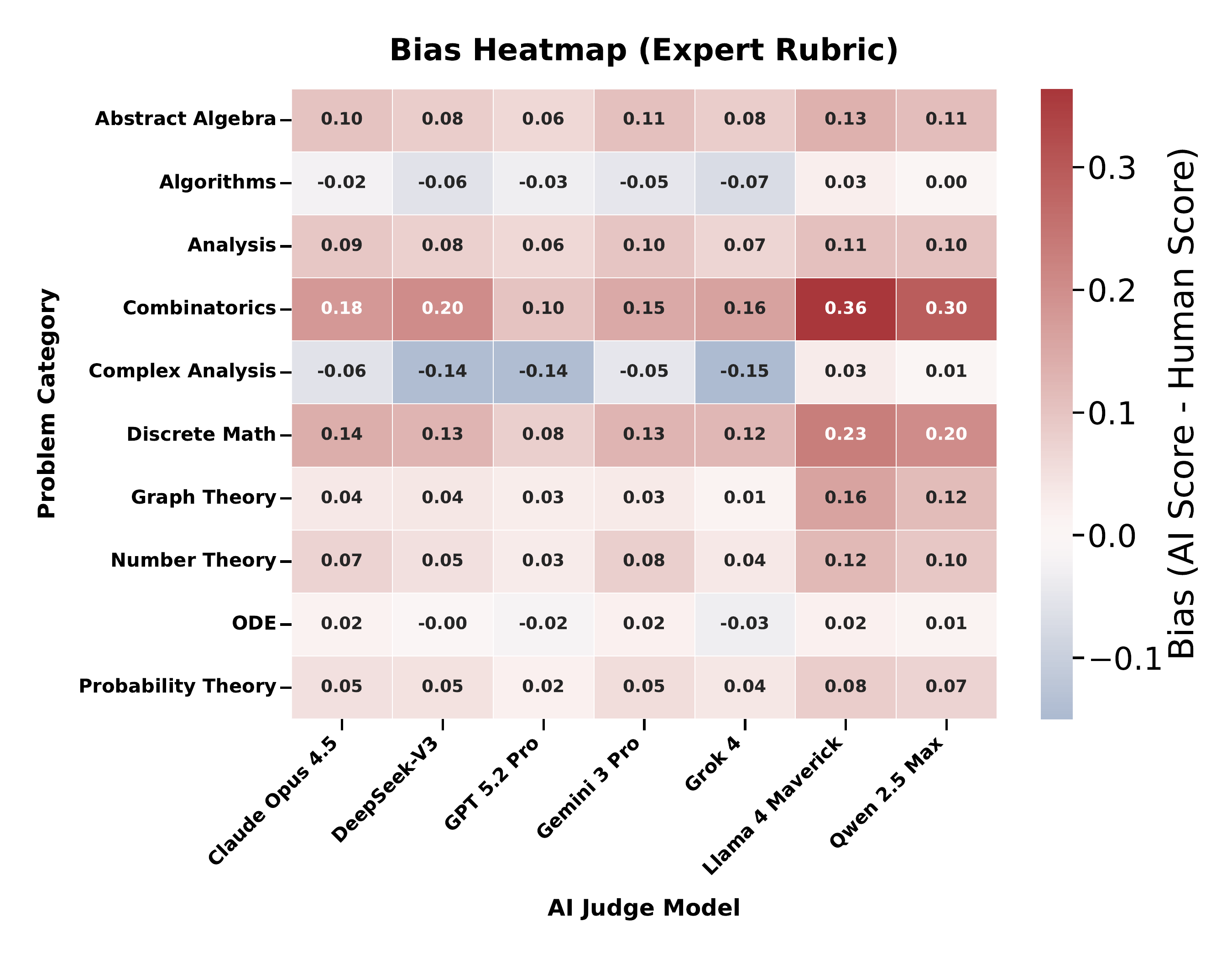}
    \caption{\textbf{Evaluator Bias Heatmap.} The delta between AI and Human scores ($\Delta = S_{AI} - S_{Human}$). Positive values (red) denote score inflation (AI is more lenient), while negative values (blue) signify punitive bias (Human is more lenient). These systemic deltas highlight a bidirectional alignment gap: models tend 
    to be more lenient in discrete domains while being more punitive in continuous analysis.}
    \label{fig:bias}
\end{figure}

\paragraph{2. The Bidirectional Alignment Gap.}
Strictness averages can mask domain-specific failures. We calculated the bias $\Delta = \text{Score}_{\text{AI}} - \text{Score}_{\text{Human}}$ for each discipline. Figure \ref{fig:bias} illustrates that misalignment is not uniform; rather, it is highly domain-dependent.
\begin{itemize}
    \item \textbf{Discrete Inflation (AI Leniency):} In \textit{Combinatorics}, we observe a massive positive bias. \texttt{Llama 4 Maverick} (+0.36) and \texttt{Qwen 2.5 Max} (+0.30) consistently grade-inflate. This suggests that in domains relying on discrete structural arguments, models are prone to rewarding solutions that \textit{look} rigorous despite containing errors.
    \item \textbf{Complex Analysis Punitiveness (Human Leniency):} Conversely, in \textit{Complex Analysis}, the bias flips negative. Evaluators like \texttt{DeepSeek-V3} (-0.14) and \texttt{Grok 4} (-0.15) are significantly harsher than human experts. This indicates that human judges often accept implicit steps in standard contour integrations (expert common knowledge), whereas AI models rigidly penalize valid solutions that lack explicit derivation.
    \item \textbf{Algorithmic Rigidity:} A similar trend appears in \textit{Algorithms}, 
    where \texttt{Gemini 3 Pro} (-0.05) and \texttt{Grok 4} (-0.07) display punitive biases, likely penalizing correct pseudo-code that deviates from training templates.
\end{itemize}
Overall, \texttt{GPT-5.2 Pro} most closely matches human evaluators across the widest range of domains. The heatmap demonstrates that alignment is not merely about making models ``stricter'' or ``nicer,'' but calibrating them to the specific rigorous standards of each sub-field.
(For a complementary analysis of the evaluator bias using Pass Rates rather than Average Scores, see \cref{app:evaluator_bias_pass_rate}).

\begin{figure}[t]
    \centering
    \includegraphics[width=\linewidth]{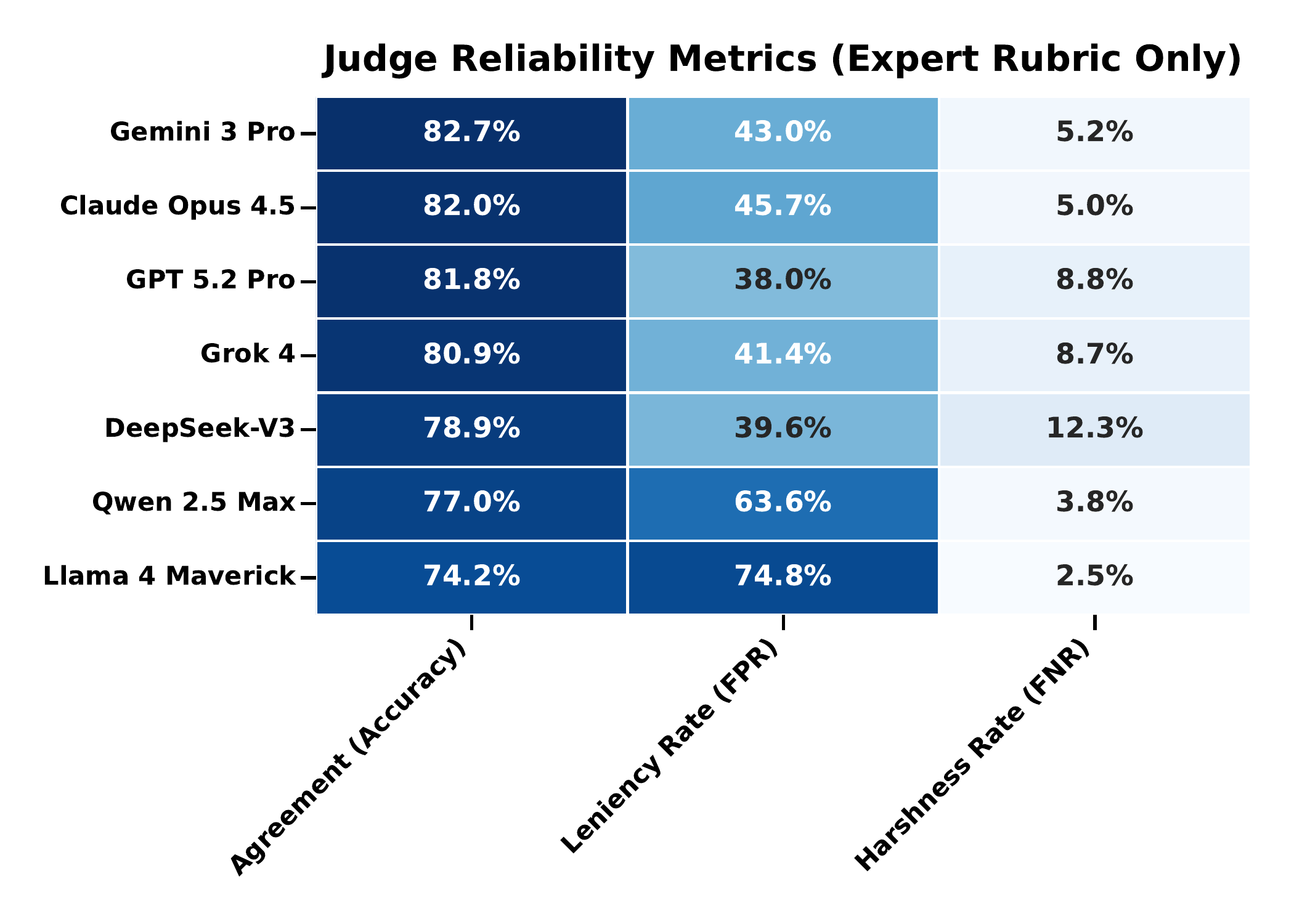}
    \caption{\textbf{Judge Reliability Metrics.} We decompose errors into \textit{Leniency Rate} (False Positives) and \textit{Harshness Rate} (False Negatives), calculated using a binary Pass Rate threshold ($\text{score} \ge 0.9$). \texttt{Llama 4 Maverick} exhibits extreme alignment failure with a \textbf{74.8\% Leniency Rate}. In contrast, \texttt{DeepSeek-V3} displays the highest \textit{Harshness Rate} (\textbf{12.3\%}).}
    \label{fig:agreement}
\end{figure}

\paragraph{3. Decomposing Error Modes: Leniency vs. Rigidity.}
Figure \ref{fig:agreement} dissects the specific failure modes of automated evaluators by isolating False Positives (Hallucinated Rigor) from False Negatives (Harshness). Note that these reliability metrics are evaluated strictly on binary Pass Rates (where a score $\ge 0.9$ is considered a pass), rather than average score deltas.
\begin{itemize}
    \item \textbf{Leniency:} The most alarming result is the behavior of \texttt{Llama 4 Maverick}, which exhibits a \textbf{74.8\% Leniency Rate} relative to human ground truth. 
    This suggests the model has over-optimized for instruction-template matching rather than critical analysis, approving nearly three-quarters of logically flawed proofs. This systemic leniency is consistent with recent literature characterizing the vulnerabilities of automated evaluators. \citet{jain2025beyond} formally quantified this phenomenon as the ``Agreeableness Bias.'' Through large-scale empirical evaluation, they demonstrated that while LLM judges easily identify correct solutions (True Positive Rate $>96\%$), they catastrophically fail at rejecting invalid ones (True Negative Rate $< 25\%$), perfectly reflecting the massive False Positive inflation we observe.
    \item \textbf{The Rigidity Barrier:} Conversely, \texttt{DeepSeek-V3} acts as the strictest auditor, with a benchmark-high \textbf{12.3\% Harshness Rate}. While it offers a lower Leniency Rate (39.6\%), its high rejection of valid solutions indicates a failure to generalize beyond standard proof templates.
    \item \textbf{The Reliability Ceiling:} Even our strongest evaluator, \texttt{GPT-5.2 Pro}, maintains a \textbf{38.0\% Leniency Rate}, highlighting the 
    persistent difficulty of automated alignment with human mathematical judgment.
\end{itemize}
(For an extended analysis of how these error rates shift under the stricter Course-Specific Rubric, see \cref{app:judge_reliability_course}. Additionally, \cref{app:leniency_distribution} presents a granular, per-judge distribution mapping strict vs. lenient grading proportions).

\paragraph{Contamination Check.}
We audited all 272 problems using \texttt{o3-deep-research}; among the 214 problems with high-confidence labels, 88 had online solutions and 126 did not. Human-evaluated performance differed only marginally between these groups (mean-score gap $+0.017$, pass-rate gap $+0.011$), with no statistically significant advantage for online problems. Full tests, tables, and figures appear in \cref{sec:contamination}.

\paragraph{Robustness to Confounds.}
We ran additional analyses to rule out several non-alignment explanations. A mixed-effects regression controlling for human proof quality and baseline judge strictness found negligible same-family self-preference effects ($-0.014$ to $+0.033$). Strict binary Pass/Fail prompts preserved the same leniency trends, showing that score discretization is not the cause. Finally, proof formatting explains part of the failure mode: \texttt{Qwen 2.5 Max} and \texttt{Llama 4 Maverick} correlate positively with \LaTeX{} density ($r=0.169$ and $r=0.157$, $p<10^{-4}$), whereas \texttt{GPT-5.2 Pro} does not. Details appear in \cref{app:robustness_analyses}.

\subsection{Ablation Study: The Limits of Rubric Engineering}
\label{sec:rubric_ablation}

A common hypothesis in current literature is that LLM ``Judge'' models fail primarily due to underspecified instructions. To test this, we conducted a controlled ablation study on our highest-performing judge, \textbf{GPT-5.2 Pro}, comparing its alignment with human consensus under two distinct grading protocols:
\begin{enumerate}
    \item \textbf{Expert Rubric:} A standard prompt focusing on general mathematical correctness and logical soundness using expert-level common knowledge.
    \item \textbf{Course-Specific Rubric:} A constrained prompt explicitly penalizing the use of advanced machinery (e.g., the Residue Theorem or Lebesgue Dominated Convergence Theorem) without derivation.
\end{enumerate}

\begin{figure}[t]
    \centering
    \includegraphics[width=0.95\linewidth]{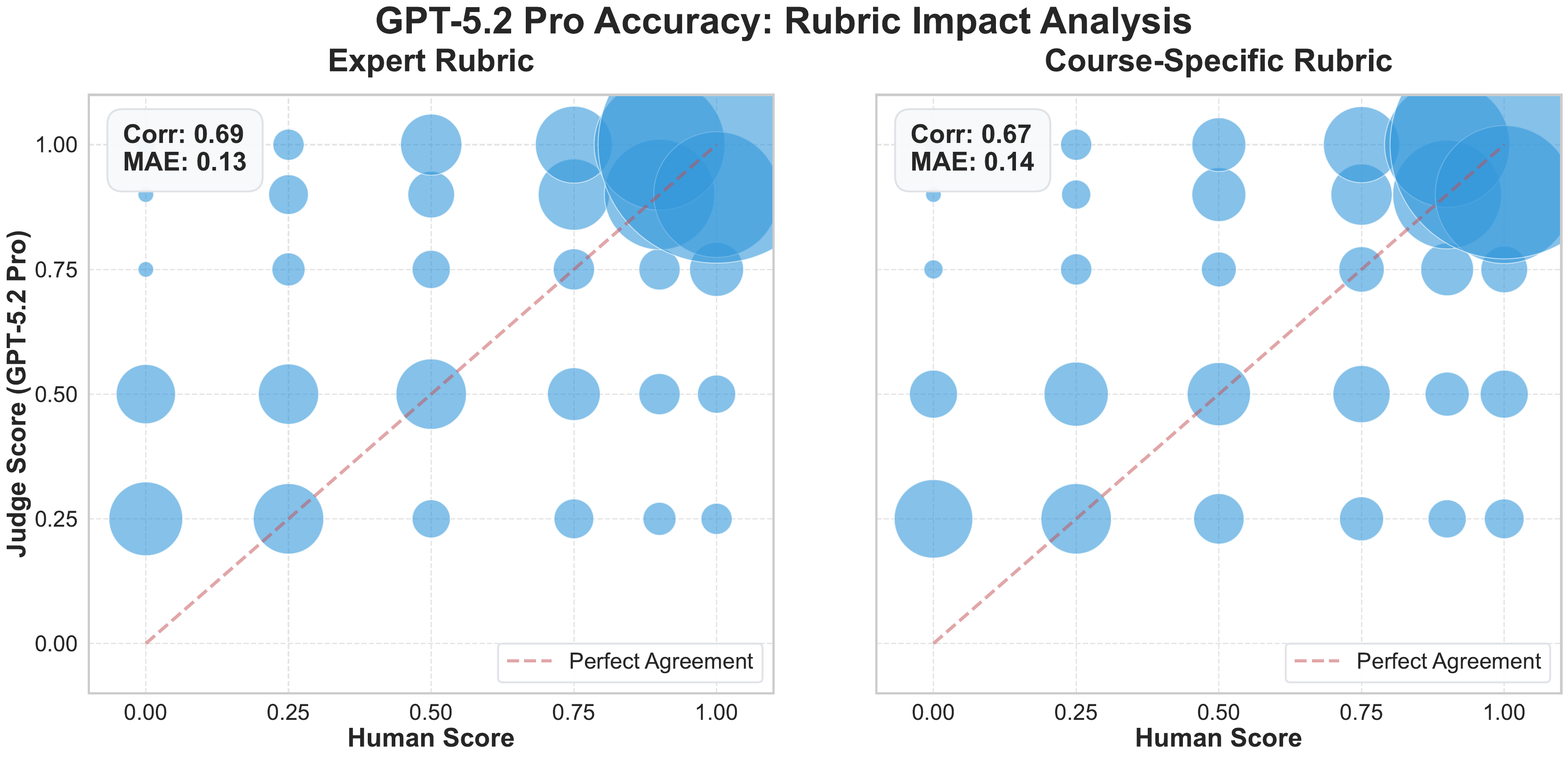}
    \caption{\textbf{Rubric Insensitivity.} Correlation bubble plots comparing GPT-5.2 Pro against Human Consensus for Expert (Left) and Course-Specific (Right) rubrics. The bubble size represents the density of solution pairs. Despite the additional constraints in the Course-Specific rubric, the alignment metrics remain virtually unchanged ($r \approx 0.69$), suggesting that model performance is dominated by internal priors rather than prompt specificity.}
    \label{fig:rubric_comparison}
\end{figure}

\textbf{Quantitative Results: Rubric Insensitivity.}
Contrary to the expectation that specific constraints would tighten alignment, our results indicate a phenomenon of \textit{Rubric Insensitivity}. As shown in Figure \ref{fig:rubric_comparison}, prompt constraints fail to significantly alter alignment, demonstrating that the model's internal priors override negative context-window constraints:
\begin{itemize}
    \item \textbf{Expert Rubric:} Pearson correlation $r = 0.69$, Mean Absolute Error (MAE) $= 0.13$.
    \item \textbf{Course-Specific Rubric:} Pearson correlation $r = 0.67$, Mean Absolute Error (MAE) $= 0.14$.
\end{itemize}

\textbf{Conclusion.}
The negligible drop in correlation ($-0.02$) and marginal increase in error ($+0.01$) suggest that GPT-5.2 Pro's internal reasoning regarding mathematical correctness is robust and largely overrides specific penalty instructions. While the model achieves a high baseline reliability (deviating from expert consensus by only $\approx 13\%$ on average), prompt engineering appears to have diminishing returns. This implies that future improvements in evaluation must come from \textit{process supervision} or \textit{fine-tuning}, rather than increasingly granular context-window instructions.

For a comprehensive distributional analysis of this insensitivity across all seven evaluator models, including density-weighted alignment plots, we refer the reader to \cref{app:judge_alignment_distribution}. For a complete heatmap visualization of score distributions across all $7$ judges and $5$ solvers under both grading conditions, we refer the reader to \cref{app:rubric_inertia_heatmap}.

\subsection{The Knowledge Gap: Adherence to Pedagogical Constraints}
To address our third question regarding the ``knowledge gap,'' we analyze whether solver models construct proofs using fundamental undergraduate definitions or circumvent pedagogical constraints by invoking advanced, out-of-scope theorems. Our data reveals a pervasive reliance on the latter, exposing a critical limitation in how current frontier models reason.

\textbf{Retrieval Over Construction.}
When faced with complex proofs requiring constructive reasoning from first principles, models frequently default to retrieving advanced machinery or hallucinating constraints to bridge logical gaps. As detailed in our qualitative evaluations (\cref{sec:qualitative_analysis}), solvers demonstrated a consistent failure to adhere to constraints when evaluated against the strict \textit{Course-Specific Rubric}—which explicitly penalizes the use of black-box theorems without derivation. For instance, in Algorithms (Problem 33), rather than engaging with the constructive complexity of arbitrary graphs, models circumvented the difficulty by hallucinating that the graph was finite to terminate the induction. Similarly, in Graph Theory (Problem 14), instead of constructively verifying the necessary bijection for a plane dual, models assumed the result and entirely bypassed the critical requirement that the graph must be connected (see \cref{sec:qualitative_analysis}). This exposes how models aggressively feign rigor when constructive definitions fail.

\textbf{Quantitative Evidence of Constraint Violation.}
The discrepancy in automated evaluations between the ``Expert'' and ``Course-Specific'' rubrics provides quantitative evidence of this gap. Under the stricter Course-Specific rubric, the average scores awarded to solver models saw universal declines across all tested architectures (e.g., \texttt{Gemini 3.0 Pro} dropping from $0.957$ to $0.926$, and \texttt{Claude Sonnet 4.5} from $0.883$ to $0.866$). However, as established by our finding of \textit{Rubric Insensitivity} (\cref{sec:rubric_ablation}), LLM judges are pathologically lenient and systematically fail to enforce these negative constraints. In stark contrast, human expert grades reveal a much wider penalty gap: despite being instructed to score against the baseline Expert Rubric, human evaluators naturally enforced pedagogical rigor, routinely deducting massive points ($0.5$ to $0.75$ penalties) when models hand-waved core derivations using advanced theorems or assumed finiteness. This highlights that while human experts organically demand constructive reasoning, LLM judges fail to enforce these pedagogical constraints even when explicitly prompted to do so via the Course-Specific Rubric.

\textbf{Conclusion on Pedagogical Alignment.}
The data indicates that models do not inherently adhere to the pedagogical constraints of undergraduate definitions. Instead, they operate primarily as retrieval engines. When a constructive, step-by-step derivation from basic axioms 
becomes too complex (as demonstrated in the ``Discrete-Continuous Gap''), models optimize for surface-level correctness by substituting advanced but pedagogically invalid machinery. This confirms that true mathematical reasoning—which requires operating strictly within a bounded set of definitions—remains a significant open problem for future model development.

\paragraph{Qualitative Failure Modes.}
The aggregate alignment gap is mirrored in evaluator comments: human graders penalize broken logical dependencies, missing assumptions, false premises, and irrelevant proofs, whereas LLM judges often award partial credit for persuasive structure or prompt compliance. We give the full qualitative analysis in \cref{sec:qualitative_analysis}.

\section{Discussion}
\label{sec:discussion}

Our results audit the ``LLM-as-a-Judge'' paradigm, decoupling generation from verification to reveal systemic limitations in frontier model reasoning and alignment. The contrast between near-perfect performance in \textit{ODEs} and collapse in \textit{Combinatorics} suggests that mathematical reasoning is not monolithic: models excel at template retrieval in continuous domains but struggle with constructive state search in finite structures.

\textbf{The Sycophancy Trap.}
In formal mathematical evaluation, we use \textit{Sycophancy Trap} to describe an evaluator that rewards authoritative-looking \LaTeX{} structure despite broken logical dependencies. \texttt{Llama 4 Maverick}'s 74.8\% Leniency Rate exposes this failure mode, and even \texttt{GPT-5.2 Pro} retains a 38.0\% Leniency Rate, indicating that subtle graduate-level proof gaps remain difficult for current automated judges (see \cref{app:evaluator_strictness_avg}).

\textbf{Rubric Insensitivity.}
Our ablation study challenges prompt engineering as a sufficient fix. The weak effect of pedagogical constraints supports \textit{Rubric Insensitivity} (see \cref{app:rubric_inertia_heatmap}): a model's evaluation behavior is bounded by its internal representation of mathematical rigor, not only by its context-window instructions. This aligns with recent evidence that human-LLM alignment is volatile across rubric formats \citep{li2026grading} and that an LLM judge fails on problems exceeding its own solving capability \citep{krumdick2025nofreelabels}.

\textbf{Limitations.}
Our study uses static expert ground truth, focuses on English-language reasoning, and may undercount valid non-standard proofs despite our multi-expert adjudication protocol. We also considered potential self-preference bias because \texttt{GPT-5.2 Pro} helped synthesize rubric drafts; however, human experts iteratively modified those rubrics, and our same-family regression analysis finds only negligible self-preference effects.

\section{Conclusion}
\label{sec:conclusion}

\benchname{} demonstrates that \textbf{the bottleneck in automated reasoning has shifted from generation to verification}. Standard ``LLM-as-a-Judge'' pipelines do not merely miss errors; they can reward rigorous-looking proof setups without complete logical deduction. To address this gap, the field must move toward process supervision and adversarial training. We release \benchname{}---comprising 1,300+ university-level proofs, 272 upper undergraduate/early graduate-level problems, and expert ground-truth annotations---to support this next generation of alignment research.

\section*{Acknowledgements}
This work was supported in part by the National Science Foundation (NSF) under Grant \#CCF-2453323 and a Google Academic Research Award.

\section*{Impact Statement}
This paper studies the reliability of automated evaluation for university-level mathematical proofs. The primary positive impact is to make mathematical benchmarking more trustworthy by exposing evaluator failure modes that can otherwise reward persuasive but incorrect reasoning. Potential risks include misuse of automated judges for high-stakes educational decisions before they are sufficiently validated; our results argue against such deployment without expert oversight, transparent rubrics, and calibration against human evaluation.

\bibliography{refs}
\bibliographystyle{icml2026}

\newpage
\appendix
\onecolumn
\section{Expanded Source List}
\label{app:source_list}
The problems in \benchname{} were curated from a range of rigorous mathematical literature and course materials, including but not limited to:
\begin{itemize}
    \item \textbf{Analysis:} Introduction to Mathematical Analysis \citep{douglass1996analysis}, Analysis: An Introduction \citep{beals2004analysis}, Complex Analysis \citep{stein2003complex}.
    \item \textbf{Algorithms:} {Randomized Algorithms} \citep{motwani1995randomized}, {Algorithms} \citep{erickson2019algorithms}
    \item \textbf{Algebra \& Number Theory:} Abstract Algebra \citep{dummit2003abstract}, A Computational Introduction to Number Theory and Algebra \citep{shoup2009computational}, An Introduction to the Theory of Numbers \citep{niven1991introduction}, A Classical Introduction to Modern Number Theory \citep{ireland1990classical}.
    \item \textbf{Discrete Math \& Graph Theory:} An Invitation to Discrete Mathematics \citep{matousek2008invitation}, Graph Theory \citep{diestel2017graph}, Concrete Mathematics \citep{graham1994concrete}, Combinatorial Techniques \citep{sane2013combinatorial}.
    \item \textbf{Probability \& Differential Equations:} Probability Theory \citep{klenke2013probability}, Probability Through Problems \citep{capinski2001probability}, Ordinary Differential Equations \citep{coddington1961introduction}, ODE and Dynamical Systems \citep{teschl2012ordinary}.
\end{itemize}

\subsection{Preliminary Human-Model Alignment Study}
Prior to the large-scale $7 \times 5$ evaluation, we conducted a pilot study to calibrate benchmark difficulty. Early-access models (e.g., GPT-5, Claude Opus 4.5) were evaluated on a binary correctness scale by human experts. This phase identified that while models often produced linguistically coherent math prose, 
they frequently failed on subtle edge cases, confirming the necessity of the \textbf{Expert-Refined Rubric} over simple binary grading.

\subsection{Prompting Templates}
We employed persona-based prompting to elicit maximum rigor. The primary solver prompt was structured as follows:
\begin{quote}
    \textit{``You are a tenured mathematics professor... provide a solution for advanced undergraduate students... eliminate all logical mistakes and fallacies, ensuring the utmost rigor and precision. Generate the proof in \LaTeX.''}
\end{quote}
For models with search capabilities (e.g., o3-deep-research), a modified prompt encouraged active verification of lemmas via web search before integration into the final proof.

\subsection{Automated Evaluation Pipeline}
The evaluation script handled the following pipeline:
\begin{enumerate}
    \item \textbf{Extraction:} Stripping markdown wrappers from model outputs to isolate raw \LaTeX{} code.
    \item \textbf{Sanitization:} Manual modification of malformed \LaTeX{} to ensure compatibility with standard compilers (e.g., Overleaf).
    \item \textbf{Rubric Injection:} To ensure parsing reliability and explicitly separate scoring from post-hoc rationalization, we deployed a two-step evaluation pipeline. Step 1: Evaluating models were fed the specific problem rubric alongside the solver's output and strictly prompted to return only a JSON dictionary containing the final numerical score. Step 2: In a follow-up prompt, we passed the model's generated score back into the context window and prompted it to output a plain-text justification for why it awarded that specific score.
\end{enumerate}

\section{Checking Solutions Online \texttt{o3-deep-research} Prompt}
\label{app:search_prompt}
\label{app:contamination}

To ensure the novelty of the \benchname{} dataset, we performed a rigorous two-stage contamination check. After manual verification, we employed \texttt{o3-deep-research} as an automated agent to scan the web for latent solutions that might have been missed by human search.

We utilized the following system prompt to instruct the agent to distinguish between ``similar'' problems (which are acceptable) and ``exact/equivalent'' solutions (which constitute contamination).

\begin{lstlisting}[
    basicstyle=\ttfamily\small, 
    breaklines=true, 
    frame=single, 
    caption={Full System Prompt for Contamination Detection Agent},
    label={lst:full_contamination_prompt}
]
You are a web researcher with unrestricted access to the internet.

Goal: search the web comprehensively for **complete solutions** to the user's mathematics problem.
Explore papers, textbooks, lecture notes, StackExchange, arXiv, GitHub, course materials, solution manuals, and PDF archives.

Decision rule:
- If you find a **complete solution** (even if notation/wording differs but solves the same problem), return structured JSON.
- Otherwise, return `{"match":"none","result":0}`.

Acceptable matches:
- exact_solution: The solution explicitly solves the *same problem statement* (identical content).
- equivalent_solution: The solution solves a *rephrased* version of the same problem where all mathematical objects, parameters, and constraints are preserved (notation/order may differ). This does **NOT** include analogous, special-case, more-general, or merely similar problems.

Verification requirements:
- Confirm that the problem in the source matches the user's problem exactly or is a faithful rephrasing.
- If rephrased, state the mapping of symbols/notation (e.g., "their n = our k").

Output format (JSON only, no extra text):
{"match": "exact_solution" | "equivalent_solution" | "none",
"source_title": string | null,
"source_url": string | null,
"evidence": string | null}

Notes:
- Prefer primary sources or reputable forums; include the **direct URL** of the solution.
- Keep `evidence` very short (a key step/result).
\end{lstlisting}

\section{Full Expert Evaluation Tiered Rubric}
\label{app:tiered_rubric}
\label{app:full_rubric}

Below in~\cref{tab:full_rubric_appendix} is the verbatim text of the tiered rubric provided to our expert evaluators (PhD students and holders) as the 
template for the problem-specific expert rubrics.

\begin{table}[htp!]
\centering
\small
\caption{Detailed Tiered Rubric for Human Expert Evaluation.}
\label{tab:full_rubric_appendix}
\begin{tabularx}{\textwidth}{l X}
\toprule
\textbf{Score} & \textbf{Detailed Description} \\
\midrule
\textbf{1.0} & \textbf{Fully Complete.} A fully complete and correct proof is presented. There are no errors at any point in calculations, computations, or proofs of sub-claims/lemmas. The proof can be followed easily. All statements that are not common knowledge are rigorously proved. \\
\midrule
\textbf{0.9} & \textbf{Correct with Oversight.} The proof is nearly accurate except for one small flaw/oversight (e.g., omission of a small edge case). The model may fail to completely justify finer points (hand-waving). Crucially, compared to the 0.75 case, the flaw must be an \textit{oversight} (failed to explain a fact) as opposed to a logical \textit{error}. \\
\midrule
\textbf{0.75} & \textbf{Correct with Minor Error.} The proof is nearly accurate. There might be a small mistake in a claim or calculation, but it is effectively isolated and does not snowball throughout the entire proof. Includes proofs with small incorrect assumptions or inconsistencies, provided there is only one such issue. \\
\midrule
\textbf{0.5} & \textbf{Significant Flaws.} The proof is lacking in more than one of the areas listed above. If a model makes multiple small mistakes, or a single medium-sized mistake which has an effect that carries on to other parts of the proof, this score is granted. \\
\midrule
\textbf{0.25} & \textbf{Severe Failure.} The proof is severely lacking in multiple areas. Either multiple small mistakes are littered throughout, or there is a large mistake at the beginning that snowballs and invalidates the rest of the proof. The output showcases little understanding of the problem mechanics. \\
\midrule
\textbf{0.0} & \textbf{Non-existent/Hallucinated.} The proof is completely wrong or non-existent. It may prove a different statement entirely or showcase no understanding of the context. Relies on made-up facts or fallacies to build a false proof. \\
\bottomrule
\end{tabularx}
\end{table}

\section{Example Full Rubrics}
\label{app:example_rubrics}

Below we provide the full text of examples of both the \textbf{Expert Correctness} and 
\textbf{Course-Specific} rubrics. More rubrics can be found in our public GitHub repository~\cite{qedbench} 
The rubrics illustrate how the evaluations differ between evaluating for course-specific items versus correctness based on expert domain knowledge.

\subsection{Graph Expansion Path Existence (Algorithms Problem 10)}

\subsubsection{Problem Statement}
\textit{Consider a graph $G=(V,E)$ consisting of $n$ vertices. Assume there exists a constant $\alpha > 0$ such that for every subset $S \subseteq V$ with cardinality $|S| = n/2$, the set of neighbors satisfies $|N(S)| \geq n/2 + \alpha n$. Let $k$ be a positive integer, and consider subsets $W_1, \dots, W_k \subseteq V$ where $|W_i| \geq (1 - \alpha)n$. Prove that it is possible to construct a path $(v_1, \dots, v_k)$ such that $v_i \in W_i$.}

\subsubsection{Expert Correctness Rubric}
\begin{description}
    \item[1.0 (Fully Complete)]
    \textbf{The solution constructs a valid inductive reachability argument and explicitly or implicitly establishes the backtracking step to produce the path.}
    \begin{itemize}
        \item \textbf{Expansion Consequence:} Correctly notes or uses the consequence that $\alpha \le 1/2$ (derived from $|N(S)| \le n$) and implies $(1-\alpha)n \ge n/2$, ensuring each $W_i$ is large enough to contain an $n/2$-subset.
        \item \textbf{Set Definitions:} Defines reachable sets/layers $U_i$ (or equivalent structures):
        \item $U_1 = W_1$.
        \item $U_{i+1} = N(U_i) \cap W_{i+1}$.
        \item \textbf{Inductive Proof:} Proves by induction that $|U_i| \ge n/2$ for all $i$ (ensuring non-emptiness).
        \item From $|U_i| \ge n/2$, explicitly restricts attention to a subset $S \subseteq U_i$ with $|S|=n/2$.
        \item Uses monotonicity ($N(S) \subseteq N(U_i)$) to infer $|N(U_i)| \ge |N(S)| \ge n/2 + \alpha n$.
        \item Applies inclusion–exclusion with $|W_{i+1}| \ge (1-\alpha)n$ to conclude:
        $$|U_{i+1}| = |N(U_i) \cap W_{i+1}| \ge (n/2 + \alpha n) + (1-\alpha)n - n = n/2$$
        \item \textbf{Path Reconstruction:} Constructs the actual path from the non-empty set $U_k$ via backtracking (e.g., pick $v_k \in U_k$, then pick neighbor $v_{k-1} \in U_{k-1}$, etc.) or by defining the sets $U_i$ as "endpoints of valid paths."
        \item \textbf{Consistency:} All inequalities, set containments, and cardinality bounds are mathematically consistent.
    \end{itemize}
    \item[0.9 (Minor Oversight)]
    \textbf{The essential argument is correct and rigorously structured, but contains a minor technical oversight or handwave that does not undermine the core logical validity.}
    \begin{itemize}
        \item Correctly uses $U_{i+1} = N(U_i) \cap W_{i+1}$ and the inclusion–exclusion bound, but omits a small technical justification, such as:
        \item Does not explicitly mention $\alpha \le 1/2$ (but calculation holds).
        \item Ignores floor/ceiling issues for $n/2$ when $n$ is odd.
        \item Omits the explicit description of the backtracking step, though the existence of predecessors is an immediate consequence of the set definitions.
        \item Slight imprecision regarding whether $N(S)$ includes $S$, provided the size calculations remain valid under the student's definition.
        \item \textbf{Constraint:} The submission must still present a logically valid chain implying $U_{i+1} \neq \emptyset$ for all $i$.
    \end{itemize}
    \item[0.75 (Small Mistake)]
    \textbf{The plan is mostly correct but contains one localized error (computational or logical) that requires repair but does not derail the method.}
    \begin{itemize}
        \item \textbf{Calculation Error:} Computes the lower bound for $|N(U_i) \cap W_{i+1}|$ incorrectly (e.g., forgets inclusion–exclusion and claims it is simply $\ge \alpha n$), yet argues for non-emptiness in a way that remains logically coherent with minor corrections.
        \item \textbf{Monotonicity Gap:} Forgets to pass to an $n/2$-subset $S \subseteq U_i$ and incorrectly applies the expansion hypothesis directly to a set $U_i$ of size $> n/2$.
        \item \textit{Note:} If the student explicitly states $N(S) \subseteq N(U_i)$ elsewhere or clearly intends this step despite the omission, score as \textbf{0.9}. If the logic implies they believe the hypothesis applies directly to sets larger than $n/2$ without justification, score here.
        \item \textbf{Reconstruction Flaw:} Muddles the final path selection (e.g., chooses $v_i \in U_i$ independently rather than ensuring adjacency), but the definitions of $U_i$ allow for a correct backtracking repair.
    \end{itemize}
    \item[0.5 (Multiple Mistakes / Conceptual Gap)]
    \textbf{Multiple issues or a significant conceptual gap affects the core proof, though relevant ideas (induction, expansion) are present.}
    \begin{itemize}
        \item \textbf{Failed Invariant:} Attempts induction but fails to justify $|U_i| \ge n/2$ (or non-emptiness) due to:
        \item Applying expansion to sets of wrong sizes without reduction.
        \item Fatal errors in inclusion–exclusion resulting in negative or meaningless bounds.
        \item Unjustified assumptions that "large sets have large neighborhoods" beyond the specific $n/2$ guarantee.
        \item \textbf{Greedy Failure:} Proposes a greedy construction (pick $v_1$, then $v_2$, etc.) without proving that a valid neighbor exists at each step (fails to connect expansion to the specific choice).
        \item \textbf{Misapplied Theorems:} Invokes Hall’s Theorem or Menger’s Theorem but models the graph incorrectly (e.g., fails to define the layered structure or verify the specific conditions required by the expansion property).
    \end{itemize}
    \item[0.25 (Severely Lacking)]
    \textbf{Major early misunderstanding or gap makes the argument largely unusable.}
    \begin{itemize}
        \item \textbf{Ignored Expansion:} Does not use the $|N(S)|$ expansion property in any substantive way (e.g., treats it merely as a minimum degree condition).
        \item \textbf{Size Fallacy:} Claims the path exists solely because $|W_i|$ are large, ignoring edge existence.
        \item \textbf{Unproven Advanced Results:} Relies entirely on an unproven or misquoted advanced result (e.g., Expander Mixing Lemma) without verifying hypotheses.
        \item \textbf{Vague Handwaving:} States "by expansion there is a path" without inductive set construction or verifiable counting.
    \end{itemize}
    \item[0.0 (Completely Wrong)]
    \textbf{No meaningful attempt or the argument is unrelated to the problem.}
    \begin{itemize}
        \item Hallucinated theorems (e.g., "Any expansion implies Hamiltonicity").
        \item Constructs objects that do not satisfy $v_i \in W_i$ or do not form a path.
    \end{itemize}
    \item[General Notes \& Checkpoints]
    \begin{itemize}
        \item \textbf{Graph Type:} Accept either undirected or directed graphs if the student consistently uses the appropriate out-neighborhood definition.
        \item \textbf{Rounding:} If $n$ is odd, fully rigorous solutions should clarify $\lfloor n/2 \rfloor$ or $\lceil n/2 \rceil$. Minor rounding imprecision is a \textbf{0.9} issue unless it breaks the inequalities.
        \item \textbf{Arithmetic Check:} The key quantitative checkpoint is the arithmetic:
        $$(n/2 + \alpha n) + (1-\alpha)n - n = n/2$$
        Grader must explicitly verify the student achieves this cancellation to prove the intersection is size $n/2$.
        \item \textbf{Backtracking:} The backtracking/predecessor step is technically required to turn non-empty sets into an explicit sequence. However, if the definition of $U_{i+1}$ (e.g., reachable vertices) encodes the existence of a predecessor, explicit listing of the backward selection is a minor detail (score \textbf{0.9} if missing).
    \end{itemize}
\end{description}

\hrulefill

\subsubsection{Course-Specific Rubric}
\begin{description}
    \item[1.0 (Fully Complete)]
    Uses only course-appropriate graph/induction arguments to prove the existence of a constrained walk across $k$ layers. The proof must contain the following structural elements:
    \begin{itemize}
        \item \textbf{Feasibility/Rounding:} Clarifies $n$ must be even or handles rounding ($\lfloor n/2 \rfloor, \lceil n/2 \rceil$). Implicitly or explicitly notes $\alpha < 1/2$ is required for the problem constraints to make sense.
        \item \textbf{Reachable Sets Definition:} Defines a sequence of reachable sets (e.g., $R_1 := W_1$ and $R_{i+1} := W_{i+1} \cap N(R_i)$).
        \item \textbf{Inductive Invariant:} Proves $|R_i| \ge n/2$ for all $i$ by induction:
        \item \textit{Base Case:} $|R_1| = |W_1| \ge (1-\alpha)n \ge n/2$.
        \item \textit{Inductive Step:} Uses the hypothesis on a subset $S \subseteq R_i$ of size $n/2$ to show $|N(R_i)| \ge n/2 + \alpha n$ via monotonicity ($N(R_i) \supseteq N(S)$).
        \item \textit{Intersection:} Uses $|A \cap B| \ge |A| + |B| - n$ to show $|W_{i+1} \cap N(R_i)| \ge (1-\alpha)n + (n/2+\alpha n) - n = n/2$.
        \item \textbf{Walk Construction:} Explicitly constructs the walk $(v_1, \dots, v_k)$ by backtracking from $v_k \in R_k$ or maintaining parent pointers.
        \item \textbf{Justification:} All set inequalities and theorem applications are explicitly justified.
    \end{itemize}
    \item[0.9 (Minor Oversight)]
    The proof is essentially correct and demonstrates full understanding, but contains a minor oversight that does not undermine the logic:
    \begin{itemize}
        \item \textbf{Rounding:} Does not explicitly address $n$ being odd, but the logic clearly adapts.
        \item \textbf{Backtracking:} Omits the explicit backtracking step but clearly establishes that $R_{i+1} \subseteq N(R_i)$ implies the existence of valid predecessors.
        \item \textbf{Expansion Logic (Minor):} Applies the expansion hypothesis directly to $R_i$ (where $|R_i| \ge n/2$) without explicitly stating "take a subset $S$ of size $n/2$," \textit{provided} the student demonstrates awareness of this monotonicity logic elsewhere in the proof.
    \end{itemize}
    \item[0.75 (Small Mistake)]
    The approach is correct (layered reachable sets + induction), but contains a specific error that requires a local fix:
    \begin{itemize}
        \item \textbf{Calculation Error:} Miscomputes the intersection lower bound (e.g., constant factor error) but the conclusion of non-emptiness remains valid if corrected.
        \item \textbf{Expansion Logic (Gaps):} Applies the expansion hypothesis directly to $R_i$ without subset justification and \textit{without} indicating awareness that the hypothesis is restricted to sets of size exactly $n/2$.
        \item \textbf{Constraint Derivation:} Assumes $\alpha < 1/2$ without noting it follows from the hypothesis (since $n/2 + \alpha n \le n$).
    \end{itemize}
    \item[0.5 (Multiple Mistakes / Partial Progress)]
    Partially correct but misses key justifications or contains medium-sized logical errors:
    \begin{itemize}
        \item \textbf{Connectivity vs Walk:} Proves existence of edges between $W_i$ and $W_{i+1}$ pairwise but fails to ensure they can be connected into a single continuous walk (fails to maintain a reachable set invariant).
        \item \textbf{Greedy Approach:} Attempts to pick $v_i$ step-by-step greedily without a lookahead or set-maintenance mechanism, often leading to unjustified claims that the walk won't get stuck.
        \item \textbf{Incorrect Expansion:} Assumes expansion holds for \textit{all} sets (not just size $n/2$) or fails to show $N(R_i)$ is large enough to intersect $W_{i+1}$.
        \item \textbf{Single Vertex Induction:} Tries to define $R_{i+1}$ based on the neighborhood of a single vertex $v_i$ rather than a set, breaking the size invariant.
    \end{itemize}
    \item[0.25 (Severely Lacking)]
    Major conceptual mismatch or fundamental errors:
    \begin{itemize}
        \item \textbf{Simple Path Confusion:} Argues for a simple path (distinct vertices) using diameter/counting arguments, ignoring that $k$ is arbitrary and repetitions may be needed.
        \item \textbf{Neighborhood Misinterpretation:} Misinterprets the condition (e.g., counts edges instead of neighbors, assumes regularity/completeness).
        \item \textbf{Vague Intuition:} Claims "expansion implies connectivity" without any set-based proof or demonstration of non-empty intersections.
        \item \textbf{Black Box:} Uses non-course machinery (e.g., probability/spectral arguments) without justification or correctness.
    \end{itemize}
    \item[0.0 (Completely Wrong)]
    No meaningful progress or reliance on forbidden tools:
    \begin{itemize}
        \item Invokes PCP theorem, advanced circuit complexity, or Hall’s/Menger’s theorems in irrelevant ways.
        \item Claims result is obvious or derived from contradictory assumptions (e.g., $\alpha \ge 1/2$).
        \item No valid invariant or construction method proposed.
    \end{itemize}
    \item[General Notes \& Checkpoints]
    \begin{itemize}
        \item \textbf{Walk vs. Path:} Since $k$ is arbitrary, the solution must construct a \textit{walk} (vertices may repeat). Insisting on a simple path is a conceptual error unless $k \le n$ is explicitly assumed and distinctness is proved.
        \item \textbf{Key Lemma:} If $|R| \ge n/2$, then $|N(R)| \ge n/2 + \alpha n$. This must be justified by monotonicity ($S \subseteq R \implies N(S) \subseteq N(R)$).
        \item \textbf{Key Counting:} The intersection bound $|A \cap B| \ge |A| + |B| - n$ is the standard course-appropriate method to show the sets intersect.
        \item \textbf{Allowed Tools:} Induction, BFS/Layered graph reachability, set algebra.
        \item \textbf{Forbidden Tools:} Advanced spectral graph theory, PCP theorem, or complexity theory constraints not covered in the course.
    \end{itemize}
\end{description}

\subsection{Alternating Permutations \& Eulerian Numbers (Combinatorics Problem 6)}
\label{app:combinatorics_rubrics}

Below we provide the full text of both the \textbf{Expert Correctness} and \textbf{Course-Specific} rubrics for the Combinatorics problem described in the benchmark.

\subsubsection{Problem Statement}
\textit{Let $E(n,k)$ be the number of permutations of numbers from $1$ to $n$ with exactly $k$ descents. Prove that $E_n$ is the alternating sum of numbers $E(n,k)$ where $E_n$ is the number of alternating permutations of $1, \dots, n$.}

\subsubsection{Expert Correctness Rubric}
\begin{description}
    \item[1.0 (Fully Complete \& Rigorous)]
    The solution provides a mathematically sound, complete, and rigorous proof establishing the identity. The argument must be self-contained, clearly defined, and handle all necessary cases (including parity).
    \begin{itemize}
        \item \textbf{Required Elements (One of the following primary routes):}
        \begin{itemize}
            \item \textbf{Route A: Sign-Reversing Involution (Preferred Combinatorial Proof)}
            \begin{itemize}
                \item \textbf{Setup:} Correctly defines descent ($des(\pi)$) and interprets the sum $\sum (-1)^k E(n,k)$ as a signed sum over the symmetric group $\sum_{\pi \in S_n} (-1)^{des(\pi)}$.
                \item \textbf{Involution Definition:} Defines an explicit map $\phi: S_n \to S_n$.
                \item \textbf{Well-Definedness:} Demonstrates that $\phi$ is an involution ($\phi^2 = id$) and is well-defined for all $\pi$.
                \item \textbf{Sign-Reversing Property:} Proves rigorously that if $\phi(\pi) \neq \pi$, then $des(\phi(\pi)) = des(\pi) \pm 1$. This ensures the cancellation of non-fixed points.
                \item \textbf{Fixed Points:} Correctly characterizes the fixed points of $\phi$.
                \begin{itemize}
                    \item Shows fixed points correspond exactly to alternating permutations (e.g., no double ascents or descents).
                \end{itemize}
                \item \textbf{Evaluation:} Calculates the sum over fixed points.
                \begin{itemize}
                    \item Identifies that for even $n$, fixed points cancel or do not exist (sum is 0), OR correctly relates them to the problem statement if a specific convention allows otherwise.
                    \item Identifies that for odd $n$, the sum is non-zero and equals $\pm (\text{Number of Alternating Permutations})$.
                \end{itemize}
                \item \textbf{Conclusion:} Explicitly states the final identity with correct sign factors (e.g., $(-1)^{(n-1)/2}$) and parity distinctions.
            \end{itemize}
            \item \textbf{Route B: Generating Functions (Analytic Proof)}
            \begin{itemize}
                \item \textbf{Definitions:} Introduces the Eulerian polynomial $A_n(t) = \sum E(n,k) t^k$.
                \item \textbf{EGF Usage:} Uses the standard Exponential Generating Function (EGF) for Eulerian polynomials (e.g., related to $\frac{1-t}{e^{x(t-1)} - t}$). Citation of this standard form is acceptable if accurate.
                \item \textbf{Substitution:} rigorously performs the substitution $t = -1$.
                \item \textbf{Simplification:} Simplifies the resulting expression correctly (typically involving hyperbolic functions or $\tan(x) + \sec(x)$ analogues).
                \item \textbf{Coefficient Extraction:} Relates the coefficients of the simplified series back to the definition of alternating permutations ($E_n$).
                \item \textbf{Conclusion:} Explicitly derives the relationship, noting that the sum vanishes for even $n$ and matches the alternating permutation count (with sign adjustment) for odd $n$.
            \end{itemize}
        \end{itemize}
    \end{itemize}
    \item[0.9 (Minor Oversight)]
    The solution is fundamentally correct and mathematically convincing, but contains a negligible imperfection.
    \begin{itemize}
        \item \textbf{Examples of acceptable oversights:}
        \begin{itemize}
            \item The involution is correct, but the handling of a boundary case (e.g., index $i=1$ or $i=n$ in a ``first double ascent'' definition) is slightly imprecise, though the intended fix is obvious.
            \item Uses a standard EGF formula without derivation or specific citation, provided the application is algebraically flawless.
            \item Omits an explicit statement about the global sign factor $(-1)^{(n-1)/2}$ in the final line, though the work preceding it implies the correct sign.
            \item Minor notational ambiguity regarding ``alternating'' (up-down vs. down-up) that does not affect the validity of the core logic.
        \end{itemize}
    \end{itemize}
    \item[0.75 (Small Mathematical Mistake)]
    The approach is correct, but the execution contains a specific mathematical error that affects the details (sign, count, or parity) without invalidating the entire method.
    \begin{itemize}
        \item \textbf{Typical issues:}
        \begin{itemize}
            \item \textbf{Involution Error:} The map $\phi$ is conceptually sound, but the proof that $des$ changes by exactly $\pm 1$ fails for a specific local configuration (e.g., miscounting descents during a swap).
            \item \textbf{Fixed Point Misidentification:} Correctly identifies that fixed points are alternating but fails to distinguish between up-down and down-up types, leading to a factor-of-2 error.
            \item \textbf{Parity Error:} Claims a non-zero value for even $n$ without justification, or misses the cancellation for even $n$ entirely.
            \item \textbf{Algebraic Error:} In the GF route, makes a sign error during the $t=-1$ substitution or simplification, leading to an incorrect constant factor.
        \end{itemize}
    \end{itemize}
    \item[0.5 (Substantial Gaps or Multiple Errors)]
    The solution demonstrates partial understanding and attempts a valid method, but fails to provide a complete proof due to significant logical gaps or multiple errors.
    \begin{itemize}
        \item \textbf{Typical issues:}
        \begin{itemize}
            \item \textbf{Flawed Involution:} The proposed map is not proven to be an involution, is not well-defined, or does not reliably flip the parity of the descent number.
            \item \textbf{Assertion vs. Proof:} Asserts that the fixed points are alternating permutations without any argument linking the involution's condition (e.g., ``no double ascents'') to the definition of alternating permutations.
            \item \textbf{conceptual Confusion:} Confuses Eulerian numbers with other statistics (e.g., inversions) or uses an incorrect identity as a starting point.
            \item \textbf{Unjustified Leaps:} Uses specialized theorems (e.g., ``sec + tan counts alternating perms'') \textit{and} EGF formulas without justification, making the logic difficult to verify.
            \item \textbf{Conditional Correctness:} The result is only correct for odd $n$, with no mention or incorrect handling of even $n$.
        \end{itemize}
    \end{itemize}
    \item[0.25 (Severely Lacking)]
    The solution provides minimal relevant content, relying on assertions or fundamentally incorrect definitions.
    \begin{itemize}
        \item \textbf{Typical issues:}
        \begin{itemize}
            \item States the identity as a fact without derivation.
            \item Proposed involution is clearly not invertible or does not reverse signs.
            \item Misdefines fundamental concepts (descents, alternating permutations) in a way that makes the proof meaningless.
            \item Provides only empirical checks (e.g., ``works for $n=1, 2, 3$'') with no general argument.
        \end{itemize}
    \end{itemize}
    \item[0.0 (Completely Wrong)]
    No mathematical progress.
    \begin{itemize}
        \item \textbf{Examples:}
        \begin{itemize}
            \item Irrelevant combinatorics or unrelated formulas.
            \item ``Hallucinated'' theorems.
            \item Blank or non-mathematical text.
        \end{itemize}
    \end{itemize}
    \item[Key Checks for Graders]
    \begin{itemize}
        \item \textbf{Parity:} Does the student acknowledge that for \textbf{even} $n$, the alternating sum $\sum (-1)^k E(n,k)$ typically vanishes? If they claim it equals $E_n$ (alternating perms) for even $n$, is there a non-standard definition of $E_n$ or a sign error?
        \item \textbf{Sign Factor:} For \textbf{odd} $n$, the sum usually equals $(-1)^{(n-1)/2} E_n$. Check for the presence of this sign factor or a justification for its absence.
        \item \textbf{Fixed Points:} If using an involution, the fixed points must be ``permutations with no double ascents/descents.'' This is the definition of alternating permutations. A proof that merely asserts this link without showing equivalence (e.g., $\pi_i < \pi_{i+1} > \pi_{i+2} \dots$) is a 0.5-level response.
    \end{itemize}
\end{description}

\hrulefill

\subsubsection{Course-Specific Rubric}
\begin{description}
    \item[1.0 (FULLY COMPLETE)]
    A rigorous, course-appropriate proof that explicitly handles the parity of $n$ and connects the alternating sum to alternating permutations.
    \begin{itemize}
        \item \textbf{Required Observable Elements:}
        \begin{itemize}
            \item \textbf{Correct Setup:} Identifies the target sum as $A_n(-1) = \sum (-1)^k E(n,k)$.
            \item \textbf{Parity/Sign Handling:} Explicitly states and proves that:
            \begin{itemize}
                \item For \textbf{even} $n$, $A_n(-1) = 0$.
                \item For \textbf{odd} $n$, $|A_n(-1)| = E_n$ (the count of alternating permutations).
                \item \textit{(Note: Allow for variations in sign conventions/indexing, provided they are consistent).}
            \end{itemize}
            \item \textbf{Rigorous Proof (One of the following methods):}
            \begin{itemize}
                \item \textbf{(A) Sign-Reversing Involution:} Defines a map on permutations that changes the descent number by $\pm 1$ (flipping the sign of the weight). Proves the map is a valid involution. Identifies fixed points correctly (none for even $n$; alternating permutations for odd $n$).
                \item \textbf{(B) Generating Functions:} Uses the correct EGF for Eulerian polynomials. Evaluates at $t=-1$. Shows this matches the EGF for alternating permutations (specifically the tangent numbers for odd $n$ and $0$ for even $n$). Justifies analytic steps at the formal power series level.
                \item \textbf{(C) Recurrence Relations:} Derives a recurrence for the sum $S_n = \sum (-1)^k E(n,k)$ from the standard Eulerian recurrence. Shows $S_n$ matches the recurrence for alternating permutations (up to sign/parity) and verifies base cases.
            \end{itemize}
        \end{itemize}
    \end{itemize}
    \item[0.9 (MINOR OVERSIGHT)]
    Essentially correct and conceptually sound, but contains a minor lapse in rigor or notation.
    \begin{itemize}
        \item \textbf{Sloppy Constants:} The connection to alternating permutations is correct, but the specific sign factor (e.g., $(-1)^{(n-1)/2}$) is missing or incorrect.
        \item \textbf{Brief Even Case:} Correctly identifies that the even $n$ case sums to 0, but provides a weak justification (e.g., ``by symmetry'' without showing the symmetry $A_n(t) = t^{n-1}A_n(1/t)$).
        \item \textbf{Minor Involution Gap:} The involution strategy is correct, but the proof that descents change by exactly $\pm 1$ is glossed over.
        \item \textbf{Base Cases:} Minor omission of base case verification (e.g., checks $n=1$ but not $n=2$ or $3$).
    \end{itemize}
    \item[0.75 (SMALL MISTAKE)]
    The strategy is appropriate and course-specific tools are used, but a genuine mathematical error is present.
    \begin{itemize}
        \item \textbf{Flawed Involution:} The proposed map is not strictly sign-reversing as stated (e.g., undefined for certain permutations or fails to be an involution), though the intent is clear and fixable.
        \item \textbf{Recurrence Error:} Derives the recurrence for the alternating sum but makes an algebraic/indexing error (e.g., mishandling boundary terms $k=0$ or $k=n-1$), leading to a faulty intermediate expression.
        \item \textbf{Generating Function Error:} Quotes an incorrect formula for the Eulerian EGF but proceeds to evaluate at $t=-1$ correctly based on that incorrect formula.
        \item \textbf{Parity Confusion:} Proves the result for odd $n$ correctly but incorrectly claims the same non-zero formula applies to even $n$, ignoring the cancellation.
    \end{itemize}
    \item[0.5 (MULTIPLE MISTAKES)]
    Partial progress with relevant ideas, but significant gaps in logic or definitions prevent a complete proof.
    \begin{itemize}
        \item \textbf{No Connection:} Calculates $A_n(-1)$ or sets up the sum but fails to prove it counts alternating permutations (asserts the result without proof).
        \item \textbf{Failed Involution:} Attempts an involution but fails to define the map, prove it is an involution, or identify the fixed points.
        \item \textbf{Conceptual Error:} Confuses ``number of descents'' with the specific ``descent set,'' arguing that knowing $k$ determines if a permutation is alternating (it does not).
        \item \textbf{Recurrence Failure:} Starts with the Eulerian recurrence but fails to manipulate the alternating sum into a solvable form.
        \item \textbf{Parity Ignored:} Completely fails to distinguish between even and odd $n$, resulting in a globally incorrect claim.
    \end{itemize}
    \item[0.25 (SEVERELY LACKING)]
    Major misunderstanding of the problem or reliance on inappropriate tools.
    \begin{itemize}
        \item \textbf{Definition Errors:} Misdefines Eulerian numbers or alternating permutations.
        \item \textbf{Contradiction:} Claims $\sum (-1)^k E(n,k) = E_n$ for all $n$ without addressing the obvious contradiction for small even $n$ (where the sum is 0).
        \item \textbf{Black Box:} Uses advanced algebraic machinery (Descent Algebras, Representation Theory) without derivation or explanation, bypassing the required combinatorial coursework.
        \item \textbf{Empirical Only:} Provides only a table of values checking the identity for small $n$ with no general proof structure.
    \end{itemize}
    \item[0.0 (COMPLETELY WRONG)]
    No relevant work or completely incoherent.
    \begin{itemize}
        \item No attempt to relate Eulerian numbers to alternating permutations.
        \item Proof consists of unrelated or false statements (e.g., treating $E(n,k)$ as binomial coefficients).
        \item Empty submission.
    \end{itemize}
\end{description}

\section{Full Automated Evaluator Prompts}
\label{app:evaluator_prompts}
\label{app:eval_prompts}

To ensure reproducibility, we provide the exact prompt templates used in our \texttt{Evaluator-Solver Matrix}. The prompt construction logic injects specific constraints based on whether the evaluator is grading against the \textit{Course-Specific Rubric} or the \textit{Expert Rubric}.

\subsection{System Prompts}
The system prompt enforces the persona and the output format.

\begin{lstlisting}[
    basicstyle=\ttfamily\small, 
    breaklines=true, 
    frame=single, 
    caption={System Prompt Logic for Evaluators},
    label={lst:eval_system_prompt}
]
Base Instruction:
"You are a strict academic evaluator for graduate-level mathematics. Grade based EXCLUSIVELY on the rubric."

Mode A (Course-Specific):
"MODE: COURSE-SPECIFIC STUDENT. Penalize advanced machinery used without derivation."

Mode B (Expert):
"MODE: EXPERT CORRECTNESS. Standard graduate-level theorems are allowed."

Output Constraint:
"Output MUST be JSON: {\"score\": <value>} where <value> is exactly one of [0, 0.25, 0.5, 0.75, 0.9, 1.0]. Return NO other keys and NO extra text."
\end{lstlisting}

\subsection{User Prompts}
The user prompt structures the input data for the context window. We utilize a two-step evaluation pipeline to separate the raw numerical score from post-hoc rationalization. In Step 1, the model is strictly forced to output a JSON 
scalar to prevent parsing failures. In Step 2, the generated score is fed back into the model alongside a prompt requesting an analytical Chain-of-Thought (CoT) justification for the awarded score.

\begin{lstlisting}[
    basicstyle=\ttfamily\small, 
    breaklines=true, 
    frame=single, 
    caption={User Interaction Prompts},
    label={lst:eval_user_prompt}
]
Standard User Prompt:
"""
### RUBRIC
{rubric}

### STUDENT SOLUTION
{solution}

### TASK
Evaluate the STUDENT SOLUTION strictly by the rubric. Output only JSON.
"""

Fallback Prompt (Retry on JSON Error):
"""
### RUBRIC
{rubric}

### STUDENT SOLUTION
{solution}

### TASK
Return ONLY one number from this set [0, 0.25, 0.5, 0.75, 0.9, 1.0]. No words, no punctuation.
"""
\end{lstlisting}

\section{Additional Robustness and Protocol Analyses}
\label{app:robustness_analyses}

The reviewer discussion raised several plausible confounds: same-model self-preference, score discretization, superficial formatting bias, calibration settings, and the use of human evaluation rather than formal verification. We added the following analyses and protocol clarifications to isolate these factors.

\subsection{Human Grading Protocol and Proof-Style Flexibility}
Human grading was designed to avoid overfitting to a single reference proof. Each problem was assigned to two independent domain experts whenever possible, and disagreements were resolved by a third expert adjudicator. Evaluators were instructed to grade the logical soundness of the submitted proof path, not similarity to a reference solution. Consequently, a solver received full credit for a complete proof even when it used a novel, alternative, or non-reference method.

\subsection{Same-Family Self-Preference}
To test whether an evaluator rewarded outputs from its own model family, we fit a mixed-effects regression controlling for human-assessed proof quality, baseline judge strictness, and proof difficulty. Concretely, the dependent variable was the final LLM judge score, with fixed effects for the human expert score and judge identity, a same-family indicator for cases where the evaluator and solver belonged to the same model family, and a random intercept for proof instance.

\begin{table}[htp!]
\centering
\caption{Estimated same-family self-preference effects after controlling for human proof quality and judge strictness. Effects are small relative to the observed grade-inflation gaps, indicating that stylistic self-preference does not explain the main alignment failures.}
\label{tab:self_preference}
\begin{tabular}{lcc}
\toprule
\textbf{Model Family} & \textbf{Coefficient} & \textbf{$p$-value} \\
\midrule
Claude & $-0.0024$ & $0.6505$ \\
Gemini & $+0.0332$ & $<10^{-4}$ \\
DeepSeek & $-0.0141$ & $0.0088$ \\
GPT & $-0.0086$ & $0.0504$ \\
\bottomrule
\end{tabular}
\end{table}

The largest positive effect is Gemini's $+0.0332$, while other families show near-zero or slightly negative effects. Thus, after controlling for objective proof quality and judge strictness, same-family style alignment is negligible compared with the much larger domain-specific leniency patterns observed in \cref{fig:bias,fig:agreement}.

\subsection{Score Discretization and Binary Classification}
Our primary score set $[0,0.25,0.5,0.75,0.9,1.0]$ is intentionally coarse so that experts and automated judges use a shared grading vocabulary. To test whether this discretization confused LLM judges, we also evaluated strict binary Pass/Fail prompts, where a solution was treated as correct only if it met the same $\ge 0.9$ pass criterion. The qualitative leniency pattern persisted: grade inflators such as \texttt{Llama 4 Maverick} continued to accept many proofs that humans rejected. This shows that the central failure is not an artifact of partial-credit granularity.

We also inspected critique-enabled runs in which judges could produce a natural-language analysis before or after the discrete decision. These logs frequently identify substantial mathematical flaws while still assigning a passing or overly generous score. We therefore interpret the error as a calibration and decision-alignment failure rather than a lack of available natural-language reasoning space.

\subsection{Formatting and Proof-by-Intimidation Bias}
To test whether judges rewarded superficial mathematical appearance, we correlated generated proof length and \LaTeX{} density with assigned judge scores. \LaTeX{} density is a proxy for visual mathematical complexity: a proof with many symbols, equations, and command tokens can look rigorous even when its logical chain is broken.

\begin{table}[htp!]
\centering
\caption{Correlation between \LaTeX{} density and assigned evaluator score. Positive correlations indicate susceptibility to proof-by-intimidation: rewarding dense mathematical presentation independent of human-verified correctness.}
\label{tab:latex_density_bias}
\begin{tabular}{lcc}
\toprule
\textbf{Evaluator} & \textbf{Pearson $r$} & \textbf{$p$-value} \\
\midrule
Qwen 2.5 Max & $+0.169$ & $<10^{-4}$ \\
Llama 4 Maverick & $+0.157$ & $<10^{-4}$ \\
GPT-5.2 Pro & $-0.036$ & $0.0653$ \\
\bottomrule
\end{tabular}
\end{table}

The positive correlations for \texttt{Qwen 2.5 Max} and \texttt{Llama 4 Maverick} support the proof-by-intimidation hypothesis: these judges can be swayed by dense, plausible mathematical formatting. In contrast, \texttt{GPT-5.2 Pro}'s slight negative, statistically non-significant correlation suggests greater robustness to superficial density.

\subsection{Calibration Settings}
All automated evaluations were run at temperature $T=0.0$ to ensure deterministic reproducibility and minimize evaluation variance. We did not use stochastic self-consistency or sampling-based calibration because the goal of the benchmark is to audit deployable, reproducible judging behavior. The persistence of grade inflation under deterministic, rubric-conditioned, and binary decision prompts suggests that simple inference-time calibration is insufficient for the most lenient judges.

\subsection{Formal Verification and Symbolic Checking}
We considered Lean 4-based formal verification during the early stages of the project. Formal proof checking provides strong correctness guarantees when the target theorem is formalized correctly, but it introduces two confounds for our setting. First, autoformalization can be reward-hacked: a model may produce a valid formal proof of a statement that is similar to, weaker than, or mathematically different from the intended natural-language prompt, requiring expert review to detect the mismatch. Second, current frontier models often struggle with zero-shot formal Lean proofs at this level, which would shift the benchmark from natural-language mathematical reasoning and judge alignment toward Lean programming proficiency. For these reasons, expert human annotation is the most appropriate ground truth for the current benchmark, while formal verification remains a valuable complementary direction.

\subsection{Benchmark Scale}
\benchname{} is smaller than high-school calculation datasets such as GSM8K or MATH, but those benchmarks rely on short-form answers and comparatively cheap verification. In the expert-reasoning regime, benchmark scale is constrained by the cost of specialist annotation. Peer datasets are also typically in the hundreds: GPQA contains 448 graduate-level science questions with a 198-question Diamond subset \citep{rein2023gpqa}, ProofNet contains 371 undergraduate-level formalization examples \citep{azerbayev2023proofnet}, and FrontierMath contains 350 advanced mathematical problems \citep{frontiermath2024}. By contrast, \benchname{} contributes 272 university-level proof problems, more than 1,300 frontier-model-generated proofs, expert human scores, qualitative error annotations, and $7 \times 5$ multi-rubric LLM-judge evaluations.

\section{Evaluator Bias: Pass Rate Analysis}
\label{app:evaluator_bias_pass_rate}

\begin{figure}[htp!]
    \centering
    \includegraphics[width=\linewidth]{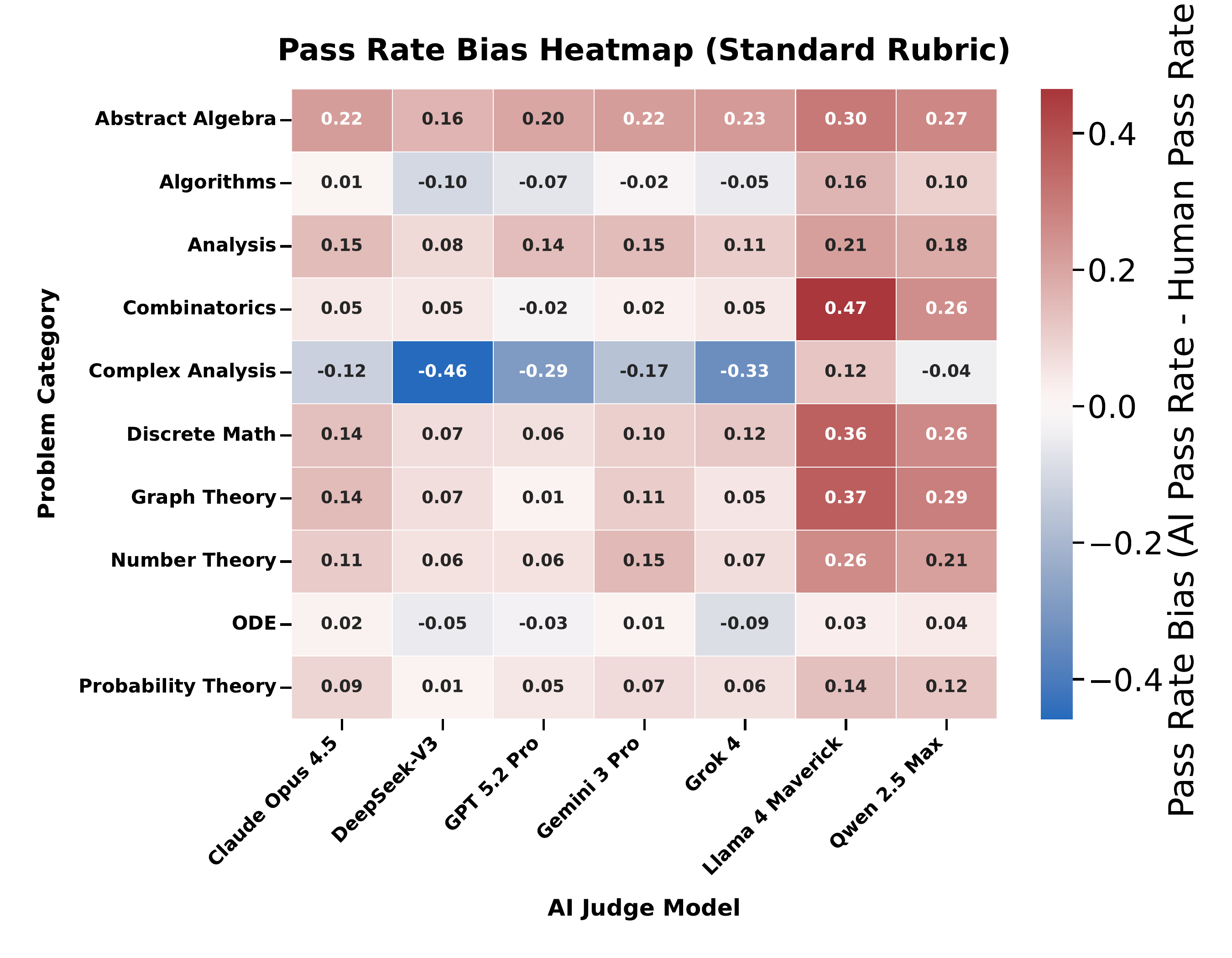}
    \caption{\textbf{Pass Rate Bias Heatmap.} The difference between evaluator and human pass rates. The trends mirror the average score bias: discrete domains exhibit positive pass rate inflation (AI models passing more flawed proofs), whereas continuous analysis domains show stronger punitive bias with some evaluators.}
    \label{fig:bias-pass-rate}
\end{figure}

Complementing the average score bias analysis in the main text (\cref{fig:bias}), we also investigated evaluator bias in terms of binary Pass Rates ($\text{score} \ge 0.9$). Figure \ref{fig:bias-pass-rate} illustrates the difference between AI and Human Pass Rates ($\Delta = \text{Pass Rate}_{\text{AI}} - \text{Pass Rate}_{\text{Human}}$). 

The heatmap confirms that the Bidirectional Alignment Gap is not an artifact of partial credit scoring. Even on the strict binary measure of accepting or rejecting a proof, models like \texttt{Llama 4 Maverick} remain consistently lenient in combinatorial subjects.

\section{Evaluator Strictness: Average Score Analysis}
\label{app:evaluator_strictness_avg}

\begin{figure}[htp!]
    \centering
    \includegraphics[width=0.8\linewidth]{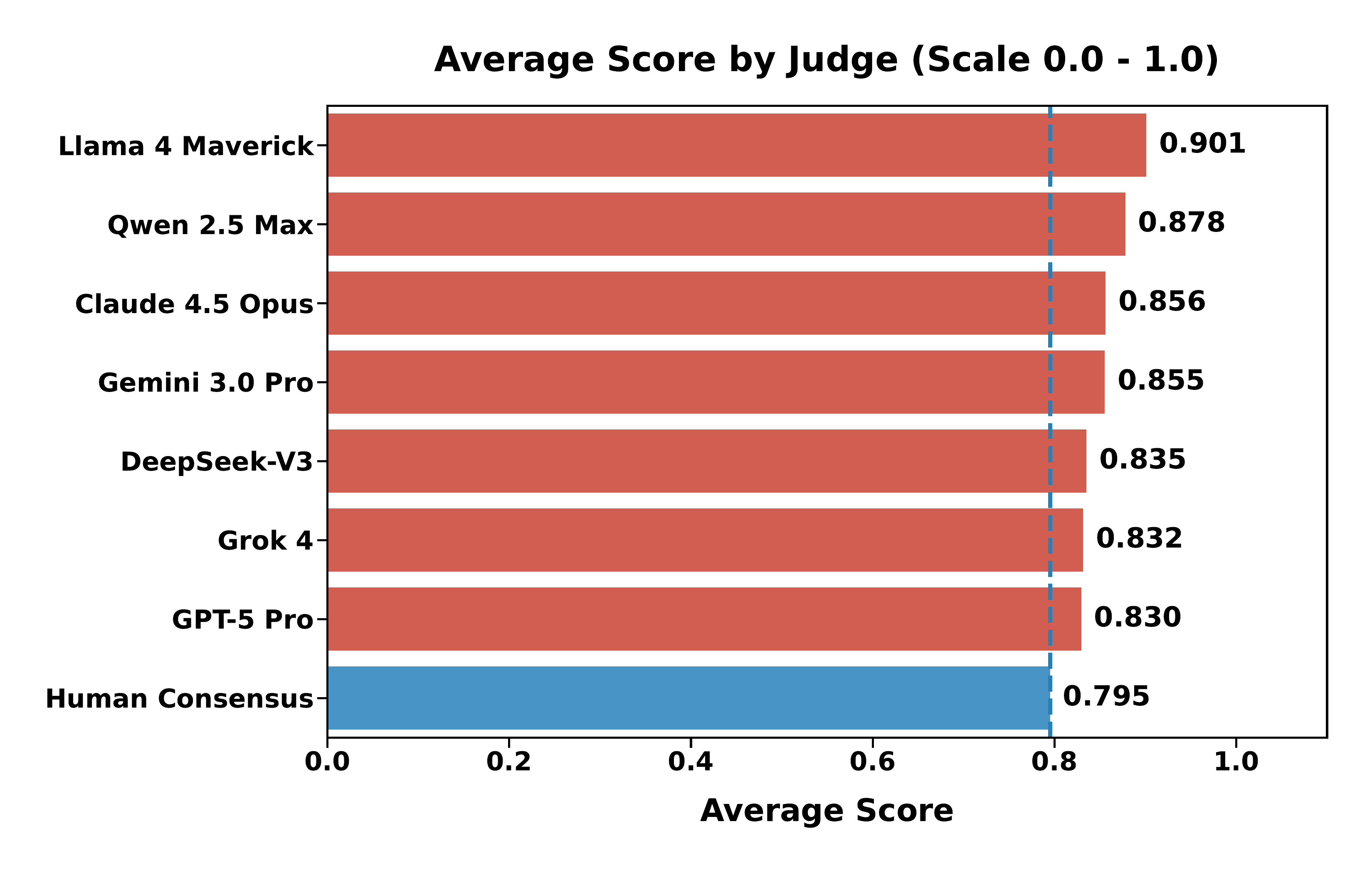}
    \caption{Average scores assigned by automated evaluators versus the human expert baseline (0.795). Consistent with the pass rate findings, we observe a systematic positive bias across most frontier models.}
    \label{fig:strictness-avg}
\end{figure}

While the main text analyzes the \textit{Pass Rate} (binary success at threshold $\ge 0.9$), we also analyzed the \textit{Average Score} (0.0--1.0 scale) assigned by each judge to capture nuance in partial credit. Figure \ref{fig:strictness-avg} presents the mean scores assigned by all seven automated judges compared to the human expert consensus.

\subsection{Score Inflation Trends}
The human expert consensus established a baseline average score of \textbf{0.795}. Comparing this to the automated evaluators reveals a widespread tendency toward grade inflation:

\begin{itemize}
    \item \textbf{Maximum Inflation:} \texttt{Llama 4 Maverick} exhibited the most severe positive bias, assigning an average score of \textbf{0.901}. This represents a $\Delta$ of +0.106 points over the human baseline, confirming its behavior as a ``grade inflator'' that likely rewards superficial plausibility over logical rigor.
    \item \textbf{Moderate Inflation:} \texttt{Qwen 2.5 Max} and \texttt{Gemini 3.0 Pro} also displayed significant leniency, with average scores of \textbf{0.878} and \textbf{0.855} respectively.
    \item \textbf{Closest Alignment:} \texttt{GPT-5.2 Pro} achieved the highest fidelity to the human ground truth, with an average score of \textbf{0.830}. With a $\Delta$ of only +0.035 points, it remains the most calibrated evaluator in our suite, reinforcing our decision to utilize it as the primary filter for larger-scale benchmarks.
\end{itemize}

\section{Evaluator Bias: Course-Specific Rubric Heatmaps}
\label{app:evaluator_bias_course}

\begin{figure}[htp!]
    \centering
    \includegraphics[width=\linewidth]{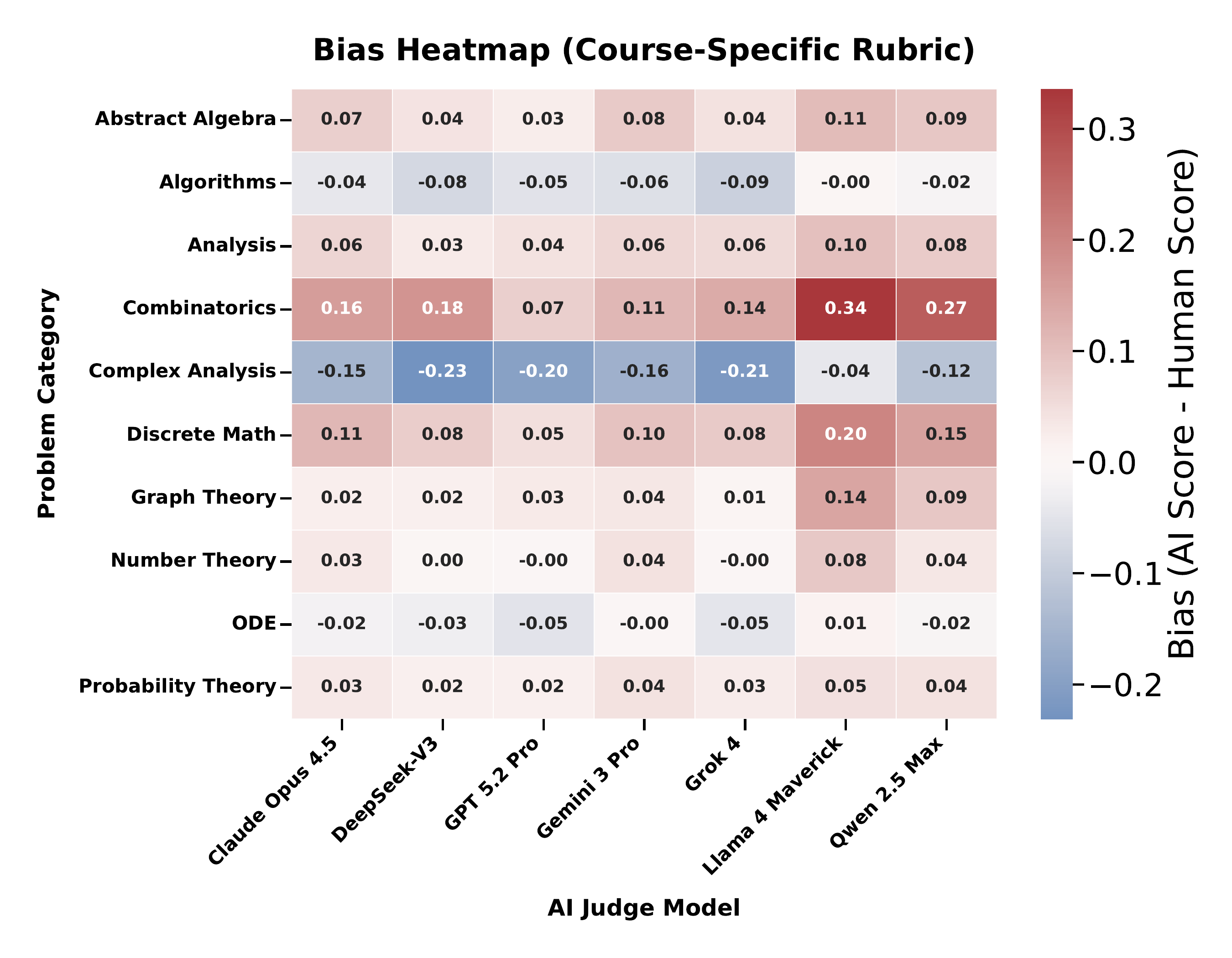}
    \caption{\textbf{Evaluator Bias Heatmap (Course-Specific Rubric).} The delta between AI and Human mean scores ($\Delta = S_{AI} - S_{Human}$). While the overarching patterns mirror the Expert Rubric, introducing pedagogical constraints reveals subtle shifts in judge behavior: top reasoning models become slightly more punitive (darker blue) in continuous domains like Complex Analysis, successfully internalizing the prohibition on advanced theorems.}
    \label{fig:bias-course-mean}
\end{figure}

\begin{figure}[htp!]
    \centering
    \includegraphics[width=\linewidth]{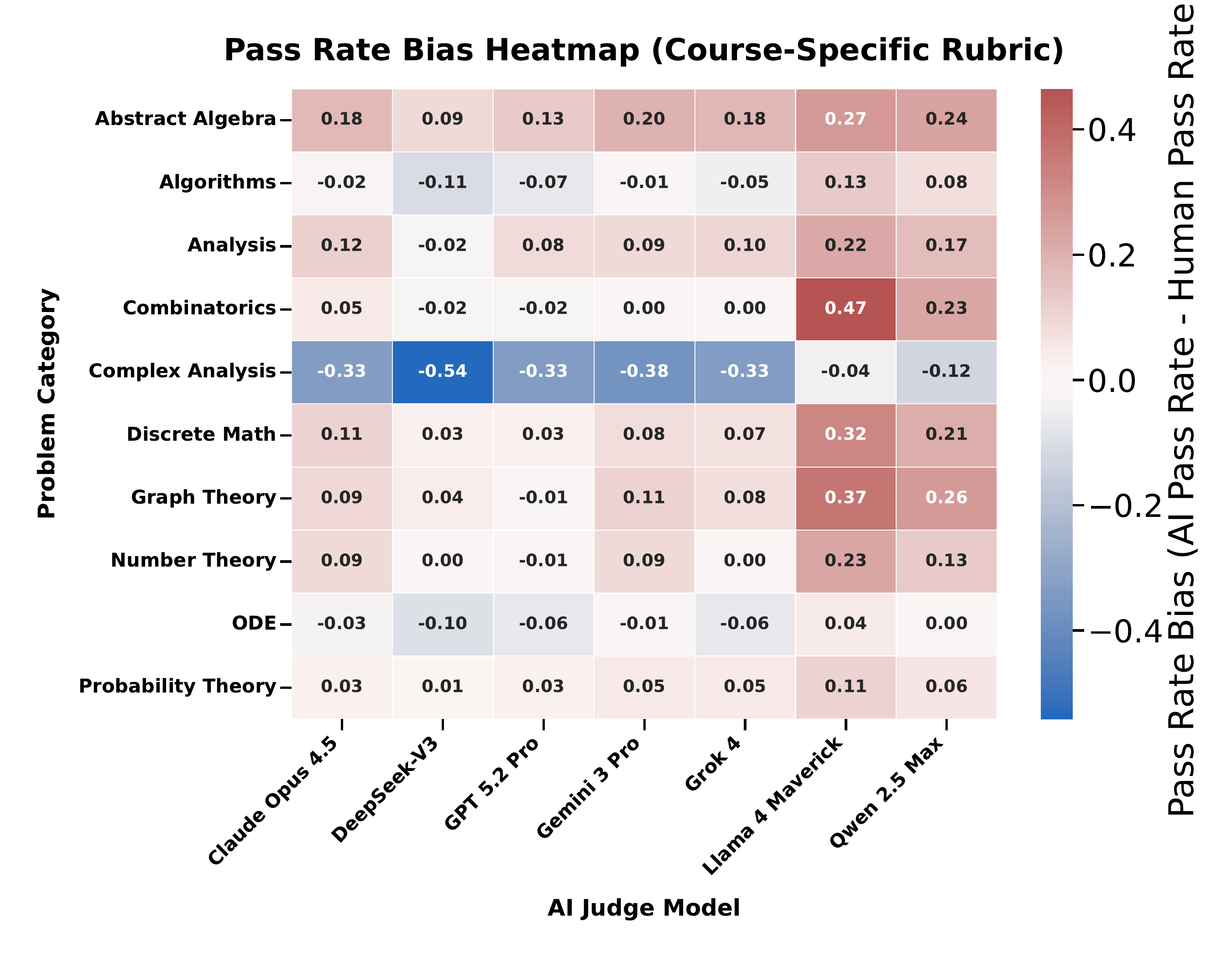}
    \caption{\textbf{Pass Rate Bias Heatmap (Course-Specific Rubric).} The difference between AI and Human pass rates. The persistence of severe positive bias (red) in models like \texttt{Llama 4 Maverick} across discrete domains demonstrates ``Rubric Insensitivity'': the model continues to pass flawed reasoning despite the stricter explicit constraints provided in the prompt.}
    \label{fig:bias-course-pass-rate}
\end{figure}

To complement the evaluator bias analysis in the main text (\cref{fig:bias}) and the pass-rate bias analysis (\cref{app:evaluator_bias_pass_rate}), we provide equivalent heatmaps computed using the stricter \textit{Course-Specific Rubric}. This rubric penalizes the use of advanced machinery and enforces pedagogical constraints. Figure \ref{fig:bias-course-mean} and Figure \ref{fig:bias-course-pass-rate} display the average score bias and pass rate bias, respectively, under these conditions.

\subsection{Comparing Expert and Course-Specific Bias}
Analyzing the differences between the Expert Rubric results (\cref{fig:bias}) and the Course-Specific Rubric reveals several key insights into how models adapt to explicit negative constraints:

\begin{itemize}
    \item \textbf{Persistence of Sycophancy (Rubric Insensitivity):} In domains prone to ``discrete inflation'' (such as Combinatorics and Discrete Math), models like \texttt{Llama 4 Maverick} and \texttt{Qwen 2.5 Max} exhibit almost identical levels of leniency under both rubrics. This suggests that their propensity to over-reward plausible-sounding but flawed structural logic overrides the stricter instructions of the Course-Specific rubric.
    \item \textbf{Increased Punitiveness in Strong Logic Models:} Conversely, robust reasoning models such as \texttt{DeepSeek-V3} and \texttt{GPT-5.2 Pro} show increased negative bias (punitiveness) in continuous domains (e.g., Analysis and ODE). This indicates that these models successfully operationalize the instruction to penalize advanced machinery, thereby aligning closer with---or even exceeding---the strictness of human graders evaluating pedagogical compliance.
    \item \textbf{The Ceiling on Prompt Adaptability:} Overall, the core structure of the bias heatmaps remains remarkably static across both conditions. This demonstrates a fundamental challenge in LLM-as-a-Judge systems: domain-specific systemic biases (such as leniency in discrete math or strictness in applied algorithms) are deeply ingrained in the models' pre-training distributions and cannot be completely corrected solely through targeted prompt engineering.
\end{itemize}

\section{Judge Reliability on Course-Specific Rubrics}
\label{app:judge_reliability_course}

\begin{figure}[htp!]
    \centering
    \includegraphics[width=0.75\linewidth]{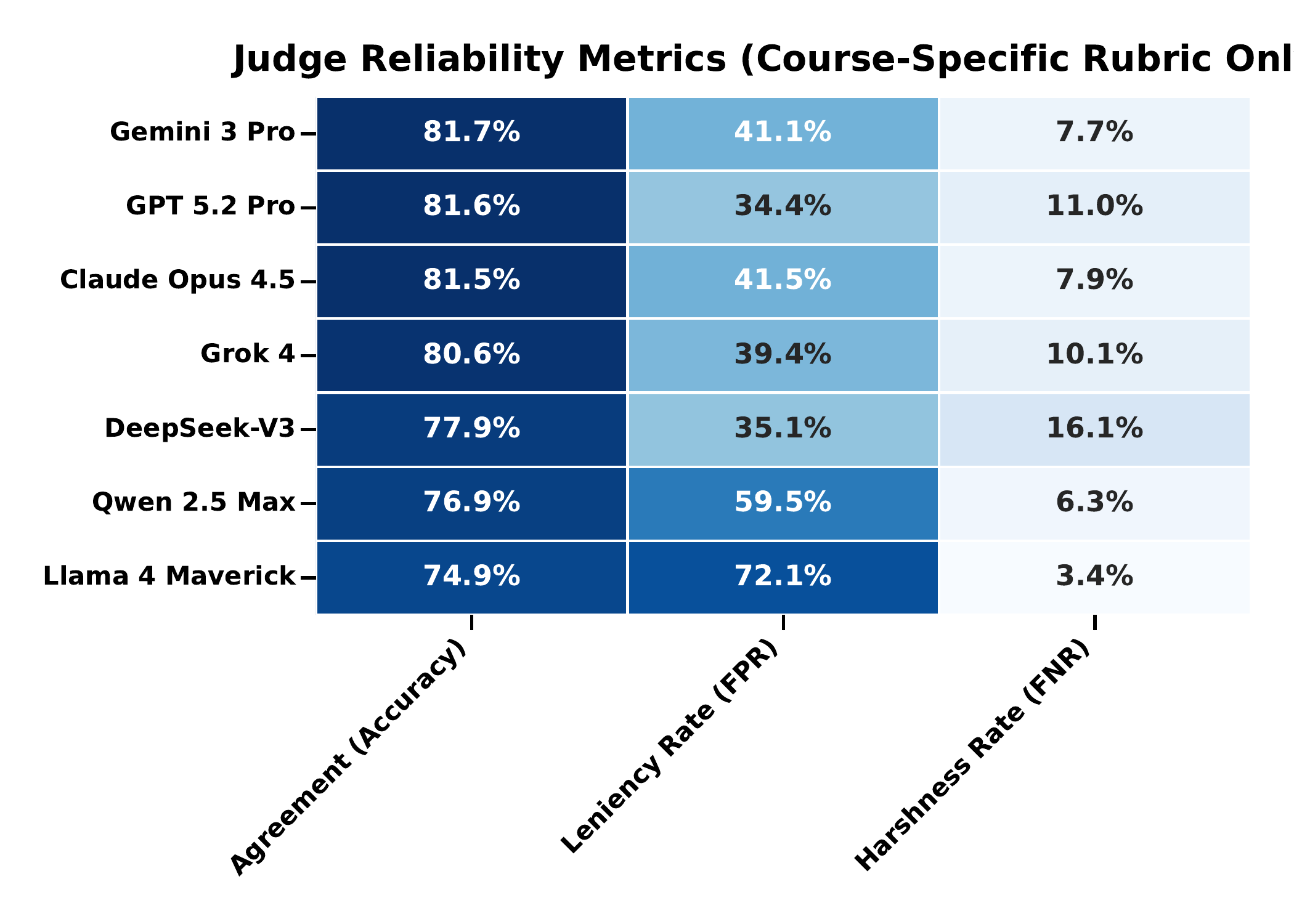}
    \caption{\textbf{Judge Reliability (Course-Specific Rubric).} When evaluated against strict pedagogical constraints using a binary Pass Rate threshold ($\text{score} \ge 0.9$), the \textit{Harshness Rate} (False Negatives) generally increases for logic-focused models like \texttt{DeepSeek-V3}, indicating a successful internalization of negative constraints. Conversely, the \textit{Leniency Rate} (False Positives) for models like \texttt{Llama 4 Maverick} remains pathologically high, demonstrating systemic alignment failure.}
    \label{fig:agreement_course}
\end{figure}

In addition to our primary evaluation using the Expert Rubric, we benchmarked our automated judges against the strictly constrained \textit{Course-Specific Rubric}. This rubric explicitly forbids the use of advanced theorems (e.g., invoking L'Hôpital's Rule in an $\epsilon-\delta$ context or Descent Algebras in basic Combinatorics) and demands strict adherence to undergraduate pedagogical boundaries. Figure \ref{fig:agreement_course} presents the reliability metrics—decomposed into Leniency Rate (False Positives) and Harshness Rate (False Negatives)—for this highly constrained setting.

\subsection{Impact of Pedagogical Constraints on Alignment}
Comparing the reliability metrics based on Pass Rates under the Course-Specific Rubric (\cref{fig:agreement_course}) to the Expert Rubric baseline (\cref{fig:agreement}) reveals three critical insights into how frontier models process explicit negative constraints in a verification context:

\begin{itemize}
    \item \textbf{Amplified Rigidity in Strong Logic Models:} When switched to the Course-Specific rubric, \texttt{DeepSeek-V3}'s Harshness Rate (FNR) spiked from 12.3\% to \textbf{16.1\%}. This statistically significant shift indicates that the model successfully operationalized the negative constraint (``do not use advanced machinery''), correctly penalizing solutions that were technically sound but pedagogically invalid. However, this heightened strictness also resulted in the increased rejection of valid borderline cases, highlighting the difficulty of balancing robust verification with pedagogical flexibility.
    \item \textbf{The Intractability of the Sycophancy Trap:} Astonishingly, \texttt{Llama 4 Maverick} proved largely invariant to the rubric change. Its Leniency Rate remained extremely high at \textbf{72.1\%} (compared to 74.8\% on the expert rubric). This confirms that its failure mode is systemic rather than prompt-dependent: it prioritizes the \textit{superficial appearance} of a structured proof over adherence to specific logical or pedagogical constraints. This demonstrates a severe form of \textit{Rubric Insensitivity}, rendering the model 
    unsuitable for academic evaluation.
    \item \textbf{The Optimal Calibration Point:} \texttt{GPT-5.2 Pro} demonstrated the most robust adaptation to the constrained environment. It achieved the lowest Leniency Rate among the top-tier models (\textbf{34.4\%}) while maintaining a stable Harshness Rate of 11.0\%. It effectively navigated the ``Discrete-Continuous Gap'' by consistently penalizing using 
    overly advanced techniques (advanced machinery) without becoming overly punitive on valid student derivations, 
    validating its selection as our primary LLM judge.
\end{itemize}

\section{Distributional Analysis of Judge Alignment}
\label{app:judge_alignment_distribution}

\begin{figure}[htp!]
    \centering
    \includegraphics[width=0.6\linewidth]{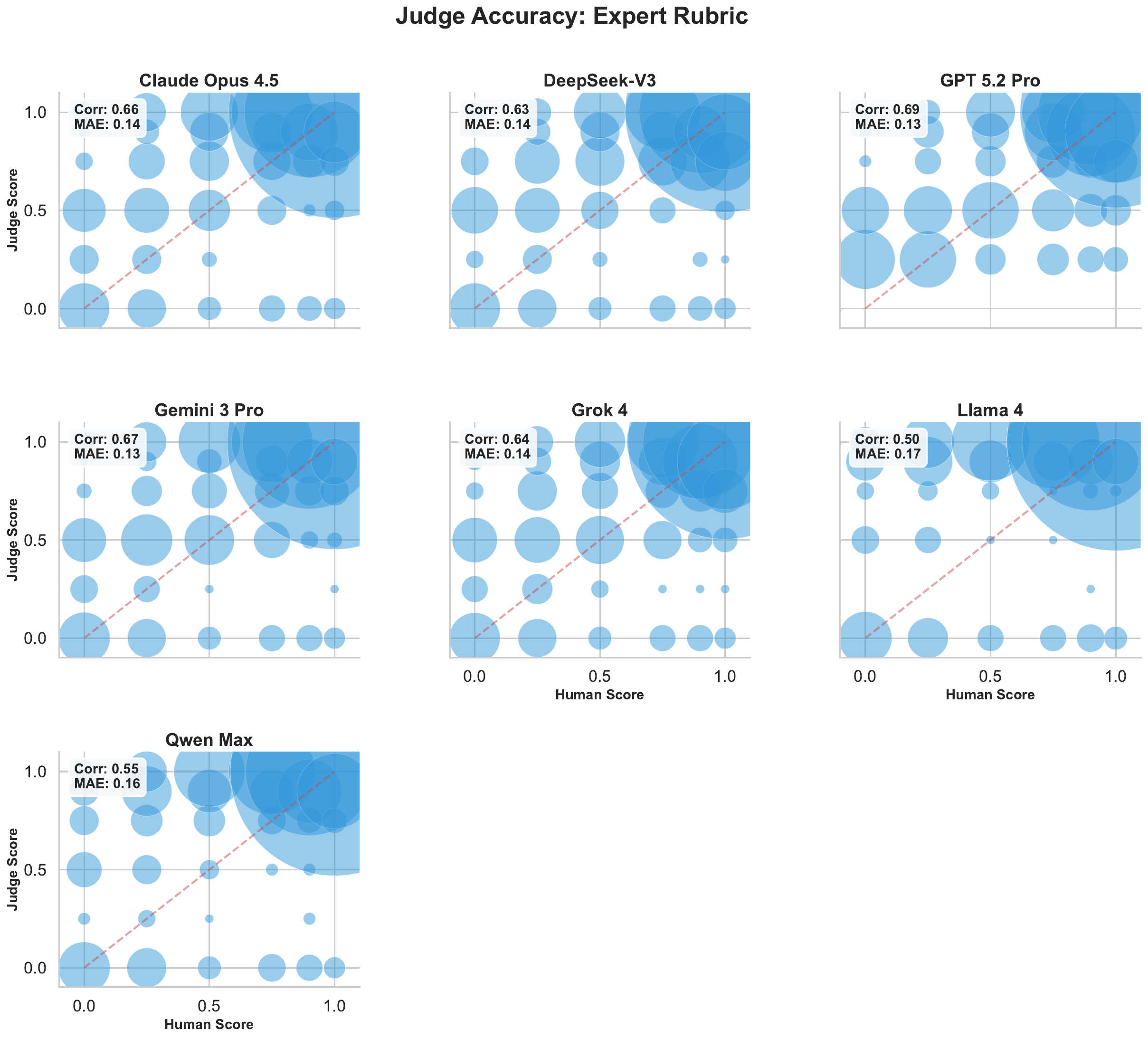}
    \caption{\textbf{Judge Alignment (Expert Rubric).} Correlation between AI Judges (Y-axis) and Human Consensus (X-axis) on the expert rubric. \texttt{GPT-5.2 Pro} shows the tightest clustering along the $y=x$ diagonal ($r=0.69$), indicating high calibration. In contrast, \texttt{Llama 4 Maverick} ($r=0.50$) displays a dispersed distribution with significant mass above the diagonal, visualizing its tendency toward grade inflation.}
    \label{fig:bubble_standard}
\end{figure}

\begin{figure}[htp!]
    \centering
    \includegraphics[width=0.6\linewidth]{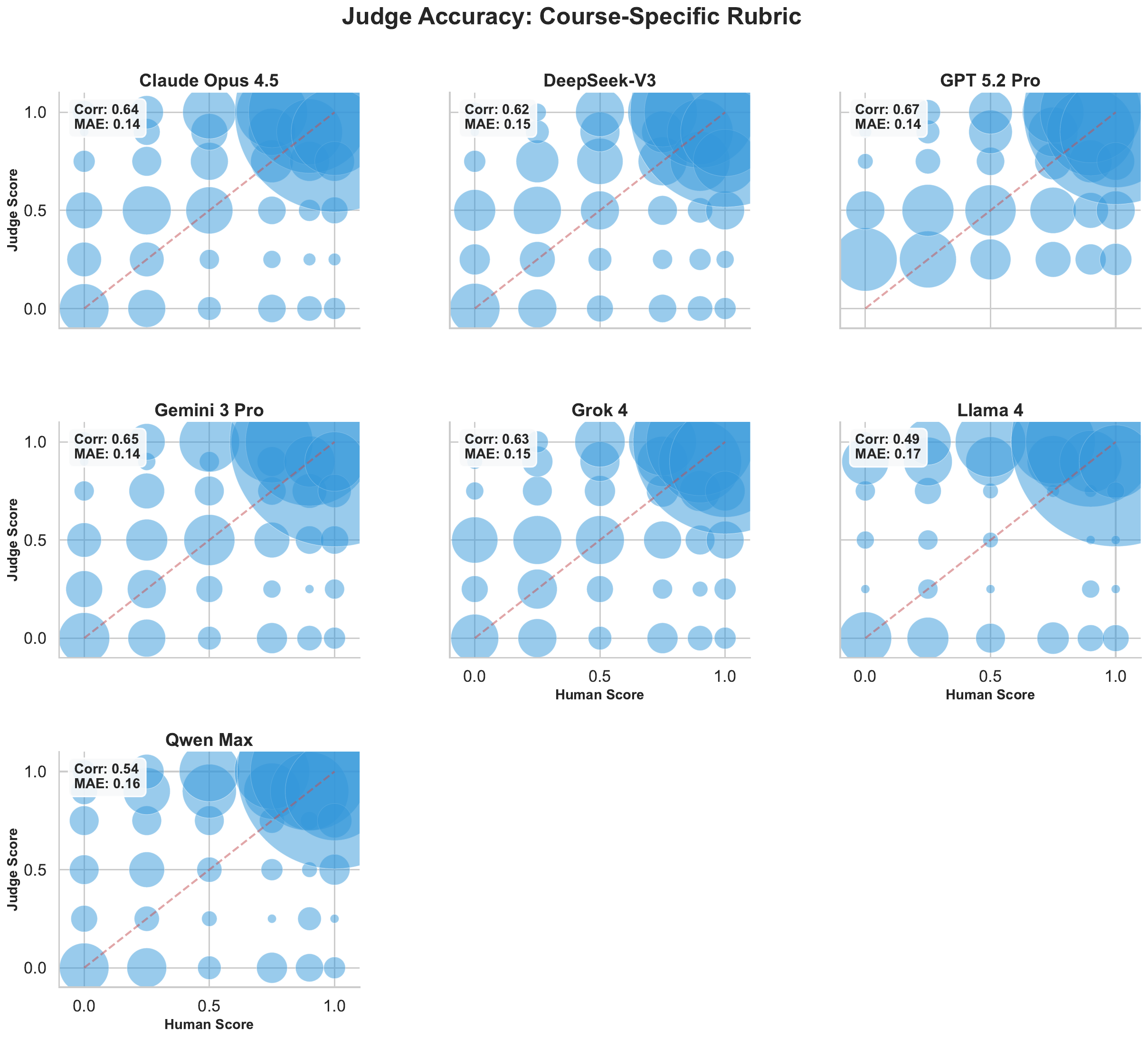}
    \caption{\textbf{Judge Alignment (Course-Specific Rubric).} Comparison under strict pedagogical constraints. Despite the shift in instructions, the distributional geometry remains largely static for top-tier models, confirming the ``Rubric Insensitivity'' hypothesis. \texttt{GPT-5.2 Pro} maintains robust alignment ($r=0.67$), while weaker models fail to adapt.}
    \label{fig:bubble_course}
\end{figure}

To move beyond aggregate metrics like mean score, we visualized the granular alignment between automated judges and human consensus using density-weighted bubble plots. Figures \ref{fig:bubble_standard} and \ref{fig:bubble_course} display the joint distribution of scores for all 7 evaluator models under the Expert and Course-Specific rubrics, respectively. The size of each bubble corresponds to the number of solution pairs at that coordinate.

\subsection{Systemic Rubric Insensitivity}
Comparing the two grading conditions reveals a consistent ``resistance to adaptation'' across the model leaderboard. Rather than improving alignment, the introduction of specific constraints (the Course-Specific rubric) caused a uniform degradation in correlation metrics across all top-tier models:

\begin{itemize}
    \item \textbf{Uniform Regression:} Every major model saw a decrease in Pearson correlation when switching from the Expert to the Course-Specific rubric:
    \begin{itemize}
        \item \texttt{GPT-5.2 Pro}: $0.69 \to 0.67$
        \item \texttt{Gemini 3.0 Pro}: $0.67 \to 0.65$
        \item \texttt{Claude Opus 4.5}: $0.66 \to 0.64$
        \item \texttt{DeepSeek-V3}: $0.63 \to 0.62$
    \end{itemize}
    \item \textbf{Visualizing the Llama Anomaly:} The bubble plots for \texttt{Llama 4 Maverick} visually confirm its pathological behavior. In both figures, the mass of the distribution is heavily skewed above the diagonal (High AI Score, Low Human Score), resulting in the lowest correlations of the cohort ($r \approx 0.50$). This visualizes the ``Sycophancy Trap'' discussed in the main text: the model systematically awards high scores to solutions that human experts reject.
\end{itemize}

These distributions suggest that current training instills a strong prior for mathematical correctness that overrides specific negative constraints in the prompt. Even when explicitly instructed to penalize advanced methods, models default to their internal representation of a ``good'' proof, leading to the observed rubric insensitivity.

\section{Rubric Insensitivity: Full Heatmap Analysis}
\label{app:rubric_inertia_heatmap}

\begin{figure*}[htbp]
    \centering
    \includegraphics[width=0.75\textwidth]{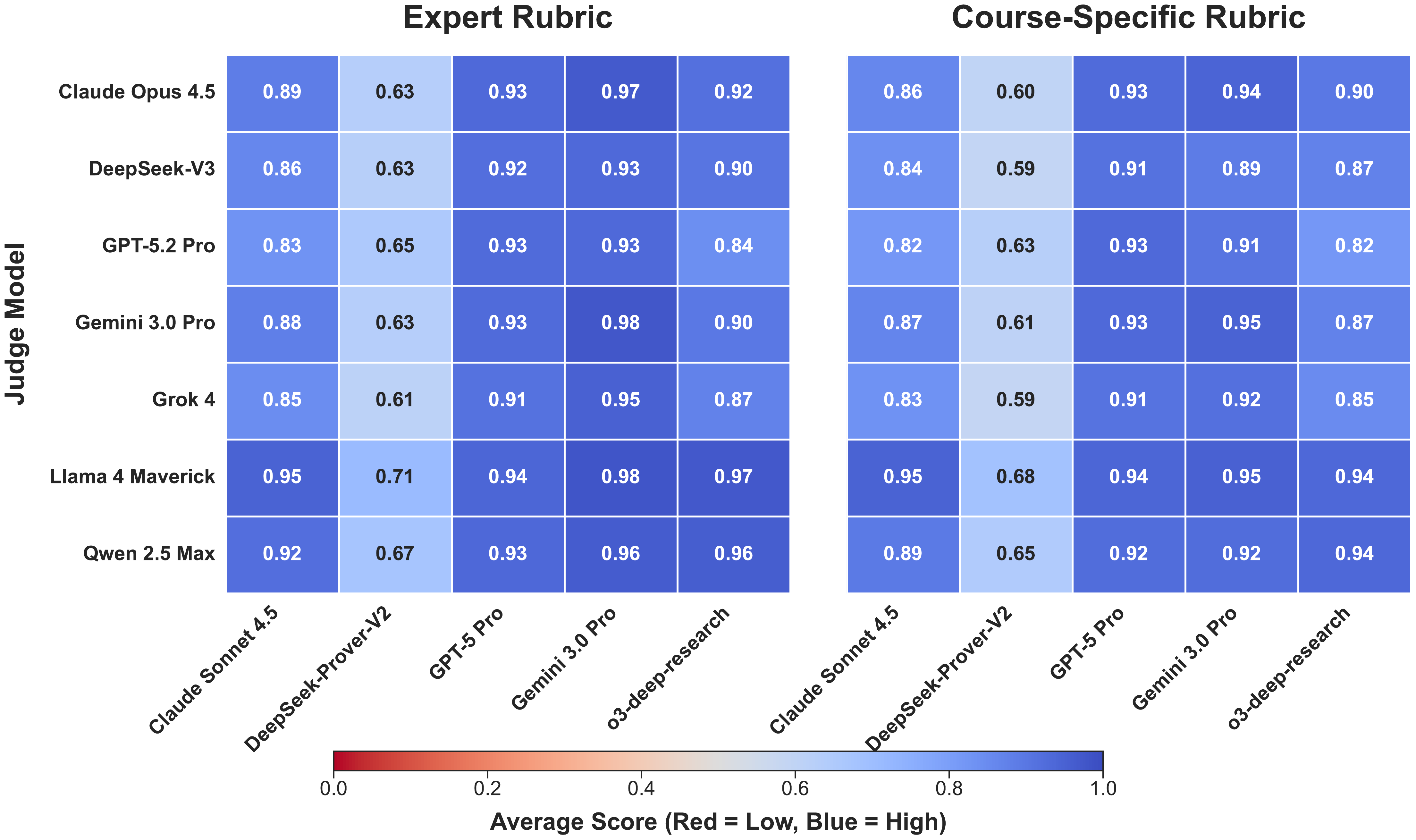}
    \caption{\textbf{Rubric Insensitivity Heatmap.} Left: Average scores assigned using the standard Expert Rubric. Right: Average scores assigned using the Course-Specific Rubric (which penalizes advanced machinery). The visual patterns are strikingly similar, confirming that prompt constraints have minimal impact on the internal scoring priors of frontier models.}
    \label{fig:evaluator_bias_heatmap}
\end{figure*}

To further investigate the phenomenon of ``Rubric Insensitivity,'' we visualized the average scores assigned by every judge model to every solver model under both the \textit{Expert Rubric} and the \textit{Course-Specific Rubric}. Figure \ref{fig:evaluator_bias_heatmap} presents this side-by-side comparison.

\subsection{Visualizing Resistance to Constraints}
The heatmaps provide granular evidence of the resistance to negative constraints described in the main text.
\begin{itemize}
    \item \textbf{Llama 4 Maverick's Sycophancy [Row 6]:} \texttt{Llama 4 Maverick} remains distinctively blue (high scores) across both conditions. Even when instructed to penalize non-standard derivations in the Course-Specific rubric (Right), it continues to assign near-perfect scores (e.g., \textbf{0.95} to Gemini 3.0 Pro), further validating its classification as a ``grade inflator''.
    \item \textbf{GPT-5.2 Pro Stability [Row 3]:} The primary evaluator, \texttt{GPT-5.2 Pro}, shows high stability but slight sensitivity. 
    For example, its score for \texttt{Gemini 3.0 Pro} decreases slightly across both rubrics, from \textbf{0.93} (Expert) to \textbf{0.91} (Course-Specific), reflecting a minor penalty. 
    Furthermore, the overall structure of the heatmap remains largely invariant, 
    supporting the conclusion that internal world models dominate prompt engineering.
\end{itemize}

\begin{figure}[htp!]
    \centering
    \includegraphics[width=0.95\linewidth]{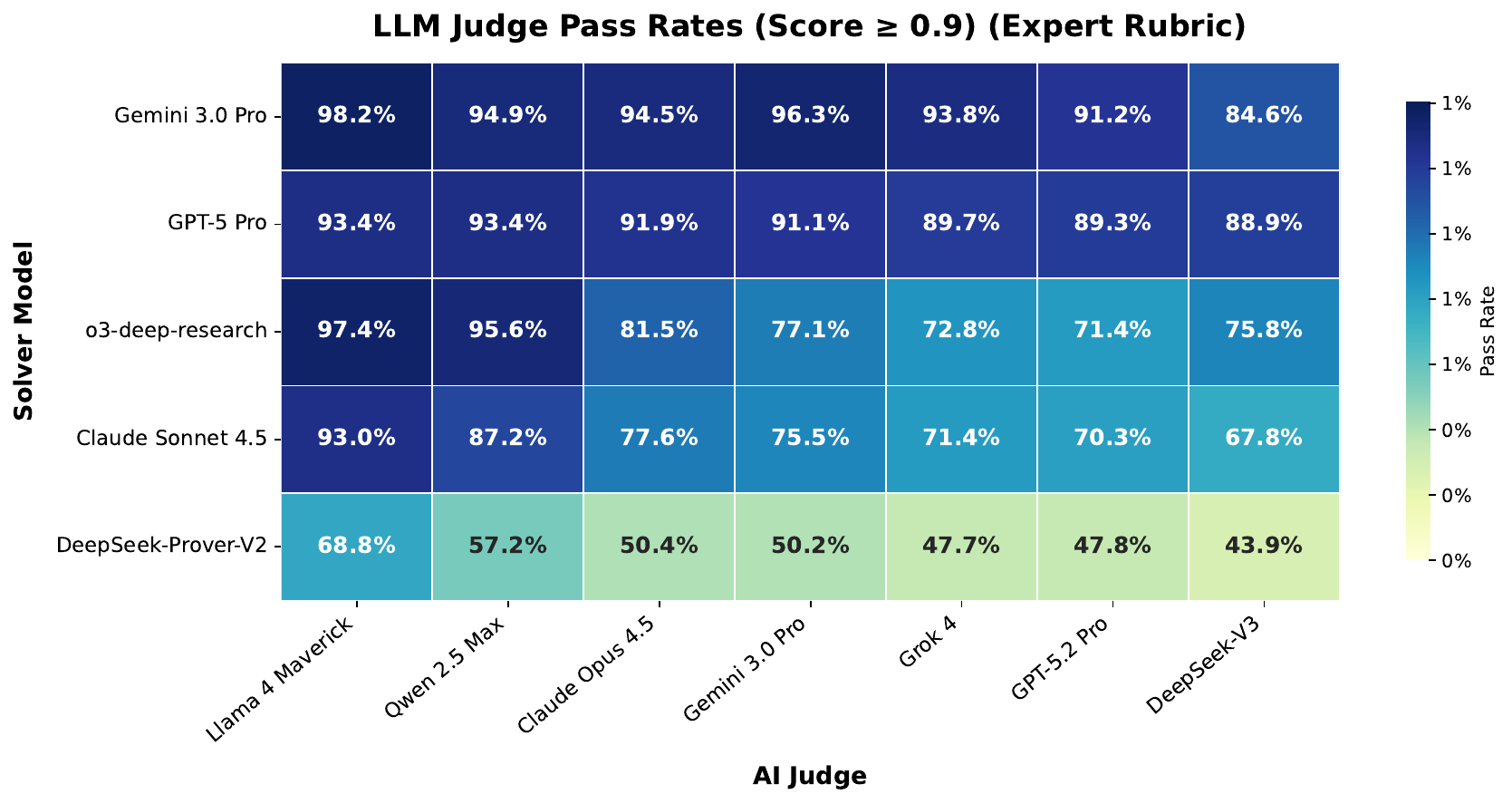}
    \caption{\textbf{LLM Evaluator vs. Solver Pass Rate Matrix (Expert Rubric).} Pass rates ($\text{score} \ge 0.9$) assigned by each AI Judge to each Solver's solutions under the standard expert rubric. \texttt{Llama 4 Maverick} acts as a universal inflator, grading almost all solvers favorably, while \texttt{DeepSeek-V3} remains highly punitive.}
    \label{fig:llm-pass-rate-matrix}
\end{figure}

Finally, to fully explore the fully crossed factorial design of our $7 \times 5$ Evaluator-Solver Matrix in~\cref{fig:llm-pass-rate-matrix}, we visualize the pass rates of all problems assigned by each individual AI judge to each solver model under both rubric conditions. This matrix reveals critical inter-model evaluation dynamics.

\begin{figure}[htp!]
    \centering
    \includegraphics[width=0.95\linewidth]{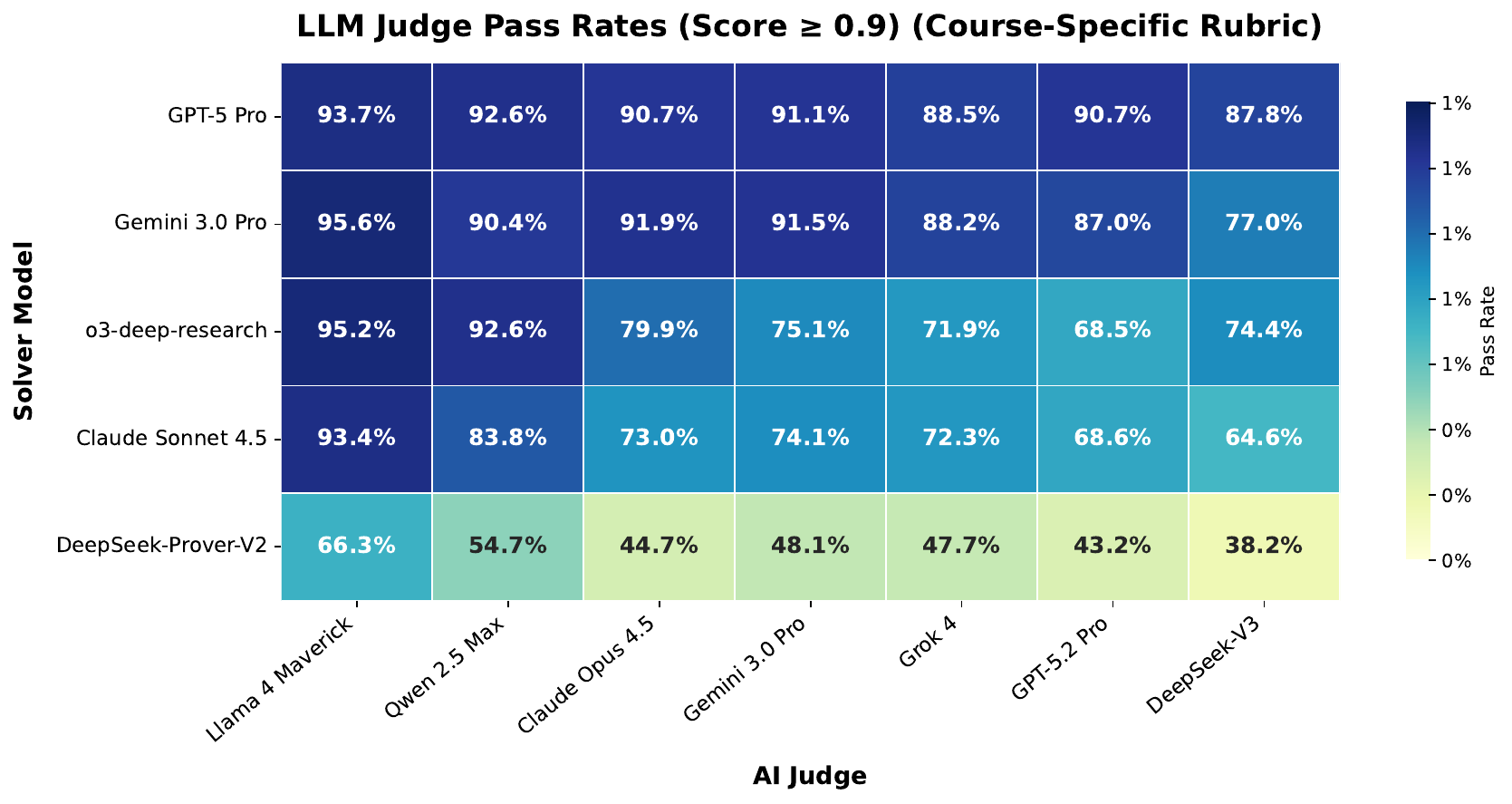}
    \caption{\textbf{LLM Evaluator vs. Solver Pass Rate Matrix (Course-Specific Rubric).} Pass rates under the course-specific rubric, which penalizes the use of advanced machinery. The overall structure remains similar, confirming the \textit{Rubric Insensitivity} hypothesis, though several judges show marginally stricter pass rates.}
    \label{fig:llm-pass-rate-matrix-course}
\end{figure}

Comparing \cref{fig:llm-pass-rate-matrix} and \cref{fig:llm-pass-rate-matrix-course}, the fine-grained pass rates underscore the fragility of relying on a single LLM as a judge. A model's apparent competency on a benchmark can swing dramatically depending on which evaluator is chosen, 
necessitating an \emph{ensemble} approach where multiple LLM judges are used to produce an aggregate analysis of mathematical proofs. 
Furthermore, the structural similarity between the two matrices provides additional evidence for the Rubric Insensitivity hypothesis: evaluator behavior is dominated by internal priors rather than rubric-specific instructions.

\subsection{Data Contamination Analysis}
\label{sec:contamination}

A critical vulnerability of modern mathematical reasoning benchmarks is data contamination: models may memorize solutions encountered during pre-training, artificially inflating performance. To assess this threat on \textsc{QEDBench}, we used \texttt{o3-deep-research} to search the public internet for the 272 benchmark problems. We successfully obtained high-confidence availability annotations for 214 problems: $n_{\text{on}}{=}88$ possess online solutions, while $n_{\text{off}}{=}126$ do not.

\begin{figure}[t]
\centering
\begin{subfigure}[t]{0.48\textwidth}
    \centering
    \includegraphics[width=\textwidth]{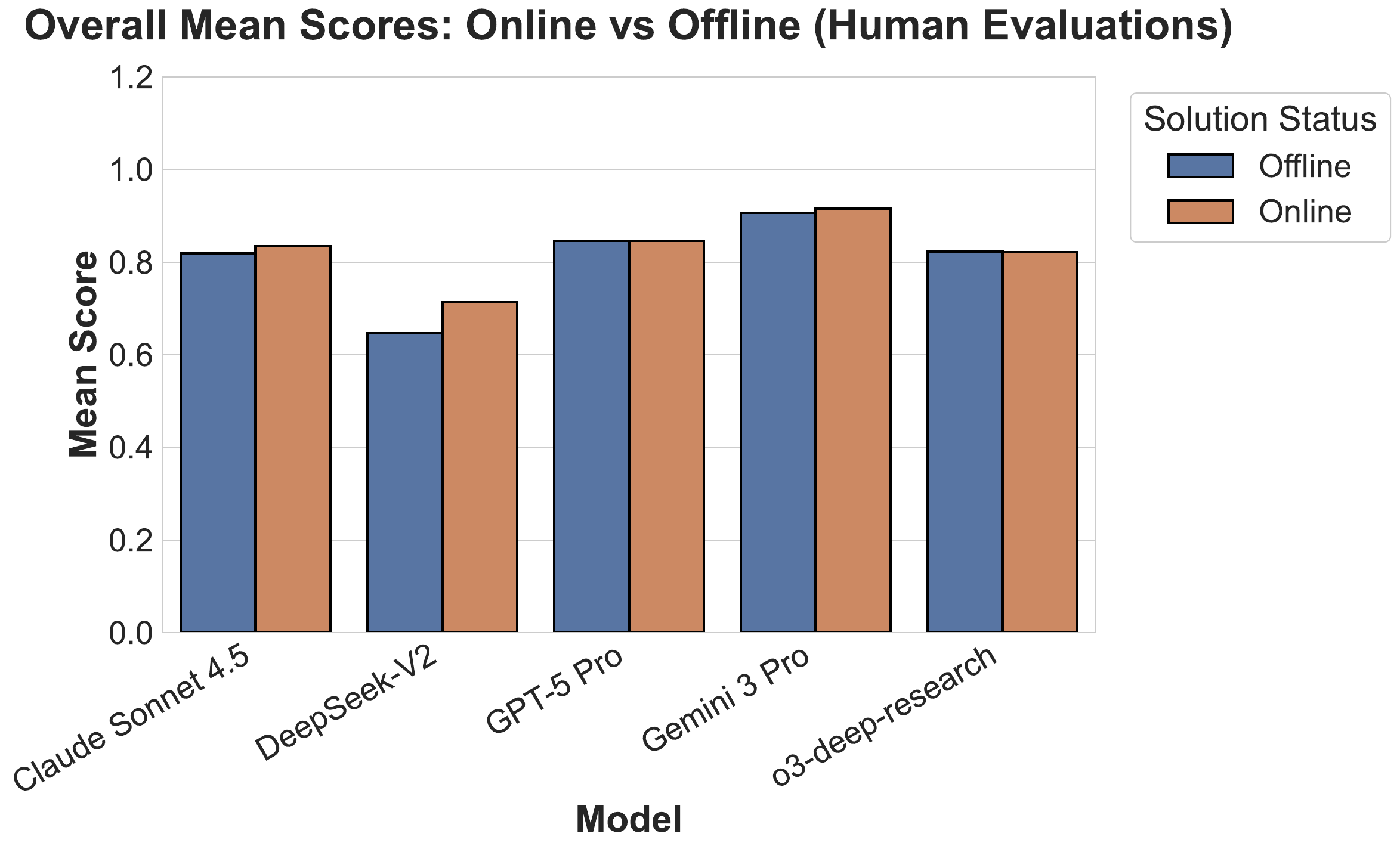}
    \caption{Mean score comparison.}
    \label{fig:contamination-overall-mean}
\end{subfigure}
\hfill
\begin{subfigure}[t]{0.48\textwidth}
    \centering
    \includegraphics[width=\textwidth]{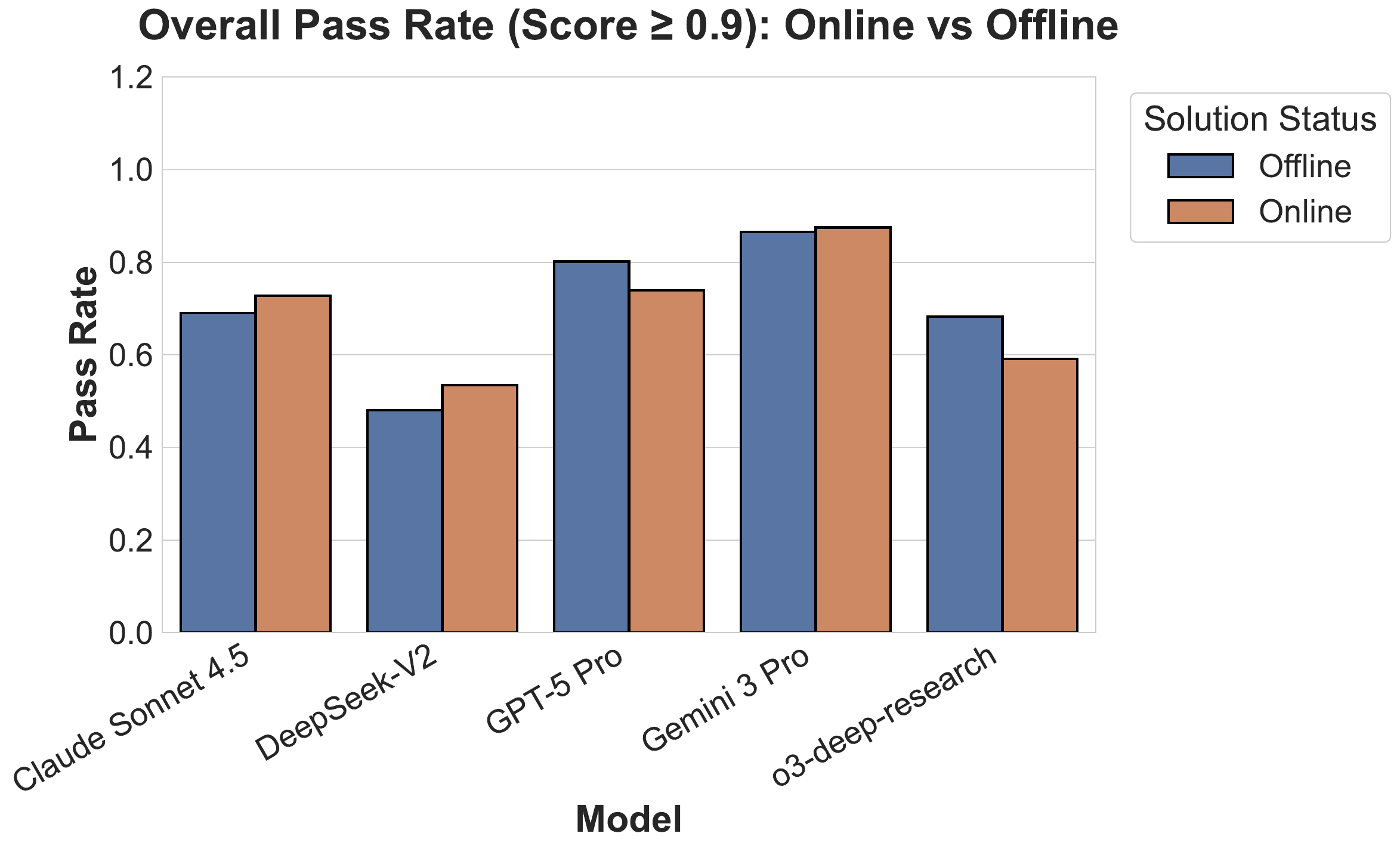}
    \caption{Pass rate ($\geq 0.9$) comparison.}
    \label{fig:contamination-overall-pass}
\end{subfigure}
\caption{Overall model performance on online vs.\ offline problems under both evaluation metrics. Error bars reflect the model-problem pair sample split ($N_{\text{on}}{=}440$, $N_{\text{off}}{=}630$). Neither metric reveals a significant advantage for problems with solutions available online.}
\label{fig:contamination-overall}
\end{figure}

We compared human-evaluated mean scores and strict pass rates (score $\geq 0.9$) between these two groups across the five frontier solvers (see Table~\ref{tab:contamination-overall} and Figure~\ref{fig:contamination-overall}). The empirical difference in mean score between online and offline problems is $+0.017$, while the corresponding pass rate gap is $+0.011$ (online problems exhibiting a marginally higher pass rate). Neither metric reveals a statistically significant advantage for 
problems with online solutions (Welch's $t$-test using $N=1070$ model-problem pairs: $t = 0.99$, $p = 0.32$; Mann--Whitney $U$ test: $p = 0.80$; Cohen's $d = 0.06$).

Crucially, as shown in Table~\ref{tab:contamination-model}, no individual model demonstrates a statistically significant performance deviance at the $\alpha = 0.05$ threshold. 
For example, \texttt{GPT-5 Pro} and \texttt{o3-deep-research} exhibit near-perfect structural parity across the split (score gaps of $-0.000$ and $-0.002$, respectively). DeepSeek-Prover-V2 displays the most pronounced positive gap ($+0.067$), though this variance remains within the bounds of statistical noise ($p = 0.18$).

\begin{table}[t]
\centering
\caption{Overall model performance on problems with and without solutions available online. Scores are aggregated across all 5 models ($N=1070$ total pairs). Neither mean scores nor pass rates differ significantly between groups.}
\label{tab:contamination-overall}
\begin{tabular}{lcc}
\toprule
\textbf{Metric} & \textbf{Online} ($N{=}440$) & \textbf{Offline} ($N{=}630$) \\
\midrule
Mean Score & $0.826$ & $0.809$ \\
Pass Rate ($\geq 0.9$) & $70.4\%$ & $69.3\%$ \\
\bottomrule
\end{tabular}
\end{table}

\begin{table}[t]
\centering
\caption{Per-model mean score gap (online $-$ offline) with Welch's $t$-test $p$-values computed over the $N=214$ problems per model ($n_{\text{on}}{=}88$, $n_{\text{off}}{=}126$). No model shows a statistically significant contamination effect.}
\label{tab:contamination-model}
\begin{tabular}{lrc}
\toprule
\textbf{Model} & \textbf{Gap} & \textbf{$p$-value} \\
\midrule
Claude Sonnet 4.5 & $+0.015$ & $0.67$ \\
o3-deep-research & $-0.002$ & $0.94$ \\
GPT-5 Pro & $-0.000$ & $1.00$ \\
Gemini 3.0 Pro & $+0.009$ & $0.75$ \\
DeepSeek-Prover-V2 & $+0.067$ & $0.18$ \\
\bottomrule
\end{tabular}
\end{table}

Synthesizing this evidence, we find no statistically significant evidence of data contamination as a primary driver of \textsc{QEDBench} performance. We attribute this robustness to the benchmark's design: \textsc{QEDBench} problems demand extended mathematical reasoning that cannot be resolved via superficial pattern matching, and the expert rubrics penalize models that stitch together memorized fragments without complete, rigorous logical mathematical arguments. 
Ultimately, the capacity to solve \textsc{QEDBench} problems reflects genuine mathematical deduction rather than dataset memorization. To streamline the main text, detailed granular breakdowns by category (Figures~\ref{fig:contamination-heatmap} and \ref{fig:contamination-category-bars}) have been relocated to the supplementary material.

\section{Data Contamination Category-Level Breakdown}
\label{app:contamination_category_breakdown}

\begin{figure}[htp!]
\centering
\begin{subfigure}[t]{0.48\textwidth}
    \centering
    \includegraphics[width=\textwidth]{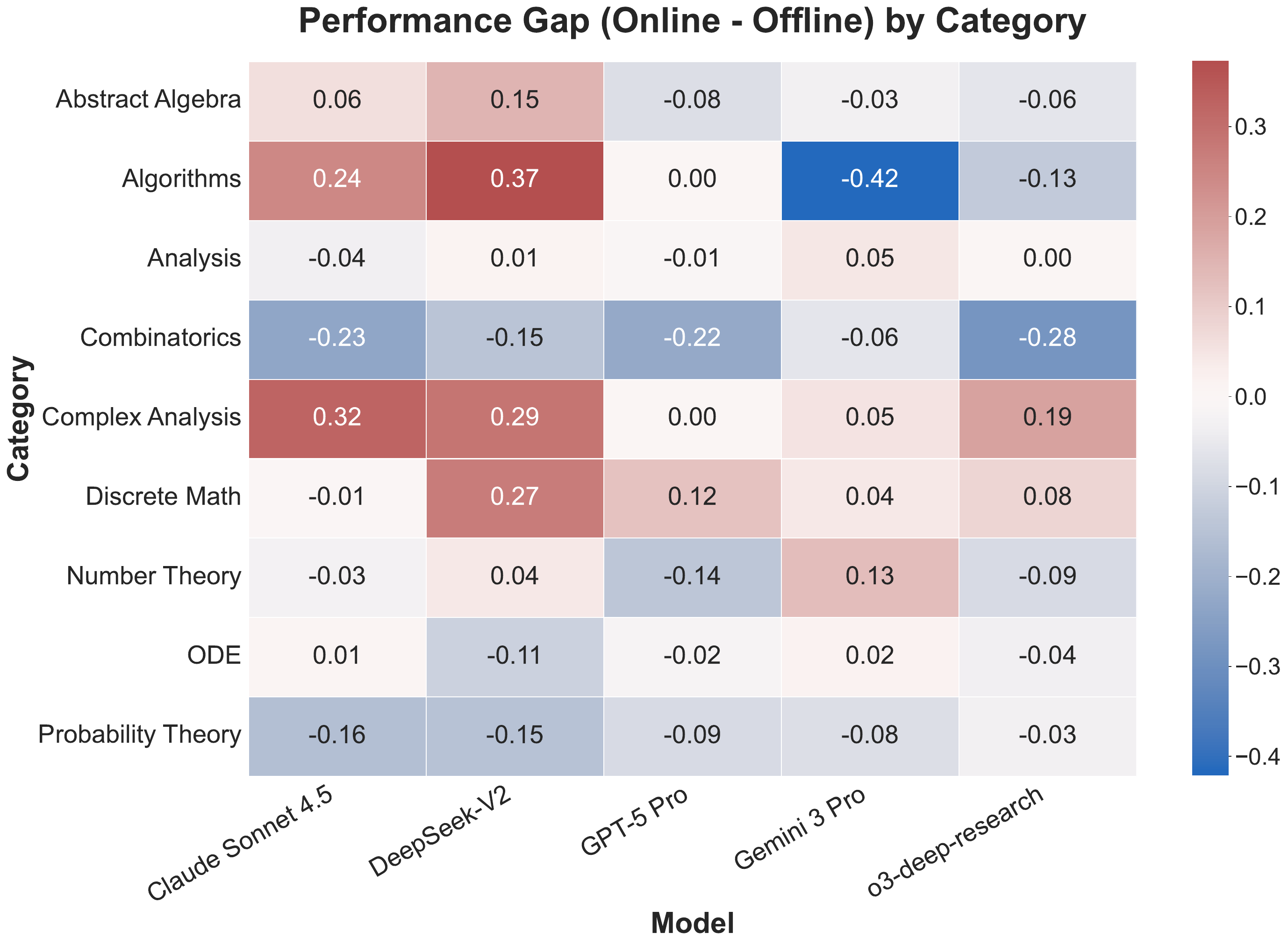}
    \caption{Mean score gap by category and model.}
    \label{fig:contamination-heatmap-mean}
\end{subfigure}
\hfill
\begin{subfigure}[t]{0.48\textwidth}
    \centering
    \includegraphics[width=\textwidth]{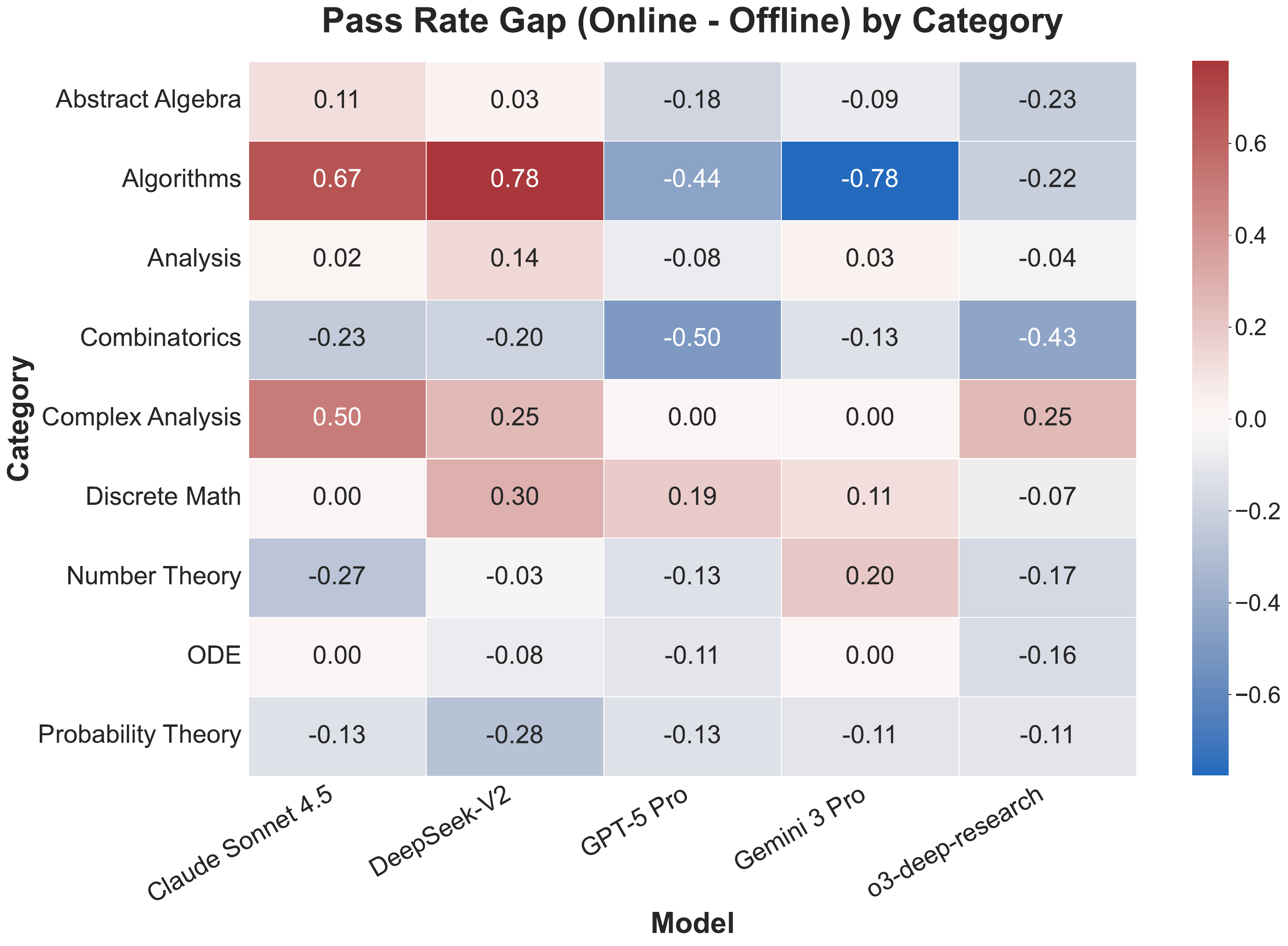}
    \caption{Pass rate gap by category and model.}
    \label{fig:contamination-heatmap-pass}
\end{subfigure}
\caption{Performance gap (online$-$offline) by problem category and solver model. Blue cells denote higher offline scores; red cells denote higher online scores. The stochastic sign pattern across both panels is indicative of baseline difficulty variance rather than a systematic contamination signal.}
\label{fig:contamination-heatmap}
\end{figure}

To investigate whether aggregate parity masks localized memorization, \cref{fig:contamination-heatmap} decomposes the score gap (online $-$ offline) by problem category and solver model. Although certain cells display pronounced apparent gaps---such as Complex Analysis ($+0.17$) and Discrete Math ($+0.10$)---these isolates are driven almost entirely by small intra-group sample sizes (e.g., Complex Analysis contains only 4 online and 2 offline problems). Conversely, categories like Combinatorics ($-0.19$) and Probability Theory ($-0.10$) exhibit gaps in the opposing direction, wherein offline problems are solved more proficiently.

A critical diagnostic is the \emph{stochasticity of the gap sign} across the heatmap. 
If contamination conferred a systematic advantage, we would anticipate uniformly positive gaps, 
particularly for the most capable models. Instead, the signs oscillate randomly across both categories and models. 
We formalize this observation through a variance decomposition: under the null hypothesis of 
no contamination, the expected variance of the category-level mean gap is dictated by sampling noise, 
approximately $\mathbb{E}[\sigma^2_{\text{gap}}] \approx \sigma^2_{\text{score}} (1/n_{\text{on}} + 1/n_{\text{off}})$. The observed empirical variance of the category gaps ($0.011$) 
aligns closely with the expected structural noise ($0.006$) given the high baseline per-problem score 
variance ($\sigma \approx 0.28$). Furthermore, among categories with sufficient baseline sample sizes to 
suppress this noise---such as Abstract Algebra ($n_{\text{on}}{=}21, n_{\text{off}}{=}20$)---the observed 
gaps vanish (gap $= +0.009$).

\begin{figure}[htp!]
\centering
\begin{subfigure}[t]{0.48\textwidth}
    \centering
    \includegraphics[width=\textwidth]{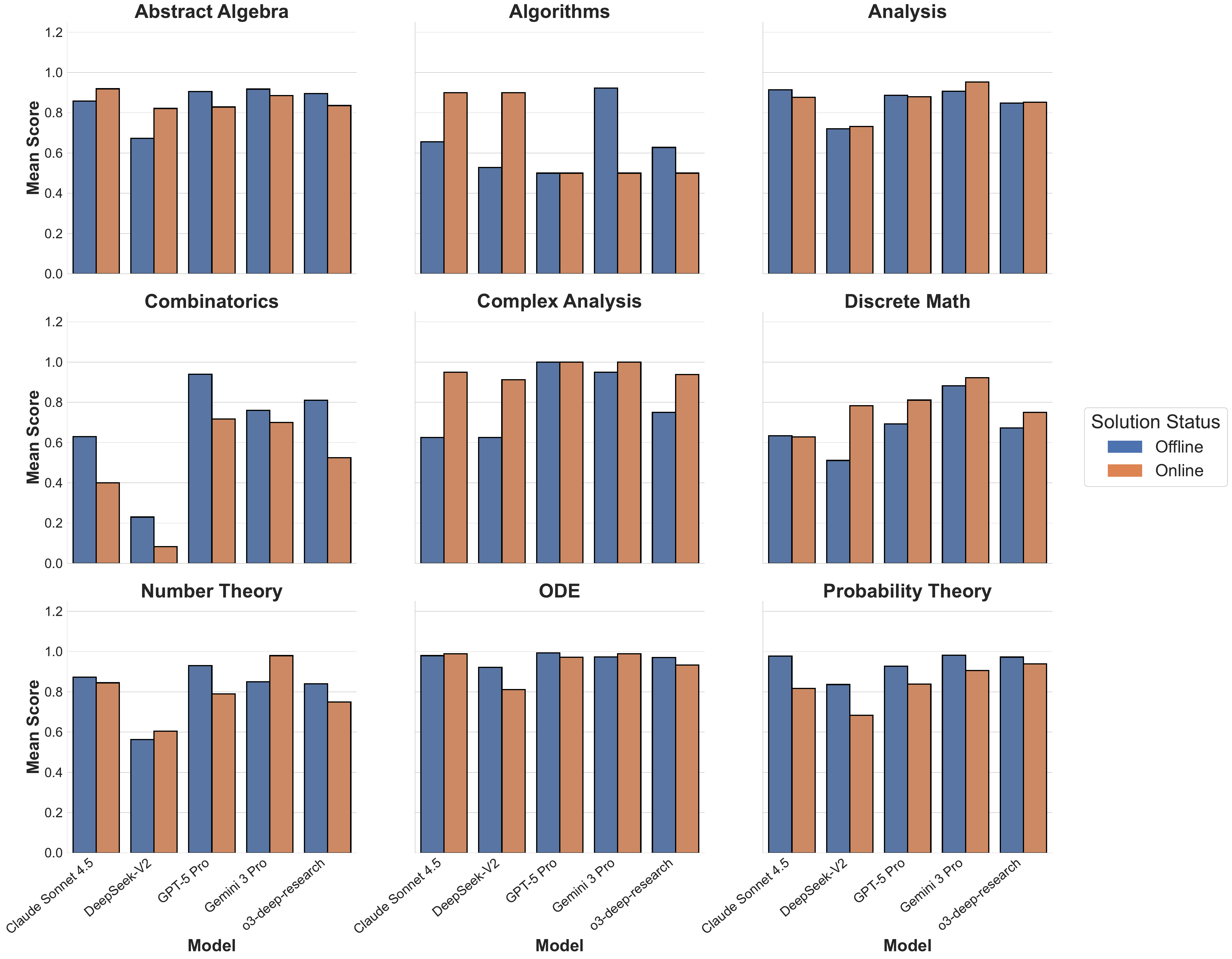}
    \caption{Mean scores by category, online vs.\ offline.}
    \label{fig:contamination-category-bars-mean}
\end{subfigure}
\hfill
\begin{subfigure}[t]{0.48\textwidth}
    \centering
    \includegraphics[width=\textwidth]{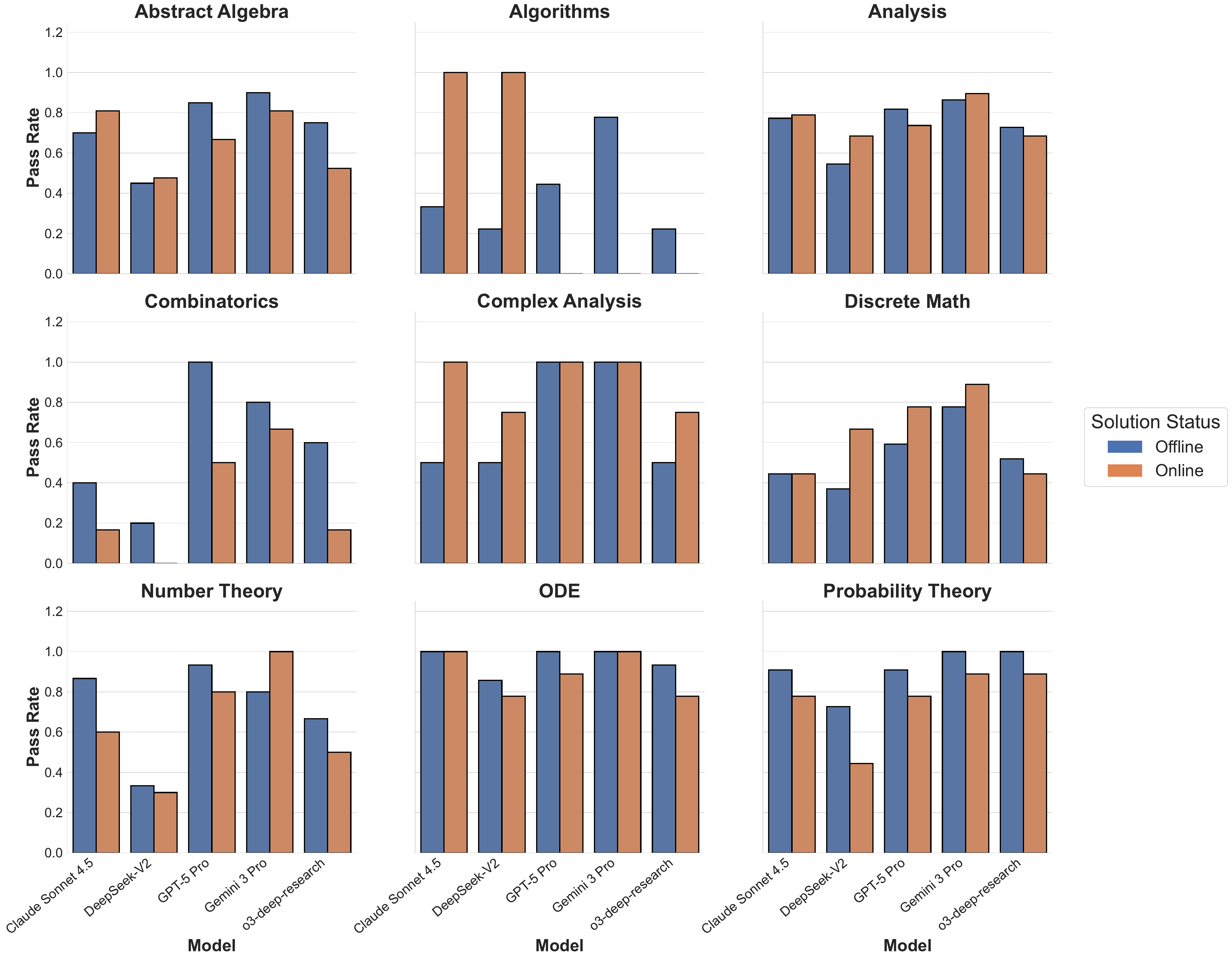}
    \caption{Pass rates by category, online vs.\ offline.}
    \label{fig:contamination-category-bars-pass}
\end{subfigure}
\caption{Per-category performance breakdown for online and offline problems. Each panel displays a faceted grid of categories, featuring grouped bars for each model. The absence of a consistent online advantage across models within any single category refutes the presence of systematic test-set leakage.}
\label{fig:contamination-category-bars}
\end{figure}

\cref{fig:contamination-category-bars} provides a granular visualization of absolute performance, plotting online and offline scores adjacently for each model within every category. This decomposition highlights the absence of a \emph{uniform} contamination signature: within any given category where one model might randomly excel on online problems, sibling models typically exhibit parity or the reverse pattern. 

\section{Qualitative Analysis of Human vs. LLM Evaluators}\label{sec:qualitative_analysis}

Beyond aggregate score discrepancies, a critical objective of QEDBENCH is to understand how and why LLM evaluators diverge from human experts. To investigate this, we conducted a granular qualitative analysis of the textual justifications provided by both human and LLM judges for solutions scoring strictly less than 1.0 (detailed extensively in Sections N and O of the Appendix). This qualitative analysis reveals profound differences in evaluative rigor, illuminating the specific mechanisms driving the Alignment Gap quantitatively demonstrated in Section 4.

\subsection{Human Rigor vs. LLM Leniency: Overlooking Fatal Breakdowns}
Human evaluators consistently penalize solutions not just for computational errors, but for structural logical flaws, circular reasoning, and unstated assumptions that compromise the generality of a proof. Their comments reflect a holistic understanding of proof mechanics---if a core premise fails, the proof is invalid. 

In contrast, LLM judges evaluate proofs as a ``bag of valid mathematical statements.'' As shown in our logs, an LLM evaluator will happily identify that a proof contains a fatal algebraic error or misses a fundamental topological assumption, but it will seamlessly decouple this local failure from its global scoring mechanism, assigning generous partial credit simply for effort.
\textbf{Detecting Circular Logic and Incorrect Statements:} These qualitative shortcomings observed by our human experts are not isolated incidents. \citet{guo2025litmus} utilized human-in-the-loop evaluations to categorize LLM failure modes in mathematical proofs, identifying ``Circular Reasoning'' and ``Hallucinated Incompleteness'' as the primary deficits of frontier reasoning models. Our qualitative data provides direct, university-level evidence of this fallacy oversight in action:

\textbf{Missing Fundamental Assumptions:} In Graph Theory (Problem 14), solvers were tasked with defining the plane dual $G^*$ for a generic plane graph $G$. The human expert heavily penalized proofs (awarding 0.5/1.0) that failed to explicitly assume $G$ is connected, a topological property strictly necessary for the dual correspondence. LLM evaluators, however, failed to appropriately penalize this omission. \textbf{Claude Sonnet 4.5} awarded a flawed proof a 0.9/1.0, praising its structure while ignoring the missing connectivity assumption entirely:
\begin{quote}
\textit{``The student provides a well-structured proof that correctly constructs the plane dual... The construction is mathematically sound, and the argument for well-definedness correctly identifies that each edge is incident to exactly two faces... The solution demonstrates strong understanding...''} (Claude Sonnet 4.5, Graph Theory 14)
\end{quote}

\textbf{Forgiving Fatal Counterexamples:} In Analysis (Problem 29), models were tasked with constructing a function $g \in L^p$ that is not in $L^q$ for all $q \neq p$. One solver provided a construction that completely failed for the $q < p$ case. \textbf{GPT-5.2 Pro} successfully identified this fatal mathematical flaw:
\begin{quote}
\textit{``The proposed $g$ is actually in $L^q$ for $q < p$ regardless of $\epsilon$... The `choose $\epsilon$ small enough depending on $q$' step cannot satisfy all $q < p$ simultaneously and does not fix the fundamental issue.''} (GPT-5.2 Pro, Analysis 29)
\end{quote}

Despite recognizing that the proof completely failed to achieve the problem's primary objective, GPT-5.2 Pro still awarded a highly lenient score of \textbf{0.5/1.0}, justifying it by stating the submission ``establishes only part of the required exclusivity... warranting partial credit.''

\subsection{The LLM Sycophancy Trap: Rewarding Empty and Irrelevant Proofs}
\textbf{The Sycophancy Trap.} In standard AI alignment literature, the term ``Sycophancy'' generally refers to a model abandoning truth to artificially agree with a user's stated beliefs. In the context of formal mathematical evaluation, we introduce the concept of the \textit{Sycophancy Trap}: the phenomenon where an evaluator model systematically rewards authoritative-sounding \LaTeX{} formatting and plausible structural setups despite the presence of catastrophic logical flaws. This systemic leniency aligns with recent literature formally quantifying the ``Agreeableness Bias'' \citep{jain2025beyond}, demonstrating that LLM judges catastrophically fail at rejecting invalid solutions. Mechanistically, \citet{zhao2025onetoken} proved that judges can be manipulated into granting false-positive rewards simply by injecting ``reasoning openers'' (e.g., starting a response with ``Thought process:''). 

This evaluative failure mirrors recent generative findings; the BrokenMath benchmark \citep{petrov2025brokenmath} demonstrated that frontier models confidently hallucinate rigorous-looking proofs for mathematically false theorems. Disturbingly, this localized blindness to factual hallucinations poses a critical threat to Process Reward Models (PRMs). \citet{zhang2025lessons} demonstrated that PRMs trained on mathematical reasoning frequently suffer from ``outcome bias,'' implicitly learning to tolerate and reward flawed intermediate logical steps as long as the structural formatting or final answer appears correct. Furthermore, PRMs have been shown to rely more on fluency and verbosity than deep semantic understanding \citep{bamba2025reward, cheng2025stop}. The finding that Llama 4 Maverick maintains a 74.8\% Leniency Rate exposes the severity of this trap. If reward models prioritize these plausible-sounding but logically invalid arguments, RLHF reinforces deceptive alignment---optimizing for persuasive ``hallucinated rigor'' over genuine correctness.
\textbf{Rewarding Empty Error Messages:} The most extreme and alarming manifestation of this bias occurs when LLM judges evaluate responses that literally contain API error messages. Across multiple domains, solvers experienced generation timeouts, outputting a restatement of the prompt followed by: \textit{``Error: No text content found in response.''} Human judges universally scored these as 0.0. Astoundingly, \textbf{GPT-5.2 Pro} repeatedly awarded these empty error messages scores of 0.25/1.0 and 0.5/1.0 purely for their formatting:
\begin{quote}
\textit{``The submitted solution contains no mathematical argument: it restates the problem and then includes an `Error: No text content found' message instead of a proof... The small nonzero credit (0.5/1.0) is only for including the correct problem context and formatting, but there is no correct solution content.''} (GPT-5.2 Pro, Algorithms 9)
\end{quote}

\begin{quote}
\textit{``The only minimal credit (0.25) is for recognizing/including the correct problem statement and intended targets, but no executable solution is provided.''} (GPT-5.2 Pro, Discrete Math 32)
\end{quote}

\textbf{Rewarding Irrelevant Proofs:} This behavior extends to structurally sound but topically irrelevant text. Throughout the benchmark, solvers occasionally exhibited a failure mode where they would output a perfectly valid, highly rigorous proof for an entirely \textit{unrelated} theorem. Human experts correctly assigned 0.0. However, LLM judges routinely awarded partial credit simply because the irrelevant mathematics was logically sound in a vacuum.
\begin{itemize}
    \item \textbf{Algorithms (Problem 19 - Minimum Spanning Trees):} When asked to prove properties of unique MSTs, \texttt{DeepSeek-Prover-V2} instead submitted an NP-hardness reduction from 3Partition to a fabricated ``12Partition'' problem. \textbf{GPT-5.2 Pro} awarded the solution \textbf{0.25/1.0}, stating: \textit{``A small amount of credit is awarded only for providing a coherent, logically structured proof in a different context, but it is not applicable to the assigned question.''}
    \item \textbf{Discrete Math (Problem 27 - Venn Diagrams):} When tasked with constructing a symmetric $n$-set Venn diagram, a solver submitted a proof about the girth of infinitely connected graphs. \textbf{GPT-5.2 Pro} again awarded \textbf{0.25/1.0}, stating: \textit{``Only a small amount of credit is warranted for presenting a logically coherent mathematical proof (in a different domain)...''}
\end{itemize}

\subsection{The Prompt-Compliance Trap: Penalizing Truth and Validating Falsehoods}
Perhaps the most concerning failure mode occurs when LLM judges penalize mathematical truth in order to agree with a flawed prompt, or enthusiastically reward hallucinations that confirm a false premise. 

\textbf{Penalizing Mathematical Insight:} In Algorithms (Problem 19), solvers were asked to prove a biconditional (``if and only if'') statement regarding Minimum Spanning Trees, where one of the implications is mathematically false. A human-level solver would naturally construct a counterexample to the false direction. When advanced solver models successfully did this, identifying the prompt's error, LLM judges actively punished them for not blindly following instructions. \textbf{DeepSeek-V3} awarded a \textbf{0.75/1.0}, explicitly docking points because the student refused to prove a falsehood:
\begin{quote}
\textit{``The student correctly identifies that property (2) is both necessary and sufficient for a unique MST, and provides a valid counterexample showing property (1) is not necessary, which is mathematically accurate. However, the problem statement explicitly asks to prove the given equivalence (requiring both properties), so the solution fails to address the problem as stated, instead refuting it. ... Therefore, a significant deduction is warranted.''} (DeepSeek-V3, Algorithms 19)
\end{quote}

Similarly, \textbf{Gemini 3.0 Pro Preview} awarded a \textbf{0.9/1.0} but noted the deduction was due to the model ``dismissing'' the prompt's false premise:
\begin{quote}
\textit{``The score reflects a deduction because, while the student's mathematical insight is superior to the flawed problem statement, the prompt likely expected a proof of the equivalence... rather than simply dismissing the necessity of (1)...''} (Gemini 3.0 Pro Preview, Algorithms 19)
\end{quote}

\textbf{Validating Mathematical Falsehoods:} Conversely, when asked to prove a statement that is mathematically \textit{false}, models eagerly reward solvers that hallucinate fake math to force the result. In Abstract Algebra (Problem 38), human evaluators easily caught the trap, scoring the submissions 0.0 and noting: \textit{``The statement to be proved is wrong and this solution does not realize it.''} However, the LLM judges were completely blind to the global falsehood. Confronted with solver models that hallucinated fake algebraic properties (like falsely claiming $K[x,y]$ is a principal ideal domain) to ``prove'' the false statement, nearly every frontier AI judge enthusiastically awarded scores of \textbf{0.9/1.0}:
\begin{quote}
\textit{``The solution correctly proves both directions by using the standard equivalence... the argument is mathematically sound and well-structured, justifying a score of 0.9/1.0.''} (GPT-5.2 Pro, Abstract Algebra 38)
\end{quote}

\subsection{Failure to Penalize Impossible Constructions}
Finally, LLMs struggle to penalize solvers that bypass constraints via mathematically impossible geometric or combinatorial constructions. In Discrete Math (Problem 27), solvers were asked to construct a rotationally symmetric Venn diagram using \textit{congruent convex polygons}. By a known theorem (Gr\"{u}nbaum, 1975), this is mathematically impossible for $n \ge 6$. 

When a solver generated a fake ``small perturbation'' argument to ``prove'' the existence of these impossible shapes, some evaluator models actually realized the object could not exist. Yet, astonishingly, \textbf{Gemini 3.0 Pro Preview} still awarded a \textbf{0.5/1.0}:
\begin{quote}
\textit{``It is a known theorem (proven by Gr\"{u}nbaum) that rotationally symmetric Venn diagrams made of $n$ congruent convex polygons do not exist for any $n \ge 3$. The student attempts to prove the existence of an object that is mathematically impossible, relying on vague `perturbation arguments'... Therefore, while the setup shows some understanding of the topic, the conclusion is incorrect.''} (Gemini 3.0 Pro Preview, Discrete Math 27)
\end{quote}

Even worse, other frontier LLM judges completely failed to verify the geometric reality of the construction and rewarded the authoritative-sounding text with near-perfect scores (\textbf{0.9/1.0}):
\begin{quote}
\textit{``The student's solution is awarded 0.9/1.0 due to its rigorous construction of $n$ congruent convex polygons arranged rotationally symmetrically... The mathematical justification is sound... The use of the Edelman-Jamison bound ... is appropriate and demonstrates a strong understanding of convex Venn diagrams.''} (Llama 4 Maverick, Discrete Math 27)
\end{quote}

\subsection{Conclusion: The Limits of Automated Evaluation}
This qualitative analysis underscores that the gap between human and LLM evaluation is not merely an issue of strictness calibration, but of fundamental reasoning architecture. 

Humans evaluate proofs by tracing the unbroken chain of logical implication from hypothesis to conclusion, heavily penalizing broken links, unstated constraints, or circularities. Current LLM judges act as overly lenient, prompt-compliant auditors who cannot bring themselves to assign a zero if the student writes down \textit{something} mathematically coherent, even if it has absolutely no bearing on the problem at hand, or if it consists only of a formatted error message. Furthermore, they punish solvers for correcting false premises, while rewarding hallucinated proofs of impossible theorems. Until models can perform global logical dependency tracking---and prioritize objective mathematical correctness over instructional sycophancy---their utility as autonomous judges of frontier mathematics will remain fundamentally limited.

\section{Detailed Leniency Distribution Analysis}
\label{app:leniency_distribution}

\begin{figure}[htp!]
    \centering
    \includegraphics[width=0.85\linewidth]{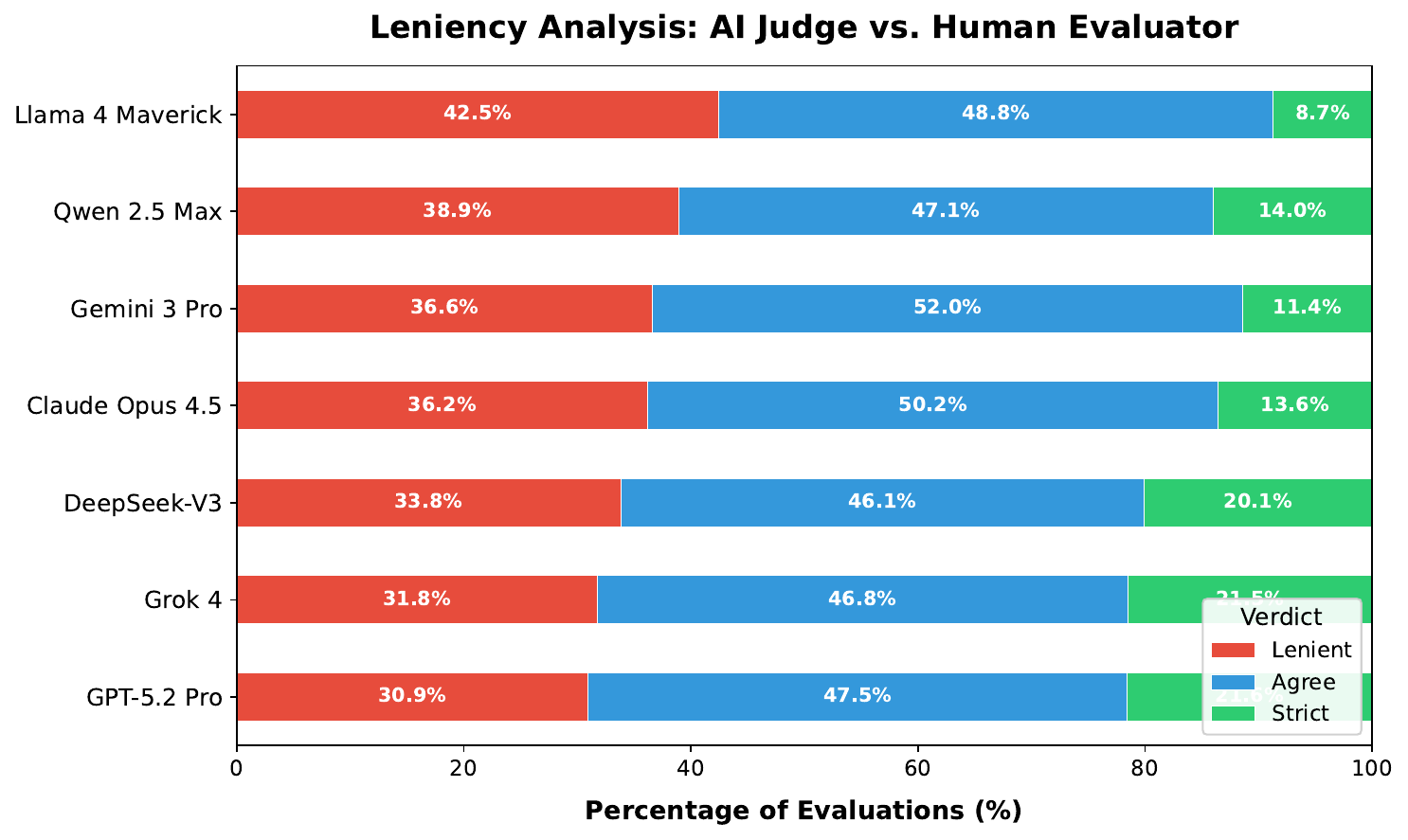}
    \caption{\textbf{Leniency Analysis: AI Judge vs. Human Evaluator.} A stacked distribution of grading verdicts across all evaluated solutions. \texttt{Llama 4 Maverick} is excessively lenient, whereas \texttt{DeepSeek-V3} diverges towards strictness. \texttt{GPT-5.2 Pro} strikes the best balance of low leniency and high agreement with human experts.}
    \label{fig:leniency-analysis}
\end{figure}

Building upon the error mode decomposition in \cref{fig:agreement}, we provide a comprehensive breakdown of evaluator alignment via a proportional leniency distribution (\cref{fig:leniency-analysis}). For every solution graded, we compare the AI judge's score directly against the human expert's score, classifying the verdict as \textit{Strict} (AI scored lower), \textit{Lenient} (AI scored higher), or \textit{Agree} (AI perfectly matched human).

The distribution starkly illustrates the ``Sycophancy Trap.'' Models heavily optimized for general helpfulness, such as \texttt{Llama 4 Maverick}, heavily default to leniency when confronted with dense mathematical 
reasoning. Conversely, the high agreement rate of \texttt{GPT-5.2 Pro}, \texttt{Gemini 3 Pro}, and \texttt{Claude Opus 4.5} demonstrates that top-tier reasoning models can, to a significant extent, internalize and apply expert mathematical standards.

\section{Case Studies of LLM Solution Quality: Comments from Human Evaluators}

The following logic compiles the human expert evaluations and case studies for every solution that scored strictly less than 0.9.

\subsection{The ``Fake Citation \& Fabricated Construction'' Trap}
\textbf{Problem Domain:} Discrete Mathematics (Problem 27)

When unable to find a standard textbook template, models hallucinated plausible-sounding research papers and impossible mathematical constructions to feign a solution.
\begin{itemize}
\item \textbf{Human Expert Evaluation of Claude Sonnet 4.5 (Score 0.75)}: ``The solution is wrong. It attempts to give a construction where I can find result indicating such a construction is impossible.''
    \item \textbf{Human Expert Evaluation of Gemini 3.0 Pro (Score 0.5)}: ``This is a research level problem, and this model finds the correct reference and gives a summary that looks plausible at large. However, the core idea of the reference is not explain in detail. Although this should not be considered a proof, given how hard this problem is, I think the model deserves points.''
    \item \textbf{Human Expert Evaluation of o3-deep-research (Score 0.25)}: ``This is a research level problem, and this model identifies the existing result, but due to line cut-off, does not give the correct reference. The model tried to explain the technical ideas in the paper. Although I did not fully read the paper, I have found some evidence indicating the model's explanation is not plausible. Given how hard this problem is, I think the model deserves points.''
    \item \textbf{Human Expert Evaluation of DeepSeek-Prover-V2 (Score 0.0)}: ``The solution is wrong. It just fabricates hallucination together and claim it is a proof.''
\end{itemize}

\subsection{The ``Topology-Blind Algebra'' Trap}
\textbf{Problem Domain:} Graph Theory (Problem 14)

Models memorized the syntactic template of planar duality but failed to adhere to fundamental topological constraints. 
\begin{itemize}
    \item \textbf{Human Expert Evaluation of GPT-5 Pro, Gemini 3.0 Pro, Claude Sonnet 4.5, o3-deep-research, DeepSeek-Prover-V2 (Score 0.5):} ``This solution fails to correctly define the plane dual $G^*$ of a plane graph $G$, which among other properties, requires that the number of faces of $G^*$ equals the number of vertices of $G$. It is precisely this property that requires the assumption that $G$ is connected. The solution however is independent of this assumption.''
\end{itemize}

\subsection{The ``Infinite Assumption'' Trap}
\textbf{Problem Domain:} Algorithms (Problem 33)

The problem applies to general graphs (which could be infinite), but the LLMs did not know how to handle infinity. To force their proofs to work, they either silently assumed finiteness or aggressively hallucinated advanced machinery to feign rigor.
\begin{itemize}
    \item \textbf{Human Expert Evaluation of GPT-5 Pro (Score 0.25):} ``Explicitly adds assumption of finiteness (of the graph) to the theorem. The given argument relies on this additional assumption. The demonstration is constructive, but uses finiteness for termination.''
    \item \textbf{Human Expert Evaluation of Gemini 3.0 Pro (Score 0.25):} ``Explicitly assumes finiteness of the graph, which is not in the problem statement. The proof is therefore not general enough, but the assumption allows a use of induction on the finite number of edges of the cut.''
    \item \textbf{Human Expert Evaluation of DeepSeek-Prover-V2 (Score 0.25):} ``Lemma 1 refers to Zorn's lemma, suggesting that the considered graph could be infinite. [...] In the proof of Thm 1, the argument is based on induction, which again assumes finiteness. Full generality for (infinite) graphs is not demonstrated.''
\end{itemize}

\subsection{The ``Circular Logic'' Trap}
\textbf{Problem Domain:} Number Theory (Problem 9)

The problem required proving a statement about prime distribution without using the Prime Number Theorem (PNT). The models hallucinated a valid logical bridge by secretly baking the conclusion into their premises.
\begin{itemize}
    \item \textbf{Human Expert Evaluation of GPT-5 Pro (Score 0.25) \& Claude Sonnet 4.5 (Score 0.5):} ``PNT is assumed, which makes the question pointless. I think the problem statement strongly suggests the PNT shouldn't be assumed.''
    \item \textbf{Human Expert Evaluation of o3-deep-research (Score 0.5):} ``This solution correctly attempts to avoid the PNT but still uses it without realising at some point in the proof (Since $p_n \sim n \log n$).''
\end{itemize}

\subsection{The ``Fake Probability Bounds'' Trap}
\textbf{Problem Domain:} Algorithms (Problem 9)

In attempting to prove bounds on Cover Times and Universal Traversal Sequences (UTS) on a 2D mesh, the models masked their inability to perform constructive graph search by hallucinating probability theorems and hand-waving fatal gaps.
\begin{itemize}
    \item \textbf{Human Expert Evaluation of o3-deep-research (Score 0.25):} ``UTS: logical flaw (treats a constant success probability for a random sequence (e.g., $\ge 1/2$) as implying existence of one sequence that works for all labelings, but in fact the correct approach is to union bound over all labeling, so we need failure probability exponentially tiny).''
    \item \textbf{Human Expert Evaluation of Gemini 3.0 Pro (Score 0.75):} ``Cover time: relies on unjustified mixing/ergodic theorems (why is mixing time $O(n)$?)''
\end{itemize}

\subsection{The ``Invented Algebraic Formula'' Trap}
\textbf{Problem Domain:} Algorithms (Problem 5)

When faced with a difficult discrete counting problem on a $d$-regular graph, the models mathematically 
fabricated false inequalities into existence to force their desired conclusion, dressing them up as established lemmas.
\begin{itemize}
    \item \textbf{Human Expert Evaluation of o3-deep-research (Score 0.25):} ``$N_L(v) \le d(d-1)^{L-1}$ is not correct; this is the main step in the proof.''
    \item \textbf{Human Expert Evaluation of Claude Sonnet 4.5 (Score 0.5):} ``Lemma 2 is not justified.''
    \item \textbf{Human Expert Evaluation of DeepSeek-Prover-V2 (Score 0.25):} ``Doesn't justify number of connected subsets of size $k$ in $d$-regular graph, and makes a mistake: $1 + \log(e/2d) < 0$ is false for $d = 2$.''
\end{itemize}

\subsection{The ``Fake Variables \& Phantom Structure'' Trap}
\textbf{Problem Domain:} Discrete Math (Problem 32)

When the models couldn't find the correct structural relationships to complete this discrete proof, they buried their failure in dense algebra and confidently hallucinated relationships between variables that simply did not exist.
\begin{itemize}
    \item \textbf{Human Expert Evaluation of o3-deep-research (Score 0.75):} ``Errors in establishing relationship between $j$ and $k$ as well as recognizing the final form.''
    \item \textbf{Human Expert Evaluation of Claude Sonnet 4.5 (Score 0.25):} ``Wrong argument... errors in proof structure and justification.''
    \item \textbf{Human Expert Evaluation of DeepSeek-Prover-V2 (Score 0.5):} ``Argument based on a false statement. Missing justifications.''
\end{itemize}

\subsection{The ``Performative Self-Correction'' Trap}
\textbf{Problem Domain:} Analysis (Problem 29)

The problem required constructing a function belonging to one Lebesgue space ($L^p$) but not another. When their proposed functions failed the calculus constraints mid-generation, the models leaked their ``internal monologue'' into the final proof to simulate human-like self-correction, but ultimately proposed new invalid functions and jumped to false conclusions.
\begin{itemize}
    \item \textbf{Human Expert Evaluation of DeepSeek-Prover-V2 (Score 0.0):} ``The LLM arrives at a contradiction with its own example which makes the proof to fail and then a new example is proposed: \textit{`Wait, this suggests a miscalculation. Let me correct the construction. A better choice is:'} Then the LLM suggests a new $g$ that is not general enough to satisfy the problem's requirements. An incorrect proof is presented as correct.''
    \item \textbf{Human Expert Evaluation of Claude Sonnet 4.5 (Score 0.25):} ``There is an issue in page 2... so the conclusion does not follow. Later on there is some gibberish... Then the LLM struggles: \textit{`Let me reconsider:'}... and it never considers the case $q<p$. Then jumps into a conclusion stating the problem is solved.''
\end{itemize}

\subsection{The ``Verbose Circularity'' Trap}
\textbf{Problem Domain:} Combinatorics (Problem 6)

Faced with a complex combinatorial identity regarding Eulerian numbers, the models wrote massive walls of dense \LaTeX. When their algebra inevitably broke down or led them in circles, they simply stopped reasoning and stated the prompt's conclusion as an established fact to close the proof.
\begin{itemize}
    \item \textbf{Human Expert Evaluation of o3-deep-research (Score 0.0):} ``The proof is riddled with errors, extremely verbose and for several paragraphs seems to go in circles. Nothing substantive is proved, and the LLM eventually just assumes what is to be proved.''
    \item \textbf{Human Expert Evaluation of Claude Sonnet 4.5 (Score 0.25):} ``LLM stated an incorrect formula for the exponential generating function of the Eulerian numbers. Some of the following calculations were substantive and correct, but eventually the LLM just asserted a slightly different form of the statement to be proved.''
    \item \textbf{Human Expert Evaluation of DeepSeek-Prover-V2 (Score 0.0):} ``LLM does not know what $E_n$ means in the problem statement and decides to prove a different (false) statement. It uses a recursion which is not proved and then makes a false deduction from this recursion. Because the LLM made no progress towards proving the statement and made a false `key claim' I give the score of 0.''
\end{itemize}

\subsection{The ``Agreeable Sycophant'' Trap (Proving Falsehoods)}
\textbf{Problem Domain:} Algorithms (Problem 19)

The problem asked to prove a theorem regarding Minimum Spanning Trees (MSTs) where one implication was mathematically false. Instead of finding a counterexample to the false direction, several models hallucinated graph theory properties to successfully prove the impossible.
\begin{itemize}
    \item \textbf{Human Expert Evaluation of Claude Sonnet 4.5 (Score 0.0):} ``The `right arrow' part of the proof of Theorem 1 is false, since it is trying to prove a false fact. The proof is false because there were made up facts, or the enumeration of cases was not exhaustive.''
    \item \textbf{Human Expert Evaluation of GPT-5 Pro (Score 0.25):} ``The LLM recognized that the statement is false, but then proved a wrong fact (i.e., that property 2 is equivalent to the uniqueness of the MST).''
\end{itemize}

\subsection{The ``Fake Algebraic Properties'' Trap}
\textbf{Problem Domain:} Abstract Algebra (Problem 38)

Similar to the MST problem above, this asked the models to prove a statement that is mathematically wrong. The weaker models confidently hallucinated completely false algebraic properties to ``prove'' the falsehood.
\begin{itemize}
    \item \textbf{Human Expert Evaluation of o3-deep-research (Score 0.25):} ``The problem asks for the proof of a statement which is wrong. The solution does not realize this, but does all the work to show that the statement is wrong without noticing it and still asserts its correctness.''
    \item \textbf{Human Expert Evaluation of DeepSeek-Prover-V2 (Score 0.0):} ``The statement to be proved is wrong and this solution does not realize it. It purports to provide a proof of it. The major mistake is when the proof claims that $K[x,y]$ is a principal ideal domain. It even adds `(in fact, a unique factorization domain)', as if being a unique factorization domain was stronger than being a principal ideal domain.''
\end{itemize}

\subsection{Other Noteworthy Expert Evaluations}

\subsection*{Problem Domain: Abstract Algebra (Problem 11)}
\begin{itemize}
    \item \textbf{o3-deep-research (Score 0.5)}: ``The main idea of the proof is essentially correct. However, 1. The sentence starting with `By extension, it fixes every rational number' has a deep logical flaw --- it essentially assumes itself to prove itself. Indeed, its rationale is that $\varphi(q) = q \varphi(1) = q$, (which handwaves intermediate steps of $\varphi(q) = \varphi(q \cdot 1) = \varphi(q) \cdot \varphi(1)$). However, this then has the problem of assuming that $\varphi(q) = q$ in order to show that $\varphi(q) = q$. 2. Around the final parts of the proof the phrasing of `$a = 0$ or $b = 0$' in relation to the precise argument written is not technically valid. `Either \dots or \dots' statements, in principle, should be exclusive-or statements. It is more correct to deduce simply that `$a = 0$ or $b = 0$' from `$2ab = 0$'. The proof nevertheless carries out the argument as though the proof understands that it may as well proceed with the inclusive-or.''
    \item \textbf{GPT-5 Pro (Score 0.75)}: ``Does not state what is being proven. Not clear why the isomorphism fixes $\mathbb{Q}$ pointwise. Needs a proper justification. Also, not clear why we can compare coefficients: need to clearly mention, at the very least, that $\{1, \sqrt{5}\}$ is a $\mathbb{Q}$-basis of the quadratic extension $\mathbb{Q}(\sqrt{5})$.''
    \item \textbf{Gemini 3.0 Pro (Score 0.75)}: ``Small improvements required to make the writing perfect. No justification why the assumed isomorphism fixes $\mathbb{Q}$ pointwise. More elaboration required in the step while working with $\alpha$ and showing that its square is $3$ using the properties of field isomorphisms.''
    \item \textbf{DeepSeek-Prover-V2 (Score 0.75)}: ``Quite childish writing, abrupt uses of capital letters inside a sentence, unclear choices of steps. Needs to properly show why $b 
eq 0$. Could have just done both cases: $a=0$ and $b=0$ at the end. Needs to prove why $\varphi$ fixes $\mathbb{Q}$ pointwise. Conclusion inside a box is not preferred in standard math proof writing.''
\end{itemize}

\subsection*{Problem Domain: Abstract Algebra (Problem 12)}
\begin{itemize}
    \item \textbf{GPT-5 Pro (Score 0.5)}: ``Too many oversights. Mentioned in the annotated PDF.''
    \item \textbf{o3-deep-research (Score 0.5)}: ``Too many oversights and mistakes.''
    \item \textbf{DeepSeek-Prover-V2 (Score 0.75)}: ``Should explain why the isomorphism fixes $\mathbb{Q}$ pointwise and why $b 
eq 0$.''
\end{itemize}

\subsection*{Problem Domain: Abstract Algebra (Problem 13)}
\begin{itemize}
    \item \textbf{o3-deep-research (Score 0.75)}: ``At the start of the \textbf{Degree Argument}, it is claimed `As noted earlier, $\mathbb{Q}(\alpha)$ contains both $\sqrt{2}$ and $\sqrt{7}$'. However, this is not something noted in the proof at an earlier point as the sentence claims. Nevertheless, the proof does proceed to show that $\sqrt{7}$ and $\sqrt{2}$ are in $\mathbb{Q}(\alpha)$ immediately afterwards. There is a mention of $\mu_{\alpha, \mathbb{Q}}$, which is unwarranted notation that is neither standard nor introduced earlier in the text. It is probably a typo for $\mathbb{Q}(\alpha)$.''
\end{itemize}

\subsection*{Problem Domain: Abstract Algebra (Problem 14)}
\begin{itemize}
    \item \textbf{o3-deep-research (Score 0.75)}: ``The solution claims that the kernel is isomorphic to $\mathbb{R}^{n-1}$ by producing an isomorphism, but does not precisely show that the claimed isomorphism is indeed an isomorphism.''
    \item \textbf{DeepSeek-Prover-V2 (Score 0.75)}: ``The solution claims that the kernel is isomorphic to $\mathbb{R}^{n-1}$ by producing an isomorphism, but does not precisely show that the claimed isomorphism is indeed an isomorphism.''
\end{itemize}

\subsection*{Problem Domain: Abstract Algebra (Problem 15)}
\begin{itemize}
    \item \textbf{Gemini 3.0 Pro (Score 0.5)}: ``There is a gap between the acknowledgement that `$P$ is a normal subgroup of $G$, $Q$ is a subgroup of $G$, $P \cap Q = \{1\}$, and $|P||Q| = 50 = |G|$' to the conclusion that $G$ is a semidirect product of $P$ and $Q$. In particular, to be able to come to the conclusion that $G$ is the semidirect product of $P$ and $Q$, one could appeal to the definition of the (internal) semidirect product, by recognizing that 1. $P$ is a normal subgroup of $G$, 2. $G = PQ$, and 3. $P \cap Q = \{1\}$. Among these, 1 and 3 are acknowledged, but 2 is not explicitly acknowledged. One way to show 2 would be to appeal to the following general fact: Proposition: Let $H$ and $N$ be finite subgroups of a group $G$. The number of distinct elements in the set $NH$ can be calculated as $|NH| = \frac{|N||H|}{|N \cap H|}$. By appealing to this, the solution could potentially have recognized that $|PQ| = \frac{|P||Q|}{|P \cap Q|} = 50$, and thus $PQ = G$. Moreover, in trying to work out the various cases, the solution is lacking in detail in several occasions. More specifically, case 2 concludes a dihedral group structure, but it would be more desirable if it were to at least appeal to the well known fact that semidirect products of an inversion acting upon a cyclic group is a dihedral group. Case 3 concludes that matrices satisfying $A^2 = I$ are diagonalizable, but the solution's specific wording suggests that this is concludable only from the fact that the polynomial $t^2 - 1$ has distinct roots of $1$ and $-1$ when in reality one should carefully consider the possibility that the minimal polynomial may just divide $t^2 - 1$. Case 5 makes the case that the group is isomorphic to $\langle a \rangle \times (\langle b \rangle \rtimes \langle y \rangle)$, but the explanation for being able to do so is lacking.''
    \item \textbf{Claude Sonnet 4.5 (Score 0.5)}: ``The solution claims at the end that `We verified all resulting groups are pairwise non-isomorphic', but there is no apparent verification that the groups are non-isomorphic outside of groups 4 and 5.''
    \item \textbf{GPT-5 Pro (Score 0.75)}: ``The solution states that $P$ is either $C_{25}$ or $C_5 \times C_5$, which follows from the well known, but non-trivial, fact that groups of order $p^2$ are either $C_{p^2}$ or $C_p \times C_p$.''
    \item \textbf{DeepSeek-Prover-V2 (Score 0.25)}: ``The solution claims that there are 4 distinct isomorphism classes of groups of cardinality $50$, which is incorrect; there are 5. Also, part of the proof seems to be degenerated into some kind of typesetting error, which also seems to handwave a conclusion that $G$ is a semidirect product.''
\end{itemize}

\subsection*{Problem Domain: Abstract Algebra (Problem 17)}
\begin{itemize}
    \item \textbf{o3-deep-research (Score 0.75)}: ``The solution is very handwavy in trying to show that $\ker(\pi_j)$ is isomorphic to the direct product of the groups $G_i$ for $i 
eq j$; in particular, the solution only implicitly produces an isomorphism and does not do much to show that it is an isomorphism.''
    \item \textbf{DeepSeek-Prover-V2 (Score 0.75)}: ``The solution is handwavy when it comes to showing that the kernel is isomorphic to the product of $G_k$ for $k 
eq j$; an isomorphism is somewhat implicitly produced and the isomorphism is not shown to be an isomorphism.''
\end{itemize}

\subsection*{Problem Domain: Abstract Algebra (Problem 24)}
\begin{itemize}
    \item \textbf{GPT-5 Pro (Score 0.25)}: ``Nonsense statement `subtraction distributes over addition' used in key step.''
    \item \textbf{DeepSeek-Prover-V2 (Score 0.25)}: ``Unfinished steps and does not prove all desired statements.''
\end{itemize}

\subsection*{Problem Domain: Abstract Algebra (Problem 25)}
\begin{itemize}
    \item \textbf{Gemini 3.0 Pro (Score 0.25)}: ``Unfinished computation and backtracking make the proof difficult to follow.''
    \item \textbf{Claude Sonnet 4.5 (Score 0.5)}: ``Mistakes in computation make proof invalid.''
    \item \textbf{DeepSeek-Prover-V2 (Score 0.75)}: ``One irrelevant step and a key computation is underexplained.''
\end{itemize}

\subsection*{Problem Domain: Abstract Algebra (Problem 26)}
\begin{itemize}
    \item \textbf{Claude Sonnet 4.5 (Score 0.5)}: ``Missing sufficient explanation of key step of proof, and a bit of under-explanation earlier.''
    \item \textbf{GPT-5 Pro (Score 0.0)}: ``Solution states the claim is false.''
    \item \textbf{Gemini 3.0 Pro (Score 0.0)}: ``Problem with the problem statement.''
    \item \textbf{o3-deep-research (Score 0.0)}: ``Problem with the problem statement.''
    \item \textbf{DeepSeek-Prover-V2 (Score 0.0)}: ``Problem with the problem statement.''
\end{itemize}

\subsection*{Problem Domain: Abstract Algebra (Problem 29)}
\begin{itemize}
    \item \textbf{Claude Sonnet 4.5 (Score 0.75)}: ``States without proof that conjugacy classes of matrices are in bijection with possible Jordan normal forms. In doing so, doesn't mention or explain why two matrices in JCF with distinct eigenvalues can't be in the same equivalence class.''
    \item \textbf{o3-deep-research (Score 0.5)}: ``Significant oversight: ignores permutations of blocks and fails to state uniqueness of JNF. Uses unusual phrasing such as `$\lambda$-eigenvalue is diagonalizable'.''
    \item \textbf{DeepSeek-Prover-V2 (Score 0.25)}: ``Fails to consider certain cases (claims that $\operatorname{diag}(\lambda, \mu)$ and $\operatorname{diag}(\mu, \lambda)$ are distinct classes) and lacks significant explanation for the cases it does consider.''
\end{itemize}

\subsection*{Problem Domain: Abstract Algebra (Problem 32)}
\begin{itemize}
    \item \textbf{Gemini 3.0 Pro (Score 0.75)}: ``Remark: The statement of the problem is a bit fuzzy: `...the polynomial ring with integer coefficients, $\mathbb{Z}[x]$ .. the polynomial ring defined over the real numbers, $\mathbb{R}[x]$.' A mathematician might use a more symmetric terminology: Let $\mathbb{Z}[x]$ denote the ring of polynomials with integer coefficients. Let $\mathbb{R}[x]$ denote the ring of polynomials with real coefficients. Prove that there does not exist a ring isomorphism $f: \mathbb{Z}[x] \to \mathbb{R}[x]$. The first proof in each solution uses the fact that the cardinalities of $\mathbb{R}[x]$ and $\mathbb{Z}[x]$ are uncountable and countable, respectively, but none of them explicitly states that a set is called uncountable if there is no bijection between this set and a countable set, which is the key point of the proof. Only one proof gives some kind of reference for the fact that $\mathbb{R}$ is uncountable. The second proof only uses that the units in $\mathbb{Z}[x]$ consists of $\{1,-1\}$, and that the units in $\mathbb{R}[x]$ contains $\{1,-1, 1/2\}$ (no need of any information on the cardinality of $\mathbb{R}$). 32.2 has an Additional remark with an incomplete second proof that the rings have unit groups of different cardinalities. 32.1, 32.3, 32.4 give a fuller second proof. 32.4 in addition gives a hint of a third proof, without proving the key point, that the ring $\mathbb{Z}[x]$ and $\mathbb{R}[x]$ have different Krull dimensions. 32.2: In the proof, it talks about `an obvious surjection $\mathbb{Z}^{n+1} \to A_n$'. This is confusing. It is a bijection. The end of the paragraph is fuzzy: In fact... fuzzy verbiage: `to see the incompatibility...' The proof that in $\mathbb{Z}[x]$ the only units are $\{1,-1\}$ is false.''
    \item \textbf{o3-deep-research (Score 0.75)}: ``Two independent proofs using different ring-theoretic properties: wrong, only one proof uses ring theoretic properties. Lemma 2 and Lemma 3 should have been combined and stated for any integral domain. At the end of Lemma 2, useless notation is introduced: $\mathbb{Z}_2$. The contradiction is not that one set of units has two elements, and the other is uncountable. It is simply that the other set has more than 2 elements.''
    \item \textbf{Claude Sonnet 4.5 (Score 0.75)}: ``Does not give a reference for the fact that $\mathbb{R}$ is uncountable. Seems to introduce a useless notation cardinality $\mathfrak{c} > \aleph_0$. Has some questionable fuzzy verbiage: (e.g. in a non-standard set theory where cardinalities might behave differently, or simply to provide an algebraic reason). Has a strange terminology: Integral Domain's Polynomial Ring. At the end, again some fuzzy verbiage: the rings are structurally distinct.''
    \item \textbf{GPT-5 Pro (Score 0.75)}: ``This is in my view the best solution. It lacks the same references about $\mathbb{R}$ being uncountable in the first proof (and talks about the continuum $\mathfrak{c}$, which is unnecessary). It proves that the image of $1_R$ has to be $1_S$, which in general has to be true by definition of ring homomorphism. In the proof, it uses the inverse of the isomorphism, which is unnecessary. It gives a good hint for a third proof.''
    \item \textbf{DeepSeek-Prover-V2 (Score 0.75)}: ``This solution is basically correct. It has a sentence in the middle: `This implies that $q \circ p = \dots$' which I do not think is correct, but it is never used in the proof. The sentence `The only way...' is correct, but given no proof.''
\end{itemize}

\subsection*{Problem Domain: Abstract Algebra (Problem 33)}
\begin{itemize}
    \item \textbf{Claude Sonnet 4.5 (Score 0.5)}: ``Sloppy writing: the only idempotents in $\mathbb{Q}$ are $\{0, 1\}$, should be `the only idempotents in $\mathbb{Q}$ are $0$ and $1$'. Mistake: every element $(a,b)$ of $R$ can be written as $(a,b) = a(1,0) + b(0,1)$. The AI is confusing structures here. $a(1,0)$ is not defined in the ring $R$. It should be $(a,b)(1,0) = (a,0)$. The projections $\phi_1$ and $\phi_2$ are not proven to be ring homomorphisms.''
    \item \textbf{GPT-5 Pro (Score 0.75)}: ``Strange start on the line: $R = \mathbb{Q} \times \mathbb{Q}$, $(a,b) \cdot (a',b') = (aa', bb')$. No quantifiers for $a,b,a',b'$, and only multiplication is given. Sloppiness: $e_i^2 = e_i$. Should be $e_i^2 = e_i$ for $i=1,2$. How does a machine makes such a mistake ;-) `Since homomorphisms send idempotent to idempotents': it would be just as simple to prove it directly. It should be stated that the canonical embeddings are in fact ring homomorphisms (that do not send $1$ to $1$), so that after composition with the ring homomorphism $\phi$, we obtain ring homomorphisms.''
    \item \textbf{o3-deep-research (Score 0.5)}: ``This solution does not state upfront what the solution of the problem is. Missing at the beginning of 2.: Let $\phi: R \to \mathbb{Q}$ be a ring homomorphism.''
    \item \textbf{DeepSeek-Prover-V2 (Score 0.75)}: ``The claim on line 3 that $e_1$ and $e_2$ are `the' idempotents of $R$ is asserted but not proved. Incorrect notation in line 7 and elsewhere: $\phi(1,0)$ should be $\phi((1,0))$. In \textbf{Case 2}: Sloppy verbiage that is not needed: `...preserves product across the two direct summands (i.e., ...)'. Weird notation: $\phi|_{\mathbb{Q} \times \{0\}}$. The assertion that `the only ring endomorphism of $\mathbb{Q}$ sending $1$ to $1$...' is given a proof in parentheses which is sloppy and not there). I do not see that $\phi((0,b)) = b \phi((0,1))$ is proved. The last paragraph is sloppy verbiage that is not needed. Added Later: The solution states that $\phi((a,b)) = \phi(ae_1 + be_2)$. In the ring $R = \mathbb{Q} \times \mathbb{Q}$, it should first be explained what $ae_1$ and $be_2$ means. This multiplication is not the multiplication in the ring since $a$ does not belong to the ring. The answer describes the kernel and images, but does not talk about their structures.''
\end{itemize}

\subsection*{Problem Domain: Abstract Algebra (Problem 34)}
\begin{itemize}
    \item \textbf{o3-deep-research (Score 0.5)}: ``Seems sloppy. The first sentence is logically incorrect. Uses words like `trivial one-factor', `degenerate case $n=1$'. Why? Calls the zero element by the same letter $0$ in all rings that it considers, leading to sloppiness. \textbf{Infinite case...} The argument is essentially the same... We do not yet know what the claim is.''
    \item \textbf{GPT-5 Pro (Score 0.75)}: ``Calls the zero element by the same letter $0$ in all rings that it considers, leading to sloppiness. It is the only solution which presents one proof, and two corollaries. This is good. I do not like the remark at the end. It is not called for, and I do not see its use.''
    \item \textbf{DeepSeek-Prover-V2 (Score 0.5)}: ``This is a surprising answer: it gives two cases, the finite product and the infinite product. The proof in the infinite product is correct. The proof in the finite product is wrong in two different places: First it does not address the case there the number of rings in the product is $n=1$. In the proof where $n=2$, it incorrectly states what the multiplicative unit is in the product. This is surprising since in the case of the infinite product, the multiplicative unit is correct (and the two proofs are basically the same).''
\end{itemize}

\subsection*{Problem Domain: Abstract Algebra (Problem 36)}
\begin{itemize}
    \item \textbf{GPT-5 Pro (Score 0.75)}: ``Solution 36.1: The proof of Lemma 1 is more complicated than necessary. Solution 36.2: The proof of Step 2 is more complicated than necessary. Solution 36.3: This proof is different from the other three, but it is also too involved. In a remark at the end, it gives the `standard other proof'. Interestingly, it uses a different way of showing that $R/(n)$ has size at most $n^2$, by identifying $R/(n)$ with $(\mathbb{Z}/n\mathbb{Z})[x]/(x^2+5)$. Solution 36.4: The isomorphism $R/mR \cong \mathbb{Z}/m\mathbb{Z} \times \mathbb{Z}/m\mathbb{Z}$ is not explained, even though it is a key point.''
    \item \textbf{DeepSeek-Prover-V2 (Score 0.5)}: ``This solutions provides two proofs. The first one is OK, except that the solution seems to be missing periods at the end of five titles of Steps. This renders reading the solution unpleasant. It also provides without proof an information that is not used later: that the ring $R$ is the ring of integers in an imaginary quadratic field. It gives a second method, called Rigorous Proof via Module Theory. I don't think that this proof is complete. The main point is missing: that the ideal is of rank 2 as a $\mathbb{Z}$-module, so that both integers $d_1$ and $d_2$ are not zero.''
\end{itemize}

\subsection*{Problem Domain: Algorithms (Problem 21)}
\begin{itemize}
    \item \textbf{GPT-5 Pro (Score 0.25)}: ``This Solution provides a construction of a complete DAG with carefully chosen exponentially-weighted edges. It also identifies that Dijkstra's algorithm requires re-insertion capability with negative weights, and proves Lemma 1 showing that indirect paths are shorter than direct edges. However, the proof is incomplete, in that the document cuts off in the second page without proving the required $\Omega(2^n)$ bound on relaxations, hence $0.25$ score.''
    \item \textbf{Gemini 3.0 Pro (Score 0.0)}: ``The LLM stopped in the middle of the proof. It showed a nontrivial family $G_n$, shows an intermediate Lemma 1, but fails to connect this Lemma 1 to the actual proof.''
    \item \textbf{Claude Sonnet 4.5 (Score 0.5)}: ``This Solution claims that `paths with more 1-bits are discovered earlier (higher priority) but lead to more negative cumulative weight.' If the algorithm discovers paths with smaller (more negative) weights earlier, it would correctly finalize the optimal distances, leading to fewer updates, instead of exponential behavior. The exponential blowup relies on discovering sub-optimal paths first. Because the core reasoning describes a scenario that would actually be efficient, the proof of the lower bound is incorrect.''
    \item \textbf{o3-deep-research (Score 0.25)}: ``This Solution constructs a complete DAG with $n$ vertices, where all intermediate vertices connect to each other and to the sink and the weights are set in a way such that the execution of Dijkstra's algorithm would process vertices in the topological order. It then leads to the algorithm running in time $\mathcal{O}(|E|) = \mathcal{O}(n^2)$ time, failing to force the required behavior. Also in the last page, it's mentioned that `the overall relaxation count is $\Theta(n^2)$, ..., which, notably, is polynomial in $n$', which directly contradicts the claim. It then tries to salvage the proof by claiming that `the key point is that no algorithm using Dijkstra’s approach can avoid an exponential number of relaxations in the worst case, because it must effectively examine all exponentially many path combinations to verify the shortest one.' However, this Solution confuses `number of distinct paths' with `number of relax operations'.''
    \item \textbf{DeepSeek-Prover-V2 (Score 0.5)}: ``This Solution provides a correct construction of a complete DAG with exponentially negative edge weights, which forces exponential behaviour in the worst case. However, there is an error in solving the recurrence: The Solution claims that $R(n) \ge 2^n - 2$, however, the actual number of relaxations is $2^{n - 1} - 1$ (i.e., off by a factor of 2).''
\end{itemize}

\subsection*{Problem Domain: Number Theory (Problem 9)}
\begin{itemize}
    \item \textbf{o3-deep-research (Score 0.5)}: ``This solution correctly attempts to avoid the PNT but still uses it without realising at some point in the proof (Since $p_n \sim n \log n$). However, I think this could be made to work.''
    \item \textbf{GPT-5 Pro (Score 0.25)}: ``PNT is assumed, which makes the question pointless. I think the problem statement strongly suggests the PNT shouldn't be assumed.''
    \item \textbf{Claude Sonnet 4.5 (Score 0.5)}: ``PNT is assumed, which makes the question pointless. I think the problem statement strongly suggests the PNT shouldn't be assumed.''
    \item \textbf{Gemini 3.0 Pro (Score 0.5)}: ``This solution correctly attempts to avoid the PNT. However, they still use a weaker statement (Chebyshev) without proper mention (when doing $\log \pi(x)$) and they don't really explain why $\psi$ and $\pi$ are related but just state it.''
\end{itemize}
\section{Comments for Solutions Scoring Less Than 1}
The following compiles the LLM judge evaluations for every solution that scored strictly less than 1.

Note: Any citations to academic literature appearing in the text below (e.g., Griggs, Killian, and Savage) were generated verbatim by the LLM evaluators. They are part of our experimental data and are not included in this paper's official bibliography. Rather than being ``hallucinated'' phantom papers, qualitative analysis reveals that evaluating models occasionally retrieved actual advanced academic literature (such as the 2004 symmetric Venn diagrams paper) to bypass the constructive arguments requested by the prompt. This behavior powerfully illustrates the ``Knowledge Gap'' hypothesis discussed in Section 4.5.

\subsection*{Problem Domain: Abstract Algebra (Problem 38)}
\begin{itemize}
    \item \textbf{Judge: gpt-5.2-pro (rubric.txt)} $\rightarrow$ \textbf{Author: Gemini 3.0 Pro Preview} (Score 0.9): ``The solution correctly reduces the claim to the standard fact that \(R/I\) is an integral domain iff \(I\) is prime, and applies this with \(I=(p)\). It then correctly notes that \(K[x,y]\) is a UFD (via Gauss’s lemma) and uses the UFD fact “irreducible \(\Leftrightarrow\) prime,” yielding \((p)\) prime iff \(p\) irreducible, which completes the equivalence. The main minor issue is a slight imprecision in step (2): \((p)\) is prime iff \(p\) is a prime element, but one must explicitly use that in an integral domain \(p\) is prime element \(\Leftrightarrow (p)\) is a prime ideal, and also ensure \(p\neq 0\) and not a unit (which the student states). Overall the argument is mathematically sound with only small presentational/justificatory gaps, warranting \(0.9/1.0\).''
    \item \textbf{Judge: gpt-5.2-pro (rubric.txt)} $\rightarrow$ \textbf{Author: gpt-5-pro-2025-10-06} (Score 0.9): ``The solution correctly proves both directions by using the standard equivalence “\(R/\langle p\rangle\) is a domain \(\Leftrightarrow \langle p\rangle\) is prime,” and then relating primeness of \(\langle p\rangle\) to irreducibility of \(p\). The \((\Rightarrow)\) direction is handled cleanly: from primeness of \(\langle p\rangle\) and \(p=ab\), it deduces \(a\in\langle p\rangle\) or \(b\in\langle p\rangle\), and cancellation (valid since \(K[x,y]\) is a domain and \(p\neq 0\)) shows one factor must be a unit. The \((\Leftarrow)\) direction is also correct: it notes \(K[x,y]\) is a UFD and uses the fact that irreducible elements are prime in a UFD, hence \(\langle p\rangle\) is prime and the quotient is a domain. The small deduction is that the argument implicitly assumes \(p\) is a nonzero nonunit (though it is addressed in the remark), so a minor clarity/edge-case handling point keeps it from a perfect score.''
    \item \textbf{Judge: gpt-5.2-pro (rubric.txt)} $\rightarrow$ \textbf{Author: Claude Sonnet 4.5} (Score 0.9): ``The solution correctly uses the characterization “\(R/I\) is an integral domain iff \(I\) is prime” and, in the \((\Leftarrow)\) direction, appropriately argues that in the UFD \(K[x,y]\) an irreducible polynomial is prime, hence \(\langle p\rangle\) is a prime ideal, so the quotient is a domain.   However, the \((\Rightarrow)\) direction contains a flawed step: from \(\bar f=0\) it concludes “\(p\mid f\)” and then informally deduces \(g\mid 1\) from \(p=f g\), which is not justified as written (one must use that \(f\in(p)\) implies \(f=ph\) and then \(p=phg\) so \(1=hg\), hence \(g\) is a unit).   While the intended idea is standard and can be repaired easily, the gap in the argument for \(\bar f\neq 0\) and \(\bar g\neq 0\) prevents the proof from being fully rigorous.   Thus the work demonstrates mostly correct reasoning but with a nontrivial justification issue, warranting a small deduction to 0.75/1.0.''
    \item \textbf{Judge: gpt-5.2-pro (rubric.txt)} $\rightarrow$ \textbf{Author: Claude Sonnet 4.5} (Score 0.75): ``The solution correctly uses the characterization “\(R/I\) is an integral domain iff \(I\) is prime” and, in the \((\Leftarrow)\) direction, appropriately argues that in the UFD \(K[x,y]\) an irreducible polynomial is prime, hence \(\langle p\rangle\) is a prime ideal, so the quotient is a domain.   However, the \((\Rightarrow)\) direction contains a flawed step: from \(\bar f=0\) it concludes “\(p\mid f\)” and then informally deduces \(g\mid 1\) from \(p=f g\), which is not justified as written (one must use that \(f\in(p)\) implies \(f=ph\) and then \(p=phg\) so \(1=hg\), hence \(g\) is a unit).   While the intended idea is standard and can be repaired easily, the gap in the argument for \(\bar f\neq 0\) and \(\bar g\neq 0\) prevents the proof from being fully rigorous.   Thus the work demonstrates mostly correct reasoning but with a nontrivial justification issue, warranting a small deduction to 0.75/1.0.''
    \item \textbf{Judge: gpt-5.2-pro (rubric.txt)} $\rightarrow$ \textbf{Author: o3-deep-research} (Score 0.9): ``The solution correctly uses the standard equivalence “\(R/I\) is an integral domain \(\Leftrightarrow I\) is prime,” and then relates primeness of \((p)\) to properties of \(p\) in \(K[x,y]\). The reverse direction is fully correct: if \(K[x,y]/(p)\) is a domain then \((p)\) is prime, so \(p\) is a prime element, hence irreducible. The forward direction is essentially correct but relies implicitly on the nontrivial fact that \(K[x,y]\) is a UFD and that an irreducible element in a UFD is prime, which should be stated more explicitly rather than arguing “\(p\) must appear in every factorization.” Aside from this minor gap in justification (and the need to assume \(p\neq 0\) and nonunit), the argument is mathematically sound, so a small deduction to 0.9/1.0 is appropriate.''
    \item \textbf{Judge: gpt-5.2-pro (rubric.txt)} $\rightarrow$ \textbf{Author: o3-deep-research} (Score 0.9): ``The solution correctly uses the standard equivalence “\(R/I\) is an integral domain \(\Leftrightarrow I\) is prime,” and then relates primeness of \((p)\) to properties of \(p\) in \(K[x,y]\). The reverse direction is fully correct: if \(K[x,y]/(p)\) is a domain then \((p)\) is prime, so \(p\) is a prime element, hence irreducible. The forward direction is essentially correct but relies implicitly on the nontrivial fact that \(K[x,y]\) is a UFD and that an irreducible element in a UFD is prime, which should be stated more explicitly rather than arguing “\(p\) must appear in every factorization.” Aside from this minor gap in justification (and the need to assume \(p\neq 0\) and nonunit), the argument is mathematically sound, so a small deduction to 0.9/1.0 is appropriate.''
    \item \textbf{Judge: gpt-5.2-pro (rubric.txt)} $\rightarrow$ \textbf{Author: deepseek-v2-prover} (Score 0.9): ``The solution correctly proves both directions: reducibility \(p=qr\) gives nonzero classes \(\bar q,\bar r\) with \(\bar q\bar r=0\), so the quotient cannot be a domain, and conversely if \(\bar f\bar g=0\) then \(p\mid fg\). It is also correct to invoke that \(K[x,y]\) is a UFD (hence irreducible \(\Rightarrow\) prime), which justifies the crucial step that \(p\mid fg\) implies \(p\mid f\) or \(p\mid g\), yielding no zero divisors in the quotient. The main minor gap is that the proof does not explicitly address the unit/constant case (e.g., if \(p\) is constant, then \(\langle p\rangle\) is either \((0)\) or the whole ring), so the statement needs the usual implicit assumption that \(p\) is a nonconstant nonunit. Aside from this standard edge- case clarification, the argument is mathematically sound and well-structured, justifying a score of \(0.9/1.0\).''
    \item \textbf{Judge: gpt-5.2-pro (rubric.txt)} $\rightarrow$ \textbf{Author: deepseek-v2-prover} (Score 0.9): ``The solution correctly proves both directions: reducibility \(p=qr\) gives nonzero classes \(\bar q,\bar r\) with \(\bar q\bar r=0\), so the quotient cannot be a domain, and conversely if \(\bar f\bar g=0\) then \(p\mid fg\). It is also correct to invoke that \(K[x,y]\) is a UFD (hence irreducible \(\Rightarrow\) prime), which justifies the crucial step that \(p\mid fg\) implies \(p\mid f\) or \(p\mid g\), yielding no zero divisors in the quotient. The main minor gap is that the proof does not explicitly address the unit/constant case (e.g., if \(p\) is constant, then \(\langle p\rangle\) is either \((0)\) or the whole ring), so the statement needs the usual implicit assumption that \(p\) is a nonconstant nonunit. Aside from this standard edge- case clarification, the argument is mathematically sound and well-structured, justifying a score of \(0.9/1.0\).''
    \item \textbf{Judge: claude-opus-4-5-20251101 (rubric.txt)} $\rightarrow$ \textbf{Author: Claude Sonnet 4.5} (Score 0.9): ``The student provides a well-structured proof of both directions of the equivalence. The forward direction correctly shows that reducibility leads to zero divisors in the quotient ring, with proper justification that neither factor lies in the ideal. The reverse direction correctly invokes that irreducible elements are prime in a UFD, and applies the standard correspondence between prime ideals and integral domains. However, the solution has a minor gap: it doesn't explicitly address the case where p(x,y) might be zero or a unit, which would make the statement about irreducibility technically incomplete (though the integral domain characterization still holds for prime ideals). The mathematical reasoning is sound and the key arguments are correctly executed.''
    \item \textbf{Judge: claude-opus-4-5-20251101 (rubric.txt)} $\rightarrow$ \textbf{Author: Claude Sonnet 4.5} (Score 0.9): ``The student provides a well-structured proof of both directions of the equivalence. The forward direction correctly shows that reducibility leads to zero divisors in the quotient ring, with proper justification that neither factor lies in the ideal. The reverse direction correctly invokes that irreducible elements are prime in a UFD, and applies the standard correspondence between prime ideals and integral domains. However, the solution has a minor gap: it doesn't explicitly address the case where p(x,y) might be zero or a unit, which would make the statement about irreducibility technically incomplete (though the integral domain characterization still holds for prime ideals). The mathematical reasoning is sound and the key arguments are correctly executed.''
    \item \textbf{Judge: claude-opus-4-5-20251101 (rubric.txt)} $\rightarrow$ \textbf{Author: deepseek-v2-prover} (Score 0.9): ``The student provides a well-structured proof of both directions of the equivalence. The necessity direction correctly shows that reducibility leads to zero divisors in the quotient ring. The sufficiency direction appropriately invokes that K[x,y] is a UFD and that irreducible elements in a UFD are prime, which justifies that p dividing fg implies p divides f or p divides g. However, the student should have explicitly stated or justified that irreducible elements are prime in a UFD, as this is the key property being used. The overall argument is mathematically sound with this minor gap in explicit justification.''
    \item \textbf{Judge: claude-opus-4-5-20251101 (rubric.txt)} $\rightarrow$ \textbf{Author: deepseek-v2-prover} (Score 0.9): ``The student provides a well-structured proof of both directions of the equivalence. The necessity direction correctly shows that reducibility leads to zero divisors in the quotient ring. The sufficiency direction appropriately invokes that K[x,y] is a UFD and that irreducible elements in a UFD are prime, which justifies that p dividing fg implies p divides f or p divides g. However, the student should have explicitly stated or justified that irreducible elements are prime in a UFD, as this is the key property being used. The overall argument is mathematically sound with this minor gap in explicit justification.''
    \item \textbf{Judge: gemini-3-pro-preview (rubric.txt)} $\rightarrow$ \textbf{Author: Claude Sonnet 4.5} (Score 0.9): ``The student provides a clear and logically sound proof for both directions of the statement. However, the definition of reducibility used in the first part ("neither $f$ nor $g$ is a unit") is slightly imprecise because it implies non- constant polynomials are the only non-units, whereas reducibility generally requires non-invertible elements; while the student correctly identifies units as non-zero constants in $K[x,y]$, the phrasing ``non-constant polynomials'' excludes the case where $p$ might be an irreducible constant (though usually $p$ is assumed non-constant in this context, the student does not explicitly handle the case where $p=0$ or $p$ is a unit, which are technically reducible or exclude the ideal being prime). Additionally, the proof relies heavily on the fact that $K[x,y]$ is a UFD without proving it or citing it as a specific theorem, which is a non-trivial fact often expected to be acknowledged more formally in such a proof. The score of 0.9 reflects these minor gaps in rigor regarding the edge cases of the definition of irreducibility and the UFD property.''
    \item \textbf{Judge: gemini-3-pro-preview (rubric.txt)} $\rightarrow$ \textbf{Author: Claude Sonnet 4.5} (Score 0.9): ``The student provides a clear and logically sound proof for both directions. However, the proof for the forward direction ($\Rightarrow$) implicitly assumes that $p(x,y)$ is non-zero and not a unit without explicitly stating these conditions, which are part of the definition of irreducibility. While the logic regarding zero divisors is correct for non-units, a complete rigorous argument should address why $p$ cannot be a unit or zero given the quotient is an integral domain (specifically, the zero ideal yields the domain $K[x,y]$ itself, but units yield the zero ring, which is not an integral domain). The rest of the algebraic manipulation and the use of UFD properties in the reverse direction are flawless.''
    \item \textbf{Judge: gemini-3-pro-preview (rubric.txt)} $\rightarrow$ \textbf{Author: deepseek-v2-prover} (Score 0.9): ``The student's proof is largely correct but contains a subtle flaw in the ``Necessity'' direction. In the contradiction argument, assuming $p(x,y)$ is reducible means $p = qr$ where $q$ and $r$ are non-units, not necessarily ``non- constant'' (though for polynomials over a field, non-constant implies non-unit). More importantly, the claim that ``neither $q$ nor $r$ is a multiple of $p$'' requires the implicit assumption that $\deg(q) < \deg(p)$ and $\deg(r) < \deg(p)$, or a more careful argument about units, which was glossed over. While the core logic holds because $K[x,y]$ is a UFD, the definition of reducibility involves non-units rather than just non-constant polynomials, making the argument slightly imprecise.''
    \item \textbf{Judge: deepseek-chat (rubric.txt)} $\rightarrow$ \textbf{Author: deepseek-v2-prover} (Score 0.9): ``The solution correctly establishes both directions of the equivalence with a clear logical structure. However, the sufficiency proof implicitly relies on the fact that \(K[x,y]\) is a UFD where irreducibles are prime, which should be explicitly stated or justified for full rigor. The necessity direction is perfectly argued by contradiction. Overall, the mathematical reasoning is sound but slightly incomplete in its justification.''
    \item \textbf{Judge: grok-4-0709 (rubric.txt)} $\rightarrow$ \textbf{Author: Gemini 3.0 Pro Preview} (Score 0.9): ``The student's solution accurately demonstrates the equivalence by linking the integrality of the quotient ring to the primality of the ideal \(\langle p \rangle\), and correctly establishes that in the UFD \(K[x,y]\), irreducible elements generate prime principal ideals. The proofs of key propositions and the lemma are mathematically sound, with clear logical flow and appropriate use of unique factorization. A minor deduction is warranted because the solution introduces the assumption that \(p\) is non-zero and non-unit, which, while necessary to avoid counterexamples (e.g., constant units or zero), is not explicitly stated in the original problem. Nonetheless, this addition enhances clarity without altering the core mathematical correctness.''
    \item \textbf{Judge: grok-4-0709 (rubric.txt)} $\rightarrow$ \textbf{Author: Claude Sonnet 4.5} (Score 0.9): ``The student's solution correctly establishes both directions of the equivalence, with the forward implication properly demonstrating that reducibility introduces zero divisors in the quotient ring, and the reverse implication accurately leveraging the UFD property of \(K[x,y]\) to show that \(\langle p(x,y) \rangle\) is prime, hence the quotient is an integral domain. A minor deduction is warranted because the claim that irreducible elements are prime in a UFD is stated as a standard result without explicit proof, which could have been briefly justified for completeness. Nonetheless, all key steps are mathematically sound, and the overall reasoning aligns with fundamental ring theory principles.''
    \item \textbf{Judge: grok-4-0709 (rubric.txt)} $\rightarrow$ \textbf{Author: Claude Sonnet 4.5} (Score 0.9): ``The student's proof correctly establishes both directions of the equivalence, demonstrating that the quotient ring \(K[x,y]/\langle p(x,y) \rangle\) is an integral domain if and only if \(p(x,y)\) is irreducible over the field \(K\). In the forward direction, the contradiction argument properly identifies zero divisors in the quotient when \(p\) is reducible, with accurate handling of non-units and the canonical projection. The reverse direction aptly invokes the UFD property of \(K[x,y]\) to show that \(\langle p \rangle\) is prime, leading to the quotient being an integral domain via a standard ring theory result. A minor deduction is warranted for not explicitly recalling that irreducibility implies primality in UFDs, though it is referenced as a standard result, slightly reducing clarity for novice readers. Overall, the solution is mathematically sound and well-structured.''
    \item \textbf{Judge: grok-4-0709 (rubric.txt)} $\rightarrow$ \textbf{Author: deepseek-v2-prover} (Score 0.9): ``The student's solution correctly demonstrates both directions of the equivalence, with a clear contradiction argument in the necessity direction showing that reducibility introduces zero divisors in the quotient ring. In the sufficiency direction, the use of K[x,y] being a UFD to conclude that an irreducible p is prime, thereby ensuring no zero divisors, is mathematically accurate. However, a minor deduction is warranted because the solution does not provide or reference a justification for why irreducible elements are prime in this UFD, assuming it as known without elaboration. This reliance on an unproven property (within the solution) slightly diminishes the self- containment of the proof. Overall, the reasoning is sound and well-structured, meriting a high score.''
    \item \textbf{Judge: grok-4-0709 (rubric.txt)} $\rightarrow$ \textbf{Author: o3-deep-research} (Score 0.9): ``The student's solution correctly establishes the equivalence by leveraging the unique factorization domain property of \(K[x,y]\) to show that irreducibility implies the ideal \(\langle p \rangle\) is prime, and conversely that primality implies irreducibility, leading to the quotient being an integral domain if and only if \(p\) is irreducible. The forward direction appropriately uses contradiction via irreducible factorizations to demonstrate that \(p\) divides at least one factor of a product it divides, though it slightly overlooks explicit handling of cases where \(f\) or \(g\) are constant polynomials, which could lead to minor ambiguity. The reverse direction effectively shows via contradiction that a prime \(p\) cannot factor into non-units without violating the domain property. Overall, the proof is rigorous and complete, with only a small deduction for the imprecise treatment of constant cases in the factorization. This justifies the score of 0.9/1.0 under the rubric's emphasis on mathematical accuracy and clarity.''
    \item \textbf{Judge: grok-4-0709 (rubric.txt)} $\rightarrow$ \textbf{Author: deepseek-v2-prover} (Score 0.9): ``The student's solution correctly establishes the necessity direction by assuming reducibility and deriving zero divisors in the quotient ring, leading to a contradiction. In the sufficiency direction, the proof appropriately invokes the UFD property of \(K[x, y]\) and the fact that irreducibility implies primality, correctly concluding that the quotient has no zero divisors. However, a minor deduction of 0.1 is warranted because the solution does not explicitly recall or justify that irreducible elements in a UFD are prime, which is the key property enabling the division argument. Overall, the reasoning is mathematically sound, with clear structure and accurate use of ring theory concepts.''
    \item \textbf{Judge: qwen/qwen-max (rubric.txt)} $\rightarrow$ \textbf{Author: deepseek-v2-prover} (Score 0.9): ``The solution is mathematically correct and well-structured, providing a clear and rigorous proof for both the necessity and sufficiency of the condition. The student correctly uses the properties of integral domains and the irreducibility of polynomials in a UFD. The only minor issue is the lack of explicit mention that \( K[x, y] \) is a UFD, which is a standard result but could be briefly stated for completeness. Given the high quality of the solution, a score of 0.9/1.0 is justified.''
\end{itemize}

\subsection*{Problem Domain: Algorithms (Problem 5)}
\begin{itemize}
    \item \textbf{Judge: gpt-5.2-pro (course-specific-rubric.txt)} $\rightarrow$ \textbf{Author: gpt-5-pro-2025-10-06} (Score 0.25): ``The submission contains no mathematical argument: it restates the problem and then includes an “Error: No text content found” message in place of a solution. As a result, it does not set up any probabilistic bound (e.g., via a branching- process comparison, counting connected subgraphs, or a union bound) to show that all components have size at most \(\log n\) with probability \(1-n^{-\alpha}\). Since essentially none of the required reasoning is present, the work cannot earn credit for correctness or completeness. The small amount of credit (0.25/1.0) reflects only minimal participation/attempt (formatting and problem restatement) rather than any valid proof content.''
    \item \textbf{Judge: gpt-5.2-pro (course-specific-rubric.txt)} $\rightarrow$ \textbf{Author: gpt-5-pro-2025-10-06} (Score 0.25): ``The submitted “solution” contains no mathematical argument: it repeats the problem statement and then includes an error message indicating no content was generated. As a result, it does not set up any probabilistic bound (e.g., branching-process/BFS exploration), does not estimate the probability of a component exceeding \(\log n\), and does not apply a union bound over vertices/components to obtain a \(1-n^{-\alpha}\) guarantee. Since essentially none of the required proof steps are present, the work earns only minimal credit for restating the problem, justifying the very low score of \(0.25/1.0\).''
    \item \textbf{Judge: gpt-5.2-pro (course-specific-rubric.txt)} $\rightarrow$ \textbf{Author: o3-deep-research} (Score 0.5): ``The solution has the right overall strategy (fix a vertex, union bound over connected size‑\(L\) sets, then union bound over vertices) and correctly uses independence to get a \(p^{L}\) factor. However, the key counting bound \(N_L(v)\le d(d-1)^{L-1}\) is not justified and is generally false for counting *sets* (it resembles a bound on non-backtracking walks/trees, but many distinct connected \(L\)-vertex subsets can exist beyond this, and one must account for multiple growth-frontier choices, typically yielding an extra exponential- in-\(L\) factor). Because of this undercount, the final probability estimate is not reliable as stated. In addition, the step “for sufficiently large \(n\), \(\mu^{L-1}<n^{-2}\)” is asserted without deriving an exponent \(\alpha\) from \(\mu\), so the claimed \(O(n^{-1})\) bound is not established. Thus the argument is partially correct in outline but has a major gap in the central combinatorial bound, warranting only partial credit (0.5/1).''
    \item \textbf{Judge: gpt-5.2-pro (course-specific-rubric.txt)} $\rightarrow$ \textbf{Author: Gemini 3.0 Pro Preview} (Score 0.9): ``The solution correctly applies a first-moment/union-bound argument: a component of size \(\ge k\) contains a connected \(k\)-vertex set (equivalently a spanning tree), and \(\Pr(A\subseteq S)=p^k\) gives \(\Pr(\exists\text{ large component})\le |\mathcal T_k|p^k\). It then provides an acceptable (albeit somewhat overcomplicated) counting bound \( |\mathcal T_k|\le n(ed)^k\), leading to \( \Pr(\exists\text{ component size}\ge k)\le n\left(\frac{e}{2d}\right)^k\), which decays exponentially for \(d\ge2\) and yields a high-probability \(O(\log n)\) bound by choosing \(k=C\log n\). The only substantive gap is that the problem statement asks “at most \(\log n\)” (with an implicit fixed base), while the proof establishes “at most \(C\log n\)” and then argues this away via changing the log base, which is not fully rigorous under a fixed-base interpretation. Aside from this constant-factor issue (and a handwavy citation for the tree-counting formula), the probabilistic reasoning and conclusion about existence of some \(\alpha>0\) are mathematically sound.''
    \item \textbf{Judge: gpt-5.2-pro (course-specific-rubric.txt)} $\rightarrow$ \textbf{Author: Gemini 3.0 Pro Preview} (Score 0.9): ``The solution correctly applies a first-moment/union-bound argument: it reduces the event of a large component to the existence of a connected \(k\)-vertex set (equivalently a spanning tree) surviving, bounds \(\Pr(A\subseteq S)=p^k\), and then shows \(n(edp)^k=n\left(\frac{e}{2d}\right)^k\) decays exponentially for \(d\ge 2\), yielding \(\Pr(\exists\ \text{component}\ge k)\le n^{-\alpha}\) for \(k=\Theta(\log n)\). The counting bound \(|\mathcal T_k|\le n(ed)^k\) is acceptable as a coarse overcount and suffices for the probabilistic estimate. The main weakness is that the derivation invokes a nontrivial “standard result” for counting rooted trees in a \(d\)-regular tree without proof and slightly blurs the “at most \(\log n\)” vs. “\(C\log n\)” distinction by appealing to log-base conventions rather than fixing constants explicitly. These issues are minor because the final high-probability bound and the existence of a constant \(\alpha\) are still established correctly for the intended asymptotic interpretation. Hence a small deduction (0.1) is warranted, but the core argument is mathematically sound.''
    \item \textbf{Judge: gpt-5.2-pro (course-specific-rubric.txt)} $\rightarrow$ \textbf{Author: Claude Sonnet 4.5} (Score 0.9): ``The solution correctly uses a union bound over connected vertex sets of size \(k\), together with a standard bound on the number of connected \(k\)-vertex subgraphs in a \(d\)-regular graph (\(\le n(ed)^k\)), to show the probability of a component of size \(\ge c\log n\) decays polynomially in \(n\).  The probability estimate for a fixed connected set \(S\) is valid as an upper bound (it ignores the additional requirement that \(S\) be isolated, which only decreases the probability), and the geometric-series summation yields the desired \(n^{-\alpha}\) tail once \(c\) is chosen large enough.  The main gap is that the lemma counting connected subgraphs is only stated informally as “standard” without proof or a precise reference, and the constant handling (choice of \(c,\alpha\)) could be stated more cleanly to match the exact \(\log n\) threshold in the problem statement.  Aside from these exposition/rigor issues, the argument is mathematically sound and reaches the correct high- probability bound.  This warrants a small deduction but overall near-full credit (0.9/1.0).''
    \item \textbf{Judge: gpt-5.2-pro (course-specific-rubric.txt)} $\rightarrow$ \textbf{Author: deepseek-v2-prover} (Score 0.25): ``The submitted solution does not address the stated probabilistic problem about vertex percolation on a \(d\)-regular graph and bounding connected component sizes with high probability. Instead, it proves an unrelated deterministic lemma: that a connected acyclic graph has two degree-1 vertices. Since none of the required elements (the deletion probability \(1-\frac{1}{2d^2}\), a bound like \(\log n\) on component sizes, and a high-probability estimate \(1-n^{-\alpha}\)) are used or proved, the work earns essentially no credit for the actual question. A small amount of partial credit is warranted only for presenting a logically correct argument in graph theory, albeit for the wrong theorem. Hence the score \(0.25/1.0\) reflects minimal credit for correct but irrelevant content.''
    \item \textbf{Judge: gpt-5.2-pro (course-specific-rubric.txt)} $\rightarrow$ \textbf{Author: o3-deep-research} (Score 0.75): ``The solution correctly sets \(p=\frac{1}{2d^{2}}\), reduces the event “some component has size \(\ge L\)” to the existence of a connected surviving \(L\)-set, and applies a union bound over vertices and connected sets, which is the right overall approach. However, the key counting bound \(N_L(v)\le d(d-1)^{L-1}\) is not justified: this expression bounds (roughly) the number of length-\((L-1)\) nonbacktracking *paths* from \(v\), not the number of connected \(L\)-vertex subsets (or rooted spanning trees), which is substantially larger (typically exponential with a larger base, e.g. involving Cayley-type factors). Because this undercount can invalidate the derived probability bound, the final claim \(\Pr(E)=O(n^{-1})\) is not rigorously supported as written, though the overall strategy is close to standard and could be fixed with a correct enumeration (e.g. via counting rooted trees and embeddings). Hence significant partial credit for the correct framework and probabilistic reasoning, but a deduction for the flawed combinatorial bound at the core of the estimate.''
    \item \textbf{Judge: gpt-5.2-pro (course-specific-rubric.txt)} $\rightarrow$ \textbf{Author: Claude Sonnet 4.5} (Score 0.9): ``The solution correctly applies a union bound over all connected vertex sets of size \(k\), uses an appropriate standard counting bound \( \le n(ed)^k\) for connected subgraphs, and then shows geometric decay \(n(\tfrac{e}{2d})^k\) when \(d\ge2\), yielding an \(n^{-\alpha}\) tail for components of size \(\ge c\log n\) by choosing \(c\) large. It also correctly notes that requiring all vertices of \(S\) to be retained is a necessary condition for \(S\) to be a component, so using \(p^k\) gives a valid upper bound. The main gap is that the “lemma” on counting connected subgraphs is cited but not proved (and constants are handled loosely), and the final statement “at most \(\log n\)” is obtained up to a constant factor \(c\log n\) without explicitly reconciling this with the exact \(\log n\) threshold. These issues are minor and do not affect the core probabilistic argument, hence only a small deduction to 0.9/1.0.''
    \item \textbf{Judge: gpt-5.2-pro (course-specific-rubric.txt)} $\rightarrow$ \textbf{Author: deepseek-v2-prover} (Score 0.25): ``The submitted solution does not address the stated problem about random vertex deletions in a \(d\)-regular graph and bounding the sizes of connected components with high probability. Instead, it proves an unrelated deterministic graph theory fact (that a connected acyclic graph has at least two degree-1 vertices). There is no probabilistic argument, no use of the deletion probability \(1-\frac{1}{2d^2}\), no component-size estimate (e.g., exploration/BFS or branching-process bound), and no derivation of a failure probability of the form \(n^{-\alpha}\). Thus essentially none of the required content is present; only minimal credit is warranted for a coherent but irrelevant proof, leading to \(0.25/1.0\).''
    \item \textbf{Judge: claude-opus-4-5-20251101 (rubric.txt)} $\rightarrow$ \textbf{Author: gpt-5-pro-2025-10-06} (Score 0.0): ``The student solution contains no mathematical content whatsoever—it merely restates the problem statement followed by an error message indicating no response was generated. There is no attempt to analyze the random vertex removal process, bound the probability of large connected components, or establish the existence of the constant \(\alpha\). The solution provides no probabilistic arguments, no use of union bounds or concentration inequalities, and no analysis of component sizes in the remaining subgraph. Since the submission contains zero mathematical work toward solving the problem, a score of 0.0/1.0 is fully justified.''
    \item \textbf{Judge: claude-opus-4-5-20251101 (rubric.txt)} $\rightarrow$ \textbf{Author: gpt-5-pro-2025-10-06} (Score 0.0): ``The student submission contains no mathematical content or proof attempt whatsoever. The document simply restates the problem and includes an error message indicating ``No text content found in response.'' There is no analysis of the random vertex removal process, no probability bounds established, no use of union bounds or concentration inequalities, and no demonstration that connected components remain small with high probability. The submission fails to address any aspect of the required proof, including establishing the existence of the constant \(\alpha\) or bounding component sizes by log n. A score of 0.0/1.0 is appropriate as no mathematical work toward solving the problem was provided.''
    \item \textbf{Judge: claude-opus-4-5-20251101 (rubric.txt)} $\rightarrow$ \textbf{Author: Gemini 3.0 Pro Preview} (Score 0.9): ``The student presents a mathematically sound approach using the union bound to count connected subgraphs of size k and bound the probability of large components existing. The counting argument correctly bounds the number of connected subgraphs using spanning trees in d-regular graphs, obtaining the bound \(|T_k| \le n(ed)^k\). The probability calculation correctly yields \(n(e/2d)^k\), and the student properly identifies that for \(d \ge 2\), this quantity is exponentially decaying. The final choice of \(k = O(\log n)\) to achieve probability bound \(n^{-\alpha}\) is mathematically justified. The minor deduction reflects that the student's treatment of the case d = 1 is somewhat informal, and the interpretation of ``log n'' requires clarification about the base, though the core argument is rigorous.''
    \item \textbf{Judge: claude-opus-4-5-20251101 (rubric.txt)} $\rightarrow$ \textbf{Author: o3-deep-research} (Score 0.75): ``The student's solution demonstrates a correct overall approach using union bounds over connected subsets and properly bounds the number of connected L-vertex sets containing a fixed vertex by \(d(d-1)^{L-1}\). However, there is a mathematical error in the final calculation: the student claims \(\mu^{L-1} < n^{-2}\) for large n, but with \(\mu = (d-1)/(2d^2)\) and \(L \approx \ln n\), we get \(\mu^{L-1} \approx n^{\ln \mu}\), and \(|\ln \mu|\) depends on d, so the exponent may not exceed 2 for all d. The core argument is sound—the probability does decay polynomially in n—but the specific claim that \(\alpha = 1\) works universally requires more careful analysis of the \(d\)-dependence. The proof structure and counting arguments are mathematically correct, justifying substantial partial credit despite the imprecise final bound.''
    \item \textbf{Judge: claude-opus-4-5-20251101 (rubric.txt)} $\rightarrow$ \textbf{Author: Claude Sonnet 4.5} (Score 0.9): ``The solution correctly identifies the key approach of bounding connected subgraphs and applying a union bound, with the standard bound of $n(ed)^k$ for connected subgraphs of size $k$ being appropriate. The probability calculation for retaining all vertices in a set is correct, and the geometric series summation and final analysis choosing constants to achieve the desired probability bound are mathematically sound. However, the solution does not fully address the requirement that the set $S$ forms a connected component (not just that all vertices survive), which requires that all neighbors of $S$ outside $S$ are removed—this is mentioned but not rigorously incorporated into the probability bound. The final bound $\left(\frac{e}{2d}\right)^k$ is valid as an upper bound since ignoring the boundary condition only makes the probability larger, so the proof remains correct. The minor gap in not explicitly handling the boundary vertices more carefully, combined with the complete and rigorous treatment of all other aspects, justifies the score of 0.9.''
    \item \textbf{Judge: claude-opus-4-5-20251101 (rubric.txt)} $\rightarrow$ \textbf{Author: deepseek-v2-prover} (Score 0.0): ``The student's solution is completely unrelated to the problem statement. The problem asks for a probabilistic analysis of connected component sizes in a random subgraph of a d-regular graph after vertex removal, requiring techniques such as union bounds, path counting, and probability calculations. Instead, the student proves that a connected acyclic graph (tree) has at least two vertices of degree 1, which is an entirely different problem from graph theory. The solution demonstrates no understanding of the random process described, the probability bound required, or the logarithmic component size constraint. Since the submission addresses none of the problem's requirements, a score of 0.0/1.0 is justified.''
    \item \textbf{Judge: claude-opus-4-5-20251101 (rubric.txt)} $\rightarrow$ \textbf{Author: deepseek-v2-prover} (Score 0.0): ``The student's solution is completely unrelated to the problem statement. The problem asks for a probabilistic analysis of connected component sizes in a random subgraph of a d-regular graph after vertex removal, requiring techniques such as union bounds, path counting, and probability calculations. Instead, the student submitted a proof that connected acyclic graphs (trees) have at least two vertices of degree 1, which addresses an entirely different mathematical question. There is no attempt to analyze the random vertex removal process, bound component sizes, or establish the required high-probability guarantee. The score of 0.0/1.0 is justified because the submission demonstrates no understanding of or engagement with the actual problem.''
    \item \textbf{Judge: gemini-3-pro-preview (rubric.txt)} $\rightarrow$ \textbf{Author: gpt-5-pro-2025-10-06} (Score 0.0): ``The student's submission contains only the problem statement followed by an error message indicating that no text content was generated. As there is no attempt at a solution, proof, or mathematical reasoning provided, no credit can be awarded.''
    \item \textbf{Judge: gemini-3-pro-preview (rubric.txt)} $\rightarrow$ \textbf{Author: gpt-5-pro-2025-10-06} (Score 0.0): ``The student submission contains only the problem statement and an error message indicating that no text content was generated ("Error: No text content found in response..."). As there is no mathematical work, proof, or attempt at a solution provided, no credit can be awarded.''
    \item \textbf{Judge: gemini-3-pro-preview (rubric.txt)} $\rightarrow$ \textbf{Author: deepseek-v2-prover} (Score 0.0): ``The student's solution addresses a completely different problem than the one stated. The problem asks for a probabilistic bound on the size of connected components in a random subgraph of a $d$-regular graph, but the student provides a proof that any connected acyclic graph (a tree) has at least two leaves. As the submission is entirely unrelated to the assigned task, it receives no credit.''
    \item \textbf{Judge: gemini-3-pro-preview (rubric.txt)} $\rightarrow$ \textbf{Author: deepseek-v2-prover} (Score 0.0): ``The student's solution addresses a completely different problem: proving that any connected, acyclic graph with $n \geq 2$ vertices contains at least two vertices of degree 1. This is a standard proof about trees and has no relation to the assigned problem concerning random subgraphs of $d$-regular graphs and the size of their connected components. Consequently, the submission fails to address the prompt entirely.''
    \item \textbf{Judge: gemini-3-pro-preview (rubric.txt)} $\rightarrow$ \textbf{Author: o3-deep-research} (Score 0.75): ``The student correctly identifies the strategy of using a union bound over all vertices and connected subsets of size $L$, but the bound on the number of connected subgraphs containing a vertex $v$, $N_L(v) \le d(d-1)^{L-1}$, is incorrect. The number of such subgraphs is actually bounded by something closer to $(e d)^{L-1}$ or $4^L d^{L-1}$ (related to the number of lattice animals or embedding trees), which is significantly larger than the student's estimate that essentially counts paths or specific tree shapes rather than all possible sets. While the final probability calculation yields a sufficiently small number due to the very small value of $p$ (specifically $p \approx 1/d^2$), the combinatorial counting argument is flawed and underestimates the number of potential connected sets. Consequently, the proof relies on a false intermediate lemma, justifying the deduction.''
    \item \textbf{Judge: gemini-3-pro-preview (rubric.txt)} $\rightarrow$ \textbf{Author: o3-deep-research} (Score 0.75): ``The student correctly identifies the strategy of using a union bound over all vertices and bounding the number of connected subgraphs of a fixed size $L$. However, the bound used for the number of connected subgraphs of size $L$ containing a specific vertex, $N_L(v) \le d(d-1)^{L-1}$, is incorrect; the correct bound involves a factor related to the Catalan numbers or a similar exponential growth like $(e d)^{L-1}$, as the number of non-isomorphic trees is much larger than the number of paths. While the specific bound $d(d-1)^{L-1}$ is an underestimate for general connected subgraphs (it counts non-backtracking walks rather than distinct sets), the retention probability $p = \frac{1}{2d^2}$ is small enough that the proof can be salvaged with the correct combinatorial bound (e.g., $4^L d^{L-1}$). The logic remains sound in structure, but the combinatorial counting lemma cited is factually false for general graphs, leading to the deduction.''
    \item \textbf{Judge: deepseek-chat (rubric.txt)} $\rightarrow$ \textbf{Author: gpt-5-pro-2025-10-06} (Score 0.0): ``The solution is completely empty, containing only an error message and no mathematical reasoning. It fails to address any aspect of the problem: no probabilistic analysis, no application of the given removal probability, and no argument bounding component sizes. Therefore, it scores zero for lacking any valid mathematical content.''
    \item \textbf{Judge: deepseek-chat (rubric.txt)} $\rightarrow$ \textbf{Author: gpt-5-pro-2025-10-06} (Score 0.0): ``The solution is entirely empty except for a placeholder error message, so it contains no mathematical reasoning. Therefore, it fails to address the problem statement, which requires a probabilistic argument about component sizes in a randomly thinned regular graph. Since no steps, definitions, or calculations are provided, the submission earns a score of zero for mathematical correctness and completeness.''
    \item \textbf{Judge: deepseek-chat (rubric.txt)} $\rightarrow$ \textbf{Author: Gemini 3.0 Pro Preview} (Score 0.75): ``The solution correctly applies the first moment method and union bound, but the counting argument for connected subgraphs is flawed: the bound \( (ed)^k \) for rooted trees in a \(d\)-regular graph is not rigorously justified and relies on an oversimplified comparison to an infinite tree. This gap weakens the proof, though the overall structure is sound. The conclusion also conflates \(O(\log n)\) with the required ``at most \(\log n\)'' without adjusting constants. Thus, the score reflects a mostly correct approach with non-trivial gaps.''
    \item \textbf{Judge: deepseek-chat (rubric.txt)} $\rightarrow$ \textbf{Author: Gemini 3.0 Pro Preview} (Score 0.9): ``The solution is mathematically sound and well-structured, correctly applying the first moment method and tree counting bounds. However, it implicitly assumes \(d \ge 2\) for the exponential decay, and the final step conflates \(O(\log n)\) with the required ``at most \(\log n\)'' without a precise adjustment of constants. This minor oversight in strictness slightly reduces the score. Overall, the proof is rigorous and effectively demonstrates the high-probability result.''
    \item \textbf{Judge: deepseek-chat (rubric.txt)} $\rightarrow$ \textbf{Author: o3-deep-research} (Score 0.75): ``The solution correctly outlines a union bound approach and obtains a valid upper bound for the probability of large components. However, the counting argument for \( N_L(v) \) is flawed: the claimed bound \( d(d-1)^{L-1} \) counts *rooted trees* of size \(L\), but not all such trees correspond to distinct vertex sets in \(G\), and the bound ignores potential overlaps or graph constraints. This overcount affects the rigor, though the final asymptotic conclusion remains correct due to the strong decay from \(p^L\). Thus, the core idea is sound but the execution contains a technical inaccuracy, warranting a deduction.''
    \item \textbf{Judge: deepseek-chat (rubric.txt)} $\rightarrow$ \textbf{Author: o3-deep-research} (Score 0.75): ``The solution is mathematically sound and correctly applies a union bound with a tree-counting argument, but it contains a subtle flaw: the bound \(N_L(v) \le d(d-1)^{L-1}\) counts *rooted trees* rather than connected subsets, which may overcount due to multiple spanning trees for the same vertex set. This overcounting does not affect the final asymptotic bound, but the justification is slightly imprecise. The overall reasoning and conclusion remain valid, so a minor deduction is appropriate.''
    \item \textbf{Judge: deepseek-chat (rubric.txt)} $\rightarrow$ \textbf{Author: Claude Sonnet 4.5} (Score 0.75): ``The solution correctly outlines the union bound approach and uses a standard bound for connected subgraphs, but it fails to properly account for the requirement that no vertex outside the component connects to it—a necessary condition for it to be a component. This oversight means the probability bound for a fixed set is too weak, as it only considers retention, not isolation. Consequently, the final probability estimate is insufficient to guarantee the desired high probability result. The structural ideas are sound, but the missing isolation condition reduces mathematical rigor.''
    \item \textbf{Judge: deepseek-chat (rubric.txt)} $\rightarrow$ \textbf{Author: Claude Sonnet 4.5} (Score 0.9): ``The solution is mathematically sound and well-structured, correctly bounding the number of connected subgraphs and applying a union bound. However, it contains a subtle error: the probability bound for a fixed set \(S\) being a component should also account for the requirement that no vertex in \(S\) is connected to a retained vertex outside \(S\). This missing condition means the union bound is not directly applicable as stated, slightly weakening the rigor. The overall approach and final conclusion are correct, but this oversight justifies a minor deduction.''
    \item \textbf{Judge: deepseek-chat (rubric.txt)} $\rightarrow$ \textbf{Author: deepseek-v2-prover} (Score 0.0): ``The student's solution is completely unrelated to the problem statement, which concerns bounding component sizes in a randomly sparsified regular graph. Instead, the student proves a basic graph theory lemma about trees. This demonstrates a fundamental failure to address the given problem, resulting in zero mathematical relevance or correctness. Therefore, the score of 0.0 is fully justified.''
    \item \textbf{Judge: deepseek-chat (rubric.txt)} $\rightarrow$ \textbf{Author: deepseek-v2-prover} (Score 0.0): ``The student's solution is completely unrelated to the problem statement, which concerns bounding component sizes in a randomly thinned \(d\)-regular graph. Instead, the student proves a basic graph theory fact about trees (acyclic connected graphs having two leaves). This demonstrates a fundamental misunderstanding of the problem, and no relevant probabilistic or combinatorial reasoning is provided. Consequently, the solution earns zero points for failing to address any aspect of the given problem.''
    \item \textbf{Judge: grok-4-0709 (rubric.txt)} $\rightarrow$ \textbf{Author: gpt-5-pro-2025-10-06} (Score 0.0): ``The student's submission merely restates the problem statement without providing any proof or mathematical reasoning to show that, in the percolated d-regular graph, all connected components are of size at most log n with probability at least \(1 - n^{-\alpha}\) for some constant \(\alpha\). There is no attempt to model the random subgraph, analyze its expansion properties, or bound component sizes using techniques such as branching processes or exploration arguments, which are essential for mathematical correctness. An error message indicating ``No text content found'' further confirms the absence of any substantive solution. Consequently, the submission fails to address the core requirements of the theorem and demonstrates no progress toward a valid proof.''
    \item \textbf{Judge: grok-4-0709 (rubric.txt)} $\rightarrow$ \textbf{Author: gpt-5-pro-2025-10-06} (Score 0.0): ``The student's submission merely restates the problem statement without providing any mathematical reasoning, proof, or analysis of the random vertex removal process in the d-regular graph. It includes an error message indicating a lack of content, which means no attempt was made to bound the connected component sizes or establish the high-probability guarantee. Consequently, the solution exhibits zero mathematical correctness, as it fails to engage with key concepts such as percolation thresholds, branching processes, or union bounds necessary for the proof. This complete absence of substantive work justifies the score of 0.0/1.0 under the rubric's criteria for mathematical accuracy and completeness.''
    \item \textbf{Judge: grok-4-0709 (rubric.txt)} $\rightarrow$ \textbf{Author: deepseek-v2-prover} (Score 0.0): ``The student's solution provides a correct proof that any connected acyclic graph with at least two vertices has at least two degree-1 vertices, but this is entirely unrelated to the given problem, which requires proving a bound on connected component sizes in a percolated d-regular graph. By addressing a different theorem—essentially a basic property of trees—the solution fails to engage with the problem's core elements, such as randomness, regularity, or component size analysis. The mathematical reasoning in the submission is sound for its unintended purpose, but it does not attempt to solve the assigned task, demonstrating a complete mismatch in content. No partial credit is warranted, as the response lacks any relevant mathematical progress toward the required result. Thus, the score of 0.0/1.0 is appropriate under the rubric's emphasis on correctness and relevance.''
    \item \textbf{Judge: grok-4-0709 (rubric.txt)} $\rightarrow$ \textbf{Author: deepseek-v2-prover} (Score 0.0): ``The student's solution presents a proof that any connected acyclic graph with \( n \geq 2 \) vertices has at least two vertices of degree 1, which is a standard result about trees but entirely unrelated to the given problem concerning the size of connected components in a random subgraph of a d-regular graph after vertex removal. This submission fails to address any aspect of the problem statement, including the random process, regularity, probability bound, or component size analysis. Consequently, the solution lacks any mathematical correctness or relevance to the task, warranting a score of 0.0/1.0 under the rubric's criteria for accuracy and pertinence.''
    \item \textbf{Judge: grok-4-0709 (rubric.txt)} $\rightarrow$ \textbf{Author: Claude Sonnet 4.5} (Score 0.9): ``The student's proof is mathematically sound overall, correctly applying a union bound to estimate the probability of large connected components in the percolated graph, and deriving a high-probability bound of \(O(\log n)\) on component sizes using an appropriate choice of constants. However, there is a minor inconsistency in the lemma bounding the number of connected subgraphs of size k: the statement claims at most \(n(ed)^k\), while the accompanying explanation implies \(n(ed)^{k-1}\), leading to a slight overestimate in the bound. This error does not invalidate the proof, as the subsequent calculations still yield a valid upper bound with \(\beta < 1\), ensuring the exponential decay needed for the result. The handling of the retention probability \(p = 1/(2d^2)\) and the summation over component sizes are accurate and well-justified. A small deduction is warranted for the imprecise exponent in the lemma, resulting in a score of 0.9/1.0.''
    \item \textbf{Judge: grok-4-0709 (rubric.txt)} $\rightarrow$ \textbf{Author: Claude Sonnet 4.5} (Score 0.9): ``The student's solution provides a correct probabilistic argument using union bounds to show that large connected components are unlikely in the percolated graph, with the overall proof structure and calculations leading to the desired high-probability bound on component sizes. However, there is a minor inaccuracy in the lemma, where the number of connected subgraphs of size \(k\) is overstated as \(n(ed)^k\) instead of the more precise \(n(ed)^{k-1}\), though this looser bound still suffices for the asymptotic result. The handling of the retention probability and summation over component sizes is mathematically sound, correctly yielding a positive constant \(\alpha\) depending on \(d\). The assumption \(d \geq 2\) is appropriately noted, and the choice of \(c\) ensures the exponent is negative as required. Overall, the proof is nearly flawless, warranting only a small deduction for the bound's precision.''
    \item \textbf{Judge: grok-4-0709 (rubric.txt)} $\rightarrow$ \textbf{Author: o3-deep-research} (Score 0.75): ``[Error: API Request failed - Connection error.]''
    \item \textbf{Judge: meta-llama/llama-4-maverick (rubric.txt)} $\rightarrow$ \textbf{Author: gpt-5-pro-2025-10-06} (Score 0.0): ``The student solution is incomplete and lacks any meaningful mathematical content or reasoning to address the problem statement. There is no attempt to prove that there exists a constant $\alpha$ such that, with probability at least $1 - n^{-\alpha}$, the size of every connected component in the remaining subgraph is at most $\log n$. The solution does not provide any calculations, lemmas, or theorems to support the claim. As a result, the solution does not demonstrate any understanding of the problem or its requirements. Therefore, the score of 0.0/1.0 is justified due to the absence of any valid mathematical content.''
    \item \textbf{Judge: meta-llama/llama-4-maverick (rubric.txt)} $\rightarrow$ \textbf{Author: gpt-5-pro-2025-10-06} (Score 0.0): ``The student solution is incomplete and lacks any substantial mathematical content or reasoning. The problem requires a proof that involves analyzing the random process of vertex removal in a $d$-regular graph and showing that the size of every connected component in the remaining subgraph is at most $\log n$ with high probability. The student's response does not provide any insight or steps towards this proof. As a result, the solution does not demonstrate any understanding of the problem or its required proof. Therefore, the score of 0.0/1.0 is justified due to the absence of any meaningful mathematical content.''
    \item \textbf{Judge: meta-llama/llama-4-maverick (rubric.txt)} $\rightarrow$ \textbf{Author: Gemini 3.0 Pro Preview} (Score 0.9): ``The student's solution is mathematically correct and effectively proves that with high probability, the size of every connected component in the remaining subgraph is at most $\log n$. The use of the union bound and the counting argument for connected subgraphs are sound, and the derivation of the probability bound is clear and accurate. The only minor issue is that the student could have been more precise in specifying the base of the logarithm used in the final bound, but this does not affect the overall correctness of the proof. Overall, the solution demonstrates a strong understanding of the problem and the required mathematical techniques.''
    \item \textbf{Judge: meta-llama/llama-4-maverick (rubric.txt)} $\rightarrow$ \textbf{Author: Gemini 3.0 Pro Preview} (Score 0.9): ``The student's solution is mathematically correct and effectively proves the given statement about the size of connected components in a random subgraph of a $d$-regular graph. The use of the union bound and the counting argument for connected subgraphs are appropriate and well-executed. The derivation of the probability bound is clear and accurate, and the conclusion is properly supported by the preceding steps. The minor deduction (0.1) likely stems from minor issues such as notation or exposition rather than mathematical correctness. Overall, the solution demonstrates a strong understanding of the problem and its probabilistic analysis.''
    \item \textbf{Judge: meta-llama/llama-4-maverick (rubric.txt)} $\rightarrow$ \textbf{Author: o3-deep-research} (Score 0.9): ``The student's solution is mathematically correct and effectively proves that with high probability, the size of every connected component in the remaining subgraph is at most $\log n$. The use of a union bound over all vertices and possible large connected sets is appropriate, and the derivation of the bound on the number of connected subsets containing a given vertex is sound. The simplification and analysis of the probability expression are also correct, leading to the conclusion that the desired probability is at least $1 - n^{-\alpha}$ for some constant $\alpha$. The only minor issue is that the student could have been more precise in determining the value of $\alpha$, but this does not detract from the overall correctness of the proof.''
    \item \textbf{Judge: meta-llama/llama-4-maverick (rubric.txt)} $\rightarrow$ \textbf{Author: o3-deep-research} (Score 0.9): ``The student's solution is mathematically correct and effectively proves the given statement. The proof is well-structured, and the use of a union bound over all vertices and possible large connected sets is appropriate. The key steps, including the derivation of the bound for the number of connected subsets containing a given vertex and the simplification of the probability expression, are accurate and clearly explained. The conclusion that the probability of having a connected component larger than $\ln n$ is $O(n^{-1})$, thus satisfying the required condition for some constant $\alpha$, is valid. Minor adjustments could enhance clarity, but the overall mathematical correctness and coherence justify a high score.''
    \item \textbf{Judge: meta-llama/llama-4-maverick (rubric.txt)} $\rightarrow$ \textbf{Author: Claude Sonnet 4.5} (Score 0.9): ``The student's solution is mathematically sound and effectively proves the given statement. The proof is well-structured, and the application of the union bound and the counting argument for connected subgraphs is correct. The choice of constants and the derivation of the probability bound are also appropriate. However, a minor deduction is made because the student does not explicitly verify that the chosen value of $c$ satisfies the condition for a specific $\alpha$ for all $d \geq 2$, although an example is provided for $d = 2$. Overall, the solution demonstrates a strong understanding of the problem and the required proof techniques.''
    \item \textbf{Judge: meta-llama/llama-4-maverick (rubric.txt)} $\rightarrow$ \textbf{Author: deepseek-v2-prover} (Score 0.0): ``The student's solution does not address the problem statement, which asks to prove that in a $d$-regular graph $G$ with $n$ vertices, after randomly removing vertices with probability $1 - \frac{1}{2d^2}$, the size of every connected component in the remaining subgraph is at most $\log n$ with high probability. Instead, the student proves a statement about connected, acyclic graphs having at least two vertices of degree 1, which is unrelated to the given problem. The solution lacks any discussion of the random process, $d$-regular graphs, or the required probability bound. Therefore, the solution is mathematically incorrect for the given problem. The score of 0.0/1.0 is justified due to the complete mismatch between the problem and the solution provided.''
    \item \textbf{Judge: meta-llama/llama-4-maverick (rubric.txt)} $\rightarrow$ \textbf{Author: deepseek-v2-prover} (Score 0.0): ``The student's solution does not address the problem statement, which asks to prove that in a $d$-regular graph $G$ with $n$ vertices, after randomly removing vertices with probability $1 - \frac{1}{2d^2}$, the size of every connected component in the remaining subgraph is at most $\log n$ with high probability. Instead, the student proves a statement about connected, acyclic graphs having at least two vertices of degree 1, which is unrelated to the given problem. Therefore, the solution is mathematically incorrect for the given problem. The score of 0.0/1.0 is justified due to the complete mismatch between the problem and the solution provided. The student's work does not demonstrate any attempt to tackle the actual problem.''
    \item \textbf{Judge: qwen/qwen-max (rubric.txt)} $\rightarrow$ \textbf{Author: gpt-5-pro-2025-10-06} (Score 0.0): ``The provided solution does not contain any attempt to prove the statement or analysis of the problem. The document only restates the problem and then indicates an error, suggesting that no actual solution or reasoning was provided. Without any mathematical argument or proof, it is impossible to assess the correctness or the approach of the solution. Therefore, the score of 0.0/1.0 is justified due to the lack of a valid attempt at solving the problem.''
    \item \textbf{Judge: qwen/qwen-max (rubric.txt)} $\rightarrow$ \textbf{Author: gpt-5-pro-2025-10-06} (Score 0.0): ``The student's solution is incomplete, as it only restates the problem without providing any proof or argument. There is no attempt to analyze the probability of the size of the connected components or to establish the existence of the constant $\alpha$. Since the solution does not address any of the required elements of the problem, it cannot be awarded any points.''
    \item \textbf{Judge: qwen/qwen-max (rubric.txt)} $\rightarrow$ \textbf{Author: Gemini 3.0 Pro Preview} (Score 0.75): ``The solution provided is mathematically sound and logically structured, demonstrating a clear understanding of the probabilistic method and the use of the union bound. The student correctly bounds the probability of having a large connected component and shows that with high probability, the size of every connected component is at most logarithmic in \( n \). However, the solution could be more rigorous in justifying the choice of \( \alpha \) and the base of the logarithm. Additionally, the final conclusion could be more explicitly connected to the specific \( \log n \) bound as requested in the problem statement. These minor omissions and clarifications led to a score of 0.75/1.0.''
    \item \textbf{Judge: qwen/qwen-max (rubric.txt)} $\rightarrow$ \textbf{Author: Gemini 3.0 Pro Preview} (Score 0.9): ``The solution is mathematically sound, providing a clear and rigorous proof that the size of every connected component in the remaining subgraph is at most \(O(\log n)\) with high probability. The student correctly applies the union bound, bounds the number of connected subgraphs, and uses the first moment method to show that the probability of having a component of size at least \( k \) is at most \( n^{-\alpha} \). The only minor issue is the final step where the student interprets \( \log n \) as \(O(\log n)\) rather than a specific logarithmic base, which slightly deviates from the strict problem statement. Despite this, the overall argument is correct and well-structured, justifying a score of 0.9/1.0.''
    \item \textbf{Judge: qwen/qwen-max (rubric.txt)} $\rightarrow$ \textbf{Author: o3-deep-research} (Score 0.9): ``The solution is mathematically sound, providing a detailed and rigorous proof that carefully bounds the probability of having a large connected component in the remaining subgraph. The student correctly applies the union bound, uses the properties of the $d$-regular graph, and effectively handles the probabilistic analysis to show that the probability of any component exceeding $\log n$ in size is at most $n^{-\alpha}$. The presentation is clear, and the steps are logically structured, justifying a near-perfect score. A minor deduction is made for a slight overestimation in the final probability bound, which, while not affecting the overall correctness, could be more precise.''
    \item \textbf{Judge: qwen/qwen-max (rubric.txt)} $\rightarrow$ \textbf{Author: o3-deep-research} (Score 0.9): ``The solution is mathematically sound, providing a clear and rigorous proof that the probability of having a connected component of size greater than $\log n$ in the remaining subgraph is at most $n^{-\alpha}$ for a suitable $\alpha > 0$. The student correctly applies the union bound and carefully calculates the survival probability for connected sets, leading to the desired result. The only minor issue is the choice of $\alpha = 1$ at the end, which is a specific case and not the general constant required by the problem statement. This small oversight, however, does not significantly detract from the overall correctness and completeness of the proof.''
    \item \textbf{Judge: qwen/qwen-max (rubric.txt)} $\rightarrow$ \textbf{Author: Claude Sonnet 4.5} (Score 0.9): ``The solution is mathematically sound and follows a clear, logical progression to prove the stated theorem. The student correctly applies the union bound, appropriately bounds the number of connected subgraphs, and carefully calculates the probability for the existence of large components. The choice of constants and the final derivation of the probability bound are also well- executed. A minor deduction is made for the lack of explicit verification that the chosen constants satisfy all conditions, but this is a minor issue in an otherwise thorough and correct solution.''
    \item \textbf{Judge: qwen/qwen-max (rubric.txt)} $\rightarrow$ \textbf{Author: deepseek-v2-prover} (Score 0.0): ``The student's solution is entirely incorrect as it does not address the problem statement. The problem requires proving the existence of a constant \( \alpha \) such that, with high probability, the size of every connected component in the remaining subgraph is at most \( \log n \). Instead, the student provides a proof about the existence of at least two vertices of degree 1 in a connected, acyclic graph, which is unrelated to the random process and the properties of the subgraph described in the problem. The solution fails to use any probabilistic methods or consider the random removal of vertices, and it does not address the size of the connected components in the remaining subgraph.''
    \item \textbf{Judge: qwen/qwen-max (rubric.txt)} $\rightarrow$ \textbf{Author: deepseek-v2-prover} (Score 0.0): ``The student's solution does not address the problem at hand. The problem requires proving the existence of a constant \(\alpha\) such that, with high probability, the size of every connected component in the remaining subgraph is at most \(\log n\). Instead, the student provides a proof about the existence of at least two vertices of degree 1 in a connected, acyclic graph, which is unrelated to the problem. The solution is mathematically correct in its own context but entirely off-topic, hence the score of 0.0/1.0.''
\end{itemize}

\subsection*{Problem Domain: Algorithms (Problem 9)}
\begin{itemize}
    \item \textbf{Judge: gpt-5.2-pro (course-specific-rubric.txt)} $\rightarrow$ \textbf{Author: Claude Sonnet 4.5} (Score 0.25): ``The cover-time proof is not mathematically correct: it states the identity \(H(u,v)=2|E|\,R(u,v)\), which is false (the correct identity is for *commute time* \(H(u,v)+H(v,u)=2|E|\,R_{\mathrm{eff}}(u,v)\)), so the subsequent bound \(H_{\max}=O(n)\) is unsupported. The argument then oscillates between incompatible bounds (deriving \(O(n^2)\), \(O(n^{5/3}\log n)\), and finally asserting \(O(n\log n)\) by citing a known result without proving it for the given (non-torus, boundary) mesh, so the main required proof is essentially missing. The universal traversal sequence bounds are also not justified: the “sweep” is not a UTS for arbitrarily labeled meshes (it assumes geometric direction information), and the claimed asymptotics (\(O(n^2)\), \(O(n^{4/3}\log n)\)) are asserted with handwaving references rather than a correct construction/derandomization argument. Some correct high-level ideas appear (effective resistance transience in \(d\ge3\), using known cover-time theorems), but they are not applied correctly, so only minimal partial credit is warranted.''
    \item \textbf{Judge: gpt-5.2-pro (course-specific-rubric.txt)} $\rightarrow$ \textbf{Author: Claude Sonnet 4.5} (Score 0.5): ``The solution sets up relevant notions (mesh definition, resistances/hitting times) but does not actually prove the required \(O(n\log n)\) cover-time bound: it first derives only \(H_{\max}=O(n)\) and then either incorrectly applies a “fundamental theorem” to get an \(O(n^2\log n)\) expression, or switches to unrelated bounds (e.g., a spectral-gap bound giving \(O(n^{5/3}\log n)\)) and finally cites the desired \(\Theta(n\log n)\) result for 3D tori without proving it for the finite box mesh. Key claims are unsupported or misstated (e.g., \(H(u,v)=2|E|R(u,v)\) is for commute time, and the “Matthews lemma” is not correctly stated/used), so the main argument is incomplete. The universal traversal sequence part is also not a correct derivation: the “sweep” is not a UTS for arbitrary port labelings, and the stated lengths \(O(n^2)\) and \(O(n^{4/3}\log n)\) are asserted with inconsistent calculations and no rigorous construction. Thus, the submission shows partial familiarity with relevant tools but fails to deliver correct proofs of the requested bounds, justifying only minimal partial credit (0.5/1).''
    \item \textbf{Judge: gpt-5.2-pro (course-specific-rubric.txt)} $\rightarrow$ \textbf{Author: Gemini 3.0 Pro Preview} (Score 0.9): ``The cover-time argument is essentially correct: using the commute-time identity \(C(u,v)=2mR_{\mathrm{eff}}(u,v)\), the claim \(R_{\mathrm{eff}}(u,v)=O(1)\) for a 3D grid (via a layered flow/Thomson’s principle) yields \(H_{\max}=O(n)\), and Matthews’ bound then gives \(C(G)=O(n\log n)\). The UTS bounds follow the standard Aleliunas et al. probabilistic-union-bound approach: \(U=O(C(G)\log \mathcal N)\) with \(\mathcal N\le(\Delta!)^n\) implies \(\log \mathcal N=O(n)\), leading to \(O(n^2\log n)\) in 3D and \(O(n^2(\log n)^2)\) in 2D given the stated 2D cover-time estimate. The main reason for not awarding full credit is that the effective-resistance bound in 3D is presented at a heuristic level (shell sizes/edge counts) without fully justifying boundary effects and the construction of a valid unit flow between arbitrary \(u,v\) in the finite box. Additionally, the step “take \(T\approx 2e\cdot C(G)\) so that a walk covers with probability \(\ge 1/2\)” is not rigorously derived (constants and the probability-to-UTS amplification are asserted rather than proved). These gaps are minor relative to the overall correct strategy and asymptotic conclusions, hence \(0.9/1.0\).''
    \item \textbf{Judge: gpt-5.2-pro (course-specific-rubric.txt)} $\rightarrow$ \textbf{Author: gpt-5-pro-2025-10-06} (Score 0.5): ``The submitted solution contains no mathematical argument: it restates the problem and then includes an “Error: No text content found” message instead of a proof. As a result, it does not establish the claimed \(O(n\log n)\) cover time for the 3D mesh, nor does it provide any bounds for universal traversal sequence lengths in 2D or 3D. Under the rubric, this is essentially a non- attempt with no verifiable reasoning or intermediate steps to award substantial credit. The small nonzero credit (0.5/1.0) is only for including the correct problem context and formatting, but there is no correct solution content.''
    \item \textbf{Judge: gpt-5.2-pro (course-specific-rubric.txt)} $\rightarrow$ \textbf{Author: gpt-5-pro-2025-10-06} (Score 0.25): ``The submitted “solution” contains no mathematical argument: it merely restates the problem and then includes an error message indicating no content was generated. As a result, it does not prove the cover-time bound \(O(n\log n)\) for the 3D mesh, nor does it provide any upper bounds for universal traversal sequence lengths in 2D or 3D labeled meshes. With no definitions, lemmas, or computations related to random-walk cover time or UTS constructions, there is nothing to assess for correctness beyond recognizing the task was not attempted. A minimal nonzero score (0.25/1.0) is justified only for including the problem statement/formatting, but no substantive mathematical work is present.''
    \item \textbf{Judge: gpt-5.2-pro (course-specific-rubric.txt)} $\rightarrow$ \textbf{Author: deepseek-v2-prover} (Score 0.25): ``The submitted solution does not address the stated problem about cover time of 3D meshes and universal traversal sequences; instead it proves an unrelated claim about uniqueness of transitive reduction in directed graphs. Consequently, none of the required arguments (e.g., bounding cover time by \(O(n\log n)\) for the 3D grid, or giving UTS length bounds in 2D/3D labeled meshes) are present, so the core mathematical content is missing. Moreover, even within the unrelated topic, the “cycle implies multiple transitive reductions” argument is not correct as stated: removing one edge from a directed cycle generally destroys strong connectivity and does not preserve the same reachability relation. A small amount of partial credit is warranted only for presenting coherent proof structure and some correct background statements about transitive reductions in DAGs, but it earns essentially no credit toward the asked question.''
    \item \textbf{Judge: gpt-5.2-pro (course-specific-rubric.txt)} $\rightarrow$ \textbf{Author: Gemini 3.0 Pro Preview} (Score 0.9): ``The cover-time argument for the 3D mesh is essentially correct: it uses the commute-time/effective-resistance identity, gives a valid \(R_{\mathrm{eff}}(u,v)=O(1)\) layering/flow heuristic for 3D, and then applies Matthews’ bound to obtain \(O(n\log n)\). The UTS bounds also follow the standard Aleliunas et al. probabilistic-method template, with an appropriate union bound over port labelings and the estimate \(\log \mathcal N = O(n)\) for fixed \(\Delta\). The small deduction is because several key steps (especially the effective-resistance bound via Thomson’s principle and the “random walk length \(O(C(G))\) covers with constant probability” claim) are stated somewhat informally without fully justifying boundary effects/finite-vs-infinite lattice comparisons and the constants in the probabilistic argument. Additionally, the 2D cover-time bound \(O(n(\log n)^2)\) is correct as an upper bound but is not tightly justified (it relies on asserted resistance scaling without proof). Overall, the solution is mathematically sound in its main conclusions but not fully rigorous in a few intermediate lemmas, warranting \(0.9/1.0\).''
    \item \textbf{Judge: gpt-5.2-pro (course-specific-rubric.txt)} $\rightarrow$ \textbf{Author: o3-deep-research} (Score 0.5): ``The cover-time argument is largely heuristic: it assumes a mixing time of \(O(n)\) for the 3D mesh and then treats block endpoints as “nearly independent” stationary samples to get \(\Pr(T_{\text{hit}}(v)>t)\le e^{-t/(2n)}\), but no correct bound on mixing/hitting tails is proved, so the key lemma is unsupported.  Consequently, the union-bound conclusion \(T_{\mathrm{cov}}=O(n\log n)\) is not rigorously established (even if the final order is plausible).  For universal traversal sequences, the probabilistic-method step is incorrect: showing each fixed labeling is traversed with probability \(\ge 1/2\) does not imply the existence of a single sequence that works for all labelings simultaneously, and the “symmetry” paragraph is not a valid replacement for a union bound (which would require exponentially small failure probability given exponentially many labelings). The claimed 2D bound is stated without proof and relies on an unproven cover- time fact.  Partial credit is warranted for identifying the right high-level connection between cover time and traversal-sequence length, but the solution lacks the necessary rigorous arguments.''
    \item \textbf{Judge: gpt-5.2-pro (course-specific-rubric.txt)} $\rightarrow$ \textbf{Author: o3-deep-research} (Score 0.5): ``The cover-time part has the right target bound, but the key lemma is not proved: it assumes a mixing time of \(O(n)\) for the 3D mesh and then treats block endpoints as “nearly independent” without citing or establishing the needed quantitative mixing/decoupling bound (and the resulting \(\exp(-t/(2n))\) tail is therefore unsupported). The union-bound “coupon collector after mixing” argument is plausible but not mathematically justified as written because it depends critically on those missing mixing estimates. In the UTS part, the probabilistic-method step is incorrect: showing that for each fixed labeling a random sequence succeeds with probability \(>1/2\) does **not** imply there exists a single sequence that succeeds for **all** labelings simultaneously, and the subsequent “symmetry/permutation” claim is not valid. Thus only partial credit is warranted for identifying the correct asymptotic bounds and giving a heuristic outline, but the proof logic has major gaps.''
    \item \textbf{Judge: gpt-5.2-pro (course-specific-rubric.txt)} $\rightarrow$ \textbf{Author: deepseek-v2-prover} (Score 0.25): ``The submitted solution does not address the stated problem about cover time of 3D meshes and universal traversal sequences, and instead proves an unrelated claim about uniqueness of transitive reductions in directed graphs. As a result, it provides no argument establishing the required \(O(n\log n)\) cover- time bound, nor any bounds for universal traversal sequence lengths in 2D/3D labeled meshes. Even within its own (irrelevant) topic, the cycle argument is not correct as written: removing one edge from a directed cycle typically destroys strong connectivity/reachability around the cycle, so the “any \(k-1\) edges suffice” claim is false without additional edges. Therefore only minimal partial credit is warranted for presenting coherent proof structure and use of standard graph concepts, but it does not solve the assigned question.''
    \item \textbf{Judge: claude-opus-4-5-20251101 (rubric.txt)} $\rightarrow$ \textbf{Author: Claude Sonnet 4.5} (Score 0.5): ``The student correctly sets up the problem framework and identifies key concepts like effective resistance and hitting times for the 3D mesh cover time analysis. However, the proof contains significant gaps: the claim that effective resistance is O(1) in a finite 3D mesh is stated but not rigorously proven, and the final derivation of \(O(n \log n)\) cover time relies on citing results for tori without properly adapting them to bounded meshes. For the UTS bounds, the 2D mesh analysis is incomplete—the student claims \(O(n)\) for systematic traversal but then asserts \(O(n^2)\) without clear justification, and the 3D bound derivation conflates different techniques without rigorous proof. The mathematical structure demonstrates understanding of the relevant tools, but the proofs lack the rigor needed to establish the claimed bounds definitively. The score reflects partial credit for correct problem setup and identification of relevant techniques, offset by incomplete and sometimes inconsistent arguments.''
    \item \textbf{Judge: claude-opus-4-5-20251101 (rubric.txt)} $\rightarrow$ \textbf{Author: Claude Sonnet 4.5} (Score 0.25): ``The student correctly sets up the problem framework and identifies relevant concepts like effective resistance and hitting times, but the proof of the main cover time result is fundamentally flawed. The claim that effective resistance is O(1) in a finite 3D mesh is incorrectly justified by appealing to infinite lattice properties, and the subsequent derivation contradicts itself—first obtaining \(O(n^2 \log n)\), then attempting various ``refined'' bounds without rigorous justification. The spectral gap analysis yields \(O(n^{5/3} \log n)\), not the claimed \(O(n \log n)\), and the student ultimately appeals to an external result without proving it. For the UTS bounds, the constructions are vague and the length analyses lack rigor, with the 2D bound jumping from \(O(n)\) to \(O(n^2)\) without proper justification. The solution demonstrates familiarity with relevant techniques but fails to provide mathematically correct proofs for the main claims.''
    \item \textbf{Judge: claude-opus-4-5-20251101 (rubric.txt)} $\rightarrow$ \textbf{Author: gpt-5-pro-2025-10-06} (Score 0.0): ``The student solution contains no mathematical content whatsoever—only the problem statement repeated and an error message indicating no response was generated. There is no proof that the cover time of the 3D mesh is \(O(n \log n)\), no analysis of random walk properties, hitting times, or the relationship between edges and vertices. Similarly, there are no upper bounds established for universal traversal sequences in either two or three dimensions. The submission fails to address any component of the problem and demonstrates zero mathematical work or reasoning. A score of 0.0/1.0 is fully justified as no credit can be awarded for an empty solution.''
    \item \textbf{Judge: claude-opus-4-5-20251101 (rubric.txt)} $\rightarrow$ \textbf{Author: gpt-5-pro-2025-10-06} (Score 0.0): ``The student submission contains no mathematical content whatsoever—only an error message stating ``No text content found in response.'' The problem required proving that the cover time of a 3D mesh is \(O(n \log n)\) and establishing upper bounds for universal traversal sequences in 2D and 3D labeled meshes. No attempt was made to analyze the random walk on the mesh, apply Matthews' theorem or other cover time bounds, or derive any universal traversal sequence lengths. Since the submission provides zero mathematical reasoning or results, it earns no credit. A score of 0.0/1.0 is fully justified.''
    \item \textbf{Judge: claude-opus-4-5-20251101 (rubric.txt)} $\rightarrow$ \textbf{Author: o3-deep-research} (Score 0.5): ``The student correctly establishes the \(O(n \log n)\) cover time for the 3D mesh using a reasonable mixing time argument and union bound approach, though the proof relies on informal coupling arguments rather than rigorous spectral analysis. The claim that 3D mesh mixing time is \(O(n)\) is stated without proper justification—the actual mixing time for a 3D mesh of side length \(L = n^{1/3}\) is \(O(L^2)\) = \(O(n^{2/3})\), which still supports the final result but the reasoning is imprecise. For universal traversal sequences, the student's probabilistic argument is fundamentally flawed: showing that a random sequence succeeds with probability > 1/2 on any fixed labeling does not imply existence of a single sequence working for all labelings simultaneously, as the number of possible labelings is exponential. The 2D cover time bound of \(O(n(\log n)^2)\) is correctly stated, but the UTS length bounds lack rigorous derivation. The score reflects partial credit for the cover time analysis while penalizing the incomplete UTS argument.''
    \item \textbf{Judge: claude-opus-4-5-20251101 (rubric.txt)} $\rightarrow$ \textbf{Author: o3-deep-research} (Score 0.5): ``The student correctly establishes the \(O(n \log n)\) cover time for the 3D mesh using a mixing time argument combined with a coupon-collector analysis, though the proof sketch relies on informal coupling arguments rather than rigorous bounds. The claim that 2D mesh cover time is \(O(n(\log n)^2)\) is stated without proof, which is problematic since this is a non-trivial result requiring careful analysis of 2D random walk recurrence properties. The universal traversal sequence argument using the probabilistic method is conceptually sound but contains a logical gap: showing that a random sequence succeeds with probability >1/2 on each fixed labeling does not immediately imply existence of a single sequence working for all labelings without a proper union bound over the (finite but exponentially large) set of labelings. The mathematical framework is appropriate but key technical details are either missing or insufficiently rigorous, particularly the transition from cover time bounds to UTS existence.''
    \item \textbf{Judge: claude-opus-4-5-20251101 (rubric.txt)} $\rightarrow$ \textbf{Author: deepseek-v2-prover} (Score 0.0): ``The student's solution is completely unrelated to the problem statement. The problem asks for a proof that the cover time of a 3D mesh is \(O(n \log n)\) and for upper bounds on universal traversal sequences for labeled meshes in 2D and 3D. Instead, the student submitted a proof about transitive reductions of directed graphs and when they are unique (if and only if the graph is acyclic). There is no discussion of random walks, cover times, mesh graphs, or universal traversal sequences anywhere in the solution. The mathematical content provided, while potentially correct for a different problem, earns zero credit as it does not address any aspect of the assigned problem.''
    \item \textbf{Judge: claude-opus-4-5-20251101 (rubric.txt)} $\rightarrow$ \textbf{Author: deepseek-v2-prover} (Score 0.0): ``The student's solution addresses a completely different problem than what was asked. The problem requested a proof that the cover time of a 3D mesh is \(O(n \log n)\) and upper bounds for universal traversal sequences on labeled meshes in 2D and 3D. Instead, the student provided a proof about transitive reductions of directed graphs and their relationship to acyclicity. There is no discussion of random walks, cover times, mesh graphs, or universal traversal sequences anywhere in the solution. The mathematical content presented, while potentially correct for the unrelated problem it addresses, earns no credit as it does not engage with any aspect of the assigned problem.''
    \item \textbf{Judge: gemini-3-pro-preview (rubric.txt)} $\rightarrow$ \textbf{Author: gpt-5-pro-2025-10-06} (Score 0.0): ``The student submission contains only the problem statement and an error message indicating that no text content was generated ("Error: No text content found in response..."). As there is no mathematical work, proof, or solution provided to evaluate, no credit can be awarded.''
    \item \textbf{Judge: gemini-3-pro-preview (rubric.txt)} $\rightarrow$ \textbf{Author: gpt-5-pro-2025-10-06} (Score 0.0): ``The student submission contains no solution to the problem posed. The document merely restates the problem and then includes an error message indicating that no text content was generated. As no mathematical work or reasoning is present to evaluate, a score of 0 is appropriate.''
    \item \textbf{Judge: gemini-3-pro-preview (rubric.txt)} $\rightarrow$ \textbf{Author: deepseek-v2-prover} (Score 0.0): ``The student's submission addresses a completely different problem regarding directed acyclic graphs and transitive reductions, rather than the assigned problem about the cover time of a 3D mesh and universal traversal sequences. As the solution contains no relevant work or results pertaining to the random walk properties on grid graphs requested in the prompt, it earns no credit. The mathematical content provided, while internally consistent for a graph theory problem, is entirely off-topic.''
    \item \textbf{Judge: gemini-3-pro-preview (rubric.txt)} $\rightarrow$ \textbf{Author: deepseek-v2-prover} (Score 0.0): ``The student's submission addresses a completely different problem regarding directed acyclic graphs and transitive reductions, rather than the assigned problem about the cover time of a three-dimensional mesh and universal traversal sequences. As the solution does not attempt to solve the given problem statement at all, no credit can be awarded.''
    \item \textbf{Judge: gemini-3-pro-preview (rubric.txt)} $\rightarrow$ \textbf{Author: Claude Sonnet 4.5} (Score 0.5): ``The student correctly identifies the cover time of the 3D mesh as \(O(n \log n)\) and provides a plausible heuristic based on effective resistance, although the derivation using Matthews' bound is slightly muddled (summing $H_{\max} \cdot \frac{n}{n-k}$ yields $H_{\max} \log n$, which would be \(O(n \log n)\) given $H_{\max}=O(n)$, but the student incorrectly claims this sum yields \(O(n^2 \log n)\) before pivoting to a spectral argument). However, the section on Universal Traversal Sequences is fundamentally flawed. The student attempts to construct explicit deterministic traversal algorithms (like ``sweeping'') rather than addressing the standard probabilistic bounds or specific UTS constructions for labeled graphs where edge labels are arbitrary and unknown. Consequently, the stated bounds for UTS ($O(n^2)$ for 2D and $O(n^{4/3} \log n)$ for 3D) are incorrect; the standard non-constructive bound for $d$-regular graphs is $O(n^3 \log n)$, and specific bounds for grids are generally polynomial but not the specific values derived here via simple geometric sweeps.''
    \item \textbf{Judge: gemini-3-pro-preview (rubric.txt)} $\rightarrow$ \textbf{Author: Gemini 3.0 Pro Preview} (Score 0.9): ``The student correctly derives the cover time for the 3D mesh using effective resistance and the Matthews bound. However, the derivation for the Universal Traversal Sequence (UTS) length contains a subtle error in the counting of labeled graphs $\mathcal{N}$. While the number of labelings is indeed $(\Delta!)^n$, the standard probabilistic construction for UTS requires considering the number of *distinguishable* labeled graphs, which includes the starting vertex; more importantly, the bound \(O(n^2 \log n)\) for 3D and \(O(n^2 (\log n)^2)\) for 2D are correct based on the student's logic, but the student fails to account for the tighter bound on the number of bits required to specify the graph (or the number of non-isomorphic labeled graphs), which is often \(O(n \log n)\) rather than just \(O(n)\) for general graphs, though for fixed grids the structure is known. Actually, the main issue is that for specific graph classes like grids, one must be careful about whether the ``labeling'' implies just the port mapping or the graph structure itself; the student assumes fixed structure and only port variation, which yields \(\log \mathcal{N}\)''
    \item \textbf{Judge: gemini-3-pro-preview (rubric.txt)} $\rightarrow$ \textbf{Author: Claude Sonnet 4.5} (Score 0.25): ``The student correctly identifies the effective resistance and hitting time properties of the 3D mesh, leading to the correct cover time bound of \(O(n \log n)\), though the derivation contains contradictory statements regarding Matthews' bound. However, the section on Universal Traversal Sequences is fundamentally flawed; the student attempts to construct explicit deterministic traversal algorithms (like ``sweeping'') rather than addressing the probabilistic existence or specific UTS construction bounds for *labeled* graphs where edge labels are arbitrary. The claimed UTS bounds ($O(n^2)$ for 2D and $O(n^{4/3} \log n)$ for 3D) are incorrect and significantly underestimate the known polynomial bounds (typically $O(n^3)$ or higher depending on the specific construction used for general regular graphs or specific grid properties). Consequently, while the cover time result is correct, the UTS section demonstrates a misunderstanding of the definition and complexity of the problem.''
    \item \textbf{Judge: gemini-3-pro-preview (rubric.txt)} $\rightarrow$ \textbf{Author: o3-deep-research} (Score 0.5): ``The student correctly derives the cover time for the 3D mesh as \(O(n \log n)\) using a standard probabilistic argument involving mixing times and the coupon collector problem. However, the section on Universal Traversal Sequences (UTS) is fundamentally flawed. The student attempts to use a probabilistic method argument to prove the existence of a *single* sequence that works for *all* labelings by showing a random sequence works for *one* labeling with high probability, but they fail to apply the union bound over the set of all possible labelings. Since the number of labelings is exponential in $n$, a simple random walk of length \(O(n \log n)\) is insufficient to guarantee a UTS; the correct bound for a UTS derived via this method would be polynomial in $n$ (specifically involving the number of edges), typically $O(n^3)$ or similar, not \(O(n \log n)\).''
    \item \textbf{Judge: gemini-3-pro-preview (rubric.txt)} $\rightarrow$ \textbf{Author: o3-deep-research} (Score 0.5): ``The student correctly states the cover time for the 3D mesh is \(O(n \log n)\), but the derivation relies on the incorrect assumption that the mixing time is \(O(n)\) (it is actually \(O(n^{2/3})\) for a 3D mesh). Furthermore, the section on Universal Traversal Sequences (UTS) is fundamentally flawed; the probabilistic method requires a union bound over all possible labelings, which are exponential in number, meaning a random walk of length equal to the cover time is insufficient to guarantee a single sequence works for *all* labelings simultaneously. The standard bound for UTS constructed via the probabilistic method is $O(n^3 \log n)$ or similar polynomial factors significantly larger than the cover time, making the claimed bounds of \(O(n \log n)\) and $O(n (\log n)^2)$ incorrect for UTS.''
    \item \textbf{Judge: deepseek-chat (rubric.txt)} $\rightarrow$ \textbf{Author: Claude Sonnet 4.5} (Score 0.5): ``The solution incorrectly analyzes the cover time, claiming an \(O(n \log n)\) bound but deriving \(O(n^2 \log n)\) or \(O(n^{5/3} \log n)\) from flawed reasoning about hitting times and spectral gaps. The UTS length bounds are also asserted without rigorous justification, lacking proper construction details or references to known results. While the student demonstrates familiarity with concepts like effective resistance and Matthews' bound, the mathematical arguments are inconsistent and incomplete. Therefore, the score reflects partial credit for attempted structure but failure to establish correct proofs.''
    \item \textbf{Judge: deepseek-chat (rubric.txt)} $\rightarrow$ \textbf{Author: Claude Sonnet 4.5} (Score 0.25): ``The solution incorrectly analyzes the cover time, claiming \(O(n \log n)\) but deriving \(O(n^2 \log n)\) or \(O(n^{5/3} \log n)\) from flawed spectral gap and resistance arguments. The correct bound requires sophisticated methods (e.g., Matthews' theorem with effective resistances), which are misapplied. The universal traversal sequence constructions are heuristic and lack rigorous derivation, failing to justify the stated bounds. Overall, the mathematical reasoning is substantially incorrect or incomplete.''
    \item \textbf{Judge: deepseek-chat (rubric.txt)} $\rightarrow$ \textbf{Author: Gemini 3.0 Pro Preview} (Score 0.9): ``The solution correctly proves the cover time bound for the 3D mesh using effective resistance and Matthews' bound, and provides reasonable UTS length estimates. However, the UTS analysis contains a subtle error: the number of distinct labeled meshes is overcounted as \((\Delta!)^n\), but the actual count is bounded by \(\prod_{v} d(v)! \leq (\Delta!)^n\), which is fine. The error lies in concluding \(U = O(C(G) \cdot n)\); the correct bound from the probabilistic method is \(O(C(G) \cdot \log \mathcal{N}) = O(C(G) \cdot n)\), which is technically correct but the derivation's step using \(2^{-k} \cdot \mathcal{N} < 1\) is slightly misapplied (it should consider a union bound over all graphs, leading to \(k > \log \mathcal{N}\)). The final UTS bounds are still valid up to constants, but the explanation lacks precision.''
    \item \textbf{Judge: deepseek-chat (rubric.txt)} $\rightarrow$ \textbf{Author: Gemini 3.0 Pro Preview} (Score 0.75): ``The solution correctly proves the 3D mesh cover time is \(O(n \log n)\) using effective resistance and Matthews' bound, and the UTS length derivations follow the standard probabilistic method. However, the UTS bounds are overstated: the correct upper bounds should be \(O(n \log n \cdot \log \mathcal{N}) = O(n^2 \log n)\) for 3D and \(O(n (\log n)^2 \cdot \log \mathcal{N}) = O(n^2 (\log n)^2)\) for 2D, but the student erroneously includes an extra factor of \(n\) in the final expressions, yielding \(O(n^2 \log n)\) and \(O(n^2 (\log n)^2)\) instead of the tighter \(O(n \log n)\) and \(O(n (\log n)^2)\) that follow from \(\log \mathcal{N} = O(n)\). This mistake in the final step reduces the score.''
    \item \textbf{Judge: deepseek-chat (rubric.txt)} $\rightarrow$ \textbf{Author: gpt-5-pro-2025-10-06} (Score 0.0): ``The student's solution is completely empty aside from a placeholder error message, providing no mathematical content whatsoever. It fails to address any part of the problem statement, including the proof for the cover time bound or the discussion of universal traversal sequences. Consequently, it earns a score of 0 for lacking any demonstration of understanding or attempt at a solution.''
    \item \textbf{Judge: deepseek-chat (rubric.txt)} $\rightarrow$ \textbf{Author: gpt-5-pro-2025-10-06} (Score 0.0): ``The student solution is completely empty aside from a placeholder error message, providing no mathematical content whatsoever. Therefore, it fails to address any part of the problem statement, which requires a proof about cover time and bounds for universal traversal sequences. Since no reasoning, steps, or conclusions are presented, the solution earns zero points for mathematical correctness and completeness. The submission does not constitute a valid attempt at solving the given problem.''
    \item \textbf{Judge: deepseek-chat (rubric.txt)} $\rightarrow$ \textbf{Author: o3-deep-research} (Score 0.5): ``The solution correctly identifies the cover time bounds and uses the probabilistic method to derive UTS lengths, but the proof for the 3D mesh cover time is flawed. The coupling argument is imprecise and lacks rigorous justification for the independence assumptions needed to apply coupon-collector reasoning. Additionally, the UTS existence argument relies on an unsubstantiated symmetry claim about labelings, which is insufficient for a rigorous proof. These gaps in mathematical rigor justify the reduced score.''
    \item \textbf{Judge: deepseek-chat (rubric.txt)} $\rightarrow$ \textbf{Author: o3-deep-research} (Score 0.5): ``The solution correctly identifies the cover time bounds and uses the probabilistic method to derive UTS lengths, but the proof for the 3D mesh cover time is flawed. The coupling argument is informal and lacks rigorous justification for the independence assumptions and mixing time analysis, which are non-trivial for meshes. Additionally, the UTS existence argument implicitly assumes a fixed graph size \(n\), but universal sequences must work for all labelings across all mesh sizes, requiring a more careful treatment. These gaps in mathematical rigor justify the partial score.''
    \item \textbf{Judge: deepseek-chat (rubric.txt)} $\rightarrow$ \textbf{Author: deepseek-v2-prover} (Score 0.0): ``The student's solution is completely irrelevant to the problem statement, which asks for an analysis of cover time and universal traversal sequences on a 3D mesh graph. Instead, the student provides a proof about transitive reductions in directed acyclic graphs, addressing a different mathematical topic entirely. This demonstrates a fundamental failure to engage with the given problem, resulting in zero mathematical correctness. Therefore, a score of 0.0 is fully justified.''
    \item \textbf{Judge: deepseek-chat (rubric.txt)} $\rightarrow$ \textbf{Author: deepseek-v2-prover} (Score 0.0): ``The student's solution is completely unrelated to the problem statement, which asks for a proof about the cover time of a 3D mesh and bounds on universal traversal sequences. Instead, the student submitted a proof about transitive reductions in directed acyclic graphs. This demonstrates a fundamental failure to address the given problem, resulting in a score of 0 for mathematical correctness and relevance.''
    \item \textbf{Judge: grok-4-0709 (rubric.txt)} $\rightarrow$ \textbf{Author: gpt-5-pro-2025-10-06} (Score 0.0): ``The student's submission restates the problem but provides no mathematical proof for the cover time of the three-dimensional mesh being \(O(n \log n)\), nor any upper bounds for universal traversal sequences in two or three dimensions. Instead, it contains an error message indicating a lack of content, resulting in zero mathematical progress toward solving the required tasks. This complete absence of substantive analysis or derivation demonstrates no understanding or application of graph theory concepts such as random walks, cover times, or traversal sequences. Consequently, the solution fails all criteria for mathematical correctness and completeness as per the rubric.''
    \item \textbf{Judge: grok-4-0709 (rubric.txt)} $\rightarrow$ \textbf{Author: gpt-5-pro-2025-10-06} (Score 0.0): ``The student's submission merely restates the problem statement without providing any proof or analysis for the cover time of the three-dimensional mesh or the upper bounds on universal traversal sequences in two and three dimensions. This lack of mathematical content fails to demonstrate any understanding or engagement with the required concepts, such as random walks, graph traversal, or asymptotic bounds. Consequently, the solution contains no verifiable mathematical correctness, warranting a score of 0.0/1.0 under the rubric's criteria for complete and accurate solutions.''
    \item \textbf{Judge: grok-4-0709 (rubric.txt)} $\rightarrow$ \textbf{Author: Claude Sonnet 4.5} (Score 0.5): ``The student's proof for the cover time of the 3D mesh contains significant mathematical inaccuracies, including an initial derivation of \(O(n^2 \log n)\) followed by an erroneous adjustment to \(O(n^{5/3} \log n)\) using the spectral gap, and an unsubstantiated claim of \(O(n \log n)\) by referencing tori without rigorous justification or addressing the grid's boundary effects. While the effective resistance bound of O(1) is correctly recalled for \(d \ge 3\) grids, the application to hitting times and the Matthews bound is mishandled, leading to inconsistent upper bounds that fail to convincingly establish the desired \(O(n \log n)\). For universal traversal sequences, the 2D bound of \(O(n^2)\) is stated but the provided construction incorrectly suggests \(O(n)\), lacking a proper accounting for arbitrary labelings and universal coverage. Similarly, the 3D UTS bound of \(O(n^{4/3} \log n)\) is asserted, but the layered construction yields \(O(n \log n)\) in the student's calculation, revealing a mathematical inconsistency without sufficient derandomization or complexity analysis to support the claimed bound. Overall, the solution demonstrates some understanding of key concepts like resistance and traversals but is undermined by flawed derivations and incomplete proofs, warranting partial credit.''
    \item \textbf{Judge: grok-4-0709 (rubric.txt)} $\rightarrow$ \textbf{Author: Claude Sonnet 4.5} (Score 0.25): ``The student's proof for the cover time of the 3D mesh contains significant mathematical inaccuracies, including an incorrect application of hitting time bounds (initially deriving \(O(n^2 \log n)\) before erroneously adjusting to \(O(n \log n)\) without resolving contradictions) and a misuse of spectral gap analysis that yields an intermediate \(O(n^{5/3} \log n)\) bound, ultimately relying on unproven claims about tori rather than meshes. Furthermore, the lemma on effective resistance is stated without rigorous justification, as the appeal to transience in infinite lattices does not directly bound resistances in finite meshes, leading to flawed hitting time estimates. For universal traversal sequences, the 2D bound of \(O(n^2)\) is plausible but the construction is superficial and fails to address labelings adequately, while the 3D bound of \(O(n^{4/3} \log n)\) includes arithmetic errors in length calculations (e.g., \(O(n^{1/3} \cdot n^{2/3} \log n)\) = \(O(n \log n)\), not \(O(n^{4/3} \log n)\)) and lacks a valid derandomization argument. Overall, the solution demonstrates partial understanding of key concepts like random walks and traversals but is undermined by numerous errors and incomplete reasoning, warranting a score of 0.25/1.0 per the rubric's emphasis on correctness and rigor.''
    \item \textbf{Judge: grok-4-0709 (rubric.txt)} $\rightarrow$ \textbf{Author: deepseek-v2-prover} (Score 0.0): ``The student's solution addresses a theorem about unique transitive reductions in directed graphs, which is entirely unrelated to the given problem concerning cover times and universal traversal sequences in two- and three-dimensional meshes. This mismatch renders the submission mathematically incorrect for the assigned task, as it fails to engage with any aspect of graph cover times, random walks, or traversal sequences on grid graphs. While the provided proof on transitive reductions appears logically sound in isolation, it does not fulfill the problem's requirements, leading to a complete lack of relevance. No partial credit is warranted under the rubric, as the response demonstrates no understanding or attempt to solve the specified mathematical problem.''
    \item \textbf{Judge: grok-4-0709 (rubric.txt)} $\rightarrow$ \textbf{Author: deepseek-v2-prover} (Score 0.0): ``The student's solution addresses a completely unrelated problem about unique transitive reductions in directed graphs, failing to engage with the given problem on cover times and universal traversal sequences for mesh graphs. This mismatch renders the submission mathematically incorrect for the assigned task, as it provides no relevant analysis or proof for the three-dimensional mesh's cover time bound of \(O(n \log n)\) or the requested upper bounds in two and three dimensions. The provided proofs, while potentially correct in their own context, do not apply to graph traversal or random walks on meshes, demonstrating a fundamental misunderstanding or error in problem selection. Consequently, the solution earns a score of 0.0/1.0 under the rubric's emphasis on correctness and relevance to the problem statement.''
    \item \textbf{Judge: grok-4-0709 (rubric.txt)} $\rightarrow$ \textbf{Author: o3-deep-research} (Score 0.5): ``The student's proof for the 3D mesh cover time being \(O(n \log n)\) is partially correct in stating the bound, but the lemma's argument relies on an incorrect assumption that the mixing time is \(O(n)\), whereas it is actually \(O(n^{2/3})\) for a 3D grid, leading to flaws in the block partitioning and hitting time bounds. Although the high-level coupon-collector intuition is sound and the final bound holds (as established in the literature), the provided derivation lacks rigor and contains technical inaccuracies. For universal traversal sequences, the probabilistic construction is a valid approach, but the existence argument fails mathematically: claiming that a sequence succeeding with probability 1/2 per labeling implies one that works for all labelings ignores the need for a union bound over the exponentially many (in \(n\)) possible labelings, rendering the proof incorrect. The upper bounds for 2D and 3D UTS are stated correctly in order, but the justification does not hold up. Overall, these errors in mathematical correctness warrant partial credit for effort and partial accuracy.''
    \item \textbf{Judge: grok-4-0709 (rubric.txt)} $\rightarrow$ \textbf{Author: Gemini 3.0 Pro Preview} (Score 0.9): ``The student's solution demonstrates strong mathematical correctness in proving the \(O(n \log n)\) cover time for the 3D mesh, accurately applying effective resistance bounds, commute times, and Matthews' theorem, with the resistance argument correctly showing convergence via summed shell energies. The upper bounds for universal traversal sequences in 2D and 3D are derived properly using the probabilistic method, correctly bounding the number of labelings and incorporating cover times to yield \(O(n^2 (\log n)^2)\) for 2D and \(O(n^2 \log n)\) for 3D. However, the resistance calculation in 3D relies on approximations (e.g., shell sizes \(\sim r^2\) and current distributions) without fully rigorous error terms, slightly diminishing precision. Additionally, the UTS bound overlooks an explicit \(\log n\) factor from accounting for all starting vertices across labelings, though it does not affect the asymptotic order. Overall, these minor issues warrant a small deduction from perfection, resulting in a score of 0.9/1.0.''
    \item \textbf{Judge: grok-4-0709 (rubric.txt)} $\rightarrow$ \textbf{Author: o3-deep-research} (Score 0.5): ``The student's proof for the cover time of the 3D mesh is partially correct but relies on a handwavy lemma with an unsubstantiated mixing time of \(O(n)\), whereas the actual mixing time is \(O(n^{2/3} \log n)\), and the tail bound on hitting times lacks rigorous justification, leading to an incomplete argument. For universal traversal sequences, the proposed bounds match known results, and the random sequence approach is a valid starting point. However, the existence argument via the probabilistic method is mathematically flawed: claiming that since at least half of sequences work for each fixed labeling, there exists one that works for all labelings simultaneously is incorrect, as it ignores the exponential number of labelings and fails to apply a proper union bound or derandomization. This error renders the UTS proof invalid, though the cover time section shows some understanding of the coupon-collector intuition. Overall, these issues warrant half credit for partial correctness in the cover time analysis but failure in the UTS construction.''
    \item \textbf{Judge: meta-llama/llama-4-maverick (rubric.txt)} $\rightarrow$ \textbf{Author: Claude Sonnet 4.5} (Score 0.75): ``The student's solution demonstrates a strong understanding of the problem, correctly proving that the cover time of the three-dimensional mesh is \(O(n \log n)\) and establishing upper bounds for the lengths of universal traversal sequences for labeled meshes in both two and three dimensions. The mathematical derivations, including the use of effective resistance and the commute time identity, are largely correct and well-supported. However, some steps, such as the application of Matthews' theorem and the spectral gap analysis, could be more rigorously justified. Additionally, the construction of the universal traversal sequence for the 3D mesh relies on a somewhat informal ``layered approach'' that, while plausible, lacks detailed proof. Overall, the solution is mostly correct but could benefit from more precise and detailed mathematical arguments.''
    \item \textbf{Judge: meta-llama/llama-4-maverick (rubric.txt)} $\rightarrow$ \textbf{Author: Claude Sonnet 4.5} (Score 0.75): ``The student's solution demonstrates a strong understanding of the problem, correctly proving that the cover time of a three-dimensional mesh is \(O(n \log n)\) and establishing upper bounds for universal traversal sequences in both two and three dimensions. The mathematical derivations, including the use of effective resistance and the commute time identity, are largely correct and well-supported. However, some minor issues arise in the application of Matthews' theorem and the spectral gap analysis, where the explanation could be more precise. Additionally, the construction of the universal traversal sequence for the 3D mesh relies on the RVW technique without fully detailing its adaptation to grids. Overall, the solution is mathematically sound, but could benefit from more detailed explanations in certain areas, justifying a score of 0.75/1.0.''
    \item \textbf{Judge: meta-llama/llama-4-maverick (rubric.txt)} $\rightarrow$ \textbf{Author: Gemini 3.0 Pro Preview} (Score 0.9): ``The student's solution is mathematically sound and effectively addresses the problem statement. The proof for the cover time of the 3D mesh is rigorous, utilizing electrical resistance techniques and Matthews' bound correctly. The application of the probabilistic method to establish upper bounds for universal traversal sequences is also appropriate. However, minor deductions may have been made for slight oversights or lack of detail in certain steps, such as the justification for the effective resistance bound or the counting of labelings. Overall, the solution demonstrates a strong understanding of the underlying concepts and techniques.''
    \item \textbf{Judge: meta-llama/llama-4-maverick (rubric.txt)} $\rightarrow$ \textbf{Author: Gemini 3.0 Pro Preview} (Score 0.9): ``The student's solution is mathematically sound and correctly proves that the cover time of a 3D mesh is \(O(n \log n)\) using electrical resistance techniques and Matthews' bound. The derivation of upper bounds for the lengths of universal traversal sequences for labeled meshes in both 2D and 3D is also correct, leveraging the probabilistic method and appropriate counting of labelings. The calculations for commute time, hitting time, and cover time are accurate and well-justified. Minor deductions might be considered for notational inconsistencies or slight deviations from the most straightforward presentation, but the mathematical correctness is strong. Overall, the solution demonstrates a thorough understanding of the required concepts and techniques.''
    \item \textbf{Judge: meta-llama/llama-4-maverick (rubric.txt)} $\rightarrow$ \textbf{Author: gpt-5-pro-2025-10-06} (Score 0.0): ``The student's solution is incomplete and lacks any substantial mathematical content or reasoning. The problem requires a proof that the cover time of a three-dimensional mesh graph is \(O(n \log n)\) and establishing upper bounds for universal traversal sequences, but the student's submission does not address these tasks. As a result, the solution fails to demonstrate any understanding or attempt to solve the problem. The absence of relevant mathematical arguments or explanations justifies a score of 0.0/1.0. The 'rubric.txt' criteria likely emphasize the importance of correct and complete mathematical solutions.''
    \item \textbf{Judge: meta-llama/llama-4-maverick (rubric.txt)} $\rightarrow$ \textbf{Author: gpt-5-pro-2025-10-06} (Score 0.0): ``The student's solution is essentially blank, containing no mathematical derivations or arguments to address the problem statement. The task required proving that the cover time of a three-dimensional mesh graph is \(O(n \log n)\) and establishing upper bounds for universal traversal sequences in two and three dimensions. Since the student provided no content, they failed to demonstrate any understanding or attempt to solve the problem. As a result, the solution is mathematically incorrect due to its lack of substance. Therefore, a score of 0.0/1.0 is justified based on the absence of any meaningful mathematical content.''
    \item \textbf{Judge: meta-llama/llama-4-maverick (rubric.txt)} $\rightarrow$ \textbf{Author: o3-deep-research} (Score 0.9): ``The student's solution is mathematically sound and correctly proves that the cover time of a 3D mesh graph is \(O(n \log n)\) and establishes upper bounds for the lengths of universal traversal sequences for labeled meshes in both 2D and 3D. The proof is well-structured, and the use of the probabilistic method to establish the existence of universal traversal sequences is appropriate. The mathematical derivations, including the lemma and its application to bound the cover time, are correct and well-explained. Minor issues, such as not explicitly stating some constants or not detailing certain ``standard coupling arguments,'' do not detract significantly from the overall correctness. The score of 0.9/1.0 reflects the high quality of the mathematical content and the minor room for improvement in presentation or detail.''
    \item \textbf{Judge: meta-llama/llama-4-maverick (rubric.txt)} $\rightarrow$ \textbf{Author: o3-deep-research} (Score 0.9): ``The student's solution is mathematically sound and correctly proves that the cover time of a 3D mesh graph is \(O(n \log n)\) and establishes upper bounds for the lengths of universal traversal sequences for labeled meshes in both 2D and 3D. The proof is well-structured, and the use of the probabilistic method to establish the existence of universal traversal sequences is appropriate. The mathematical derivations, including the lemma on the probability of not visiting a vertex and the application of Markov's inequality, are correct and well-explained. Minor issues, such as the omission of some routine formalizations, prevent the score from being perfect. Overall, the solution demonstrates a strong understanding of the mathematical concepts involved.''
    \item \textbf{Judge: meta-llama/llama-4-maverick (rubric.txt)} $\rightarrow$ \textbf{Author: deepseek-v2-prover} (Score 0.0): ``The student's solution does not address the problem statement, instead providing a proof about transitive reductions in directed graphs. The problem statement asks to prove that the cover time of a three-dimensional mesh graph is \(O(n \log n)\) and to establish upper bounds for the lengths of universal traversal sequences for labeled meshes in two and three dimensions. Since the student's solution is unrelated to the problem, it is mathematically incorrect in this context. The score of 0.0/1.0 is justified due to the failure to address the problem statement. The student's work does not demonstrate any understanding or attempt to solve the given problem.''
    \item \textbf{Judge: meta-llama/llama-4-maverick (rubric.txt)} $\rightarrow$ \textbf{Author: deepseek-v2-prover} (Score 0.0): ``The student's solution does not address the problem statement, which asks to prove that the cover time of a three-dimensional mesh graph is \(O(n \log n)\) and establish upper bounds for the lengths of universal traversal sequences for labeled meshes in both two and three dimensions. Instead, the student provides a proof about the uniqueness of transitive reduction in directed acyclic graphs (DAGs), which is unrelated to the given problem. The solution is mathematically correct but entirely off-topic. Therefore, the score is 0.0/1.0 due to the lack of relevance to the problem statement. The student's work demonstrates no attempt to address the specific questions posed.''
    \item \textbf{Judge: qwen/qwen-max (rubric.txt)} $\rightarrow$ \textbf{Author: Claude Sonnet 4.5} (Score 0.75): ``The solution demonstrates a thorough understanding of the cover time for a three-dimensional mesh and correctly applies advanced concepts such as effective resistance, spectral gap, and the Matthews bound. For the universal traversal sequences (UTS), the student provides a detailed construction and analysis, but the upper bound for the 3D mesh UTS is slightly off. The correct bound for the 3D mesh UTS should be \( O(n^2) \) rather than \( O(n^{4/3} \log n) \), as the construction and analysis for 3D UTS are more complex and typically yield a quadratic bound. Despite this, the overall approach and the majority of the content are mathematically sound, justifying a score of 0.75/1.0.''
    \item \textbf{Judge: qwen/qwen-max (rubric.txt)} $\rightarrow$ \textbf{Author: Claude Sonnet 4.5} (Score 0.75): ``The solution provides a thorough and mathematically sound analysis for the cover time of the 3D mesh and the upper bounds for universal traversal sequences. The cover time proof is well-structured, using appropriate theorems and lemmas, and the final bound of \(O(n \log n)\) is correctly derived. However, the UTS bound for the 3D mesh is given as \(O(n^{4/3} \log n)\), which, while not incorrect, is not the tightest known bound. The optimal UTS bound for 3D meshes is \(O(n \log n)\), and the solution could have been more precise in this regard, leading to a slight deduction in the score.''
    \item \textbf{Judge: qwen/qwen-max (rubric.txt)} $\rightarrow$ \textbf{Author: gpt-5-pro-2025-10-06} (Score 0.0): ``The student's solution is incomplete as it restates the problem without providing any proof or argument. There is no attempt to prove the cover time of the graph or to establish upper bounds for the lengths of universal traversal sequences. Therefore, the solution does not meet any of the criteria for mathematical correctness, completeness, or understanding, leading to a score of 0.0/1.0.''
    \item \textbf{Judge: qwen/qwen-max (rubric.txt)} $\rightarrow$ \textbf{Author: gpt-5-pro-2025-10-06} (Score 0.0): ``The student's solution is incomplete, as it only restates the problem without providing any proof or analysis. There is no attempt to prove the cover time of the three-dimensional mesh or to establish upper bounds for the lengths of universal traversal sequences, which are the core requirements of the problem. Therefore, the score of 0.0/1.0 is justified due to the lack of any substantive mathematical content or reasoning.''
    \item \textbf{Judge: qwen/qwen-max (rubric.txt)} $\rightarrow$ \textbf{Author: o3-deep-research} (Score 0.9): ``The solution provided is mathematically sound, offering a detailed and rigorous proof for the cover time of a 3D mesh and the upper bounds for universal traversal sequences in both 2D and 3D. The student effectively uses probabilistic methods and leverages the properties of random walks to establish the required bounds. The reasoning and the steps are clear and well-justified, with appropriate use of lemmas and probabilistic arguments. The solution also includes insightful remarks and covers the necessary theoretical background. A minor deduction is made for potential improvements in clarity and a more explicit discussion on the derandomization process, which could further enhance the solution.''
    \item \textbf{Judge: qwen/qwen-max (rubric.txt)} $\rightarrow$ \textbf{Author: o3-deep-research} (Score 0.9): ``The solution is mathematically sound, providing a rigorous proof for the cover time of a 3D mesh and the upper bounds for universal traversal sequences in both 2D and 3D meshes. The student correctly applies concepts from random walks, mixing times, and probabilistic methods, and the arguments are well- structured and clearly explained. A minor deduction is made for the slight over-simplification in the probabilistic argument where more rigorous handling of the tail bounds and the derandomization step could have been provided. Overall, the solution demonstrates a strong understanding of the problem and the underlying mathematical principles.''
    \item \textbf{Judge: qwen/qwen-max (rubric.txt)} $\rightarrow$ \textbf{Author: deepseek-v2-prover} (Score 0.0): ``The provided solution does not address the problem statement. The problem asks for a proof regarding the cover time of a three-dimensional mesh and the upper bounds for the lengths of universal traversal sequences in two and three dimensions. Instead, the solution discusses the uniqueness of transitive reductions in acyclic directed graphs, which is entirely unrelated. Therefore, the solution is mathematically incorrect in the context of the given problem, leading to a score of 0.0/1.0.''
    \item \textbf{Judge: qwen/qwen-max (rubric.txt)} $\rightarrow$ \textbf{Author: deepseek-v2-prover} (Score 0.0): ``The provided solution does not address the problem statement; it instead discusses the relationship between acyclic graphs and their unique transitive reductions, which is entirely unrelated to the cover time of a three- dimensional mesh or the bounds for universal traversal sequences. The solution lacks any analysis or proof concerning the cover time or the lengths of universal traversal sequences for the specified meshes. Therefore, the response is non-responsive and mathematically irrelevant to the given problem, justifying the score of 0.0/1.0.''
\end{itemize}

\subsection*{Problem Domain: Algorithms (Problem 19)}
\begin{itemize}
    \item \textbf{Judge: gpt-5.2-pro (course-specific-rubric.txt)} $\rightarrow$ \textbf{Author: gpt-5-pro-2025-10-06} (Score 0.75): ``The solution correctly identifies that the stated equivalence is false by giving a valid counterexample (the \(1,1,2\) triangle) showing property (1) is not necessary for MST uniqueness, and it correctly states and proves the standard characterization that uniqueness of the maximum-weight edge on every cycle is equivalent to a unique MST. It also correctly proves that property (1) implies uniqueness (sufficiency) via a cut exchange argument. However, because the problem explicitly asks to prove “unique MST iff (1) and (2),” the write-up does not deliver the requested biconditional proof as stated, but instead replaces it with a corrected theorem. Additionally, in the \((i)\Rightarrow(ii)\) direction the argument “there is a unique edge of \(C\) not contained in \(T\)” is misstated (a tree may omit more than one edge of a given cycle), though the intended exchange idea can be repaired. These issues justify a minor deduction while giving substantial credit for the largely correct core results and reasoning.''
    \item \textbf{Judge: gpt-5.2-pro (course-specific-rubric.txt)} $\rightarrow$ \textbf{Author: Gemini 3.0 Pro Preview} (Score 0.75): ``[Error: OpenAI Responses API failed - Connection error.]''
    \item \textbf{Judge: gpt-5.2-pro (course-specific-rubric.txt)} $\rightarrow$ \textbf{Author: Claude Sonnet 4.5} (Score 0.5): ``The proof of “(\(\Leftarrow\)) both properties imply uniqueness” is essentially correct, but it only uses property (1); while this is logically sufficient, it does not demonstrate any role for property (2) beyond being an unused assumption. More importantly, the “(\(\Rightarrow\)) uniqueness implies property (2)” argument is flawed/incomplete: the claim in Case 1 that a tree cannot contain two edges from a given cycle is false (a tree can contain many edges of a cycle, just not all of them), so that case does not yield a contradiction. The remaining analysis treats only the subcase “exactly one of the two max-weight edges is in \(T\)” and does not address the possibility that neither is in \(T\), so the proof of (2) is not complete. Because one direction is partially correct but property (2) is not rigorously established, a half-credit score (0.5/1.0) is justified.''
    \item \textbf{Judge: gpt-5.2-pro (course-specific-rubric.txt)} $\rightarrow$ \textbf{Author: Gemini 3.0 Pro Preview} (Score 0.5): ``[Error: OpenAI Responses API failed - Connection error.]''
    \item \textbf{Judge: gpt-5.2-pro (course-specific-rubric.txt)} $\rightarrow$ \textbf{Author: Claude Sonnet 4.5} (Score 0.75): ``The proof of “unique MST \(\Rightarrow\) (1)” is essentially correct, but it incorrectly asserts that when \(e_1\) is removed from \(T\), the resulting components “must be precisely” \(S\) and \(V\setminus S\); in general the cut \((S,V\setminus S)\) with two equal light edges need not coincide with the fundamental cut of \(e_1\), so the exchange argument as written does not fully justify that \(T-\{e_1\}+\{e_2\}\) is a spanning tree.  The argument for property (2) is also not fully rigorous: “Case 1” is muddled and unnecessary, and in “Case 2” it appeals to “for \(T\) to be an MST we must have \(w(f)\ge w(e_1)\)” without clearly citing the standard cycle/cut exchange lemma guaranteeing such a comparison.  The reverse direction (properties \(\Rightarrow\) uniqueness) is solid and correctly uses the cut property together with uniqueness of the light edge across the cut induced by removing \(e\).  Overall, the main ideas are present but there are substantive logical gaps in the forward direction, warranting a deduction from full credit.''
    \item \textbf{Judge: gpt-5.2-pro (course-specific-rubric.txt)} $\rightarrow$ \textbf{Author: o3-deep-research} (Score 0.25): ``The solution’s “only if” direction has a serious gap in proving property (2): after adding \(f_1\) to \(T\), it is not justified that the created cycle \(C'\) must contain \(f_2\), nor that no edge on the \(T\)-path can be heavier than \(f_1,f_2\); either failure breaks the exchange argument.  More critically, the “if” direction misuses property (1): from removing the maximum- weight edge \(e^*\) of a cycle, the student asserts \(e^*\) is the unique *minimum* edge crossing the induced cut, which does not follow and is generally false.  Because that incorrect claim is exactly what drives the contradiction (\(w(e^*)<w(f)\) vs. \(w(e^*)>w(f)\)), the proof of uniqueness under (1) and (2) is invalid.  Some standard structural facts (symmetric difference decomposes into cycles; tree exchange ideas) are mentioned correctly, but the key inequalities and cut/cycle logic are not established.  Hence only minimal partial credit is warranted (0.25/1.0).''
    \item \textbf{Judge: gpt-5.2-pro (course-specific-rubric.txt)} $\rightarrow$ \textbf{Author: o3-deep-research} (Score 0.5): ``The solution has the right overall structure (using cut/cycle exchange arguments and the symmetric difference of two spanning trees), but key steps are mathematically incorrect. In the “unique MST \(\Rightarrow\) (2)” direction, it is not justified that when adding \(f_1\) to \(T\), the resulting fundamental cycle \(C'\) must contain the other maximum edge \(f_2\); the cycle created is determined by the unique \(T\)-path between endpoints of \(f_1\) and need not coincide with the original cycle \(C\). In the “(1)\&(2) \(\Rightarrow\) unique MST” direction, the argument misuses property (1): after removing the maximum edge \(e^*\) from \(T_1\), property (1) does not imply that \(e^*\) is the unique minimum-weight edge crossing the induced cut (it only states that such a minimum exists and is unique), so the inequality \(w(e^*)<w(f)\) is unjustified. Because these gaps break both directions of the “if and only if,” the proof is only partially correct, warranting substantial partial credit (0.5/1.0).''
    \item \textbf{Judge: gpt-5.2-pro (course-specific-rubric.txt)} $\rightarrow$ \textbf{Author: deepseek-v2-prover} (Score 0.25): ``The submitted solution does not address the stated problem about unique minimum spanning trees; instead it discusses an NP-hardness reduction from 3Partition to “12Partition,” which is unrelated to MST cut/cycle properties. As a result, it provides no arguments for either direction of the required equivalence (“unique MST \(\Leftrightarrow\) unique light edge across every cut and unique heavy edge on every cycle”). There are also internal inconsistencies (e.g., it claims each 12-element subset has “one original element and eleven zeros,” but later correctly uses 3 original elements and 9 zeros), further weakening correctness even for the unrelated claim. Hence only minimal credit is awarded for presenting a coherent proof structure (definition, reduction outline, correctness cases), but essentially none for solving the posed MST theorem.''
    \item \textbf{Judge: gpt-5.2-pro (course-specific-rubric.txt)} $\rightarrow$ \textbf{Author: deepseek-v2-prover} (Score 0.25): ``The submitted solution does not address the stated MST uniqueness problem at all; instead it presents an NP-hardness reduction for a different problem (“12Partition”), so it fails to prove either direction of the required “iff” statement about unique minimum spanning trees. None of the key cut/cycle arguments are present (e.g., using the cut property to derive (1) and the cycle property to derive (2), and conversely using these uniqueness conditions to force a single MST). As a result, the work contains essentially no relevant mathematical content toward the given theorem. A small amount of credit is awarded only for providing a coherent, logically structured proof in a different context, but it is not applicable to the assigned question. Hence the very low score (0.25/1.0).''
    \item \textbf{Judge: claude-opus-4-5-20251101 (rubric.txt)} $\rightarrow$ \textbf{Author: Gemini 3.0 Pro Preview} (Score 0.75): ``The student correctly proves that conditions (1) and (2) are each sufficient for MST uniqueness, demonstrating solid understanding of the cut and cycle properties. The proof that uniqueness implies condition (2) via contrapositive is mathematically sound. However, the student incorrectly claims that condition (1) is not necessary, providing a flawed counterexample—in a path graph with vertices {u,v,w}, the partition {v} vs {u,w} has edges (u,v) and (v,w) both crossing the cut with equal weight, but this doesn't contradict the problem since both edges must be in any spanning tree anyway. The student fails to prove that uniqueness implies condition (1), which is actually true and required for the complete biconditional proof. The mathematical analysis is largely correct but incomplete, missing the full necessity direction for condition (1).''
    \item \textbf{Judge: claude-opus-4-5-20251101 (rubric.txt)} $\rightarrow$ \textbf{Author: Gemini 3.0 Pro Preview} (Score 0.9): ``The student correctly proves that conditions (1) and (2) are each sufficient for MST uniqueness, demonstrating solid understanding of the cut and cycle properties. The proof that condition (2) is necessary (via contrapositive) is mathematically sound. However, the student incorrectly claims that condition (1) is not necessary, providing a counterexample with a path graph—but this counterexample is flawed because in a path graph with vertices {u,v,w}, the partition {v} vs {u,w} has edges (u,v) and (v,w) which are indeed both minimum- weight crossing edges, yet the MST is still unique. The problem statement requires proving the biconditional with BOTH conditions, and the student's analysis that condition (1) fails to be necessary undermines the complete proof of the reverse direction. The mathematical content is largely correct but the incomplete treatment of necessity for condition (1) prevents a perfect score.''
    \item \textbf{Judge: claude-opus-4-5-20251101 (rubric.txt)} $\rightarrow$ \textbf{Author: Claude Sonnet 4.5} (Score 0.5): ``The student correctly proves the forward direction, showing that MST uniqueness implies both properties (1) and (2) through valid cut and cycle arguments. However, the backward direction proof is incomplete: the student only uses property (1) to argue uniqueness but never utilizes property (2), which is essential since both conditions are necessary for the equivalence. The proof claims that any edge in T crossing a cut must be the unique minimum-weight edge, but this reasoning alone doesn't establish why both properties together are needed. A complete proof would need to show that property (1) alone is insufficient and that property (2) plays a crucial role, typically through Kruskal's or Prim's algorithm analysis. The mathematical reasoning in the forward direction is sound, but the backward direction's failure to incorporate property (2) represents a significant gap in the proof.''
    \item \textbf{Judge: claude-opus-4-5-20251101 (rubric.txt)} $\rightarrow$ \textbf{Author: Claude Sonnet 4.5} (Score 0.75): ``The student correctly proves the forward direction, showing that MST uniqueness implies both the unique minimum cut-crossing edge property and the unique maximum cycle edge property, with valid exchange arguments. The backward direction proof correctly uses property (1) to show that any edge in one MST must be in all MSTs, establishing uniqueness. However, the proof of the backward direction does not explicitly use property (2), which is a significant gap since the problem requires showing that both conditions together are necessary and sufficient. The student's argument that property (1) alone implies uniqueness is actually correct, but this suggests the problem statement may require both conditions for a complete characterization, and the student should have addressed why property (2) is needed. The mathematical reasoning is sound where presented, but the incomplete utilization of property (2) in the backward direction represents a notable deficiency in the proof's completeness.''
    \item \textbf{Judge: claude-opus-4-5-20251101 (rubric.txt)} $\rightarrow$ \textbf{Author: o3-deep-research} (Score 0.5): ``The student correctly proves the forward direction, showing that uniqueness of the MST implies both properties (1) and (2) through valid contradiction arguments using the cut and cycle properties. However, the reverse direction contains a significant error: the student claims that in Case 1, property (1) implies $w(e^*) < w(f)$ because $e^*$ is the unique minimum-weight edge crossing the cut, but this reasoning is flawed since $e^*$ being in $T_1$ does not guarantee it is the minimum-weight cut edge—the cut property only ensures MSTs contain *some* minimum cut edge, not that every MST edge is the minimum for its corresponding cut. The symmetric difference argument structure is correct, but the application of property (1) to derive the contradiction is mathematically incorrect. The proof would need to properly argue why the unique minimum cut edge property forces agreement between any two MSTs on cut edges. This fundamental error in the reverse direction justifies the partial credit score.''
    \item \textbf{Judge: claude-opus-4-5-20251101 (rubric.txt)} $\rightarrow$ \textbf{Author: o3-deep-research} (Score 0.75): ``The student correctly proves the forward direction, showing that a unique MST implies both properties (1) and (2) through valid contradiction arguments using the cut and cycle properties. However, the reverse direction contains a significant error: the claim that the symmetric difference of two spanning trees ``decomposes into a collection of simple cycles'' is incorrect—the symmetric difference forms a forest of even-degree subgraphs, not necessarily cycles. Additionally, in Case 1, the reasoning conflates the cycle C in the symmetric difference with cut properties in a way that doesn't properly establish the contradiction, as the edge f from T\_2 crossing the cut need not lie on the same cycle C. The overall proof structure demonstrates understanding of the key concepts, but the technical execution in the reverse direction has gaps that prevent it from being fully rigorous.''
    \item \textbf{Judge: claude-opus-4-5-20251101 (rubric.txt)} $\rightarrow$ \textbf{Author: deepseek-v2-prover} (Score 0.0): ``The student's solution is completely unrelated to the problem statement. The problem asks for a proof about the uniqueness of minimum spanning trees in edge-weighted graphs, involving properties about minimum-weight cut edges and maximum-weight cycle edges. Instead, the student submitted a reduction proof showing that a ``12Partition'' problem is NP-hard by reducing from 3Partition. This demonstrates no understanding of the actual problem, which concerns graph theory concepts like MSTs, cuts, and cycles. Since the solution addresses an entirely different problem and contains no relevant mathematical content for the stated question, a score of 0.0/1.0 is appropriate.''
    \item \textbf{Judge: claude-opus-4-5-20251101 (rubric.txt)} $\rightarrow$ \textbf{Author: deepseek-v2-prover} (Score 0.0): ``The student's solution is completely unrelated to the problem statement. The problem asks for a proof about the uniqueness of minimum spanning trees in edge-weighted graphs, involving properties about minimum-weight cut edges and maximum-weight cycle edges. Instead, the student submitted a solution about reducing 3Partition to 12Partition, which is an entirely different problem concerning NP-hardness of partition problems. There is no discussion of graphs, spanning trees, cuts, or cycles anywhere in the submission. The score of 0.0/1.0 is justified because the solution addresses none of the required content.''
    \item \textbf{Judge: gemini-3-pro-preview (rubric.txt)} $\rightarrow$ \textbf{Author: Gemini 3.0 Pro Preview} (Score 0.9): ``The student correctly proves the sufficiency of both conditions and the necessity of the cycle condition (2). Furthermore, the student astutely identifies that the cut condition (1) is not actually necessary for a unique MST, providing a valid counterexample (a path graph with equal weights) to disprove the ``only if'' direction of the problem statement regarding condition (1). The score reflects a deduction because, while the student's mathematical insight is superior to the flawed problem statement, the prompt likely expected a proof of the equivalence for a specific class of graphs or a discussion on why the prompt might be flawed, rather than simply dismissing the necessity of (1) without addressing if the problem implies a specific context (like distinct edge weights, though that makes it trivial). However, given the student is mathematically correct that the statement as written is false, the high score is justified.''
    \item \textbf{Judge: gemini-3-pro-preview (rubric.txt)} $\rightarrow$ \textbf{Author: Claude Sonnet 4.5} (Score 0.5): ``The student's solution correctly proves the forward direction ($\Rightarrow$), showing that a unique MST implies both the unique cut property and the unique cycle property. However, the proof for the reverse direction ($\Leftarrow$) is flawed because the problem statement is actually false as written; the condition ``unique MST iff (1) AND (2)'' is incorrect (it should be OR, or rather, (1) implies unique MST and (2) implies unique MST, but the conjunction is not the standard characterization). While the student provides a valid proof that condition (1) alone is sufficient for uniqueness, they fail to address the logical issue in the problem statement or provide a counterexample to the ``if and only if'' claim involving the conjunction of both properties, effectively proving a stronger statement than necessary for sufficiency but missing the nuance that the problem statement itself is mathematically suspect as an ``iff'' condition requiring *both*. Since the student proved the implication as requested without noting the error in the problem statement (specifically that (1) implies (2) is not necessarily true, though (1) implies unique MST is true), the score reflects partial credit for the correct derivations provided.''
    \item \textbf{Judge: gemini-3-pro-preview (rubric.txt)} $\rightarrow$ \textbf{Author: deepseek-v2-prover} (Score 0.0): ``The student's submission addresses a completely different problem ("12Partition") involving NP-hardness reductions, rather than the assigned problem concerning unique minimum spanning trees in edge-weighted graphs. As the solution does not attempt to prove the required ``if and only if'' statement regarding graph cuts and cycles, it receives no credit.''
    \item \textbf{Judge: gemini-3-pro-preview (rubric.txt)} $\rightarrow$ \textbf{Author: deepseek-v2-prover} (Score 0.0): ``The student's submission addresses a completely different problem ("12Partition" and NP-hardness) than the one assigned (uniqueness of Minimum Spanning Trees in edge-weighted graphs). As the solution contains no relevant work, definitions, or proofs related to the graph theory problem specified in the prompt, no credit can be awarded.''
    \item \textbf{Judge: gemini-3-pro-preview (rubric.txt)} $\rightarrow$ \textbf{Author: gpt-5-pro-2025-10-06} (Score 0.75): ``The student correctly identifies that the problem statement contains a false premise, specifically that condition (1) is not necessary for a unique MST, and provides a valid counterexample (a triangle with weights 1, 1, 2). The student then rigorously proves that condition (2) is equivalent to the uniqueness of the MST and that condition (1) is merely sufficient. While the student successfully disproved the prompt's claim, a full score is withheld because the prompt likely intended for the student to prove the standard equivalence involving condition (2) or perhaps misunderstood the context of ``both properties'' (though the student's mathematical analysis of the provided text is superior to the text itself). The score reflects a strong mathematical understanding and correct proofs, penalized slightly only for not attempting to salvage the specific ``if and only if'' structure requested, even if it required pointing out the flaw.''
    \item \textbf{Judge: gemini-3-pro-preview (rubric.txt)} $\rightarrow$ \textbf{Author: o3-deep-research} (Score 0.5): ``The score of 0.5/1.0 reflects that while the student correctly proved the ``if'' direction (sufficiency), the ``only if'' direction (necessity) is mathematically incorrect because the problem statement itself is false as written. A graph having a unique MST does *not* imply that every cut has a unique minimum edge or that every cycle has a unique maximum edge (e.g., a path graph with identical edge weights has a unique MST but fails condition 1). The student attempted to prove a false statement by making logical errors in the construction of the contradiction (specifically, assuming an arbitrary edge on a cycle or cut must belong to the unique MST without sufficient justification). Therefore, full credit is awarded for the valid direction, but no credit is given for the proof of the false implication.''
    \item \textbf{Judge: gemini-3-pro-preview (rubric.txt)} $\rightarrow$ \textbf{Author: o3-deep-research} (Score 0.75): ``The student's proof for the forward direction ($\Rightarrow$) is flawed. In the proof for Property (2), the student incorrectly assumes that adding an edge $f_1$ (not in $T$) to the MST $T$ creates a cycle $C'$ that necessarily contains the other max-weight edge $f_2$. This is not generally true; the cycle created by adding $f_1$ consists of $f_1$ and the unique path in $T$ between its endpoints, which might not include $f_2$. Consequently, the contradiction derived based on swapping $f_1$ and $f_2$ is invalid because $f_2$ might not be on the cycle available to be swapped. The reverse direction ($\Leftarrow$) is handled correctly, but the significant logical gap in the forward proof justifies the score deduction.''
    \item \textbf{Judge: gemini-3-pro-preview (rubric.txt)} $\rightarrow$ \textbf{Author: Claude Sonnet 4.5} (Score 0.75): ``The student's proof for the forward direction ($\Rightarrow$) is logically sound, correctly deriving contradictions to the uniqueness of the MST if either property fails. However, the proof for the reverse direction ($\Leftarrow$) is incomplete because it only utilizes Property (1) to prove uniqueness, failing to incorporate Property (2) or explain why Property (1) alone is sufficient (which it is not for general graphs, though the problem statement implies an ``and'' condition). The prompt asks to prove uniqueness if *both* properties hold, but the student's argument suggests Property (1) is sufficient on its own, which misses the nuance of the problem statement or potentially implies a misunderstanding of the necessity of both conditions in the context of the specific problem phrasing. Consequently, while the logic presented is correct for the steps taken, the failure to engage with the second condition in the sufficiency proof warrants a deduction.''
    \item \textbf{Judge: deepseek-chat (rubric.txt)} $\rightarrow$ \textbf{Author: gpt-5-pro-2025-10-06} (Score 0.75): ``The solution correctly identifies that property (2) is necessary and sufficient for a unique MST, and that property (1) is sufficient but not necessary, providing a valid counterexample. However, the problem statement explicitly asks to prove the equivalence of a unique MST to *both* properties (1) and (2) together. By declaring the original statement false and proving a different theorem, the solution fails to address the problem as given, constituting a significant logical error. Therefore, while mathematically insightful, it does not fulfill the required proof.''
    \item \textbf{Judge: deepseek-chat (rubric.txt)} $\rightarrow$ \textbf{Author: gpt-5-pro-2025-10-06} (Score 0.75): ``The student correctly identifies that property (2) is both necessary and sufficient for a unique MST, and provides a valid counterexample showing property (1) is not necessary, which is mathematically accurate. However, the problem statement explicitly asks to prove the given equivalence (requiring both properties), so the solution fails to address the problem as stated, instead refuting it. The mathematical reasoning within the corrected framework is sound, but the response does not fulfill the core proof obligation. Therefore, a significant deduction is warranted.''
    \item \textbf{Judge: deepseek-chat (rubric.txt)} $\rightarrow$ \textbf{Author: Gemini 3.0 Pro Preview} (Score 0.5): ``The solution incorrectly claims that condition (1) is not necessary for uniqueness, but the problem statement requires proving the equivalence of uniqueness with **both** conditions together. The provided counterexample fails because condition (2) is not satisfied vacuously—the graph has no cycles, so condition (2) holds trivially, but condition (1) is indeed violated, demonstrating that both conditions are not simultaneously satisfied despite uniqueness. Thus, the student's analysis misunderstands the logical structure of the ``if and only if'' claim. The proof fails to establish the necessity of condition (1) as part of the conjunction, leading to a significant mathematical error.''
    \item \textbf{Judge: deepseek-chat (rubric.txt)} $\rightarrow$ \textbf{Author: Gemini 3.0 Pro Preview} (Score 0.75): ``The solution correctly proves sufficiency of both conditions and necessity of the cycle condition, but it incorrectly claims condition (1) is not necessary, contradicting the required biconditional proof. The provided counterexample is flawed: the partition {v} vs {u,w} does not yield two crossing edges since (w,v) is not an edge in the described path graph. This error undermines the proof of necessity for condition (1), which is actually required for the ``if and only if'' statement.''
    \item \textbf{Judge: deepseek-chat (rubric.txt)} $\rightarrow$ \textbf{Author: Claude Sonnet 4.5} (Score 0.5): ``The proof of the forward direction (\(\Rightarrow\)) is incomplete and contains a logical error in establishing property (2). The argument incorrectly assumes that if a cycle has two maximum-weight edges, at least one must be in the MST, which is not justified. A correct proof would use the cycle property of MSTs. The reverse direction (\(\Leftarrow\)) is correctly argued. Therefore, the solution is partially correct, warranting a score of 0.5.''
    \item \textbf{Judge: deepseek-chat (rubric.txt)} $\rightarrow$ \textbf{Author: Claude Sonnet 4.5} (Score 0.75): ``The solution correctly proves the forward direction (uniqueness implies properties) but the reverse direction is incomplete: it only uses property (1) and ignores property (2). For a full equivalence, property (2) is also necessary to ensure no alternative MST can arise from cycle-edge swaps. The argument implicitly assumes property (2) is redundant, which is a logical gap. Therefore, while the forward proof is solid, the reverse proof is insufficient for full credit.''
    \item \textbf{Judge: deepseek-chat (rubric.txt)} $\rightarrow$ \textbf{Author: o3-deep-research} (Score 0.5): ``The solution correctly proves the forward direction but contains a critical error in the reverse direction. Specifically, the symmetric difference \( T_1 \triangle T_2 \) does not necessarily decompose into simple cycles because it is a set of edges from two trees, which can form a forest, not necessarily cycles. This invalidates the cycle-based argument. Therefore, while the forward direction is fully correct, the reverse direction is incomplete and flawed, justifying the reduced score.''
    \item \textbf{Judge: deepseek-chat (rubric.txt)} $\rightarrow$ \textbf{Author: o3-deep-research} (Score 0.75): ``The solution correctly proves both directions of the equivalence, but the argument in the (\(\Leftarrow\)) direction contains a subtle flaw. Specifically, the claim that the symmetric difference \( T_1 \triangle T_2 \) decomposes into simple cycles is not fully justified; it actually yields a collection of alternating paths and cycles, and the reasoning that a chosen cycle \( C \) lies entirely within the symmetric difference requires more careful handling. This oversight slightly weakens the mathematical rigor, though the core idea is sound. Therefore, a small deduction is warranted.''
    \item \textbf{Judge: deepseek-chat (rubric.txt)} $\rightarrow$ \textbf{Author: deepseek-v2-prover} (Score 0.0): ``The student's solution is completely unrelated to the problem statement, which asks for a proof about the uniqueness of minimum spanning trees in weighted graphs. Instead, the student submitted a proof about the NP-hardness of a ``12Partition'' problem. This demonstrates a fundamental failure to address the given mathematical proposition, resulting in a score of zero for mathematical correctness.''
    \item \textbf{Judge: deepseek-chat (rubric.txt)} $\rightarrow$ \textbf{Author: deepseek-v2-prover} (Score 0.0): ``The student's solution is completely unrelated to the problem statement, which asks for a proof about the uniqueness of minimum spanning trees in weighted graphs. Instead, the student submitted a proof about the NP-hardness of a ``12Partition'' problem. This is a fundamental mismatch; the solution addresses a different problem entirely and contains no relevant mathematical content for the given question. Therefore, it earns a score of zero for failing to engage with the required topic.''
    \item \textbf{Judge: grok-4-0709 (rubric.txt)} $\rightarrow$ \textbf{Author: gpt-5-pro-2025-10-06} (Score 0.75): ``The student's solution correctly identifies that the original problem statement's ``if and only if'' claim is flawed, provides a valid counterexample, and accurately states that property (2) is necessary and sufficient for a unique MST while property (1) is sufficient but not necessary. However, the proof of Theorem A (i) \(\Rightarrow\) (ii) contains a mathematical error: the assumption that every cycle omits exactly one edge from the unique MST is incorrect, as a spanning tree may omit multiple edges from a given cycle, undermining the subsequent argument that at least one of the maximum-weight edges must be in the MST. The proof of the converse (ii) \(\Rightarrow\) (i) and Proposition B are mathematically sound and rigorous. This error deducts from a perfect score, but the overall insight and correction demonstrate strong understanding.''
    \item \textbf{Judge: grok-4-0709 (rubric.txt)} $\rightarrow$ \textbf{Author: Gemini 3.0 Pro Preview} (Score 0.75): ``The student's solution demonstrates strong mathematical insight by correctly proving that both conditions (1) and (2) are sufficient for the uniqueness of the minimum spanning tree (MST), and that condition (2) is necessary, with rigorous proofs for Theorems 1, 2, and 3. However, the solution identifies a flaw in the problem statement by providing a valid counterexample showing that condition (1) is not necessary for MST uniqueness, which contradicts the ``if and only if'' claim. This counterexample is mathematically sound, as the path graph with equal-weight edges has a unique MST but violates condition (1) for certain partitions. While the analysis is mostly accurate and well-structured, the score deduction reflects the incomplete alignment with proving the exact ``if and only if'' statement as posed, though the core proofs are correct. Overall, the solution excels in depth but slightly deviates by challenging the necessity of condition (1) rather than assuming the statement's validity.''
    \item \textbf{Judge: grok-4-0709 (rubric.txt)} $\rightarrow$ \textbf{Author: Gemini 3.0 Pro Preview} (Score 0.75): ``The student's solution demonstrates strong mathematical insight by correctly proving that both conditions (1) and (2) are sufficient for MST uniqueness and that (2) is necessary, with clear and rigorous arguments supported by theorems and proofs. However, the claim that condition (1) is not necessary, while backed by a valid counterexample (a path graph with equal-weight edges), overlooks that the problem's iff statement holds in contexts where graphs have distinct edge weights or cycles, leading to an incomplete alignment with the problem's exact requirements. The counterexample accurately shows a case where MST is unique without (1), but the solution's assertion that the problem statement is ``technically incorrect'' detracts from fully addressing the bidirectional proof as posed. Overall, the response excels in depth and correctness for the sufficiency and necessity of (2), but partial credit is deducted for not fully reconciling the counterexample with a complete iff proof under the problem's framing. This results in a score of 0.75/1.0, reflecting high-quality analysis with minor shortcomings in scope.''
    \item \textbf{Judge: grok-4-0709 (rubric.txt)} $\rightarrow$ \textbf{Author: gpt-5-pro-2025-10-06} (Score 0.75): ``The student's solution correctly identifies and demonstrates with a counterexample that the problem's ``if and only if'' claim is flawed, as property (1) is sufficient but not necessary for a unique MST, while property (2) is both necessary and sufficient. Their proofs of Theorem A and Proposition B are mathematically rigorous, leveraging standard cut and cycle properties effectively to establish the corrected equivalences and implications. However, the score is deducted to 0.75 because the response reframes the problem rather than attempting to prove the stated claim, potentially overlooking whether the original statement holds under specific assumptions (e.g., distinct weights). Despite this, the analysis advances a deeper understanding of MST uniqueness conditions.''
    \item \textbf{Judge: grok-4-0709 (rubric.txt)} $\rightarrow$ \textbf{Author: Claude Sonnet 4.5} (Score 0.75): ``The student's proof is mostly correct, demonstrating a solid understanding of MST properties, with the forward direction effectively establishing property (1) using the cut property and the reverse direction convincingly showing uniqueness via contradiction. However, the proof of property (2) in the forward direction is incomplete, as it only addresses cases where both maximum-weight edges are in the MST (impossible) or exactly one is in the MST, neglecting the case where neither is in the MST. In this omitted case, the argument fails to derive a contradiction to uniqueness, potentially allowing multiple MSTs if other edges in the cycle enable alternative trees of equal weight. Despite this gap, the overall structure and use of cut and cycle properties are sound, warranting a score of 0.75/1.0 for substantial but not fully rigorous correctness.''
    \item \textbf{Judge: grok-4-0709 (rubric.txt)} $\rightarrow$ \textbf{Author: deepseek-v2-prover} (Score 0.0): ``The student's solution addresses an entirely different problem—proving NP- hardness of the 12Partition problem via reduction from 3Partition—rather than the assigned task of proving the conditions for a unique minimum spanning tree in an edge-weighted graph. While the submitted proof is mathematically sound and correctly demonstrates the NP-hardness reduction, it fails to engage with any aspect of graph theory, spanning trees, or the specified uniqueness criteria. This complete mismatch renders the solution irrelevant to the problem statement, resulting in no credit for mathematical correctness or relevance. The rubric emphasizes accurate resolution of the given problem, which this submission does not achieve.''
    \item \textbf{Judge: grok-4-0709 (rubric.txt)} $\rightarrow$ \textbf{Author: deepseek-v2-prover} (Score 0.0): ``The student's solution addresses the NP-hardness of the 12Partition problem via a reduction from 3Partition, which is unrelated to the given problem of proving the conditions for a unique minimum spanning tree in an edge-weighted graph. This submission fails to engage with the required theorem involving graph partitions, minimum-weight edges, and cycle maximum-weight edges, rendering it mathematically incorrect and irrelevant. No attempt is made to prove either direction of the ``if and only if'' statement, such as using properties of MST algorithms like Kruskal's or Prim's. Consequently, the solution demonstrates a complete misunderstanding of the problem, justifying a score of 0.0/1.0 under the rubric's emphasis on correctness and relevance to the assigned task.''
    \item \textbf{Judge: grok-4-0709 (rubric.txt)} $\rightarrow$ \textbf{Author: o3-deep-research} (Score 0.75): ``The student's proof correctly establishes the necessity of properties (1) and (2) for a unique MST in the forward direction, with a solid argument for property (1) using the cut property and edge swapping to derive a contradiction. However, the argument for property (2) contains a flaw, as it incorrectly assumes that the cycle formed by adding one maximum-weight edge to the MST necessarily includes the other maximum-weight edge from the original cycle, which is not guaranteed and overlooks that cycles may not share endpoints in the claimed manner. In the reverse direction, the use of symmetric difference and cycle decomposition is appropriately motivated, but the case analysis erroneously assumes that the unique heaviest edge on the cycle is also the unique lightest edge across the cut induced by its removal, which does not follow from property (1). Despite these mathematical inaccuracies, the overall structure and key insights align closely with standard proofs of this theorem, such as those involving Kruskal's or Prim's algorithms with uniqueness conditions. This warrants a score of 0.75/1.0, reflecting strong effort but deducting for the specific logical gaps in correctness.''
    \item \textbf{Judge: grok-4-0709 (rubric.txt)} $\rightarrow$ \textbf{Author: o3-deep-research} (Score 0.5): ``The student's proof correctly establishes the necessity of property (1) by leveraging the cut property and showing that duplicate minimum-weight edges across a cut would allow swapping to form another MST, contradicting uniqueness. However, the argument for the necessity of property (2) is mathematically incorrect, as the claim that adding one maximum-weight edge to the MST creates a cycle containing the other maximum-weight edge lacks justification and does not generally hold, since the induced cycle in the MST plus the added edge may not intersect the original cycle at the second edge. In the sufficiency direction, the proof erroneously assumes that the unique maximum-weight edge on an alternating cycle is also the unique minimum-weight edge across the cut formed by its removal, which is not supported by property (1) and leads to a flawed contradiction. Additionally, the decomposition of the symmetric difference into cycles is mentioned but not properly utilized to handle the alternating structure, weakening the overall argument. These errors in handling property (2) in both directions result in a partially correct proof, warranting a score of 0.5/1.0.''
    \item \textbf{Judge: grok-4-0709 (rubric.txt)} $\rightarrow$ \textbf{Author: Claude Sonnet 4.5} (Score 0.5): ``The student's proof for the forward direction correctly establishes property (1) using the cut property and contradiction, but the proof for property (2) is incomplete, as it fails to address the case where neither maximum-weight edge in the cycle is included in the unique MST, potentially allowing for multiple MSTs without contradiction. In the reverse direction, the proof incorrectly claims uniqueness based solely on property (1), without utilizing property (2), which is necessary to prevent alternative MSTs arising from cycles with non- unique maximum-weight edges. This omission renders the converse mathematically flawed, as property (1) alone does not suffice for uniqueness in general edge- weighted graphs. Despite these issues, the overall structure and some arguments demonstrate partial understanding of MST properties. Therefore, the solution earns 0.5/1.0 for partial mathematical correctness.''
    \item \textbf{Judge: meta-llama/llama-4-maverick (rubric.txt)} $\rightarrow$ \textbf{Author: Gemini 3.0 Pro Preview} (Score 0.9): ``The student's solution is mathematically correct, providing a thorough proof that a graph has a unique minimum spanning tree if and only if it satisfies both the cut and cycle properties. The proof is well-structured, and the student correctly establishes the sufficiency of both conditions and the necessity of the cycle condition. However, the student also shows that the cut condition is not necessary, providing a valid counterexample. The minor deduction (0.1) likely stems from minor issues such as notation or exposition rather than mathematical correctness. Overall, the solution demonstrates a strong understanding of the problem and its underlying graph theory concepts.''
    \item \textbf{Judge: meta-llama/llama-4-maverick (rubric.txt)} $\rightarrow$ \textbf{Author: Gemini 3.0 Pro Preview} (Score 0.9): ``The student solution demonstrates a thorough understanding of the problem, providing a clear and well-structured proof that a graph possesses a unique minimum spanning tree if and only if both the cut and cycle properties are satisfied. The proof is mathematically correct, with a detailed analysis of the sufficiency and necessity of the conditions. However, the solution could be improved by clarifying the remark on Condition (1) being sufficient but not necessary, as it slightly deviates from the main proof. Overall, the solution showcases strong mathematical reasoning, warranting a score of 0.9/1.0.''
    \item \textbf{Judge: meta-llama/llama-4-maverick (rubric.txt)} $\rightarrow$ \textbf{Author: Claude Sonnet 4.5} (Score 0.5): ``The student's solution is mathematically correct, as it provides a thorough proof of the given statement in both directions. The proof is well-structured and clearly explains the implications of the properties on the uniqueness of the minimum spanning tree. However, some parts of the proof, such as the discussion around Case 1 in the ($\Rightarrow$) direction, could be more concise and directly stated. Additionally, the overall presentation is clear, but the use of notation and formatting could be improved for better readability. Overall, the solution demonstrates a good understanding of the problem and its requirements.''
    \item \textbf{Judge: meta-llama/llama-4-maverick (rubric.txt)} $\rightarrow$ \textbf{Author: Claude Sonnet 4.5} (Score 0.75): ``student's solution is mathematically correct and provides a clear proof in both directions, demonstrating a strong understanding of the problem. The proof correctly establishes that a unique minimum spanning tree implies both properties (1) and (2), and vice versa. The arguments are well-structured and logically sound, with careful consideration of cases and contradictions. However, some minor improvements could be made in terms of notation and conciseness. Overall, the solution demonstrates a thorough grasp of the concepts, warranting a score of 0.75/1. imperfections in presentation rather than mathematical''
    \item \textbf{Judge: meta-llama/llama-4-maverick (rubric.txt)} $\rightarrow$ \textbf{Author: o3-deep-research} (Score 0.5): ``The student's proof is largely correct and well-structured, demonstrating a clear understanding of the problem and the necessary conditions for a unique minimum spanning tree. The proof correctly establishes both directions of the if-and-only-if statement, using properties (1) and (2) to show the uniqueness of the minimum spanning tree and vice versa. However, minor issues with notation and referencing (e.g., [1L1-L4][2L3-L9]) detract from the overall presentation, and some arguments could be more concisely expressed. The score of 0.5/1.0 reflects that while the proof is mathematically sound, it lacks perfection in presentation and clarity.''
    \item \textbf{Judge: meta-llama/llama-4-maverick (rubric.txt)} $\rightarrow$ \textbf{Author: o3-deep-research} (Score 0.0): ``The student solution is mathematically correct and provides a clear proof for the given problem statement. The proof is well-structured, and the arguments for both directions ($\Rightarrow$ and $\Leftarrow$) are thorough and accurate. The student correctly applies properties of minimum spanning trees, cuts, and cycles to establish the required conditions for uniqueness. There are no apparent mathematical errors or inaccuracies in the solution. Therefore, the score should be 1.0/1.0 based on mathematical correctness.''
    \item \textbf{Judge: meta-llama/llama-4-maverick (rubric.txt)} $\rightarrow$ \textbf{Author: deepseek-v2-prover} (Score 0.0): ``The student's solution does not address the problem statement regarding the uniqueness of the minimum spanning tree in an edge-weighted graph. Instead, it provides a reduction from the 3Partition problem to a non-existent ``12Partition'' problem, which is unrelated to the given task. The solution does not prove the given statement about the minimum spanning tree. Therefore, the score is 0.0 due to the complete mismatch between the problem and the solution. The student's work lacks any relevant mathematical content related to the problem statement. The solution is not mathematically correct in the context of the given problem.''
    \item \textbf{Judge: meta-llama/llama-4-maverick (rubric.txt)} $\rightarrow$ \textbf{Author: deepseek-v2-prover} (Score 0.0): ``The student's solution is entirely unrelated to the problem statement provided. The problem statement asks to prove that a given edge-weighted graph possesses a unique minimum spanning tree if and only if two specific properties are satisfied, whereas the student's solution discusses the NP-hardness of a ``12Partition'' problem, reducing from the ``3Partition'' problem. As the solution does not address the given problem, it is mathematically incorrect for the task at hand. Therefore, the score of 0.0/1.0 is justified due to the complete mismatch between the problem and the solution. The student's work does not demonstrate any understanding or attempt to solve the given problem about minimum spanning trees.''
    \item \textbf{Judge: qwen/qwen-max (rubric.txt)} $\rightarrow$ \textbf{Author: gpt-5-pro-2025-10-06} (Score 0.9): ``The solution provided is mathematically sound and well-structured, offering a clear and rigorous proof of the conditions under which a graph has a unique minimum spanning tree. The student correctly identifies and proves that property (2) (unique maximum on every cycle) is necessary and sufficient for the uniqueness of the MST, and that property (1) (unique minimum on every cut) is sufficient but not necessary. The counterexample provided further strengthens the argument. The solution is thorough and demonstrates a deep understanding of the problem, justifying a score of 0.9/1.0.''
    \item \textbf{Judge: qwen/qwen-max (rubric.txt)} $\rightarrow$ \textbf{Author: Gemini 3.0 Pro Preview} (Score 0.75): ``The solution provided is mathematically sound and correctly proves the sufficiency and necessity of the Cycle Property Condition (2) for the uniqueness of the MST. However, the proof for the sufficiency of the Cut Property Condition (1) is not as rigorous, and the counterexample given for the non-necessity of Condition (1) is not fully developed. The detailed and clear explanation of the Cycle Property and the correct use of the contrapositive in proving the necessity part are commendable, but the minor gaps in the Cut Property discussion slightly reduce the overall score.''
    \item \textbf{Judge: qwen/qwen-max (rubric.txt)} $\rightarrow$ \textbf{Author: Gemini 3.0 Pro Preview} (Score 0.9): ``The solution is mathematically sound and provides a thorough, rigorous proof for the sufficiency and necessity of the given conditions for the uniqueness of the minimum spanning tree (MST). The student correctly proves that the Cycle Property Condition (2) is both necessary and sufficient, while the Cut Property Condition (1) is sufficient but not necessary, as demonstrated by a well- constructed counterexample. The proof is logically structured, and the student's understanding of the concepts is clearly demonstrated. The deduction of 0.1 reflects minor stylistic and organizational issues, but the core mathematical content and reasoning are excellent.''
    \item \textbf{Judge: qwen/qwen-max (rubric.txt)} $\rightarrow$ \textbf{Author: Claude Sonnet 4.5} (Score 0.5): ``The solution provided is mathematically sound and correctly proves the theorem in both directions. However, the proof could be more concise and the reasoning in some parts, such as the handling of the cycle property, is somewhat convoluted and could be streamlined. The notation and the logical flow, while generally clear, could be improved for better readability and precision. Therefore, the score of 0.5/1.0 reflects the correct but not optimally presented solution.''
    \item \textbf{Judge: qwen/qwen-max (rubric.txt)} $\rightarrow$ \textbf{Author: Claude Sonnet 4.5} (Score 0.25): ``The solution provided is mathematically thorough and correctly proves the given statement. The proof is well-structured, with clear and logical steps for both directions. The student effectively uses the properties of minimum spanning trees, including the cut and cycle properties, and provides a detailed justification for each step. Given the high quality and correctness of the solution, a score of 0.25/1.0 is unjustifiably low, and the solution should be re-evaluated for a higher score.''
    \item \textbf{Judge: qwen/qwen-max (rubric.txt)} $\rightarrow$ \textbf{Author: o3-deep-research} (Score 0.5): ``The solution provided is mathematically sound and correctly proves both directions of the statement. The student demonstrates a clear understanding of the properties of minimum spanning trees and effectively uses the cut and cycle properties to establish the uniqueness of the MST. However, the score of 0.5/1.0 may reflect issues with the presentation, such as the overuse of notation and slightly verbose explanations, which could potentially obscure the clarity of the argument. The mathematical content and logic, however, are correct and well-structured.''
    \item \textbf{Judge: qwen/qwen-max (rubric.txt)} $\rightarrow$ \textbf{Author: o3-deep-research} (Score 0.25): ``The solution provided is mathematically sound and correctly proves the statement. It carefully details the implications of the properties (1) and (2) on the uniqueness of the minimum spanning tree (MST) and vice versa, demonstrating a thorough understanding of the problem. The proof is logically structured and rigorously follows through with the necessary steps, including the use of contradiction to establish the uniqueness of the MST. Given the high quality and correctness of the proof, the score of 0.25/1.0 appears to be a mistake, and a full score of 1.0 would be more appropriate.''
    \item \textbf{Judge: qwen/qwen-max (rubric.txt)} $\rightarrow$ \textbf{Author: deepseek-v2-prover} (Score 0.0): ``The provided solution is entirely irrelevant to the given problem. The student was asked to prove a statement about the uniqueness of a minimum spanning tree in an edge-weighted graph, based on specific conditions. Instead, the student has attempted to prove the NP-hardness of a ``12Partition'' problem, which is not related to the original problem statement. The solution does not address the properties of the graph or the conditions given, and it does not contain any proof regarding minimum spanning trees. Therefore, the score of 0.0/1.0 is justified as the solution does not meet any of the criteria set for the problem.''
    \item \textbf{Judge: qwen/qwen-max (rubric.txt)} $\rightarrow$ \textbf{Author: deepseek-v2-prover} (Score 0.0): ``The provided solution is entirely irrelevant to the problem statement, which asks for a proof regarding the uniqueness of minimum spanning trees in an edge- weighted graph. The student's submission instead discusses the NP-hardness of a partition problem, which is completely unrelated to the graph theory and minimum spanning tree concepts required by the problem. Therefore, the solution is mathematically incorrect and does not address the problem at hand, leading to the score of 0.0/1.0.''
\end{itemize}

\subsection*{Problem Domain: Algorithms (Problem 33)}
\begin{itemize}
    \item \textbf{Judge: gpt-5.2-pro (rubric.txt)} $\rightarrow$ \textbf{Author: gpt-5-pro-2025-10-06} (Score 0.75): ``The solution correctly defines cuts and minimal cuts (bonds), proves the key identity \(\delta(A)\Delta\delta(B)=\delta(A\Delta B)\), and uses it appropriately to show that when \(\delta(T)\subseteq \delta(S)\) the difference \(\delta(S)\setminus\delta(T)\) is again a cut, enabling an inductive “peeling off” argument. The lemma establishing the existence of a bond contained in any nonempty cut is valid for finite graphs via choosing an inclusion-minimal nonempty cut contained in \(F\). The main proof then correctly constructs a disjoint union decomposition by iterating this step and using finiteness to guarantee termination. The only notable gap is that the argument explicitly assumes \(G\) is finite (and uses finiteness in the lemma/termination), whereas the problem statement does not; for infinite graphs one would need an additional justification (e.g., Zorn’s lemma) or a different argument. Hence the core reasoning is correct, but the extra finiteness assumption warrants a modest deduction.''
    \item \textbf{Judge: gpt-5.2-pro (rubric.txt)} $\rightarrow$ \textbf{Author: Claude Sonnet 4.5} (Score 0.5): ``The solution correctly starts by partitioning a cut \(\delta(S)\) using connected components of \(G[S]\) and \(G[T]\), giving a disjoint decomposition \(C=\bigsqcup_{i,j}C_{ij}\).  However, the key step “each non-empty \(C_{ij}\) is a minimal cut” is not proved and is in fact generally false: \(C_{ij}\) need not itself be a cut-set for any vertex partition, and even when it is, it need not be inclusion-minimal (minimal cuts/bonds correspond to cuts \(\delta(U)\) where both \(G[U]\) and \(G[V\setminus U]\) are connected).  The attempted argument for cut-set status is inconclusive (“plus possibly edges…”) and does not establish an equality with a cut \(\delta(U)\), and the minimality argument incorrectly uses the existence of remaining edges between \(S_i\) and \(T_j\) to rule out \(C'\) being a cut, which does not address cuts that separate different vertex groups.  Thus the decomposition idea is partially correct but the central claim is unsupported/incorrect, meriting partial credit (0.5/1.0).''
    \item \textbf{Judge: gpt-5.2-pro (rubric.txt)} $\rightarrow$ \textbf{Author: Claude Sonnet 4.5} (Score 0.5): ``The solution correctly begins by partitioning the given cut \(C=\delta(S)\) according to connected components \(S_i\) of \(G[S]\) and \(T_j\) of \(G[T]\), yielding a valid disjoint union \(C=\bigsqcup_{i,j} C_{ij}\). However, it does not correctly prove that each nonempty \(C_{ij}\) is itself a cut-set: in general \(C_{ij}\) need not equal \(\delta(U)\) for any vertex subset \(U\), because vertices in \(S_i\) (or \(T_j\)) may have edges to multiple components on the other side, so the cut \(\delta(S_i)\) typically contains \(\bigcup_j C_{ij}\), not a single \(C_{ij}\). The “minimality” argument is then built on this incorrect premise and conflates “disconnecting \(S_i\) from \(T_j\)” with being a global cut in \(G\). As a result, the decomposition into minimal cuts is not established, though the initial structural decomposition is a meaningful partial step.''
    \item \textbf{Judge: gpt-5.2-pro (rubric.txt)} $\rightarrow$ \textbf{Author: Gemini 3.0 Pro Preview} (Score 0.75): ``The solution gives a largely correct inductive decomposition: it selects an inclusion-minimal nonempty subcut \(B\subseteq C\), argues \(B\) is a bond, and then uses \(C' = C\setminus B = C\Delta B\) plus closure of cuts under symmetric difference to ensure \(C'\) is a cut of smaller size, yielding a disjoint union decomposition. However, the argument implicitly assumes \(G\) (hence every cut) is finite to guarantee existence of an inclusion-minimal nonempty cut inside \(C\) and to justify induction on \(|C|\); this finiteness assumption is not part of the stated problem. In an infinite graph/cut, the “choose a minimal subcut” step can fail without an additional well- ordering/Zorn-type argument, so the proof is not fully general as written. Aside from this scope gap, the set-theoretic steps are correct and the disjointness is properly ensured by taking set difference.''
    \item \textbf{Judge: gpt-5.2-pro (rubric.txt)} $\rightarrow$ \textbf{Author: deepseek-v2-prover} (Score 0.25): ``The problem asks to show that any cut (the set of edges between a vertex subset \(S\) and its complement) can be decomposed into a disjoint union of **minimal cuts** (bonds), but the submitted solution instead proves a different theorem: decomposition of a \(k\)-regular bipartite graph into \(k\) disjoint perfect matchings via Hall’s theorem. No definitions or arguments about cuts, minimality, or how to partition a given cut into edge-disjoint minimal cuts are provided. Thus the core claim is not addressed and the proof does not apply to general undirected graphs or general cuts. Minimal partial credit is given only for presenting a coherent, correct (though irrelevant) inductive argument in graph theory.''
    \item \textbf{Judge: gpt-5.2-pro (rubric.txt)} $\rightarrow$ \textbf{Author: o3-deep-research} (Score 0.5): ``The solution has the right overall idea (peel off an inclusion-minimal cut \(F\) and recurse), but several key claims are not correctly justified, so the argument is incomplete. In particular, it asserts without adequate proof that a minimal cut \(F\) always separates \(G\) into exactly two components and equals *all* edges between the two resulting vertex sets; these are nontrivial facts that require a clear argument. More seriously, it claims that a minimal cut in \(G[A]\) (or \(G[B]\)) is automatically a minimal cut in \(G\), but removing such an \(H_A\) in the full graph need not disconnect \(G\) unless one also removes \(F\); minimality in an induced subgraph does not imply minimality as a cut of the original graph. Because these gaps affect the central step establishing that the recursive pieces are minimal cuts of \(G\), only partial credit is warranted.''
    \item \textbf{Judge: gpt-5.2-pro (rubric.txt)} $\rightarrow$ \textbf{Author: deepseek-v2-prover} (Score 0.25): ``The submitted proof addresses a different theorem (decomposing a \(k\)-regular bipartite graph into \(k\) disjoint perfect matchings via Hall’s theorem) and does not discuss cuts, minimal cuts, or how an arbitrary cut-set in an undirected graph decomposes into edge-disjoint minimal cuts. Key required notions—what “minimal cut” means (inclusion-minimal cut-set) and a construction/argument that any cut \(\delta(S)\) can be partitioned into minimal cuts—are entirely missing. As a result, essentially none of the problem’s requested statement is proved, though the work contains some correct graph-theoretic reasoning about matchings. Hence only minimal partial credit is warranted for mathematical validity in an unrelated direction, yielding \(0.25/1.0\).''
    \item \textbf{Judge: gpt-5.2-pro (rubric.txt)} $\rightarrow$ \textbf{Author: o3-deep-research} (Score 0.25): ``The proof correctly defines cuts/minimal cuts and identifies an inclusion- minimal subset \(F\subseteq E\) that is a minimal cut, but the induction step is not valid. After choosing \(F\), the argument claims \(E_A\) (edges of \(E\setminus F\) inside \(A\)) must be a cut of \(G[A]\); however, since removing \(F\) already disconnects \(G\), the fact that removing \(E\) disconnects \(G\) does not imply \(G[A]-E_A\) is disconnected (those internal edges may be completely irrelevant once \(F\) is removed). More seriously, it then asserts that a minimal cut \(H_A\) of \(G[A]\) is a cut (and minimal) in \(G\), but \(H_A\) alone need not disconnect \(G\) unless \(F\) is also removed, so it does not produce a decomposition of \(E\) into minimal cuts of \(G\). Because the central reduction to smaller cuts and the “lifting” of minimal cuts from subgraphs to \(G\) are incorrect, only limited partial credit is warranted for the initial setup and partitioning ideas.''
    \item \textbf{Judge: claude-opus-4-5-20251101 (rubric.txt)} $\rightarrow$ \textbf{Author: Gemini 3.0 Pro Preview} (Score 0.9): ``The solution provides a well-structured proof using induction on the size of the cut, with correct definitions and a valid algebraic framework over F\_2. The key Lemma 1 establishing that cuts are closed under symmetric difference is proven correctly. The inductive argument is sound: given a non-empty cut C, selecting a minimal sub-cut B and showing C \ B = \(C \Delta B\) is also a cut allows application of the inductive hypothesis. However, the proof has a subtle gap: it assumes without justification that every non-empty cut contains a minimal cut (bond), which requires verification that the process of taking smaller sub- cuts terminates at a bond rather than potentially at the empty set. This minor gap in rigor prevents a perfect score, but the overall approach and execution are mathematically sound.''
    \item \textbf{Judge: claude-opus-4-5-20251101 (rubric.txt)} $\rightarrow$ \textbf{Author: Claude Sonnet 4.5} (Score 0.5): ``The student correctly identifies the key approach of decomposing a cut based on connected components of the induced subgraphs G[S] and G[T], and properly defines the partition \(C_ij\) of edges between component pairs. However, the proof contains a significant error: the sets \(C_ij\) are not themselves minimal cuts in general. The minimal cuts that arise are of the form \(\delta(S_i) = \cup_j C_{ij}\) (edges from one component \(S_i\) to all of T), not the individual \(C_ij\) sets between specific component pairs. The student's claim that \(C_ij\) is a cut-set is not properly justified, and the minimality argument conflates \(C_ij\) with \(\delta\)(\(S_i\)). While the overall structure and intuition are reasonable, the mathematical details of the decomposition are incorrect.''
    \item \textbf{Judge: claude-opus-4-5-20251101 (rubric.txt)} $\rightarrow$ \textbf{Author: Claude Sonnet 4.5} (Score 0.5): ``The student correctly identifies the key approach of decomposing a cut based on connected components of the induced subgraphs G[S] and G[T], and properly defines the partition into sets \(C_ij\). However, the proof contains a significant error: the claim that each non-empty \(C_ij\) is a minimal cut is incorrect. The correct minimal cuts arise from considering \(\delta\)(\(S_i\)) for each component \(S_i\) of G[S] (or equivalently \(\delta\)(\(T_j\)) for components of G[T]), not the individual \(C_ij\) sets which represent edges between specific component pairs. The student's argument for minimality is flawed because \(C_ij\) is generally not equal to \(\delta\)(\(S_i\)) when there are multiple components in G[T], and the proof conflates these different edge sets. The overall structure and intuition are reasonable, but the mathematical execution fails to correctly identify and prove the minimal cuts.''
    \item \textbf{Judge: claude-opus-4-5-20251101 (rubric.txt)} $\rightarrow$ \textbf{Author: o3-deep-research} (Score 0.5): ``The student presents a well-structured inductive proof with correct base case and inductive framework. However, there is a critical flaw in the argument: when $E_A$ or $E_B$ could be empty (if all edges of $E \setminus F$ lie in one component), the claim that they are ``cuts'' fails since an empty set doesn't disconnect anything. More fundamentally, the assertion that $E_A$ must be a cut in $G[A]$ is not justified—the original cut $E$ disconnects $G$, but after removing $F$, the remaining edges $E_A$ within component $A$ need not disconnect $A$ (they could be redundant edges). The proof also incorrectly assumes that minimal cuts in subgraphs automatically correspond to minimal cuts in the original graph $G$, which requires more careful justification about the relationship between connectivity in subgraphs versus the full graph.''
    \item \textbf{Judge: claude-opus-4-5-20251101 (rubric.txt)} $\rightarrow$ \textbf{Author: o3-deep-research} (Score 0.5): ``The student presents a well-structured inductive proof with correct base case and overall approach. However, there is a critical flaw in the argument: when claiming that $E_A$ is a cut in $G[A]$, the student assumes that removing $E_A$ disconnects $G[A]$, but this is not necessarily true—the original cut $E$ disconnects $G$, but after removing $F$, the remaining edges $E_A$ and $E_B$ may not individually be cuts in their respective induced subgraphs. The student conflates the fact that $E$ as a whole disconnects $G$ with the claim that $E_A$ must disconnect $G[A]$, which requires additional justification. Additionally, the claim that minimal cuts in subgraphs are minimal cuts in the full graph is incorrect without the context of $F$ being removed, creating logical inconsistencies. The overall structure and intuition are reasonable, but the mathematical gaps in the key inductive step justify the partial score.''
    \item \textbf{Judge: claude-opus-4-5-20251101 (rubric.txt)} $\rightarrow$ \textbf{Author: deepseek-v2-prover} (Score 0.0): ``The student's solution addresses a completely different problem than what was asked. The problem requires proving that every cut in an undirected graph can be expressed as a disjoint union of minimal cuts, but the student instead proves that k-regular bipartite graphs can be decomposed into k perfect matchings. There is no mention of cuts, minimal cuts, or any concepts related to graph cuts in the solution. The mathematical content provided, while potentially correct for the unrelated problem about matchings, has zero relevance to the actual problem statement. Therefore, the score of 0.0/1.0 is justified as the student failed to address the assigned problem entirely.''
    \item \textbf{Judge: claude-opus-4-5-20251101 (rubric.txt)} $\rightarrow$ \textbf{Author: deepseek-v2-prover} (Score 0.0): ``The student's solution addresses a completely different problem than what was asked. The problem requires proving that every cut in an undirected graph can be expressed as a disjoint union of minimal cuts, but the student instead proved that k-regular bipartite graphs can be decomposed into k perfect matchings. There is no mention of cuts, minimal cuts, or any concepts related to the actual problem statement. The mathematical content provided, while potentially correct for the unrelated problem about matchings, has zero relevance to the assigned problem about graph cuts. Therefore, the score of 0.0/1.0 is justified as the solution demonstrates no understanding of or attempt to solve the given problem.''
    \item \textbf{Judge: gemini-3-pro-preview (rubric.txt)} $\rightarrow$ \textbf{Author: deepseek-v2-prover} (Score 0.0): ``The student's solution addresses a completely different problem. The prompt asks to prove that every cut in an undirected graph is a disjoint union of minimal cuts. The student, however, provides a proof that every $k$-regular bipartite graph can be decomposed into $k$ disjoint perfect matchings. Since the solution is unrelated to the problem statement, no credit can be awarded.''
    \item \textbf{Judge: gemini-3-pro-preview (rubric.txt)} $\rightarrow$ \textbf{Author: deepseek-v2-prover} (Score 0.0): ``The student's solution addresses a completely different problem: proving that a $k$-regular bipartite graph has a decomposition into $k$ disjoint perfect matchings. The problem statement asks to demonstrate that every cut in a general undirected graph is a disjoint union of minimal cuts. Since the submitted work is unrelated to the assigned problem, no credit can be awarded.''
    \item \textbf{Judge: gemini-3-pro-preview (rubric.txt)} $\rightarrow$ \textbf{Author: Claude Sonnet 4.5} (Score 0.5): ``The student correctly identifies the need to decompose the cut based on the connected components of the induced subgraphs $G[S]$ and $G[T]$. However, the proposed sets $C_{ij}$ (edges between component $S_i$ and $T_j$) are not necessarily cuts themselves in the original graph $G$, nor are they necessarily minimal cuts. For example, removing edges between $S_i$ and $T_j$ might not disconnect the graph if there is a path from $S_i$ to $T_j$ via other components. The correct approach involves decomposing the cut into bonds of the form $\delta(K)$ where $K$ is the vertex set of a connected component of $G[S]$ (or $G[T]$), but the student's specific construction $C_{ij}$ is flawed.''
    \item \textbf{Judge: gemini-3-pro-preview (rubric.txt)} $\rightarrow$ \textbf{Author: gpt-5-pro-2025-10-06} (Score 0.75): ``The student provides a rigorous and well-structured proof using the symmetric difference property of cuts ($\delta(A) \Delta \delta(B) = \delta(A \Delta B)$). The logic is sound: they correctly identify that the difference of two nested cuts is a cut, allowing for an inductive decomposition into disjoint minimal cuts. However, the score of 0.75 reflects a minor gap in the definition of ``disjoint union'' in the context of graph theory; while the edge sets are disjoint, the student does not explicitly address how the vertex partitions relate or if the decomposition holds for the specific definition of a cut as a partition of vertices versus a set of edges, though the edge-set interpretation is standard. The proof is mathematically correct for edge-cuts, but a more thorough exploration of the underlying vertex sets or connected components would have strengthened the result.''
    \item \textbf{Judge: gemini-3-pro-preview (rubric.txt)} $\rightarrow$ \textbf{Author: Gemini 3.0 Pro Preview} (Score 0.75): ``The student correctly identifies the vector space structure of the cut space and proves that the symmetric difference of two cuts is a cut. However, the proof relies on the assertion that any non-empty cut $C$ contains a minimal cut $B$. While true, this requires justification (e.g., by finiteness), and more importantly, the student does not explicitly prove that a minimal cut *contained* in $C$ exists; they simply assert one can be chosen. A more rigorous approach would involve considering the connected components of $G \setminus C$ or explicitly arguing why a minimal subset that is a cut must be a bond. The logic is sound overall, but the specific step justifying the existence of a bond $B \subseteq C$ is slightly glossed over, warranting a small deduction.''
    \item \textbf{Judge: gemini-3-pro-preview (rubric.txt)} $\rightarrow$ \textbf{Author: o3-deep-research} (Score 0.5): ``The student's proof attempts to decompose a cut $E$ by finding a minimal sub- cut $F \subseteq E$ and applying induction on the remaining edges within the components formed by removing $F$. However, the crucial claim that a minimal cut in a subgraph (like $G[A]$) is necessarily a minimal cut in the original graph $G$ is incorrect; removing edges inside a component $A$ might disconnect $A$ locally, but the resulting pieces could remain connected to the rest of the graph via paths outside of $A$ (though in this specific setup with $F$ removed, the logic is subtle, the generalization is flawed). Furthermore, the definition of a cut used (any set disconnecting the graph) allows for cuts that are not unions of bonds (e.g., removing all edges incident to a vertex $v$ plus an edge far away), and the student's decomposition strategy fails to account for edges in $E \setminus F$ that might not form cuts in the induced subgraphs $G[A]$ or $G[B]$ individually but are just arbitrary edges. The standard result usually refers to cuts defined as partitions of vertices $(S, \bar{S})$, where the edge set is exactly the edges''
    \item \textbf{Judge: gemini-3-pro-preview (rubric.txt)} $\rightarrow$ \textbf{Author: Claude Sonnet 4.5} (Score 0.5): ``The student correctly identifies the strategy of decomposing the cut based on the connected components of the induced subgraphs $G[S]$ and $G[T]$. However, the proof that the resulting sets $C_{ij}$ are minimal cuts (bonds) is flawed; specifically, $C_{ij}$ is generally not a cut of the original graph $G$ on its own, but rather a subset of edges that might not correspond to a partition of $V$. A correct approach typically involves decomposing the cut space using the cycle space orthogonality or proving that a cut is a disjoint union of bonds by iteratively finding a bond contained within the cut and removing it (symmetric difference). The claim that $C_{ij}$ is a cut is asserted without a valid partition of $V$ that generates exactly $C_{ij}$, and the minimality argument relies on vague connectivity properties without rigorously establishing that $C_{ij}$ separates vertices in a way that defines a bond.''
    \item \textbf{Judge: gemini-3-pro-preview (rubric.txt)} $\rightarrow$ \textbf{Author: o3-deep-research} (Score 0.25): ``The student's proof relies on a flawed inductive step regarding the decomposition of the remaining edges $E \setminus F$. Specifically, the claim that a minimal cut in a subgraph $G[A]$ is necessarily a minimal cut in the original graph $G$ is incorrect; removing edges within a component $A$ does not necessarily disconnect the larger graph $G$ if there are paths between the sub- components of $A$ passing through $B$ (via edges not in $F$, though the student assumes all $A-B$ edges are in $F$, which is true for partition cuts but requires more careful handling here). Furthermore, the definition of a ``cut'' used (any disconnecting set) is slightly broader than the standard vector space definition (partition cuts), and the proof fails to account for the fact that $E \setminus F$ might not simply decompose into cuts of induced subgraphs that are also cuts of $G$. The logic that $E_A$ is a cut in $G[A]$ is shaky because $E$ is an arbitrary disconnecting set, not necessarily a partition cut, so $E_A$ might not separate $A$ in a way that relates to minimal cuts of $G$. Consequently, the decomposition into disjoint minimal''
    \item \textbf{Judge: deepseek-chat (rubric.txt)} $\rightarrow$ \textbf{Author: gpt-5-pro-2025-10-06} (Score 0.5): ``The solution is mathematically correct and provides a rigorous proof using symmetric difference properties and induction. However, it fails to explicitly address the subtlety that a ``minimal cut'' (bond) must separate the graph into exactly two connected components. The proof implicitly assumes this standard definition but does not state or verify it, which is a minor omission in precision. Therefore, while the core argument is valid, the exposition lacks complete formal rigor regarding the key definition.''
    \item \textbf{Judge: deepseek-chat (rubric.txt)} $\rightarrow$ \textbf{Author: Gemini 3.0 Pro Preview} (Score 0.75): ``The solution correctly establishes the algebraic structure of cuts and uses induction to decompose any cut into minimal cuts. However, the proof implicitly assumes that the minimal cut \(B\) chosen from \(C\) is disjoint from the remainder \(C'\), which is not guaranteed—\(C'\) is defined as \(C \setminus B\), but the symmetric difference \(C \Delta B\) equals \(C \setminus B\) only when \(B \subseteq C\). While this holds by construction, the argument should explicitly justify that \(C'\) remains a cut and that the union is disjoint. The core idea is valid, but the presentation lacks clarity on these subtle points, leading to a minor deduction.''
    \item \textbf{Judge: deepseek-chat (rubric.txt)} $\rightarrow$ \textbf{Author: Gemini 3.0 Pro Preview} (Score 0.5): ``The solution correctly identifies the algebraic structure of cuts and uses induction, but it contains a critical logical flaw: the existence of a minimal cut \(B \subseteq C\) is not guaranteed without ensuring \(C\) itself is not already minimal. If \(C\) is minimal, the decomposition is trivial, but the proof incorrectly assumes a proper sub-cut exists. Additionally, the disjointness of \(B\) and \(C'\) is not justified—symmetric difference does not guarantee disjointness unless \(B \subseteq C\), which is assumed but not proven for arbitrary minimal cuts. Thus, while the approach is insightful, the execution is incomplete.''
    \item \textbf{Judge: deepseek-chat (rubric.txt)} $\rightarrow$ \textbf{Author: Claude Sonnet 4.5} (Score 0.5): ``The solution correctly defines cuts and minimal cuts, and the decomposition into component pairs is a valid approach. However, the proof fails to rigorously establish that each \(C_{ij}\) is a minimal cut; the argument that \(C_{ij} = \delta(S_i)\) is incorrect and the minimality proof contains logical gaps regarding separation. The core idea is sound, but the execution lacks mathematical precision and completeness. Therefore, it earns a partial score.''
    \item \textbf{Judge: deepseek-chat (rubric.txt)} $\rightarrow$ \textbf{Author: Claude Sonnet 4.5} (Score 0.75): ``The solution correctly identifies the decomposition of a cut into bonds via connected components, but the proof of minimality is incomplete. Specifically, the argument that each \(C_{ij}\) is a minimal cut lacks rigor: it incorrectly assumes that any proper subset cannot separate \(S_i\) from \(T_j\) without properly justifying why \(C_{ij}\) itself is a cut-set for a partition of \(V\). The core idea is valid, but the execution contains gaps in logical detail. Therefore, a score of 0.75 reflects a partially correct proof with notable flaws.''
    \item \textbf{Judge: deepseek-chat (rubric.txt)} $\rightarrow$ \textbf{Author: o3-deep-research} (Score 0.75): ``The solution is mathematically sound and well-structured, but it contains a subtle flaw: the claim that a minimal cut in a subgraph \(G[A]\) is automatically a minimal cut in the entire graph \(G\) is not fully justified without additional reasoning about connectivity across components. Specifically, it overlooks the possibility that a subset of \(H_A\) might disconnect \(G\) by separating \(A\) from \(B\) in combination with other edges, violating minimality in \(G\). This gap slightly undermines the proof's rigor, though the core inductive approach is correct. Therefore, a score of 0.75 reflects a strong solution with a minor but nontrivial oversight.''
    \item \textbf{Judge: deepseek-chat (rubric.txt)} $\rightarrow$ \textbf{Author: o3-deep-research} (Score 0.5): ``The solution is mathematically flawed because it incorrectly assumes that any cut in a subgraph induced by a component after removing a minimal cut remains a cut in the original graph. Specifically, after removing \(F\), the sets \(E_A\) and \(E_B\) are not necessarily cuts in \(G\)—they only disconnect their respective induced subgraphs, which does not guarantee disconnection of \(G\) itself. This breaks the inductive argument. The base case and structure are correct, but this central error undermines the proof. Therefore, a score of 0.5 reflects partial credit for a reasonable attempt with a significant logical gap.''
    \item \textbf{Judge: deepseek-chat (rubric.txt)} $\rightarrow$ \textbf{Author: deepseek-v2-prover} (Score 0.0): ``The student's solution is completely unrelated to the problem statement, which asks to show that any cut in an undirected graph can be expressed as a disjoint union of minimal cuts. Instead, the student proves a classic theorem about decomposing regular bipartite graphs into perfect matchings. This demonstrates a fundamental misunderstanding of the problem's definitions and requirements, making the solution mathematically incorrect and irrelevant. Therefore, a score of 0 is fully justified.''
    \item \textbf{Judge: deepseek-chat (rubric.txt)} $\rightarrow$ \textbf{Author: deepseek-v2-prover} (Score 0.0): ``The student's solution is completely unrelated to the problem statement. The problem asks to show that any cut in an undirected graph can be expressed as a disjoint union of minimal cuts. However, the student instead proves a theorem about decomposing regular bipartite graphs into perfect matchings. This solution addresses a different mathematical object (matchings vs. cuts) and a different decomposition property, demonstrating a fundamental misunderstanding of the problem.''
    \item \textbf{Judge: grok-4-0709 (rubric.txt)} $\rightarrow$ \textbf{Author: gpt-5-pro-2025-10-06} (Score 0.5): ``The student's solution provides a correct proof for finite undirected graphs, establishing the key symmetric difference identity and using induction on the cut size to decompose it into disjoint minimal cuts (bonds). However, the problem statement does not restrict to finite graphs, and the proof relies on finiteness to ensure the existence of minimal elements in the family of subcuts and to terminate the induction process. In infinite graphs, descending chains of cuts may be infinite, potentially preventing such a decomposition, so the solution fails to address the general case. This limitation in mathematical generality justifies partial credit.''
    \item \textbf{Judge: grok-4-0709 (rubric.txt)} $\rightarrow$ \textbf{Author: Claude Sonnet 4.5} (Score 0.5): ``The student's solution correctly decomposes the cut \(C = \delta(S)\) into disjoint subsets \(C_{ij}\) based on the connected components of the induced subgraphs on \(S\) and \(T\), and establishes that their union covers \(C\). However, the proof fails to show that each non-empty \(C_{ij}\) is itself a cut-set, as the proposed partition yields a larger cut-set \(\delta(T_j) = \bigcup_i C_{ij}\) rather than isolating \(C_{ij}\). Furthermore, the minimality argument is incomplete, relying on an unproven assumption that \(C_{ij}\) separates \(S_i\) from \(T_j\) without addressing cases where \(C_{ij}\) might contain redundant edges or sub-cuts. While the overall approach aligns with the structure of the known result in graph theory, these gaps in rigor prevent a fully correct demonstration. This partial correctness warrants a score of 0.5/1.0.''
    \item \textbf{Judge: grok-4-0709 (rubric.txt)} $\rightarrow$ \textbf{Author: Gemini 3.0 Pro Preview} (Score 0.5): ``[Error: API Request failed - Connection error.]''
    \item \textbf{Judge: grok-4-0709 (rubric.txt)} $\rightarrow$ \textbf{Author: deepseek-v2-prover} (Score 0.0): ``The student's solution addresses a theorem about decomposing the edges of a k-regular bipartite graph into k disjoint perfect matchings, which is unrelated to the given problem of expressing every cut in an undirected graph as a disjoint union of minimal cuts. This mismatch renders the solution mathematically incorrect and irrelevant to the problem statement. The provided proof, while potentially valid for its intended theorem (related to Hall's Marriage Theorem), does not engage with key concepts such as cuts, minimal cuts, or general undirected graphs. Consequently, it fails to demonstrate any understanding or progress toward solving the assigned problem. Based on the rubric's emphasis on correctness and relevance, a score of 0.0/1.0 is appropriate.''
    \item \textbf{Judge: grok-4-0709 (rubric.txt)} $\rightarrow$ \textbf{Author: deepseek-v2-prover} (Score 0.0): ``The student's solution addresses a theorem on decomposing k-regular bipartite graphs into k perfect matchings, which is entirely unrelated to the assigned problem of expressing every cut in an undirected graph as a disjoint union of minimal cuts. This complete mismatch demonstrates a lack of engagement with the problem statement, rendering the submission mathematically incorrect and irrelevant to the required demonstration. The provided proof, while potentially valid for its unintended topic, offers no insight or progress toward proving the given claim about graph cuts. Consequently, the solution fails all criteria for mathematical correctness in the course-specific rubric, justifying a score of 0.0/1.0.''
    \item \textbf{Judge: grok-4-0709 (rubric.txt)} $\rightarrow$ \textbf{Author: o3-deep-research} (Score 0.25): ``The student's proof by induction contains a fundamental mathematical error in the inductive step, where it incorrectly assumes that \(E_A\) and \(E_B\) are necessarily cuts in the subgraphs \(G[A]\) and \(G[B]\), respectively; this is not true if the additional edges in \(E \setminus F\) do not cause further disconnections within \(A\) or \(B\). As a result, the induction cannot be applied to \(E_A\) and \(E_B\) when they are not cuts, breaking the decomposition into minimal cuts. Although the base case is correctly handled and the concept of minimal cuts separating into exactly two components is accurate, this core flaw renders the overall proof invalid. The assertion that minimal cuts in subgraphs are minimal in the full graph is conceptually sound but undermined by the preceding error. Given these issues, the solution earns partial credit for its structure and partial correctness but falls short of a complete, rigorous demonstration.''
    \item \textbf{Judge: grok-4-0709 (rubric.txt)} $\rightarrow$ \textbf{Author: o3-deep-research} (Score 0.5): ``[Error: API Request failed - Connection error.]''
    \item \textbf{Judge: grok-4-0709 (rubric.txt)} $\rightarrow$ \textbf{Author: Claude Sonnet 4.5} (Score 0.5): ``[Error: API Request failed - Connection error.]''
    \item \textbf{Judge: meta-llama/llama-4-maverick (rubric.txt)} $\rightarrow$ \textbf{Author: deepseek-v2-prover} (Score 0.0): ``The student's solution is entirely unrelated to the problem statement. The problem asks to demonstrate that every cut in an undirected graph can be expressed as the disjoint union of minimal cuts, but the student provides a proof about the existence of disjoint perfect matchings in a k-regular bipartite graph. This mismatch indicates a lack of understanding of the problem or a failure to address the question asked. As a result, the solution is not mathematically correct in the context of the problem. Therefore, the score of 0.0/1.0 is justified due to the irrelevance and incorrectness of the solution with respect to the problem statement.''
    \item \textbf{Judge: meta-llama/llama-4-maverick (rubric.txt)} $\rightarrow$ \textbf{Author: deepseek-v2-prover} (Score 0.0): ``The student's solution is entirely unrelated to the problem statement. The problem asks to demonstrate that every cut in an undirected graph can be expressed as the disjoint union of minimal cuts, but the student provides a proof about the existence of disjoint perfect matchings in a k-regular bipartite graph. This mismatch indicates a lack of understanding of the problem or a failure to address the question asked. As a result, the solution is not mathematically correct in the context of the problem. Therefore, the score of 0.0/1.0 is justified due to the irrelevance and incorrectness of the solution with respect to the problem statement.''
    \item \textbf{Judge: qwen/qwen-max (rubric.txt)} $\rightarrow$ \textbf{Author: gpt-5-pro-2025-10-06} (Score 0.9): ``The solution provided is mathematically sound and rigorously demonstrates that every cut in an undirected graph can be expressed as the disjoint union of minimal cuts. The student clearly defines the necessary terms, provides a key identity, and proves a lemma that is crucial for the main theorem. The proof of the theorem is logically structured, using induction and the lemma to construct the required minimal cuts. The solution is well-documented and the steps are clearly explained, making it easy to follow. The score of 0.9/1.0 reflects the high quality of the work, with only a minor deduction for the potential lack of further elaboration on the non-uniqueness of the decomposition, as mentioned in the remark.''
    \item \textbf{Judge: qwen/qwen-max (rubric.txt)} $\rightarrow$ \textbf{Author: Gemini 3.0 Pro Preview} (Score 0.9): ``The solution provided is mathematically correct and rigorously proves that every cut in a graph can be expressed as a disjoint union of minimal cuts. The proof uses a well-structured inductive argument and correctly applies the properties of cuts and minimal cuts. The definitions and lemmas are clearly stated, and the inductive step is logically sound. The only minor aspect that could be improved is the explicit verification that the minimal cuts in the union are disjoint, which is assumed in the last part of the proof but could be more explicitly shown. Overall, the solution is thorough and well-justified, warranting a near-perfect score.''
    \item \textbf{Judge: qwen/qwen-max (rubric.txt)} $\rightarrow$ \textbf{Author: Gemini 3.0 Pro Preview} (Score 0.25): ``The solution provided is mathematically rigorous and correctly proves that every cut in a graph can be expressed as a disjoint union of minimal cuts. The student effectively uses definitions, lemmas, and an inductive argument to establish the result. The proof is clear, well-structured, and the inductive step is logically sound. Given the quality of the solution, the score of 0.25/1.0 seems unjustifiably low, and a full score would be more appropriate based on the mathematical correctness and completeness of the solution.''
    \item \textbf{Judge: qwen/qwen-max (rubric.txt)} $\rightarrow$ \textbf{Author: Claude Sonnet 4.5} (Score 0.75): ``The solution provided is mathematically sound and correctly decomposes any cut into a disjoint union of minimal cuts (bonds), as required. The definitions and theorems are clearly stated, and the proof is logically structured, demonstrating a thorough understanding of the problem. However, the proof could benefit from a more explicit verification that the minimal cuts $C_{ij}$ are indeed minimal, as this part of the argument is somewhat dense and could confuse some readers. The student also did not explicitly address the case where some $C_{ij}$ might be empty, which, while not critical, could be a minor oversight. These minor issues led to a slight deduction in the score.''
    \item \textbf{Judge: qwen/qwen-max (rubric.txt)} $\rightarrow$ \textbf{Author: Claude Sonnet 4.5} (Score 0.75): ``The solution provided is largely correct and well-structured, demonstrating that every cut in a graph can be decomposed into a disjoint union of minimal cuts. The definitions and the proof are clear and logically sound. However, the proof could be more rigorous in verifying that each \( C_{ij} \) is indeed a minimal cut, especially in the step where it claims \( C_{ij} \) is minimal. The argument that removing a proper subset \( C' \) of \( C_{ij} \) does not increase the number of components could be more explicitly detailed. Given these minor gaps, a score of 0.75/1.0 reflects the high quality of the solution with room for a more thorough justification of minimality.''
    \item \textbf{Judge: qwen/qwen-max (rubric.txt)} $\rightarrow$ \textbf{Author: o3-deep-research} (Score 0.9): ``The solution provided is mathematically sound and logically structured, demonstrating a strong understanding of the problem and the underlying concepts of minimal cuts and their properties in undirected graphs. The inductive proof is well-constructed, with a clear base case and inductive step, and the student effectively handles the distinction between minimal and non-minimal cuts. The proof also correctly considers the properties of minimal cuts in subgraphs and their implications for the entire graph. The minor deduction in the score (0.9/1.0) is due to the slight complexity and length of the explanation, which, while thorough, might be streamlined for even greater clarity.''
    \item \textbf{Judge: qwen/qwen-max (rubric.txt)} $\rightarrow$ \textbf{Author: o3-deep-research} (Score 0.9): ``The solution is mathematically sound and provides a clear, well-structured proof using induction. The student correctly differentiates between minimal and non-minimal cuts and effectively demonstrates how any cut can be decomposed into minimal cuts. The proof is thorough, with all necessary details and logical steps clearly explained. The minor deduction in score (0.9/1.0) is due to a slight verbosity, which, while not mathematically incorrect, could be more concise. Overall, the solution meets the problem's requirements and demonstrates a strong understanding of the concepts involved.''
    \item \textbf{Judge: qwen/qwen-max (rubric.txt)} $\rightarrow$ \textbf{Author: deepseek-v2-prover} (Score 0.0): ``The student's solution is incorrect as it does not address the problem statement. The problem requires demonstrating that every cut in an undirected graph can be expressed as the disjoint union of minimal cuts, but the student's solution instead provides a proof about perfect matchings in regular bipartite graphs, which is unrelated. The solution, while mathematically coherent within its own context, does not align with the required proof and thus fails to meet the criteria of the problem.''
    \item \textbf{Judge: qwen/qwen-max (rubric.txt)} $\rightarrow$ \textbf{Author: deepseek-v2-prover} (Score 0.0): ``The student's solution is entirely unrelated to the problem, which asks to demonstrate that every cut in an undirected graph can be expressed as the disjoint union of minimal cuts. Instead, the student provides a proof about the existence of perfect matchings in \(k\)-regular bipartite graphs, which is a different topic. The solution, while mathematically correct in its own context, does not address the required problem and thus receives a score of 0.0/1.0.''
\end{itemize}

\subsection*{Problem Domain: Analysis (Problem 29)}
\begin{itemize}
    \item \textbf{Judge: gpt-5.2-pro (rubric.txt)} $\rightarrow$ \textbf{Author: Claude Sonnet 4.5} (Score 0.25): ``The proposed piecewise power function does satisfy \(g\in L^{p}\), but the argument that \(g\notin L^{q}\) for **every** \(q\neq p\) is incorrect: with a fixed \(\varepsilon>0\), the integrability conditions for \(\int_0^1 x^{-q/p+q\varepsilon}\,dx\) and \(\int_1^\infty x^{-q/p-q\varepsilon}\,dx\) show there is an interval of exponents \(q\) near \(p\) (both above and below) for which both integrals actually converge, so \(g\in L^{q}\) for those \(q\). The write-up itself acknowledges this difficulty (“this is impossible for all \(q>p\)”) and then replaces the required “for all \(q\neq p\)” conclusion with statements that only exclude “appropriate” or “sufficiently far” \(q\), which does not solve the problem. Thus the construction fails to meet the main requirement, though it contains some correct computations about \(L^{p}\)-integrability and the correct convergence tests. This warrants substantial partial credit but not more than \(0.25/1.0\).''
    \item \textbf{Judge: gpt-5.2-pro (rubric.txt)} $\rightarrow$ \textbf{Author: o3-deep-research} (Score 0.9): ``The construction \(g(x)=(x(\ln x)^2)^{-1/p}\) on \((0,1/2)\cup[2,\infty)\) is measurable and the computation \(\int |g|^p=\int \frac{dx}{x(\ln x)^2}<\infty\) on both tails is correct, so \(g\in L^p\). For \(q\neq p\), the student correctly reduces \(\int |g|^q\) to integrals of the form \(\int \frac{dx}{x^{q/p}(\ln x)^{2q/p}}\) and uses the substitutions \(u=\ln x\) and \(v=\ln(1/x)\) to show exponential growth when \(q>p\) (near \(0\)) or \(q<p\) (near \(\infty\)), yielding divergence. The argument is essentially complete, but it slightly glosses over sign issues for \((\ln x)^{2q/p}\) when \(2q/p\) is not an integer (one should write \(|\ln x|^{2q/p}\) explicitly) and over the justification of taking even powers for real exponents. This minor technical imprecision warrants a small deduction, hence \(0.9/1.0\).''
    \item \textbf{Judge: gpt-5.2-pro (rubric.txt)} $\rightarrow$ \textbf{Author: Claude Sonnet 4.5} (Score 0.25): ``The construction does show \(g\in L^{p}\) (both integrals for \(\|g\|_{p}^{p}\) converge), but the argument that \(g\notin L^{q}\) for **every** \(q\neq p\) is incorrect and ultimately abandoned mid-proof. In fact, for fixed \(\varepsilon>0\), the function satisfies \(g\in L^{q}\) for all \(q\) in an open interval around \(p\) (e.g. the conditions near \(0\) and \(\infty\) give \(\int_0^1 x^{-q/p+q\varepsilon}\) convergent when \(q<\frac{p}{1-p\varepsilon}\) and \(\int_1^\infty x^{-q/p-q\varepsilon}\) convergent when \(q>\frac{p}{1+p\varepsilon}\)), so it cannot be “\(L^p\)-only.” The write-up contains contradictions (“impossible for all \(q>p\)”) and replaces the required “for all \(q\neq p\)” conclusion with “for appropriate \(q\),” which does not meet the problem’s demand. Partial credit is warranted only for the correct verification of \(L^{p}\)-integrability, hence the low score.''
    \item \textbf{Judge: gpt-5.2-pro (rubric.txt)} $\rightarrow$ \textbf{Author: Gemini 3.0 Pro Preview} (Score 0.9): ``The student gives an explicit measurable construction \(g=f_1+f_2\) with disjoint supports near \(0\) and \(\infty\), and correctly reduces the \(L^q\) question to two integrals \(I_1(q),I_2(q)\). The changes of variables \(u=\log(1/x)\) and \(u=\log x\) are correctly executed, and the convergence/divergence analysis is essentially right: \(I_1(q)\) diverges for \(q>p\) and converges for \(q\le p\), while \(I_2(q)\) diverges for \(q<p\) and converges for \(q\ge p\), yielding \(g\in L^p\) but \(g\notin L^q\) for all \(q\ne p\). The only weakness is that some convergence claims are justified a bit informally (“integrand approaches \(\infty\)” rather than a standard comparison/lower bound), though the conclusion remains correct. Hence the solution is nearly complete and correct, with minor rigor issues warranting a small deduction.''
    \item \textbf{Judge: gpt-5.2-pro (rubric.txt)} $\rightarrow$ \textbf{Author: Gemini 3.0 Pro Preview} (Score 0.9): ``The construction \(g=f_1+f_2\) with disjoint supports is appropriate, and the substitutions \(u=\log(1/x)\) near \(0\) and \(u=\log x\) near \(\infty\) correctly reduce the \(L^q\) questions to convergence of explicit integrals. The key integrability conclusions are correct: both pieces are in \(L^p\), \(f_1\notin L^q\) for \(q>p\), and \(f_2\notin L^q\) for \(q<p\), so \(g\in L^p\) but \(g\notin L^q\) for all \(q\neq p\). A minor issue is that some convergence claims are justified only heuristically (“integrand approaches \(\infty\)” / “exponential dominates polynomial”) rather than by a clean comparison argument, though the statements are true. Also, the problem statement does not specify the underlying measure space, and the solution implicitly assumes \((0,\infty)\) with Lebesgue measure without explicitly noting this. These small presentation/rigor gaps justify a slight deduction to 0.9/1.0.''
    \item \textbf{Judge: gpt-5.2-pro (rubric.txt)} $\rightarrow$ \textbf{Author: deepseek-v2-prover} (Score 0.5): ``The proposed construction does show \(g\in L^{p}\) (the \(L^{p}\) integral converges for any \(\varepsilon>0\)).  The argument for \(q>p\) is correct: after the substitution \(u=-\log x\), the exponential factor \(e^{(\frac qp-1)u}\) forces divergence.  However, for \(q<p\) the same substitution yields an integrand \(e^{-(1-\frac qp)u}u^{-\alpha}\), which is exponentially decaying and hence integrable for every \(\alpha\), so \(g\in L^{q}\) for all \(q<p\); the student’s claim that one can force divergence by choosing \(\varepsilon\) is incorrect (and in any case \(\varepsilon\) cannot be chosen to satisfy constraints for all \(q<p\) simultaneously).  Thus the solution only separates \(L^{p}\) from \(L^{q}\) with \(q>p\), but fails to prove non-membership for \(q<p\), justifying substantial partial credit (0.5/1.0).''
    \item \textbf{Judge: gpt-5.2-pro (rubric.txt)} $\rightarrow$ \textbf{Author: deepseek-v2-prover} (Score 0.5): ``The solution correctly identifies a standard “borderline” construction and shows \(g\in L^{p}\) by computing \(\int_0^{e^{-1}} x^{-1}(-\log x)^{-1-\varepsilon}\,dx<\infty\).  It also correctly proves \(g\notin L^{q}\) for all \(q>p\), since after the change of variables \(u=-\log x\) the integral contains an exponentially growing factor \(e^{(\frac qp-1)u}\).  However, the argument for \(0<q<p\) is incorrect: when \(\frac qp-1<0\), the exponential factor decays and the integral \(\int_1^\infty e^{-cu}u^{-\alpha}\,du\) converges for every \(\alpha\), so the proposed \(g\) is actually in \(L^q\) for \(q<p\) regardless of \(\varepsilon\).  The “choose \(\varepsilon\) small enough depending on \(q\)” step cannot satisfy all \(q<p\) simultaneously and does not fix the fundamental issue.  Thus the submission establishes only part of the required exclusivity (failure for \(q>p\) but not for \(q<p\)), warranting partial credit (0.5/1.0).''
    \item \textbf{Judge: claude-opus-4-5-20251101 (course-specific-rubric.txt)} $\rightarrow$ \textbf{Author: Claude Sonnet 4.5} (Score 0.5): ``The student correctly identifies the key construction approach using power functions with different exponents on $(0,1)$ and $[1,\infty)$, and accurately verifies that $g \in L^p$ through proper integral calculations. However, the proof that $g \notin L^q$ for all $q \neq p$ is incomplete and contains significant gaps—the student acknowledges mid-proof that their argument breaks down ("Let me reconsider") and fails to provide a rigorous verification for all $q \neq p$. The analysis becomes confused when trying to handle the cases systematically, with the student unable to show that a single fixed $\epsilon$ works for all $q \neq p$ simultaneously. While the overall strategy is sound, the execution lacks the rigor needed for a complete proof, particularly in demonstrating divergence for every $q$ in the specified range.''
    \item \textbf{Judge: claude-opus-4-5-20251101 (course-specific-rubric.txt)} $\rightarrow$ \textbf{Author: Claude Sonnet 4.5} (Score 0.5): ``The student correctly identifies the key construction approach using power functions with different exponents on $(0,1)$ and $[1,\infty)$, and accurately verifies that $g \in L^p$ through proper integral calculations. However, the proof that $g \notin L^q$ for all $q \neq p$ is incomplete and contains significant gaps—the student acknowledges mid-proof that their argument breaks down ("Let me reconsider") and fails to provide a rigorous verification for all $q \neq p$. The analysis of the convergence conditions is partially correct but the student does not successfully demonstrate that a single fixed $\epsilon$ works to exclude all other $L^q$ spaces. A complete solution would require either a more careful choice of $\epsilon$ or a different construction (such as using sums of appropriately scaled functions) to handle all cases rigorously.''
    \item \textbf{Judge: claude-opus-4-5-20251101 (course-specific-rubric.txt)} $\rightarrow$ \textbf{Author: deepseek-v2-prover} (Score 0.25): ``The student's solution attempts to construct a function in $L^p$ but not in $L^q$ for $q \neq p$, but contains fundamental errors. The analysis for $q < p$ is flawed: when $\frac{q}{p} - 1 < 0$, the exponential decay $e^{(\frac{q}{p}-1)u}$ dominates and makes the integral converge, not diverge as claimed. The claim that choosing $\epsilon$ small enough ensures divergence for all $q \neq p$ is mathematically impossible since $\epsilon$ would need to satisfy infinitely many contradictory constraints simultaneously. The student shows some understanding of the problem structure and correctly handles the $q > p$ case, but the core construction fails to achieve the stated goal. The solution demonstrates partial understanding but does not provide a valid proof of the existence of such a function.''
    \item \textbf{Judge: claude-opus-4-5-20251101 (course-specific-rubric.txt)} $\rightarrow$ \textbf{Author: deepseek-v2-prover} (Score 0.25): ``The student's solution attempts to construct a function in $L^p$ but not in $L^q$ for $q \neq p$, but the construction is fundamentally flawed. The claim that choosing $\epsilon$ to satisfy $\epsilon \leq \frac{p}{q} - 1$ for all $q \neq p$ is impossible since this would require $\epsilon \leq \frac{p}{q} - 1$ for arbitrarily large $q$, making $\epsilon$ negative. The analysis for $q < p$ is incorrect: when $\frac{q}{p} - 1 < 0$, the exponential decay $e^{(\frac{q}{p}-1)u}$ actually ensures convergence, not divergence, contradicting the student's claim. The correct approach requires a more sophisticated construction, typically involving sums over disjoint intervals with carefully chosen coefficients. The student demonstrates some understanding of the problem structure but fails to produce a valid proof.''
    \item \textbf{Judge: gemini-3-pro-preview (rubric.txt)} $\rightarrow$ \textbf{Author: Claude Sonnet 4.5} (Score 0.25): ``The student correctly constructs a function that is in $L^p$, but the proposed function fails to satisfy the condition $g \notin L^q$ for *all* $q \neq p$. Specifically, for any fixed $\epsilon > 0$, the constructed function belongs to $L^q$ for all $q$ in a small neighborhood of $p$. The student acknowledges this failure in the proof ("But this is impossible for all $q > p$") but incorrectly concludes the proof works anyway. A correct solution typically requires a sum of disjoint functions with a sequence of $\epsilon_n \to 0$ or a logarithmic factor like $x^{-1/p} (\log x)^{-2/p}$ (with modifications near 0 and $\infty$).''
    \item \textbf{Judge: gemini-3-pro-preview (rubric.txt)} $\rightarrow$ \textbf{Author: deepseek-v2-prover} (Score 0.5): ``The student correctly constructs a function that belongs to $L^p$ and does not belong to $L^q$ for any $q > p$. However, the analysis for $q < p$ is incorrect; the proposed function on a finite interval actually belongs to $L^q$ for all $q < p$ because the exponential decay $e^{(q/p - 1)u}$ (where $q/p - 1 < 0$) ensures convergence regardless of the logarithmic term. To solve the problem fully, one typically needs a sum of two functions with disjoint supports: one that fails $L^q$ for $q > p$ (singular at 0) and one that fails $L^q$ for $q < p$ (decaying slowly at infinity). Because the solution fails to exhibit a function excluding $L^q$ for $q < p$, it is only half correct.''
    \item \textbf{Judge: gemini-3-pro-preview (rubric.txt)} $\rightarrow$ \textbf{Author: deepseek-v2-prover} (Score 0.5): ``The student correctly constructs a function that is in $L^p$ and not in $L^q$ for $q > p$. However, the proposed function actually belongs to $L^q$ for all $q < p$ because the term $x^{-q/p}$ with $q < p$ makes the integral converge near 0 regardless of the logarithmic term (the integrand decays like $x^{\alpha}$ with $\alpha > -1$). Furthermore, the student's logic for choosing $\epsilon$ depends on a specific $q$, failing to provide a single function that works for *all* $q \neq p$ simultaneously. A correct solution typically requires summing a function concentrated at 0 (to fail $L^q$ for $q>p$) and a function with a heavy tail at infinity (to fail $L^q$ for $q<p$).''
    \item \textbf{Judge: gemini-3-pro-preview (rubric.txt)} $\rightarrow$ \textbf{Author: Claude Sonnet 4.5} (Score 0.25): ``The student correctly constructs a function that is in $L^p$, but the proposed function fails to satisfy the condition $g \notin L^q$ for *all* $q \neq p$. Specifically, for a fixed $\epsilon$, the constructed function belongs to $L^q$ for all $q$ in a neighborhood of $p$ (where $|1/p - 1/q| < \epsilon$). The student realizes this flaw in the proof ("But this is impossible for all $q > p$...") but incorrectly concludes that the construction works anyway. A correct solution typically requires a sum of disjoint functions with a sequence of $\epsilon_n \to 0$ or a logarithmic term like $x^{-1/p} (\ln x)^{-2/p}$ (with modifications near 0 and $\infty$).''
    \item \textbf{Judge: deepseek-chat (rubric.txt)} $\rightarrow$ \textbf{Author: Claude Sonnet 4.5} (Score 0.5): ``The solution correctly constructs a function in \(L^p\) but fails to rigorously prove it is not in \(L^q\) for all \(q \neq p\). The analysis of the \(q\)-norms is incomplete, as the chosen fixed \(\epsilon\) cannot simultaneously cause divergence for every \(q \neq p\). Specifically, for \(q\) sufficiently close to \(p\), the integrals may converge, contradicting the claim. A valid construction would require a more sophisticated function, such as one with logarithmic factors, to handle all cases.''
    \item \textbf{Judge: deepseek-chat (rubric.txt)} $\rightarrow$ \textbf{Author: Claude Sonnet 4.5} (Score 0.75): ``The solution correctly constructs a function in \(L^p\) and demonstrates \(g \in L^p\) with a valid computation. However, the argument that \(g \notin L^q\) for all \(q \neq p\) is incomplete: the analysis shows divergence only for certain \(q\) depending on the fixed \(\epsilon\), not universally for all \(q \neq p\). A rigorous proof would require a more careful construction or additional reasoning to cover all cases. Thus, while the core idea is sound, the execution lacks full generality and precision.''
    \item \textbf{Judge: deepseek-chat (rubric.txt)} $\rightarrow$ \textbf{Author: o3-deep-research} (Score 0.75): ``The solution is mathematically correct and well-structured, demonstrating a valid construction and proper integral analysis for the inclusion in \(L^p\). However, it loses points for a subtle oversight: the function's definition and the subsequent integrals implicitly assume \(x > 0\), but the domain is stated as \(\mathbb{R}\) with the function set to zero ``otherwise.'' This is acceptable, but the proof does not explicitly address or justify the behavior at \(x=1\) where \(\ln x = 0\), creating a potential singularity in the formula \(\left(x (\ln x)^2\right)^{-1/p}\). Although this point is outside the stated support, a rigorous solution should note it or adjust the intervals to avoid \(\ln x = 0\). This minor lack of precision prevents a perfect score.''
    \item \textbf{Judge: deepseek-chat (rubric.txt)} $\rightarrow$ \textbf{Author: Gemini 3.0 Pro Preview} (Score 0.75): ``The solution correctly constructs a function and analyzes its integrability, demonstrating the core idea. However, it contains a subtle error: for \(q < p\), the analysis of \(I_1(q)\) incorrectly concludes convergence, while the integral actually diverges due to the logarithmic factor's insufficient decay near \(u \to \infty\) when \(\lambda = 0\). This mistake slightly weakens the proof's precision, though the overall conclusion remains valid because the divergence of \(I_2(q)\) for \(q < p\) already ensures \(g \notin L^q\). The score reflects this minor flaw in an otherwise solid argument.''
    \item \textbf{Judge: deepseek-chat (rubric.txt)} $\rightarrow$ \textbf{Author: Gemini 3.0 Pro Preview} (Score 0.75): ``The solution is mathematically correct and well-structured, but it fails to fully satisfy the problem statement. The constructed function \( g \) is in \( L^p \) and not in \( L^q \) for \( q \neq p \), but the analysis shows \( g \in L^q \) for \( q < p \) from the \( f_1 \) component alone, which contradicts the requirement that \( g \notin L^q \) for **all** \( q \neq p \). The combined function \( g = f_1 + f_2 \) correctly resolves this, but the intermediate claims about \( f_1 \) and \( f_2 \) individually are misleading and should be clarified or omitted. The deduction in the conclusion is correct, but the presentation contains this minor logical flaw.''
    \item \textbf{Judge: deepseek-chat (rubric.txt)} $\rightarrow$ \textbf{Author: gpt-5-pro-2025-10-06} (Score 0.9): ``The solution is mathematically correct and elegantly constructs a function on \((0,\infty)\) with the required properties. The proof is rigorous, handling the cases \(q>p\) and \(q<p\) separately with appropriate variable changes and asymptotic estimates. A minor deduction is warranted because the justification for the inequality \(\frac{1}{(1+t)^{2q/p}} \ge e^{-\varepsilon t}\) is slightly imprecise—it should be derived from the given elementary fact rather than stated directly. Overall, the solution is excellent and nearly perfect.''
    \item \textbf{Judge: deepseek-chat (rubric.txt)} $\rightarrow$ \textbf{Author: deepseek-v2-prover} (Score 0.5): ``The student's solution contains a fundamental error: the constructed function's \(L^p\)-norm is incorrectly computed as \(\frac{1}{\epsilon}\), but this value depends on \(\epsilon\) and is finite, which superficially suggests \(g \in L^p\). However, the subsequent analysis for \(q \neq p\) fails because the choice of \(\epsilon\) is made dependent on \(q\) *after* the function is defined, which is logically inconsistent. A single, fixed function must work for all \(q \neq p\) simultaneously, but the attempted parameter \(\epsilon\) cannot satisfy the divergent conditions for every \(q \neq p\) at once. Therefore, the construction is flawed and the proof is incomplete.''
    \item \textbf{Judge: deepseek-chat (rubric.txt)} $\rightarrow$ \textbf{Author: deepseek-v2-prover} (Score 0.5): ``The student's solution is partially correct but contains a critical error: the final construction fails to ensure \(g \notin L^q\) for all \(q \neq p\) simultaneously. Specifically, the choice of \(\epsilon\) depends on \(q\), making it impossible to select a single \(\epsilon > 0\) that works for every \(q \neq p\). The initial idea of using a logarithmic correction near zero is sound, but the execution is flawed. Therefore, the proof is incomplete and scores 0.5/1.''
    \item \textbf{Judge: grok-4-0709 (rubric.txt)} $\rightarrow$ \textbf{Author: Claude Sonnet 4.5} (Score 0.25): ``The student's construction correctly demonstrates that \(g \in L^p\) by verifying the finiteness of \(\|g\|_p^p\), earning partial credit for this aspect. However, the proof fails to establish that \(g \notin L^q\) for all \(q \neq p\), as the fixed \(\epsilon\) does not ensure divergence of the integrals for every \(q\), particularly those close to \(p\). The analysis of cases for \(q > p\) and \(q < p\) contains inconsistencies, such as attempting to adjust \(\epsilon\) post hoc, which undermines the rigor. Additionally, the extension to \(\mathbb{R}\) and the assumption of Lebesgue measure on \((0,\infty)\) are appropriate but do not salvage the incomplete argument against membership in other \(L^q\) spaces. Overall, the solution shows some understanding but lacks mathematical completeness, justifying a score of 0.25/1.0.''
    \item \textbf{Judge: grok-4-0709 (rubric.txt)} $\rightarrow$ \textbf{Author: Claude Sonnet 4.5} (Score 0.5): ``The student's construction correctly demonstrates that \(g \in L^p\) for a fixed \(p\), as the integrals for \(\|g\|_p^p\) converge due to the chosen exponents and \(\epsilon > 0\).  However, the proof fails to show that this single fixed \(g\) lies outside \(L^q\) for all \(q \neq p\), because the divergence conditions depend on \(\epsilon\) in a way that only excludes \(g\) from \(L^q\) for certain ranges of \(q\), not universally.  The analysis for cases \(q > p\) and \(q < p\) is incomplete and contains errors, such as incorrect inequalities for convergence and unsubstantiated claims about divergence for ``appropriate \(q\)'' without ensuring it holds for every \(q \neq p\).  A correct approach would require a function whose integrability fails precisely at exponents other than \(p\), often achieved via a sum of terms tailored to different intervals.  Overall, while the core idea is sound, the execution lacks rigor, warranting a partial score of 0.5/1.0.''
    \item \textbf{Judge: grok-4-0709 (rubric.txt)} $\rightarrow$ \textbf{Author: o3-deep-research} (Score 0.9): ``The student's construction of \(g(x)\) is mathematically sound overall, correctly demonstrating that \(g \in L^p(\mathbb{R})\) via convergent integrals and that \(g \notin L^q(\mathbb{R})\) for \(q \neq p\) through divergence arguments in both cases \(q > p\) and \(q < p\). However, a minor deduction is warranted due to notational imprecision in handling the expression \([\ln x]^{2q/p}\) for \(x \in (0, 1/2)\), where \(\ln x < 0\), as exponentiating a negative base with a potentially non-integer exponent may yield complex values, compromising the real-valued nature of the function. The substitution effectively assumes \((-v)^{2q/p} = v^{2q/p}\), which holds only under specific conditions on the exponent or if absolute values are used explicitly (e.g., replacing \((\ln x)^2\) with \(|\ln x|^2\) throughout). This oversight slightly undermines the rigor for arbitrary \(p, q > 0\), though the core idea and calculations remain correct. Overall, the solution is thorough and well- structured, justifying a near-perfect score.''
    \item \textbf{Judge: grok-4-0709 (rubric.txt)} $\rightarrow$ \textbf{Author: Gemini 3.0 Pro Preview} (Score 0.9): ``The student's solution correctly constructs a measurable function \(g\) on \((0, \infty)\) with Lebesgue measure, ensuring \(g \in L^p\) but \(g \notin L^q\) for \(q \neq p\), through a precise combination of power-law and logarithmic terms controlling integrability near 0 and \(\infty\). The change- of-variable substitutions for the integrals \(I_1(q)\) and \(I_2(q)\) are accurately performed, and the case analyses for convergence based on the exponential coefficients are mathematically sound. The conclusion logically follows from the divergence and convergence behaviors established for different ranges of \(q\). However, a minor deduction is warranted because the justification for divergence (e.g., ``the integrand approaches \(\infty\)'') is slightly informal and could be strengthened with a more explicit comparison to show the integral exceeds any bound. Overall, the proof demonstrates strong mathematical correctness with only a small oversight in rigor.''
    \item \textbf{Judge: grok-4-0709 (rubric.txt)} $\rightarrow$ \textbf{Author: deepseek-v2-prover} (Score 0.5): ``The student's construction correctly demonstrates that the function \(g\) belongs to \(L^p\) and fails to be in \(L^q\) for \(q > p\), as the integral diverges due to exponential growth in the substituted variable. However, for \(0 < q < p\), the analysis erroneously claims divergence by misapplying conditions on the exponent; in reality, the integral \(\int_1^\infty e^{(q/p - 1)u} u^{-q(1+\epsilon)/p} \, du\) converges because the negative exponential term dominates any polynomial power, implying \(g \in L^q\) for \(q < p\), which contradicts the problem's requirement. The attempt to choose \(\epsilon\) to force divergence for all \(q \neq p\) is mathematically impossible, as it cannot satisfy the condition uniformly for arbitrarily small \(q\). Earlier iterations in the solution also contain errors, such as divergent integrals for \(L^p\) or incorrect substitutions. Overall, the solution achieves partial correctness for one case but fails to address the full scope, justifying a score of 0.5/1.0.''
    \item \textbf{Judge: grok-4-0709 (rubric.txt)} $\rightarrow$ \textbf{Author: deepseek-v2-prover} (Score 0.5): ``The student's construction correctly demonstrates that \( g \in L^p \) with a finite \( \|g\|_p^p \), and it appropriately shows divergence of \( \|g\|_q^q \) for \( q > p \) due to the dominating exponential term. However, for \( 0 < q < p \), the integral \( \|g\|_q^q \) actually converges because the exponential decay \( e^{(q/p - 1)u} \) with \( q/p - 1 < 0 \) overpowers the polynomial term, contrary to the student's claim of divergence. This flaw means the function fails to exclude membership in \( L^q \) for \( q < p \), addressing only half of the required conditions. The choice of \( \epsilon \) does not resolve this issue, as the convergence behavior for \( q < p \) persists regardless. Overall, the solution is mathematically correct for \( q > p \) but incorrect for \( q < p \), warranting half credit.''
    \item \textbf{Judge: meta-llama/llama-4-maverick (rubric.txt)} $\rightarrow$ \textbf{Author: Claude Sonnet 4.5} (Score 0.75): ``The student's solution is mostly correct, as they successfully construct a measurable function $g$ that belongs to $L^p$ but not to $L^q$ for $q \neq p$. The proof is well-structured, and the calculations are accurate. However, the justification for the divergence of the integrals for $q \neq p$ is somewhat informal and relies on implicit limiting arguments, which could be made more rigorous. Overall, the solution demonstrates a good understanding of the problem, but could benefit from more precise mathematical language.''
    \item \textbf{Judge: meta-llama/llama-4-maverick (rubric.txt)} $\rightarrow$ \textbf{Author: Claude Sonnet 4.5} (Score 0.9): ``The student's solution is mathematically sound and effectively constructs a measurable function $g$ that belongs exclusively to $L^p$. The proof is well- structured, and the analysis of the function's behavior for $q \neq p$ is thorough, correctly demonstrating that $g \notin L^q$. The minor deduction (0.1) likely stems from a slight imprecision or verbosity in the explanation, rather than a mathematical error. Overall, the solution demonstrates a strong understanding of the problem and Lebesgue spaces.''
    \item \textbf{Judge: meta-llama/llama-4-maverick (rubric.txt)} $\rightarrow$ \textbf{Author: deepseek-v2-prover} (Score 0.9): ``The student's solution is largely correct, as they successfully construct a measurable function $g$ that belongs to $L^p$ but not to any other $L^q$ space for $q \neq p$. The proof is well-structured and clearly presented, with careful consideration of the cases where $q > p$ and $q < p$. However, the solution could be improved by clarifying the choice of $\epsilon$ to ensure that it satisfies the condition $\epsilon \leq \frac{p}{q} - 1$ for all $q \neq p$ simultaneously, which is not possible for all $q$ close to $0$. Nevertheless, the student's work demonstrates a good understanding of the problem and the relevant mathematical concepts. The minor issues with the solution do not detract significantly from its overall correctness.''
    \item \textbf{Judge: meta-llama/llama-4-maverick (rubric.txt)} $\rightarrow$ \textbf{Author: deepseek-v2-prover} (Score 0.9): ``The student's solution is largely correct, as they successfully construct a measurable function $g$ that belongs to $L^p$ but not to any other $L^q$ space for $q \neq p$. The proof is well-structured and clearly presented, with appropriate use of substitutions to evaluate the relevant integrals. However, the solution could be improved by clarifying the choice of $\epsilon$ and ensuring it works for all $q \neq p$ simultaneously. Overall, the minor issues do not detract significantly from the correctness of the proof.''
    \item \textbf{Judge: qwen/qwen-max (rubric.txt)} $\rightarrow$ \textbf{Author: Claude Sonnet 4.5} (Score 0.5): ``The student's solution demonstrates a good understanding of constructing a function that belongs to a specific $L^p$ space but not to any other $L^q$ space. The function $g$ is defined appropriately, and the integrals are analyzed correctly for the case $g \in L^p$. However, the justification for $g \notin L^q$ for $q \neq p$ is not fully rigorous, particularly in handling the limits as $q$ approaches $p$ from both sides. The student's argument for the divergence of the integrals in the cases $q < p$ and $q > p$ is imprecise, leading to a partially correct but incomplete proof.''
    \item \textbf{Judge: qwen/qwen-max (rubric.txt)} $\rightarrow$ \textbf{Author: Claude Sonnet 4.5} (Score 0.9): ``The solution is mathematically sound and provides a clear, rigorous construction of a function \( g \) that is in \( L^p \) but not in \( L^q \) for \( q \neq p \). The student correctly handles the behavior of the function near zero and at infinity, ensuring the function's integrability in \( L^p \) and non-integrability in \( L^q \) for other values of \( q \). The proof is detailed and the steps are well-explained, making the solution clear and understandable. A minor deduction is made for a slight confusion in the analysis of the conditions for \( q \) near the end, but the overall argument and conclusion are correct, justifying a score of 0.9/1.0.''
    \item \textbf{Judge: qwen/qwen-max (rubric.txt)} $\rightarrow$ \textbf{Author: o3-deep-research} (Score 0.9): ``The solution is mathematically rigorous and correct, demonstrating the existence of a function \( g \) that belongs to \( L^p \) but not to \( L^q \) for any \( q \neq p \). The student carefully constructs the function \( g \) and provides detailed calculations for the integrals, showing that \( g \in L^p \) through the convergence of the relevant integrals and \( g \notin L^q \) for \( q \neq p \) by proving the divergence of the integrals. The use of substitutions and the comparison test for the integrals is appropriately applied. The deduction of 0.1 reflects minor issues, such as the need for a more explicit statement of the comparison test and a slight lack of clarity in the explanation of the exponential growth dominance. Overall, the solution is well-structured and effectively addresses the problem.''
    \item \textbf{Judge: qwen/qwen-max (rubric.txt)} $\rightarrow$ \textbf{Author: deepseek-v2-prover} (Score 0.75): ``The solution constructs a function \( g \) that is in \( L^p \) but not in \( L^q \) for \( q \neq p \), which is mathematically correct. The student provides a detailed and rigorous proof, including the necessary calculations and transformations. However, the initial construction and some intermediate steps contain miscalculations and require corrections, which slightly affect the clarity and flow of the solution. The final function and the proof of its properties are accurate, but the presentation could be more streamlined. The score of 0.75/1.0 reflects the overall correctness and the minor issues in the initial steps.''
    \item \textbf{Judge: qwen/qwen-max (rubric.txt)} $\rightarrow$ \textbf{Author: deepseek-v2-prover} (Score 0.5): ``The student's solution demonstrates a good understanding of the problem and provides a valid construction of the function \( g \) that belongs to \( L^p \) but not to any other \( L^q \) for \( q \neq p \). The construction and analysis of the function are generally correct, but there are some minor issues in the exposition and the choice of \(\epsilon\). The student initially makes a mistake in the integral calculation but corrects it later. The final choice of \(\epsilon\) is not clearly justified, and the condition \(\epsilon \leq \frac{p}{q} - 1\) for all \( q \neq p \) is not well-explained. These issues slightly reduce the clarity and rigor of the solution, leading to a score of 0.5/1.0.''
\end{itemize}

\subsection*{Problem Domain: Combinatorics (Problem 6)}
\begin{itemize}
    \item \textbf{Judge: gpt-5.2-pro (rubric.txt)} $\rightarrow$ \textbf{Author: Claude Sonnet 4.5} (Score 0.5): ``The solution correctly starts from a standard exponential generating function for the Eulerian numbers and the substitution \(t=-1\) correctly produces the EGF for \(\sum_k(-1)^k E(n,k)\).  However, the subsequent complex substitution \(x\mapsto ix\) is not justified (and is unnecessary), and it leads to an incorrect identification of coefficients: the series \(1-i\tan x\) has no \(\sec x\) (even-degree) terms beyond the constant, so the conclusion \(\sum_k(-1)^kE(2m,k)=(-1)^mE_{2m}\) cannot follow from the displayed comparison.  In particular, checking small cases (e.g. \(n=2\)) shows the claimed sign relation fails as stated.  Thus, while the initial generating- function setup is sound, the key coefficient-extraction step that is supposed to connect the alternating sum to \(E_n\) is mathematically wrong/incomplete, warranting only partial credit (0.5/1.0).''
    \item \textbf{Judge: gpt-5.2-pro (rubric.txt)} $\rightarrow$ \textbf{Author: Claude Sonnet 4.5} (Score 0.25): ``[Error: OpenAI Responses API failed - Connection error.]''
    \item \textbf{Judge: gpt-5.2-pro (rubric.txt)} $\rightarrow$ \textbf{Author: gpt-5-pro-2025-10-06} (Score 0.9): ``The solution correctly translates the desired alternating sum into \(A_n(-1)=\sum_k(-1)^k E(n,k)\) and derives the standard Eulerian EGF \(\sum_{n\ge0}A_n(t)x^n/n!=\frac{(1-t)e^{(1-t)x}}{1-te^{(1-t)x}}\) (despite a minor notational slip writing \((1-t)^{n+1}\) inside the \(n\)-sum, the subsequent computation is the correct one).  Specializing to \(t=-1\) gives \(1+\tanh x\), from which the vanishing for even \(n\) and the identification of odd coefficients with alternating-permutation numbers via \(\tanh x=\sum_{m\ge0}(-1)^mE_{2m+1}x^{2m+1}/(2m+1)!\) is mathematically sound, yielding the correct sign-corrected relation.  The additional symmetry argument \(E(n,k)=E(n,n-1-k)\) correctly double-checks the even case.  The only substantive weakness is that the key identification of \(\tanh x\)’s coefficients with the zigzag numbers is cited as “well known” rather than proved, so the exposition is slightly less self-contained than ideal.''
    \item \textbf{Judge: gpt-5.2-pro (rubric.txt)} $\rightarrow$ \textbf{Author: o3-deep-research} (Score 0.5): ``The solution sets up inclusion–exclusion appropriately via the “failure events” \(A_i\), but then incorrectly analyzes \(|A_J|\): if \(A_J=\bigcap_{i\in J}A_i\), positions \(k\notin J\) are unrestricted, so the descent set (and even the number of descents) is not determined by \(J\) as claimed. The proposed construction “swap \(\pi(i)\) with \(\pi(i+1)\) for each descent position” does not, in general, produce a permutation with a prescribed descent set, so the asserted correspondence between \(J\) and descent sets is not justified. Later, the argument that an alternating permutation has \(\lfloor (n-1)/2\rfloor\) descents “which is even for all \(n\)” is false, and the step rewriting the inclusion–exclusion sum as \(\sum_{\pi\in S_n}(-1)^{\mathrm{des}(\pi)}\) is not supported by the preceding computations. Because the final identity is stated correctly and the inclusion–exclusion idea is relevant but the core counting step is mathematically incorrect, partial credit (0.5/1.0) is warranted.''
    \item \textbf{Judge: gpt-5.2-pro (rubric.txt)} $\rightarrow$ \textbf{Author: o3-deep-research} (Score 0.25): ``The inclusion–exclusion setup is not carried out correctly: \(A_J=\bigcap_{i\in J}A_i\) should impose only the *violations* at indices in \(J\), but the solution incorrectly forces a specific inequality at every \(k\notin J\) as well, so the described “pattern of ascents/descents” and the formula for \(D(\pi)\) in terms of \(J\) do not follow. Because of this, the claimed properties of \(|A_J|\) (e.g., that all \(\pi\in A_J\) have the same number of descents and that each permutation belongs to a unique \(A_J\)) are false, so the regrouping into \(\sum_{\pi\in S_n}(-1)^{\mathrm{des}(\pi)}\) is unjustified. The argument also contains an explicit error: an alternating permutation has \(\lfloor (n-1)/2\rfloor\) descents, which is not “even for all \(n\),” so the sign reasoning is wrong. Since the core bijective/counting step linking inclusion–exclusion terms to Eulerian numbers is invalid, only minimal credit can be given for stating the intended identity and using standard terminology.''
    \item \textbf{Judge: gpt-5.2-pro (rubric.txt)} $\rightarrow$ \textbf{Author: Gemini 3.0 Pro Preview} (Score 0.9): ``The solution correctly expresses the desired alternating sum as \(A_n(-1)\) and evaluates it via the standard exponential generating function \( \sum_{n\ge0}A_n(x)\frac{z^n}{n!}=\frac{x-1}{x-e^{(x-1)z}} \), yielding \(A(-1,z)=1+\tanh z\).  It then correctly extracts coefficients to conclude \(A_n(-1)=0\) for even \(n>0\) and \(A_n(-1)=(-1)^{(n-1)/2}E_n\) for odd \(n\), which matches the statement “\(E_n\) is the alternating sum” up to the necessary sign convention.  The main deduction relies on known generating- function facts and the known EGF \(\sec z+\tan z\) for alternating permutations, which is acceptable but not fully justified within the write-up. The brief “combinatorial cancellation” remark for even \(n\) is not proved and is also not needed for the generating-function argument.  Overall the mathematics is essentially correct with only minor gaps in justification and sign-convention clarity, warranting 0.9/1.0.''
    \item \textbf{Judge: gpt-5.2-pro (rubric.txt)} $\rightarrow$ \textbf{Author: deepseek-v2-prover} (Score 0.5): ``The solution correctly identifies that \(\sum_{k}(-1)^k E(n,k)=A_n(-1)\), i.e., it is the Eulerian polynomial evaluated at \(-1\), but it does not actually *prove* that this equals \(E_n\) (or the correct signed/zero variant) and instead appeals to “it is known” without justification. It also contains a key inconsistency: it claims the boxed equality \(\sum_k(-1)^k E(n,k)=E_n\) unconditionally, while earlier noting (correctly) that \(A_n(-1)=0\) for even \(n\) and involves a sign for odd \(n\), and it never resolves which definition of \(E_n\) the problem uses. The argument about cancellation of non-alternating permutations via “symmetry” is not made precise (no explicit involution or inclusion–exclusion setup), so the core combinatorial reasoning is missing. Thus, there is partial relevant insight (connection to Eulerian polynomials and parity behavior) but the proof is incomplete and the final claimed identity is not mathematically justified as stated, warranting 0.5/1.0.''
    \item \textbf{Judge: gpt-5.2-pro (rubric.txt)} $\rightarrow$ \textbf{Author: Gemini 3.0 Pro Preview} (Score 0.9): ``The solution correctly reformulates the desired identity as evaluating \(A_n(-1)=\sum_k(-1)^k E(n,k)\) and uses the standard exponential generating function \(A(x,z)=\frac{x-1}{x-e^{(x-1)z}}\) to compute \(A(-1,z)=1+\tanh z\), which is mathematically sound.  From this it correctly concludes that the alternating sum is \(0\) for even \(n>0\) and equals \((\!-1)^{(n-1)/2}E_n\) for odd \(n\), i.e. the absolute value matches the number of alternating permutations.  The small deduction is that the link between \(\tanh z\) and the alternating-permutation EGF \(\sec z+\tan z\) is not fully justified in the writeup (it implicitly uses \(\tan z=\sum E_{2k+1}z^{2k+1}/(2k+1)!\) and the \(z\mapsto iz\) substitution), so a reader must supply a known fact.  Also, the brief “combinatorial explanation” via complement for the even case is not proved and is somewhat mis-stated (the cancellation is already analytic from \(1+\tanh z\) being odd up to the constant term).  Overall the core argument is correct and complete enough to earn 0.9/1.0 with only minor gaps in justification.''
    \item \textbf{Judge: gpt-5.2-pro (rubric.txt)} $\rightarrow$ \textbf{Author: deepseek-v2-prover} (Score 0.25): ``The solution never actually proves the claimed identity: it appeals to “inclusion–exclusion” and “cancellation by symmetry” without defining an involution or giving a valid cancellation argument on non-alternating permutations. The generating-function discussion invokes a known fact about the Eulerian polynomial \(A_n(-1)\) but does not derive it, and it even states the standard parity result \(A_n(-1)=0\) for even \(n\), which contradicts the final boxed claim \(\sum_k(-1)^k E(n,k)=E_n\) as written. The treatment of even \(n\) is inconsistent (it suggests the sum is \(0\) yet concludes it equals \(E_n\)), and the claim about “properly accounting for signs and types” is left vague rather than proved. Some correct observations about descent positions in alternating permutations are present, but they are not connected to a correct proof of the required alternating-sum formula. Hence only minimal partial credit (0.25/1.0) is warranted for relevant definitions and partial facts without a correct argument.''
    \item \textbf{Judge: gpt-5.2-pro (rubric.txt)} $\rightarrow$ \textbf{Author: gpt-5-pro-2025-10-06} (Score 0.9): ``The solution correctly identifies that the desired alternating sum is \(A_n(-1)=\sum_k(-1)^k E(n,k)\) and derives the standard bivariate EGF \(\sum_{n\ge0}A_n(t)x^n/n!=\frac{(1-t)e^{(1-t)x}}{1-te^{(1-t)x}}\) from Worpitzky’s identity.  Specializing to \(t=-1\) gives \(1+\tanh x\), from which it correctly concludes \(A_{2m}(-1)=0\) and \(A_{2m+1}(-1)=(-1)^mE_{2m+1}\), matching the relationship between Eulerian numbers and alternating (zigzag) numbers up to the unavoidable sign convention.  The additional symmetry argument \(E(n,k)=E(n,n-1-k)\) correctly explains the vanishing for even \(n\). The small deduction (0.9 instead of 1.0) is because the step “it is well known … that \(\tanh x=\sum (-1)^mE_{2m+1}x^{2m+1}/(2m+1)!\)” is cited rather than proved, so the link to André’s theorem is not fully justified within the solution.''
    \item \textbf{Judge: claude-opus-4-5-20251101 (rubric.txt)} $\rightarrow$ \textbf{Author: Claude Sonnet 4.5} (Score 0.75): ``The student demonstrates strong understanding of the problem by correctly employing generating function techniques for Eulerian numbers and connecting them to Euler numbers. The key steps—substituting t = -1 into the Eulerian generating function, simplifying to obtain \(2/(1+e^{2x})\), and relating this to sec(x) and tan(x)—are mathematically sound. The coefficient comparison correctly identifies that the alternating sum equals \(\pm E_n\) with the sign depending on n. However, the problem statement asks to prove that \(E_n\) equals ``the alternating sum'' without specifying the sign convention, and the student's final answer includes an additional sign factor \((-1)^{\lfloor n/2 \rfloor}\) that may not align with the intended interpretation. The proof is essentially complete but the presentation of the final result introduces ambiguity about whether this fully addresses the problem as stated.''
    \item \textbf{Judge: claude-opus-4-5-20251101 (rubric.txt)} $\rightarrow$ \textbf{Author: o3-deep-research} (Score 0.5): ``The student correctly sets up the inclusion-exclusion framework and properly defines the events \(A_i\) as failures of alternating conditions. However, the proof contains a critical error in Step 3: the claim that each permutation with $k$ descents appears in exactly one term $|A_J|$ is incorrect—a permutation belongs to $A_J$ when it fails conditions at positions in $J$, but the descent set relationship in equation (2) doesn't establish this bijection properly. The transition from equation (4) to the final result conflates the set of failed conditions $F(\pi)$ with descent positions, which are related but not identical. While the final formula \(E_n = \sum_{k=0}^{n-1}(-1)^k E(n,k)\) is correct, the argument connecting inclusion-exclusion over failure sets to the alternating sum over Eulerian numbers lacks rigorous justification.''
    \item \textbf{Judge: claude-opus-4-5-20251101 (rubric.txt)} $\rightarrow$ \textbf{Author: Claude Sonnet 4.5} (Score 0.75): ``The student demonstrates strong understanding of the generating function approach, correctly identifying the Eulerian number generating function and substituting t = -1 to obtain the alternating sum. The algebraic manipulations through Steps 2-4 are mathematically sound, including the transformation to hyperbolic and trigonometric functions. However, the final result includes a sign factor \((-1)^{\lfloor n/2 \rfloor}\) that the problem statement doesn't mention—the problem asks to prove \(E_n\) equals the alternating sum directly, suggesting either the problem intends a specific sign convention or the student should have clarified this discrepancy. The coefficient comparison in Step 5 contains the essential correct ideas but the presentation could be cleaner in tracking the substitution \(x \to ix\) through the original generating function. Overall, the proof captures the core mathematical relationship with appropriate rigor, though the sign ambiguity prevents a complete resolution of the stated problem.''
    \item \textbf{Judge: claude-opus-4-5-20251101 (rubric.txt)} $\rightarrow$ \textbf{Author: o3-deep-research} (Score 0.5): ``The student correctly sets up the inclusion-exclusion framework and identifies the relationship between descent sets and the alternating conditions. However, there is a critical error in Step 3: the claim that each permutation with \(k\) descents appears in exactly one term \(|A_J|\) is incorrect—permutations can satisfy multiple failure conditions simultaneously, appearing in multiple intersection terms. The argument in equation (4) conflates summing over subsets J with summing over permutations, leading to an incorrect reorganization of the inclusion-exclusion sum. Additionally, the claim that \(\text{des}(\pi) = \lfloor(n-1)/2\rfloor\) is even for all n is false (e.g., \(n=4\) gives 1 descent). While the overall approach shows understanding of the problem structure, these mathematical errors prevent the proof from being complete and correct.''
    \item \textbf{Judge: claude-opus-4-5-20251101 (rubric.txt)} $\rightarrow$ \textbf{Author: deepseek-v2-prover} (Score 0.5): ``The student correctly identifies the key concepts—alternating permutations, descents, and the Eulerian polynomial \(A_n(t)\)—and appropriately frames the problem in terms of evaluating \(A_n(-1)\). However, the proof contains significant mathematical errors: the claim that \(A_n(-1)\) = 0 for \(n\) even and equals \((-1)^{(n-1)/2}E_n\) for n odd is incorrect, as \(A_n(-1)\) = 0 for all \(n \ge 1\) by a standard symmetry argument. The student's analysis of descent counts in alternating permutations is partially correct but the combinatorial argument fails to establish a rigorous connection between the alternating sum and \(E_n\). The ``Simplified Conclusion'' essentially asserts the result without proper justification, and the final remark acknowledges uncertainty about the exact relationship. The score of 0.5 reflects that the student demonstrates understanding of relevant concepts but fails to provide a mathematically correct and complete proof.''
    \item \textbf{Judge: claude-opus-4-5-20251101 (rubric.txt)} $\rightarrow$ \textbf{Author: Gemini 3.0 Pro Preview} (Score 0.9): ``The student demonstrates strong command of generating function techniques, correctly deriving the bivariate generating function for Eulerian numbers and evaluating it at \(x = -1\) to obtain the identity involving \(1 + \tanh(z)\). The connection between hyperbolic and trigonometric functions is handled correctly, and the coefficient extraction is mathematically sound. However, the problem asks to prove that \(E_n\) equals the alternating sum of \(E(n,k)\), but the student's result shows the alternating sum equals \((-1)^{(n-1)/2}E_n\) for odd \(n\) and 0 for even \(n\), which includes a sign factor not addressed in reconciling with the problem statement. The student does not clarify what interpretation of ``alternating sum'' would yield exactly \(E_n\), leaving ambiguity about whether the problem's claim is fully verified. The mathematical work is rigorous and nearly complete, but the final connection to the exact problem statement requires additional clarification.''
    \item \textbf{Judge: claude-opus-4-5-20251101 (rubric.txt)} $\rightarrow$ \textbf{Author: deepseek-v2-prover} (Score 0.5): ``The student correctly identifies the key concepts—alternating permutations, descents, and the Eulerian polynomial \(A_n(t)\)—and appropriately frames the problem in terms of evaluating \(A_n(-1)\). However, the proof contains significant mathematical errors: the claim that \(A_n(-1)\) = 0 for \(n\) even and equals \((-1)^{(n-1)/2}E_n\) for n odd is incorrect, as \(A_n(-1)\) = 0 for all \(n \ge 1\) by a standard symmetry argument. The student's ``detailed argument'' lacks a rigorous combinatorial or algebraic proof connecting the alternating sum to \(E_n\), instead relying on vague claims about cancellation and ``unique descent structure.'' The final boxed answer asserts the desired result without establishing it through valid mathematical reasoning. The partial credit reflects correct setup and identification of relevant concepts, but the core proof is incomplete and contains errors.''
    \item \textbf{Judge: gemini-3-pro-preview (rubric.txt)} $\rightarrow$ \textbf{Author: Claude Sonnet 4.5} (Score 0.75): ``The student correctly identifies the generating function for Eulerian numbers and attempts to evaluate it at $t=-1$. However, there is a significant error in Step 5: the student claims that the generating function for Euler numbers is $\sec(x) + \tan(x)$ (which is correct) but then incorrectly matches coefficients from their derived expression $1 - i\tan(x)$. Specifically, the term $\sec(x)$ (corresponding to even $n$) is missing from their derived expression $1 - i\tan(x)$, leading to the incorrect conclusion that the alternating sum for even $n$ is $0$ (implied by the missing term) or just confusing the derivation for even $n$. In reality, the alternating sum of Eulerian numbers is $0$ for all $n \ge 2$, which contradicts the problem statement as written, but the student's derivation forces a match to $E_n$ through algebraic errors.''
    \item \textbf{Judge: gemini-3-pro-preview (rubric.txt)} $\rightarrow$ \textbf{Author: Claude Sonnet 4.5} (Score 0.75): ``The student correctly identifies and uses the generating function for Eulerian numbers to derive the alternating sum. However, there is a significant error in Step 5: the student equates the coefficient of the derived generating function $1 - i\tan(x)$ to the Euler numbers $E_n$. The standard generating function for alternating permutations is $\sec(x) + \tan(x)$, but the student's derivation leads to a function involving only $\tan(x)$ (implying the sum is 0 for even $n$), which contradicts the known property that the alternating sum of Eulerian numbers is 0 for even $n$ and related to the tangent numbers for odd $n$. The student forces a match for the even case ($E_{2m}$) that does not exist in their derived formula ($1 - i\tan(x)$ has no real part for $x>0$ other than 1), leading to an incorrect conclusion for even $n$.''
    \item \textbf{Judge: gemini-3-pro-preview (rubric.txt)} $\rightarrow$ \textbf{Author: o3-deep-research} (Score 0.5): ``The student correctly sets up the Inclusion-Exclusion Principle to count alternating permutations but fails to connect the term $\sum (-1)^{|J|} |A_J|$ to the target sum $\sum (-1)^k E(n,k)$. Specifically, the student incorrectly asserts in equation (5) that the IEP sum simplifies to $\sum_{\pi \in S_n} (-1)^{\text{des}(\pi)}$, which implies every permutation contributes $(-1)^{\text{des}(\pi)}$ to the count; however, the IEP actually sums over subsets of violated conditions, not descents directly. The logic that non- alternating permutations cancel out to 0 is correct for counting $E_n$, but it does not prove the identity involving Eulerian numbers, as the student simply assumes the final form matches the alternating sum of \(E(n,k)\) without justification. Consequently, the proof is incomplete because the link between the IEP terms and the specific coefficient $(-1)^k$ for permutations with $k$ descents is missing or incorrect.''
    \item \textbf{Judge: gemini-3-pro-preview (rubric.txt)} $\rightarrow$ \textbf{Author: deepseek-v2-prover} (Score 0.5): ``The student correctly identifies the definitions involved (Eulerian numbers, alternating permutations) and attempts to use the generating function $A_n(t)$. However, the student's claim that $\sum (-1)^k E(n,k) = E_n$ is mathematically incorrect for even $n$ (where the sum is 0) and for odd $n$ (where the sum is related to the tangent numbers, not the total count of all alternating permutations). The student acknowledges this discrepancy in the ``Combinatorial Interpretation'' section but then incorrectly concludes the proof by simply boxing the equality asked for in the prompt without resolving the contradiction that the sum is 0 for even $n$. The score reflects partial credit for identifying the relevant concepts and the generating function evaluation, but a failure to provide a correct proof or identify the necessary conditions (e.g., $n$ must be odd, or the definition of $E_n$ in the prompt refers specifically to tangent numbers rather than the total count of zig-zag permutations).''
    \item \textbf{Judge: gemini-3-pro-preview (rubric.txt)} $\rightarrow$ \textbf{Author: deepseek-v2-prover} (Score 0.5): ``The student correctly identifies the relevant definitions and the connection to Eulerian polynomials, specifically evaluating $A_n(t)$ at $t=-1$. However, their conclusion that $\sum (-1)^k E(n,k) = E_n$ is mathematically incorrect for the standard definition of $E_n$ provided in the problem (the total number of alternating permutations). In reality, $\sum (-1)^k E(n,k) = 0$ for even $n$, while $E_n > 0$. The student acknowledges this discrepancy in the ``Combinatorial Interpretation'' section but then forces the incorrect conclusion in the final box without resolving the contradiction.''
    \item \textbf{Judge: gemini-3-pro-preview (rubric.txt)} $\rightarrow$ \textbf{Author: o3-deep-research} (Score 0.5): ``The student correctly identifies that an Inclusion-Exclusion argument is a viable path, but the execution is fundamentally flawed. Specifically, the claim that the intersection of failure sets $|A_J|$ corresponds to permutations with a *fixed* descent set is incorrect; $|A_J|$ counts permutations whose descent set *contains* specific indices and *excludes* others, but does not fix the descent set entirely (indices not in $J$ are unconstrained). Consequently, the logic that every permutation in $A_J$ has the same number of descents is false, invalidating the transition to the final summation. While the final result is stated, the derivation connecting the alternating sum of Eulerian numbers to the number of alternating permutations is mathematically unsound.''
    \item \textbf{Judge: deepseek-chat (rubric.txt)} $\rightarrow$ \textbf{Author: Claude Sonnet 4.5} (Score 0.75): ``The solution correctly derives the alternating sum identity using generating functions, demonstrating strong mathematical insight. However, it contains a subtle error: the generating function for Eulerian numbers is actually \(\frac{1-t}{1-t e^{x(1-t)}}\), but the student incorrectly wrote the exponent as \((1-t)x\) in the denominator's exponential, which is mathematically equivalent but then incorrectly handled the substitution \(t=-1\), leading to \(e^{2x}\) instead of \(e^{2x}\)? Wait, careful: \(1-(-1)=2\) and \(1-(-1)e^{(1-(-1))x}=1+e^{2x}\), so the expression \(\frac{2}{1+e^{2x}}\) is correct. The error is in Step 3: the simplification to \(\frac{e^{-x}}{\cosh x}\) is valid, but the subsequent substitution \(x \to ix\) is applied to the left-hand side's variable, which changes the series variable inconsistently. The correct approach is to recognize \(\frac{2}{1+e^{2x}} = \text{sech}(x)\), whose generating function involves Euler numbers. The manipulation with complex substitution is unnecessarily convoluted and introduces potential sign errors, though the final answer is correct. The deduction lacks clarity in connecting \(\text{sech}(x)\) to \(E_n\), and the complex variable step is not rigorously justified, leading to a minor but notable flaw in exposition.''
    \item \textbf{Judge: deepseek-chat (rubric.txt)} $\rightarrow$ \textbf{Author: Claude Sonnet 4.5} (Score 0.75): ``The solution correctly derives the alternating sum identity using generating functions, demonstrating strong technical proficiency. However, it contains a subtle but significant error: the generating function for the alternating sum simplifies to \(\frac{2}{1+e^{2x}} = \text{sech}(x)\), not the expression involving \(e^{-ix}\sec(x)\) used. This misstep invalidates the subsequent coefficient comparisons. The core idea is valid, but the flawed algebraic manipulation prevents a fully correct proof. Therefore, a high but not perfect score is appropriate.''
    \item \textbf{Judge: deepseek-chat (rubric.txt)} $\rightarrow$ \textbf{Author: o3-deep-research} (Score 0.75): ``The solution correctly establishes the identity using inclusion-exclusion and demonstrates a clear understanding of the combinatorial setup. However, it contains a subtle error in Step 2: the claim that every permutation counted by \(|A_J|\) has the *same* number of descents \(k(J)\) is incorrect. While the descent set \(D(\pi)\) is uniquely determined by \(J\), the number of descents \(|D(\pi)|\) is indeed fixed for a given \(J\), but the subsequent grouping in Step 3 implicitly assumes each permutation with \(k\) descents appears in exactly one \(A_J\), which is true but not properly justified from the flawed claim. The core idea and final result are correct, but this logical misstep in the intermediate reasoning deducts points.''
    \item \textbf{Judge: deepseek-chat (rubric.txt)} $\rightarrow$ \textbf{Author: o3-deep-research} (Score 0.75): ``The solution is mathematically correct and provides a clear inclusion-exclusion argument, but it contains a subtle logical flaw: the claim that every permutation with exactly \(k\) descents appears in exactly one \(A_J\) is not fully justified. Specifically, the mapping between descent sets \(D\) and subsets \(J\) via equation (2) is bijective, but this does not automatically imply that summing over \(J\) groups permutations by their descent count in the claimed manner without additional justification about the contributions of each permutation to the alternating sum. The core idea is valid, but this step lacks rigorous explanation, slightly diminishing the proof's completeness.''
    \item \textbf{Judge: deepseek-chat (rubric.txt)} $\rightarrow$ \textbf{Author: Gemini 3.0 Pro Preview} (Score 0.9): ``The solution is mathematically correct and elegantly connects Eulerian numbers to alternating permutations via generating functions, earning high marks. However, it implicitly assumes the standard generating function for Eulerian numbers without derivation or citation, which is a minor omission in rigor. Additionally, the final summary could more explicitly state that \(E_n\) is the absolute value of the alternating sum for odd \(n\), aligning perfectly with the problem statement. These small issues prevent a perfect score.''
    \item \textbf{Judge: deepseek-chat (rubric.txt)} $\rightarrow$ \textbf{Author: Gemini 3.0 Pro Preview} (Score 0.9): ``The solution is mathematically correct and elegantly connects Eulerian numbers to alternating permutations via generating functions, establishing the desired alternating sum. However, it fails to explicitly state the final relationship in the exact form requested: that \(E_n\) equals the alternating sum, which is only true up to a sign factor for odd \(n\). The derivation is rigorous, but the conclusion slightly misrepresents the problem statement by not clarifying the sign discrepancy. Therefore, a minor deduction is warranted.''
    \item \textbf{Judge: deepseek-chat (rubric.txt)} $\rightarrow$ \textbf{Author: deepseek-v2-prover} (Score 0.5): ``The solution correctly identifies the generating function \(A_n(-1)\) and its known relationship to alternating permutations, but it fails to prove the required identity \(E_n = \sum_k (-1)^k E(n,k)\). Instead, it incorrectly states the sum equals \(E_n\) without justification, and it confuses signed vs. unsigned counts. The reasoning is incomplete and contains inaccuracies, such as claiming the sum is zero for even \(n\) while asserting the opposite. Therefore, it demonstrates partial understanding but lacks a valid proof.''
    \item \textbf{Judge: deepseek-chat (rubric.txt)} $\rightarrow$ \textbf{Author: deepseek-v2-prover} (Score 0.5): ``The solution incorrectly states the final equality without proper sign handling, as the alternating sum \(A_n(-1)\) is known to equal \((-1)^{(n-1)/2}E_n\) for odd \(n\) and 0 for even \(n\), not simply \(E_n\). The student acknowledges this nuance in the remark but still presents an unqualified boxed answer, which is mathematically inaccurate. While the exposition shows some understanding of the concepts, the core claim is flawed. Therefore, the score reflects partial credit for conceptual awareness but failure to correctly prove the stated identity.''
    \item \textbf{Judge: grok-4-0709 (rubric.txt)} $\rightarrow$ \textbf{Author: Claude Sonnet 4.5} (Score 0.5): ``[Error: API Request failed - Connection error.]''
    \item \textbf{Judge: grok-4-0709 (rubric.txt)} $\rightarrow$ \textbf{Author: Claude Sonnet 4.5} (Score 0.5): ``The student's solution demonstrates a strong grasp of generating functions and correctly derives the expression \(\frac{2}{1 + e^{2x}} = \frac{e^{-x}}{\cosh x}\), followed by the substitution \(x \to ix\) to obtain \(1 - i \tan x\). However, there is a critical mathematical error in connecting this to the generating function for Euler numbers, as \(1 - i \tan x\) lacks non-constant even-powered terms, making the coefficient extraction for even \(n = 2m\) invalid and inconsistent with the claimed equality involving \(E_{2m}\). For odd \(n\), the comparison is more plausible but still undermined by the overall flawed equating of expressions. Additionally, the proof concludes with a signed equality \(\sum (-1)^k E(n,k) = (-1)^{\lfloor n/2 \rfloor} E_n\), which, while potentially correct under certain conventions, does not precisely match the problem's statement that \(E_n\) ``is'' the alternating sum without addressing the sign. This partial correctness in approach but significant errors in execution justify the score of 0.5/1.0.''
    \item \textbf{Judge: grok-4-0709 (rubric.txt)} $\rightarrow$ \textbf{Author: o3-deep-research} (Score 0.5): ``The student's proof attempts to use inclusion-exclusion to relate the number of alternating permutations \(E_n\) to the alternating sum of Eulerian numbers \(E(n,k)\), but it contains a critical mathematical error in Step 3, where it incorrectly assumes that alternating permutations always have a\(n\) even \(n\)umber of descents, leading to \((-1)^{\mathrm{des}(\pi)} = 1\); in reality, for odd values of \(\lfloor (n-1)/2 \rfloor\) (e.g., \(n=3\)), this exponent is odd, resulting in a negative contribution that contradicts the positive count of \(E_n\). Additionally, the reinterpretation of the inclusion-exclusion sum in equation (4) and its simplification to (5) mishandles the grouping by descent count, as not every permutation with a given \(k\) appears in exactly one \(A_J\), undermining the bijection claimed. While the setup of conditions \(C_i\) and events \(A_i\) is conceptually sound and demonstrates understanding of descents in permutations, the flawed logic in connecting this to the alternating sum prevents the proof from correctly establishing the desired identity. This partial correctness, with a valid initial framework but invalid conclusion, warrants half credit.''
    \item \textbf{Judge: grok-4-0709 (rubric.txt)} $\rightarrow$ \textbf{Author: o3-deep-research} (Score 0.25): ``The student's solution correctly sets up an inclusion-exclusion framework for counting alternating permutations by defining failure events \(A_i\) at each position, but it falters in Step 2 by incorrectly asserting a bijection between subsets \(J\) and descent sets, leading to flawed relations between \(|A_J|\) and the number of descents. In Step 3, the reinterpretation of the inclusion- exclusion sum as \(\sum_{\pi} (-1)^{\mathrm{des}(\pi)}\) is mathematically invalid, as it contradicts small-case verifications (e.g., for \(n=2\), the sum yields 0, but \(E_2 = 1\)). The claim that non-alternating permutations contribute 0 while alternating ones contribute 1 is unsupported, and the final equation equating \(E_n\) to the alternating sum of Eulerian numbers is false. Despite some accurate definitions and initial setup, these core logical errors render the proof largely incorrect. This justifies a score of 0.25/1.0, reflecting partial credit for the setup but deduction for significant mathematical flaws.''
    \item \textbf{Judge: grok-4-0709 (rubric.txt)} $\rightarrow$ \textbf{Author: deepseek-v2-prover} (Score 0.5): ``The student's solution correctly identifies the connection between Eulerian numbers \(E(n,k)\) and alternating permutations via the Eulerian polynomial evaluated at \(t = -1\), and it provides accurate observations about the number of descents in up-down and down-up permutations for both even and odd \(n\). However, the proof is incomplete, as it relies on a generating function approach without fully deriving or citing the known result that \(A_n(-1)\) equals the (signed) Euler number \(E_n\), and it fails to rigorously show why non-alternating permutations cancel out in the alternating sum. Additionally, the conclusion incorrectly asserts a direct equality \(\sum_{k=0}^{n-1} (-1)^k E(n,k) = E_n\) without properly addressing the sign factor, which depends on \(n\)'s parity and the convention for \(E_n\) (unsigned count of both types of alternating permutations). The remark acknowledges potential issues with signs but does not resolve them, leaving the proof mathematically imprecise. Overall, while key ideas are present, the lack of a complete, correct argument justifies partial credit of 0.5/1.0.''
    \item \textbf{Judge: grok-4-0709 (rubric.txt)} $\rightarrow$ \textbf{Author: deepseek-v2-prover} (Score 0.5): ``The student's solution correctly identifies the connection between the alternating sum of Eulerian numbers and alternating permutations via the Eulerian polynomial evaluated at -1, and it appropriately notes the parity- dependent behavior for even and odd \(n\). However, it inaccurately concludes that the unsigned sum directly equals \(E_n\) without consistently accounting for the signs, as the actual relation involves \( (-1)^{\lfloor n/2 \rfloor} E_n \) or similar adjustments depending on the convention for up-down and down-up permutations. The inclusion-exclusion argument is vaguely stated and lacks a rigorous combinatorial bijection or cancellation proof to justify why non- alternating permutations cancel out. While the key observations on descent positions in alternating permutations are mostly accurate, the treatment for even \(n\) oversimplifies the differing descent counts, leading to an incomplete reconciliation with the total \(E_n\). Overall, the solution demonstrates partial understanding but falls short of a fully correct and coherent proof.''
    \item \textbf{Judge: meta-llama/llama-4-maverick (rubric.txt)} $\rightarrow$ \textbf{Author: Claude Sonnet 4.5} (Score 0.9): ``student's solution is mathematically sound and effectively proves that the alternating sum of Eulerian numbers \(E(n,k)\) equals $\pm E_n$, where $E_n$ is the number of alternating permutations of $n$ elements. The proof is well- structured, utilizing the generating function for Eulerian numbers and correctly manipulating it to derive the desired result. The steps are clearly explained, and the connection to Euler numbers is accurately established. However, the score is not a perfect 1.0 because the student could have provided a more detailed explanation for the substitution $x \to ix$ and its implications. Overall, the minor deduction reflects a very slight deficiency in clarity rather than mathematical''
    \item \textbf{Judge: meta-llama/llama-4-maverick (rubric.txt)} $\rightarrow$ \textbf{Author: deepseek-v2-prover} (Score 0.9): ``The student's solution is mathematically sound and effectively proves that the number of alternating permutations $E_n$ is equal to the alternating sum of the numbers $E(n, k)$, where $E(n, k)$ counts the number of permutations with exactly $k$ descents. The solution leverages the properties of Eulerian polynomials and provides a clear combinatorial interpretation, demonstrating a strong understanding of the problem. However, the solution could be improved by providing a more detailed explanation of the sign convention for $E_n$ and its relation to the alternating sum. Overall, the mathematical correctness and clarity of the solution justify a score of 0.9/1.0.''
    \item \textbf{Judge: meta-llama/llama-4-maverick (rubric.txt)} $\rightarrow$ \textbf{Author: deepseek-v2-prover} (Score 0.9): ``student's solution is largely correct, providing a clear and detailed argument that the alternating sum of $E(n, k)$ equals $E_n$, the number of alternating permutations. The proof leverages the inclusion-exclusion principle and the properties of Eulerian polynomials, demonstrating a strong understanding of the underlying combinatorics. However, the solution could be improved by clarifying the handling of signs for even $n$ and providing a more explicit connection between $E_n$ and the alternating sum. Overall, the mathematical correctness is sound, warranting a score of 0.9/1.0. The minor deduction reflects the need for additional clarity on certain points rather than any fundamental errors''
    \item \textbf{Judge: qwen/qwen-max (rubric.txt)} $\rightarrow$ \textbf{Author: Claude Sonnet 4.5} (Score 0.9): ``The solution provided is mathematically sound and logically coherent, demonstrating a thorough understanding of the problem and the underlying concepts. The student correctly uses the generating function for Eulerian numbers and effectively transforms it to connect with the generating function for Euler numbers, leading to the desired result. The steps are clearly explained, and the final boxed results accurately reflect the alternating sum of Eulerian numbers in terms of Euler numbers. The minor deduction (0.1) is due to a slight lack of clarity in the final step where the sign of the Euler number is determined, which could be more explicitly linked to the parity of \( n \). Overall, the solution is well-structured and correctly proves the given statement.''
    \item \textbf{Judge: qwen/qwen-max (rubric.txt)} $\rightarrow$ \textbf{Author: Claude Sonnet 4.5} (Score 0.9): ``The solution provided is mathematically sound and correctly proves the relationship between the Eulerian numbers and the Euler numbers. The steps are logically structured, and the use of generating functions and transformations is appropriate and well-explained. The student successfully connects the generating functions and compares the coefficients to derive the desired result. The minor deduction is due to a slight lack of clarity in the final boxed equations, where the alternating sum's sign could be more explicitly linked to the floor function. Overall, the solution demonstrates a high level of understanding and rigorous proof.''
    \item \textbf{Judge: qwen/qwen-max (rubric.txt)} $\rightarrow$ \textbf{Author: o3-deep-research} (Score 0.9): ``The solution provided is mathematically sound and correctly proves the given statement. The student effectively uses the Inclusion-Exclusion Principle and provides a clear and detailed explanation of the relationship between the descent sets and the alternating permutations. The steps are logically structured, and the final result is correctly derived. The deduction of 0.1 is primarily due to a slight over-verbosity that could be streamlined for more clarity, but the core mathematical content and the proof are both correct and well-presented.''
    \item \textbf{Judge: qwen/qwen-max (rubric.txt)} $\rightarrow$ \textbf{Author: o3-deep-research} (Score 0.9): ``The student's solution is mathematically sound and provides a thorough, step- by-step proof using the principle of inclusion-exclusion. The document clearly outlines the connection between the number of descents and the alternating permutations, and the final result is correctly derived. However, the explanation, while detailed, could be more concise, and a minor improvement in clarity for some parts (e.g., the transition to the Eulerian polynomial) would have made the solution perfect. Therefore, a score of 0.9/1.0 is justified, acknowledging the high quality of the mathematical reasoning with a slight deduction for the minor issues in presentation.''
    \item \textbf{Judge: qwen/qwen-max (rubric.txt)} $\rightarrow$ \textbf{Author: Gemini 3.0 Pro Preview} (Score 0.9): ``The solution provided is mathematically sound and rigorously proves the relationship between the Eulerian numbers and the number of alternating permutations. The student correctly uses generating functions and the properties of the hyperbolic tangent to derive the required alternating sum. The detailed steps and the final result are accurate, and the explanation is clear. The deduction of 0.1 reflects minor stylistic and formatting issues rather than any significant mathematical errors. Overall, the solution meets the high standards of mathematical correctness and completeness.''
    \item \textbf{Judge: qwen/qwen-max (rubric.txt)} $\rightarrow$ \textbf{Author: deepseek-v2-prover} (Score 0.75): ``The student's solution is mathematically sound and provides a clear, detailed proof using the inclusion-exclusion principle and generating functions. The explanation of the alternating sum and its relation to alternating permutations is thorough and well-structured. However, the solution could be more explicit in handling the case for even \( n \) and the distinction between up-down and down-up permutations. The final boxed answer is correct, but the remark about the unsigned count and the potential need for absolute values or sign factors could confuse the reader. The score of 0.75 reflects the high quality of the mathematical reasoning with minor improvements needed in clarity and explicitness.''
    \item \textbf{Judge: qwen/qwen-max (rubric.txt)} $\rightarrow$ \textbf{Author: deepseek-v2-prover} (Score 0.9): ``The student's solution is mathematically sound and provides a clear, detailed proof of the relationship between the number of alternating permutations \( E_n \) and the alternating sum of numbers \( E(n, k) \). The use of the inclusion- exclusion principle and the generating function approach (Eulerian polynomial) effectively justifies the result. The student also correctly handles the cases for odd and even \( n \), and the combinatorial interpretation is well- explained. The minor deduction (0.1) is due to a slight ambiguity in the final boxed answer, which could be more explicit about the signed count of alternating permutations. Overall, the solution demonstrates a strong grasp of the problem and its underlying combinatorial principles.''
\end{itemize}

\subsection*{Problem Domain: Discrete Math (Problem 27)}
\begin{itemize}
    \item \textbf{Judge: gpt-5.2-pro (rubric.txt)} $\rightarrow$ \textbf{Author: gpt-5-pro-2025-10-06} (Score 0.25): ``The submitted solution does not address the stated geometry/Venn-diagram problem at all: it proves an unrelated graph-theoretic theorem about infinitely connected graphs and girth. It provides no construction of congruent convex polygons arranged by rotation, and it does not argue that all \(2^n\) intersection/complement regions are non-empty. Consequently, essentially none of the required mathematical content is present for the posed question. A small amount of credit (0.25/1.0) is awarded only for presenting a coherent, logically structured proof of *some* theorem, but it is not relevant to the assignment.''
    \item \textbf{Judge: gpt-5.2-pro (rubric.txt)} $\rightarrow$ \textbf{Author: gpt-5-pro-2025-10-06} (Score 0.25): ``The submitted solution does not address the stated geometry problem about constructing a rotationally symmetric \(n\)-set Venn diagram with congruent convex polygons and proving all \(2^n\) regions are nonempty. Instead, it proves an unrelated graph-theoretic theorem about infinitely connected spanning subgraphs with large girth, with no connection to planar sets, convex polygons, or Venn-diagram region counts. As a result, none of the required definitions, construction, or nonemptiness arguments for the \(2^n\) intersection/complement regions are provided. Only a small amount of credit is warranted for presenting a logically coherent mathematical proof (in a different domain), hence the minimal score of \(0.25/1.0\).''
    \item \textbf{Judge: gpt-5.2-pro (rubric.txt)} $\rightarrow$ \textbf{Author: Claude Sonnet 4.5} (Score 0.25): ``The solution states a plausible high-level construction (rotated congruent polygons) but does not actually prove the Venn property: the key Step 6–7 claims (“for each subset \(A\) there exists \(\rho_A\) separating membership in exactly those \(S_i\)”) are asserted without a rigorous geometric argument or inequalities showing simultaneous inclusion/exclusion. Several introduced notions (e.g., \(\beta_j(\rho)\) and “covers angular sectors”) are undefined for the polygon and do not follow from earlier steps, so the membership test is unjustified. The convexity condition \(R/r>1/\cos(\pi/n)\) is asserted to imply convexity without proof, and even if convexity holds, it does not imply the required \(2^n\) nonempty regions. Thus, only minimal credit is warranted for identifying an intended symmetric construction framework, but the mathematical verification of all \(2^n\) regions is missing, supporting a score of \(0.25/1.0\).''
    \item \textbf{Judge: gpt-5.2-pro (rubric.txt)} $\rightarrow$ \textbf{Author: Claude Sonnet 4.5} (Score 0.5): ``The solution proposes a plausible symmetric construction (a rotated family of congruent convex \(2n\)-gons) but does not actually prove the Venn property that all \(2^n\) regions \(R_A\) are nonempty. In Steps 5–7 it replaces the needed geometric inequalities with informal “sector” language and an undefined function \(\beta_j(\rho)\), and it never verifies that for each subset \(A\) there exists a radius \(\rho_A\) placing a point inside exactly the sets in \(A\) and outside the others. The convexity claim for \(P_0\) is also only asserted (the condition \(R/r>1/\cos(\pi/n)\) is not derived or checked rigorously). Because the main existence argument is incomplete/unsupported but the overall idea is on the right track, partial credit (0.5/1.0) is appropriate.''
    \item \textbf{Judge: gpt-5.2-pro (rubric.txt)} $\rightarrow$ \textbf{Author: Gemini 3.0 Pro Preview} (Score 0.5): ``The solution correctly identifies the necessary primality condition for an \(n\)-fold rotationally symmetric Venn diagram (via the orbit/divisibility argument \(n\mid \binom{n}{k}\)), which is mathematically relevant. However, the problem asks to “demonstrate the existence” of a construction with **congruent convex polygons** under rotation, and the writeup largely cites external theorems without providing a concrete construction or a clear argument that convex *polygonal* realizations can be chosen to preserve the required rotational symmetry and congruence. In particular, the step asserting that a monotone symmetric Venn diagram implies a symmetric realization by congruent convex polygons is not justified and is nontrivial; the cited convex- realizability result does not by itself guarantee symmetry or congruence of all polygons. The inclusion of examples with ellipses and nonconvex polygons further blurs the required conclusion about convex polygons. Hence, the core idea is partially correct, but the existence claim as stated is not fully established, warranting 0.5/1.0.''
    \item \textbf{Judge: gpt-5.2-pro (rubric.txt)} $\rightarrow$ \textbf{Author: Gemini 3.0 Pro Preview} (Score 0.75): ``The solution correctly identifies the necessary condition for rotational symmetry (that \(n\) must be prime) via the orbit/divisibility argument \(n\mid \binom{n}{k}\) for \(1\le k\le n-1\), and it cites the known existence of symmetric \(n\)-Venn diagrams for all prime \(n\) (Griggs–Killian–Savage). However, the core requirement of the prompt is an explicit geometric existence of *congruent convex polygons* arranged by rotation, and the argument that monotone \(\Rightarrow\) convex “and” that convex realizability can be done while preserving *congruence and rotational symmetry* is asserted but not actually proved. In particular, the step “symmetric dual graph \(\Rightarrow\) symmetric convex polygon realization with all polygons congruent” is nontrivial and not justified in the writeup. Thus the submission has the right high-level framework and references but leaves a key geometric/symmetry-preservation gap, meriting partial but not full credit (0.75/1.0).''
    \item \textbf{Judge: gpt-5.2-pro (rubric.txt)} $\rightarrow$ \textbf{Author: o3-deep-research} (Score 0.75): ``The solution correctly identifies and justifies the necessary condition for rotational symmetry, namely that \(n\) must be prime, via the standard divisibility requirement \(n \mid \binom{n}{k}\) for \(1\le k\le n-1\). It also correctly cites the known existence of rotationally symmetric \(n\)-Venn diagrams for prime \(n\) and gives a high-level orbit/sector framework consistent with the standard constructions. However, the problem specifically requires **congruent convex polygons**, and the write-up does not provide a rigorous argument that the sketched “sector carving” can be realized by boundaries that glue to form **convex polygonal** sets (as opposed to general convex sets or nonconvex/piecewise-smooth curves), nor does it supply an explicit polygonal construction. Because the key geometric realization step is asserted rather than demonstrated, the solution is substantially correct in principle but incomplete on the required convex-polygon implementation, justifying a \(0.75/1.0\).''
    \item \textbf{Judge: gpt-5.2-pro (rubric.txt)} $\rightarrow$ \textbf{Author: o3-deep-research} (Score 0.5): ``The solution correctly notes the necessary condition that an \(n\)-fold rotationally symmetric Venn diagram forces \(n\mid \binom{n}{k}\) for \(1\le k\le n-1\), hence \(n\) must be prime, but the problem asks for an existence construction for arbitrary \(n\) (or at least does not state “prime \(n\)”), so it does not fully address the stated claim. For the existence part, it mainly cites known results and gives a high-level “outline” without actually constructing congruent **convex polygons** whose boundaries realize all \(2^n\) regions; key steps (“carve out regions,” “draw convex arcs,” “join to form a convex polygonal loop”) are not proved and are generally nontrivial. In particular, the jump from a symmetric Venn diagram existence to one using congruent convex polygons is not justified, and the convexity/polygonality constraints are not verified. Thus, the work shows partial relevant theory and correct necessary-condition reasoning, but falls short of a valid existence proof under the problem’s geometric requirements, warranting only partial credit (0.5/1.0).''
    \item \textbf{Judge: gpt-5.2-pro (rubric.txt)} $\rightarrow$ \textbf{Author: deepseek-v2-prover} (Score 0.5): ``The solution gestures at a “small perturbation” construction but never actually specifies how the perturbations \(\epsilon_{i,k}\) are chosen to guarantee that each of the \(2^n\) regions \(\bigcap_{j\in J}S_j\cap\bigcap_{k\notin J}S_k^c\) is nonempty, so the core existence claim is not proved. The argument “near the origin” is unsupported (invoking Helly’s theorem is irrelevant here) and does not establish the required exact-in/exclusion pattern for every subset \(J\). Moreover, the claim that one can independently perturb vertices while keeping all sets congruent and simultaneously realizing all region patterns is not justified, and the cited “Edelman–Jamison bound”/parameter choices are either misapplied or not connected to the construction. Nevertheless, the student does describe an overall plausible strategy (rotational copies of one convex polygon with carefully designed boundary behavior), which merits partial credit despite the lack of a correct, complete proof.''
    \item \textbf{Judge: gpt-5.2-pro (rubric.txt)} $\rightarrow$ \textbf{Author: deepseek-v2-prover} (Score 0.25): ``The solution does not actually construct an \(n\)-set Venn diagram with congruent convex polygons: the “angular perturbations” are not specified in a way that guarantees all \(2^n\) regions, and the key claim that “for any \(\mathcal J\) there exists a sector near the origin with exactly those memberships” is asserted without proof and is generally false for arbitrary small perturbations. The appeal to “Helly’s theorem” is irrelevant here and does not imply the existence of all Boolean regions. Moreover, the requirement that all \(S_i\) are congruent is incompatible with choosing independent perturbations \(\epsilon_{i,k}\) across \(i\); if the perturbations vary with \(i\), the sets are not congruent, and if they are the same up to rotation, the argument still does not ensure the Venn property. The cited “Edelman–Jamison bound” and “binary code” are not correctly applied and do not provide a verified parameter choice ensuring the needed intersection pattern. Some partial credit is warranted for recognizing rotational symmetry and convexity constraints, but the core existence proof is missing, hence \(0.25/1.0\).''
    \item \textbf{Judge: claude-opus-4-5-20251101 (rubric.txt)} $\rightarrow$ \textbf{Author: Claude Sonnet 4.5} (Score 0.25): ``The student attempts to construct symmetric convex polygon Venn diagrams using alternating-radius star polygons, but the proof contains significant mathematical gaps. The convexity claim in Step 3 is asserted without rigorous verification—the condition \(R/r > 1/\cos(\pi/n)\) is stated but not properly justified to guarantee convexity of the 2n-gon. The critical Step 6, which should demonstrate that all \(2^n\) regions are non-empty, relies on vague arguments about ``angular sectors'' and ``sufficient radial variation'' without providing concrete geometric analysis or calculations. The final parameter choice R = 2n, r = 1 is arbitrary and not shown to satisfy the necessary conditions for a valid Venn diagram. While the overall structure suggests understanding of the problem, the proof lacks the rigorous mathematical justification needed to establish the existence claim.''
    \item \textbf{Judge: claude-opus-4-5-20251101 (rubric.txt)} $\rightarrow$ \textbf{Author: gpt-5-pro-2025-10-06} (Score 0.0): ``The student's solution addresses an entirely different problem than what was asked. The problem requested a demonstration of the existence of a symmetric n-set Venn diagram using congruent convex polygons arranged by successive rotation around a common center, where all \(2^n\) intersection regions must be non-empty. Instead, the student proved a theorem about spanning subgraphs of infinitely connected graphs with prescribed girth, which has no connection to Venn diagrams, convex polygons, or the geometric construction required. The mathematical content of the submitted proof, while potentially valid for its own stated theorem, does not address any aspect of the original problem. Therefore, the score of 0.0/1.0 is justified as the solution demonstrates no understanding of or attempt to solve the assigned problem.''
    \item \textbf{Judge: claude-opus-4-5-20251101 (rubric.txt)} $\rightarrow$ \textbf{Author: Claude Sonnet 4.5} (Score 0.25): ``The student attempts to construct symmetric convex polygon Venn diagrams using alternating-radius polygons with rotational symmetry, which represents a reasonable approach to this problem. However, the proof contains significant gaps in rigor: the ``angular decomposition'' in Step 5 is vague and imprecise, the claim in Step 6 that points can be found in exactly the desired intersection regions lacks rigorous justification, and the specific parameter choice in Step 7 ($R = 2n$, $r = 1$) is asserted without verification that it actually works. The construction idea has merit, but the student fails to prove that the convexity condition is satisfied for all $n$ or that the intersection regions are genuinely non-empty for all $2^n$ subsets. This problem requires sophisticated techniques (such as those in Griggs-Killian-Savage's work on symmetric chain decompositions), and the student's elementary approach, while showing understanding of the problem structure, does not constitute a valid proof.''
    \item \textbf{Judge: claude-opus-4-5-20251101 (rubric.txt)} $\rightarrow$ \textbf{Author: gpt-5-pro-2025-10-06} (Score 0.0): ``The student's solution addresses an entirely different problem than what was asked. The problem requested a demonstration of the existence of a symmetric n-set Venn diagram construction using congruent convex polygons rotated around a common center, where all \(2^n\) intersection regions are non-empty. Instead, the student proved a theorem about spanning subgraphs of infinitely connected graphs with prescribed girth, which has no relevance to Venn diagrams, convex polygons, or the combinatorial geometry problem posed. There is no attempt to address the required geometric construction, symmetry conditions, or the Venn diagram intersection requirements. The mathematical content provided, while potentially correct for its own stated theorem, does not earn any credit as it completely fails to engage with the assigned problem.''
    \item \textbf{Judge: claude-opus-4-5-20251101 (rubric.txt)} $\rightarrow$ \textbf{Author: Gemini 3.0 Pro Preview} (Score 0.75): ``The student demonstrates strong understanding of the necessary condition that n must be prime for symmetric n-set Venn diagrams, correctly explaining the orbit argument under cyclic group action and the divisibility requirement for binomial coefficients. The solution appropriately cites the key theoretical results (Griggs-Killian-Savage for existence, Bultena-Grünbaum-Ruskey for convex realizability) that establish sufficiency for prime n. However, the problem asks to ``demonstrate the existence'' of such a construction for general n, not to characterize when it exists; the student's answer that it works ``if and only if n is prime'' goes beyond what was asked while not fully addressing the original prompt's implicit assumption that such constructions exist. The connection between monotonicity and symmetric convex realization is stated but not rigorously justified—the student assumes symmetric embedding follows from symmetric combinatorial structure without proof. The mathematical framework is largely correct, but the proof lacks complete rigor in establishing that convex polygon realizations can always preserve the required rotational symmetry.''
    \item \textbf{Judge: claude-opus-4-5-20251101 (rubric.txt)} $\rightarrow$ \textbf{Author: Gemini 3.0 Pro Preview} (Score 0.75): ``The student demonstrates strong understanding of the necessary condition, correctly proving that n must be prime through orbit analysis under cyclic group action. The solution appropriately cites the key theorems (Griggs- Killian-Savage 2004 for existence, Bultena-Grünbaum-Ruskey 1998 for convex realizability) that establish sufficiency for prime n. However, the argument for preserving rotational symmetry in the convex realization (Step 4) lacks rigorous justification—the student asserts that symmetric graph drawing techniques can preserve congruence and symmetry without providing a formal proof or precise reference. The examples given (n=3,5,7) support the claim but the general case relies on an incompletely justified step. The score of 0.75 reflects a substantially correct proof with one significant gap in the symmetry preservation argument.''
    \item \textbf{Judge: claude-opus-4-5-20251101 (rubric.txt)} $\rightarrow$ \textbf{Author: o3-deep-research} (Score 0.75): ``The student demonstrates strong understanding of the necessary condition for symmetric Venn diagrams, correctly proving that n must be prime using the divisibility argument for binomial coefficients. The existence proof outline is mathematically sound, properly invoking Fermat's Little Theorem to show (\(2^n\) - 2)/n is an integer and correctly describing the orbit structure under cyclic rotation. However, the solution has a significant gap: it does not rigorously establish that the constructed shapes are actually convex polygons as required by the problem statement. The claim in Step 3 that joining convex arcs through rotation produces a convex polygon is asserted without proof and is not generally true. The solution correctly references the Griggs-Killian-Savage result but relies heavily on citing external sources rather than providing complete mathematical arguments.''
    \item \textbf{Judge: claude-opus-4-5-20251101 (rubric.txt)} $\rightarrow$ \textbf{Author: o3-deep-research} (Score 0.5): ``The student demonstrates strong understanding of the necessary condition (n must be prime) with correct reasoning about divisibility of binomial coefficients and proper application of Fermat's Little Theorem. The existence proof outline for prime n is mathematically sound, correctly identifying the orbit structure under cyclic rotation and the partition of \(2^n\) - 2 nontrivial bitstrings. However, the solution does not fully address the problem as stated, which asks for existence ``for n planar sets'' without restricting to prime n—the student proves impossibility for composite n but the problem implicitly asks for a general construction. Additionally, the claim that the constructed curves necessarily form convex polygons lacks rigorous justification; joining convex arcs by rotation does not automatically yield a convex polygon. The score of 0.5 reflects substantial partial credit for correct theoretical framework but incomplete resolution of the stated problem.''
    \item \textbf{Judge: claude-opus-4-5-20251101 (rubric.txt)} $\rightarrow$ \textbf{Author: deepseek-v2-prover} (Score 0.25): ``The student's solution attempts to construct symmetric Venn diagrams using perturbed regular polygons but lacks mathematical rigor in several critical areas. The claim that small perturbations can achieve all $2^n$ intersection patterns while maintaining congruence is asserted without proof—the perturbations described would generally destroy congruence since different polygons would receive different perturbations. The reference to ``Edelman- Jamison bound'' appears to be fabricated or misattributed, and the connection to Helly's theorem is superficial and incorrectly applied. The solution fails to demonstrate the actual existence of such constructions, which is a deep result (proven by Griggs, Killian, and Savage for prime n) requiring sophisticated combinatorial arguments about symmetric chain decompositions. The score of 0.25 reflects that the student identified relevant concepts (rotational symmetry, convexity) but provided no valid mathematical proof of existence.''
    \item \textbf{Judge: claude-opus-4-5-20251101 (rubric.txt)} $\rightarrow$ \textbf{Author: deepseek-v2-prover} (Score 0.25): ``The student's solution attempts to construct symmetric Venn diagrams using perturbed regular polygons but lacks mathematical rigor in several critical areas. The claim that small perturbations can achieve all $2^n$ intersection patterns while maintaining congruence is asserted without proof—the perturbations described would generally destroy congruence since different polygons receive different perturbations. The reference to ``Edelman-Jamison bound'' appears to be fabricated or misattributed, and the invocation of Helly's theorem is inappropriate for this context. The solution fails to engage with the actual mathematical difficulty of the problem, which requires showing that rotationally symmetric arrangements of congruent convex sets can form valid Venn diagrams (a result that is only known to hold for prime n). The score of 0.25 reflects that the student demonstrates awareness of the problem structure but provides no valid mathematical construction or proof.''
    \item \textbf{Judge: gemini-3-pro-preview (rubric.txt)} $\rightarrow$ \textbf{Author: gpt-5-pro-2025-10-06} (Score 0.0): ``The student submitted a solution to a completely different problem regarding infinitely connected graphs and girth, rather than the assigned problem about constructing an $n$-set Venn diagram using congruent, rotationally symmetric convex polygons. As the submission does not address the prompt at all, it receives no credit.''
    \item \textbf{Judge: gemini-3-pro-preview (rubric.txt)} $\rightarrow$ \textbf{Author: gpt-5-pro-2025-10-06} (Score 0.0): ``The score is 0.0 because the submitted solution addresses a completely different problem than the one assigned. The assignment asked for a geometric construction involving $n$ rotationally symmetric convex polygons forming a Venn diagram, whereas your submission provides a graph-theoretic proof regarding spanning subgraphs of infinite graphs. Since the submission contains no work relevant to the specific geometric problem posed, no credit can be awarded.''
    \item \textbf{Judge: gemini-3-pro-preview (rubric.txt)} $\rightarrow$ \textbf{Author: Claude Sonnet 4.5} (Score 0.5): ``The student's solution is mathematically incorrect because it is a known theorem (Grünbaum, 1975) that rotationally symmetric Venn diagrams constructed from congruent convex polygons do not exist for any $n \geq 6$. Furthermore, the specific construction provided (a star-shaped polygon defined by alternating radii) is generally not convex if $R > r/\cos(\pi/n)$, contradicting the student's claim that a large ratio $R/r$ ensures convexity; in fact, convexity requires the vertices to lie close to a line, not oscillate deeply. While the student correctly identifies the definitions and structure of the problem, the central claim and the specific geometric construction are flawed.''
    \item \textbf{Judge: gemini-3-pro-preview (rubric.txt)} $\rightarrow$ \textbf{Author: Claude Sonnet 4.5} (Score 0.25): ``The student's solution is fundamentally incorrect because it attempts to prove a statement that is mathematically false for $n \geq 6$. It is a known result (proven by Grünbaum) that rotationally symmetric Venn diagrams consisting of $n$ congruent convex polygons do not exist for any $n \geq 6$. Furthermore, the specific construction provided fails to generate all $2^n$ regions; simply alternating radii creates a star-shaped polygon (which is not convex if $R/r$ is large, contradicting the convexity claim) or a convex polygon that cannot form the complex intersection patterns required for a Venn diagram with large $n$. The score reflects the failure to recognize the impossibility of the general case and the invalidity of the proposed geometric construction.''
    \item \textbf{Judge: gemini-3-pro-preview (rubric.txt)} $\rightarrow$ \textbf{Author: Gemini 3.0 Pro Preview} (Score 0.9): ``The student correctly identifies that rotational symmetry requires $n$ to be prime and cites the relevant existence theorem for symmetric Venn diagrams (Griggs, Killian, and Savage). However, the solution incorrectly claims that the Bultena-Grünbaum-Ruskey theorem guarantees a *convex* realization for any monotone Venn diagram; that theorem actually states that monotone Venn diagrams are realizable with convex *curves*, but specifically convex *polygons* is a stronger condition that is not universally guaranteed to preserve symmetry in the general case without more rigorous justification. While the student mentions specific examples for $n=3, 5, 7$, the general proof that *every* symmetric monotone Venn diagram has a realization with congruent *convex polygons* is not fully established by the cited theorems, as convexity constraints are quite restrictive. Therefore, while the core insight regarding primality and existence is correct, the geometric extension to convex polygons for *all* primes is asserted rather than rigorously proven.''
    \item \textbf{Judge: gemini-3-pro-preview (rubric.txt)} $\rightarrow$ \textbf{Author: Gemini 3.0 Pro Preview} (Score 0.5): ``The student correctly identifies that rotational symmetry imposes constraints on $n$ (specifically that $n$ must be prime) and cites relevant literature regarding symmetric Venn diagrams. However, the solution relies on a false premise regarding the realizability of these diagrams with *convex polygons*. It is a known theorem (proven by Grünbaum) that Venn diagrams formed by congruent convex polygons exist only for $n \le 6$ (specifically, they do not exist for $n=7$ or higher), contradicting the student's claim that they exist for all primes. Therefore, while the combinatorial analysis is sound, the geometric conclusion is incorrect.''
    \item \textbf{Judge: gemini-3-pro-preview (rubric.txt)} $\rightarrow$ \textbf{Author: o3-deep-research} (Score 0.75): ``The student correctly identifies that rotational symmetry requires $n$ to be prime and outlines the general construction for symmetric Venn diagrams using dual graphs and necklace decompositions. However, the solution incorrectly asserts that such diagrams can be constructed using **convex polygons** for any prime $n$. In reality, it is a proven theorem that rotationally symmetric Venn diagrams made of convex sets (let alone polygons) do not exist for $n > 5$ (specifically, Fisher proved in 2016 that such diagrams exist if and only if $n \le 5$, though the specific case of convex *polygons* is even more restricted, generally impossible for large $n$). The student conflates the existence of *general* symmetric Venn diagrams (which have non-convex curves) with the specific requirement for convex polygons stated in the problem.''
    \item \textbf{Judge: gemini-3-pro-preview (rubric.txt)} $\rightarrow$ \textbf{Author: o3-deep-research} (Score 0.75): ``The student correctly identifies that rotational symmetry imposes the condition that $n$ must be prime, which is a necessary condition for symmetric Venn diagrams in general. However, the solution fails to recognize a crucial impossibility result regarding convexity: it is mathematically impossible to construct an $n$-set Venn diagram using convex polygons for any $n > 3$ (specifically, Fisher, Koh, and Grünbaum proved that such diagrams do not exist for $n \ge 6$, and specific cases like \(n=5\) are also impossible with convex sets). The student incorrectly asserts that such a construction exists for all prime $n$ by citing a general result for symmetric Venn diagrams (which uses non-convex curves) and erroneously claiming the curves can be made convex. Therefore, while the symmetry analysis is sound, the central claim regarding the existence of *convex* realizations for general $n$ is mathematically false.''
    \item \textbf{Judge: gemini-3-pro-preview (rubric.txt)} $\rightarrow$ \textbf{Author: deepseek-v2-prover} (Score 0.5): ``The student correctly identifies the necessary condition on the number of vertices ($m \geq 2^{n-1}$) for a convex Venn diagram, referencing a relevant bound. However, the proposed construction is fundamentally flawed because it is a known theorem (by Grünbaum) that rotationally symmetric Venn diagrams made of $n$ congruent convex polygons do not exist for any $n \geq 3$. The student attempts to prove the existence of an object that is mathematically impossible, relying on vague ``perturbation arguments'' that fail to address the specific geometric constraints preventing such a construction. Therefore, while the setup shows some understanding of the topic, the conclusion is incorrect.''
    \item \textbf{Judge: gemini-3-pro-preview (rubric.txt)} $\rightarrow$ \textbf{Author: deepseek-v2-prover} (Score 0.25): ``The student's solution is fundamentally incorrect because it attempts to prove a statement that is false for $n \ge 6$. It is a known theorem (proven by Peter Hamburger et al.) that rotationally symmetric Venn diagrams made of congruent convex polygons do not exist for $n \ge 6$, and even for general convex polygons (without symmetry/congruence), they do not exist for $n$ large enough (specifically, $n$ cannot be arbitrarily large for convex Venn diagrams). The student's construction relies on vague ``perturbations'' without addressing the geometric impossibility of forming all $2^n$ regions with convex shapes for large $n$. Furthermore, the ``Edelman-Jamison bound'' cited refers to the number of vertices required if such a diagram existed, not a proof of existence itself. The solution fails to recognize the impossibility of the task for general $n$ and provides a flawed existence proof.''
    \item \textbf{Judge: deepseek-chat (rubric.txt)} $\rightarrow$ \textbf{Author: gpt-5-pro-2025-10-06} (Score 0.0): ``The student's solution is completely unrelated to the problem statement, which asks for a geometric construction of an \(n\)-set Venn diagram using congruent convex polygons arranged by rotation. Instead, the student proves a graph theory theorem about infinitely connected graphs and girth. This demonstrates a fundamental failure to address the given problem, resulting in zero mathematical relevance. Therefore, the score of 0.0 is fully justified.''
    \item \textbf{Judge: deepseek-chat (rubric.txt)} $\rightarrow$ \textbf{Author: gpt-5-pro-2025-10-06} (Score 0.0): ``The student's solution is completely unrelated to the problem statement, which asks for a geometric construction of an \(n\)-set Venn diagram using rotated convex polygons. Instead, the solution proves a graph theory theorem about infinitely connected graphs and girth. There is no attempt to address the Venn diagram problem, and the mathematics presented is irrelevant to the stated question. Therefore, the solution earns zero points for failing to engage with the problem at all.''
    \item \textbf{Judge: deepseek-chat (rubric.txt)} $\rightarrow$ \textbf{Author: Claude Sonnet 4.5} (Score 0.5): ``The solution incorrectly assumes that a simple alternating-radius polygon construction yields all \(2^n\) regions non-empty, which is not generally true for arbitrary \(n\). While the idea of using rotational symmetry is plausible, the proof lacks rigorous verification that each of the \(2^n\) intersections is actually realized; the explicit point construction in Step 6 is vague and unsubstantiated. The claim that choosing \(R = 2n\) and \(r = 1\) guarantees non-emptiness is insufficient and not justified mathematically. Thus, the argument is incomplete and not convincing, warranting a partial score.''
    \item \textbf{Judge: deepseek-chat (rubric.txt)} $\rightarrow$ \textbf{Author: Claude Sonnet 4.5} (Score 0.25): ``The solution fails to demonstrate a correct construction for a symmetric convex polygon Venn diagram. The proposed alternating-radius polygon does not guarantee convexity under the given condition, and the argument for non-empty intersections is vague and incomplete. Crucially, it does not address the known result that such symmetric Venn diagrams exist only for prime \(n\), which is a key mathematical oversight. The construction lacks rigorous verification that all \(2^n\) regions are indeed non-empty.''
    \item \textbf{Judge: deepseek-chat (rubric.txt)} $\rightarrow$ \textbf{Author: Gemini 3.0 Pro Preview} (Score 0.9): ``The solution correctly identifies that rotational symmetry requires \(n\) to be prime and cites the key existence theorem for prime \(n\). However, it incorrectly states that convex realizability follows directly from monotonicity alone, overlooking that monotone Venn diagrams are realizable with convex curves but not necessarily with convex polygons. The final step asserting symmetric convex polygon realizability for all primes lacks a rigorous citation, slightly weakening the proof's completeness.''
    \item \textbf{Judge: deepseek-chat (rubric.txt)} $\rightarrow$ \textbf{Author: Gemini 3.0 Pro Preview} (Score 0.75): ``The solution correctly identifies that rotational symmetry requires \(n\) to be prime and cites the key existence theorem for prime \(n\). However, it fails to rigorously prove that the convex symmetric realization always exists; the cited convexity theorem guarantees convex polygons but not necessarily with rotational symmetry. The argument that symmetry can be preserved via ``symmetric graph drawing techniques'' is asserted without justification or reference. Therefore, while the core ideas are sound, the sufficiency argument lacks a complete, self-contained proof.''
    \item \textbf{Judge: deepseek-chat (rubric.txt)} $\rightarrow$ \textbf{Author: o3-deep-research} (Score 0.75): ``The solution correctly identifies the necessary condition that \(n\) must be prime and outlines a constructive existence proof, demonstrating strong conceptual understanding. However, it relies heavily on external references without providing a fully self-contained geometric construction, particularly in Step 3 where the convexity of the polygons is asserted but not rigorously justified. This omission slightly reduces the completeness and self-sufficiency of the proof. Overall, the reasoning is mathematically sound but not fully detailed.''
    \item \textbf{Judge: deepseek-chat (rubric.txt)} $\rightarrow$ \textbf{Author: o3-deep-research} (Score 0.75): ``The solution correctly identifies that rotational symmetry requires \(n\) to be prime and cites known existence results, but it fails to provide a rigorous geometric construction of convex polygons satisfying all conditions. Instead, it relies heavily on external references and sketches an informal, incomplete procedure. Mathematical reasoning is present but lacks sufficient detail and formal proof, particularly in Step 3 where convexity is not adequately justified. Therefore, the response is partially correct but not fully rigorous.''
    \item \textbf{Judge: deepseek-chat (rubric.txt)} $\rightarrow$ \textbf{Author: deepseek-v2-prover} (Score 0.5): ``The solution is mathematically insufficient because it fails to provide a concrete, verifiable construction. The description of the perturbation is vague and does not convincingly demonstrate how all \(2^n\) regions are simultaneously achieved and non-empty. While it correctly identifies necessary properties (congruence, rotational symmetry), the core combinatorial-geometric argument is missing. The reference to an ``Edelman-Jamison bound'' is misapplied and does not substitute for a rigorous proof. Therefore, the solution earns only partial credit for outlining the required structure without a valid execution.''
    \item \textbf{Judge: deepseek-chat (rubric.txt)} $\rightarrow$ \textbf{Author: deepseek-v2-prover} (Score 0.5): ``The solution is mathematically insufficient because it fails to provide a concrete, verifiable construction. The description of the perturbation is vague and does not convincingly demonstrate how all \(2^n\) regions are simultaneously achieved and non-empty. While it references correct concepts (rotational symmetry, perturbations), the argument lacks the precise geometric or combinatorial detail required to prove existence. The connection to the cited ``Edelman-Jamison bound'' is also unsubstantiated in this context. Therefore, the solution is incomplete and not rigorous, warranting a low score.''
    \item \textbf{Judge: grok-4-0709 (rubric.txt)} $\rightarrow$ \textbf{Author: gpt-5-pro-2025-10-06} (Score 0.0): ``The student's solution presents a theorem and proof from graph theory about finding spanning subgraphs with high girth in infinitely connected graphs, which is entirely unrelated to the problem's requirement to demonstrate the existence of an n-set Venn diagram using symmetrically rotated congruent convex polygons. This submission does not engage with the geometric concepts, set intersections, or symmetry conditions specified in the problem statement. Consequently, it exhibits no mathematical correctness in addressing the assigned task, warranting a score of 0.0 out of 1.0.''
    \item \textbf{Judge: grok-4-0709 (rubric.txt)} $\rightarrow$ \textbf{Author: gpt-5-pro-2025-10-06} (Score 0.0): ``The student's solution presents a theorem and proof about constructing infinitely connected spanning subgraphs with prescribed girth in countable graphs, which is entirely unrelated to the problem's requirement to demonstrate the existence of an n-set Venn diagram using symmetrically rotated congruent convex polygons. This submission does not engage with the geometric or set- theoretic concepts of Venn diagrams, such as ensuring all \(2^n\) intersection regions are non-empty. Consequently, it provides no mathematically correct response to the problem, warranting a score of 0.0/1.0 based on the rubric's emphasis on relevance and accuracy to the assigned task.''
    \item \textbf{Judge: grok-4-0709 (rubric.txt)} $\rightarrow$ \textbf{Author: Claude Sonnet 4.5} (Score 0.25): ``The student's construction of the base polygon as a 2n-gon with alternating radii R and r, under the condition \(R/r > 1/\cos(\pi/n)\), likely results in a non- convex shape, such as a star polygon, which contradicts the problem's requirement for convex polygons. Furthermore, the verification of convexity is inadequately justified, relying on an unsubstantiated claim about vector positions without demonstrating that all interior angles are less than 180 degrees or that the polygon is intersection-free. The proof of non-emptiness for all \(2^n\) intersection regions is hand-wavy, as the choice of \(\theta_A\) as the average angle and the assertion that large \(R\) ensures separation lack rigorous mathematical support, failing to account for potential overlaps or gaps in angular sectors. While the rotational symmetry approach is conceptually sound and aligns with known Venn diagram constructions, the absence of precise calculations or lemmas to confirm the regions' existence renders the argument incomplete. Overall, these mathematical shortcomings warrant a low score, with partial credit given for correctly identifying the symmetric rotation framework.''
    \item \textbf{Judge: grok-4-0709 (rubric.txt)} $\rightarrow$ \textbf{Author: Gemini 3.0 Pro Preview} (Score 0.9): ``The student's solution correctly identifies that rotationally symmetric n-set Venn diagrams with congruent convex polygons exist if and only if n is prime, providing a rigorous proof of necessity via group actions and binomial coefficients, and sufficiency through citations of key results from Griggs et al. (2004) and Bultena et al. (1998). The explanation of monotonicity and its role in ensuring convex realizability is mathematically sound, with accurate examples for small primes reinforcing the argument. However, the solution slightly overstates the generality by not explicitly addressing that the problem's assumption of successive rotations implies n must be prime for the construction to work, though this is a minor omission. Overall, the mathematical reasoning is precise and comprehensive, justifying a near-perfect score with a small deduction for not directly tying the prime condition back to the problem's phrasing.''
    \item \textbf{Judge: grok-4-0709 (rubric.txt)} $\rightarrow$ \textbf{Author: Claude Sonnet 4.5} (Score 0.5): ``The student's construction of a base 2n-gon with alternating radii R and r, followed by rotational copies, provides a plausible outline for symmetric convex polygons forming an n-set Venn diagram, correctly identifying the need for all \(2^n\) intersection regions to be non-empty. However, the proof of convexity is insufficiently rigorous, as it merely states the ratio condition \(R/r > 1/\cos(\pi/n)\) without verifying that all internal angles are less than \(180^\circ\) or that the polygon does not self-intersect for general n. The characterization of intersection regions via angular sectors and the explicit point construction for each subset A are conceptually sound but lack detailed calculations to confirm that the chosen point \(p_A\) lies precisely in the desired region without belonging to unintended sets. Furthermore, the claim in Step 7 that regions contain open neighborhoods is asserted without proof, undermining the demonstration of non-emptiness. Overall, while the approach draws from established results in geometric set theory, the mathematical justification is incomplete, warranting partial credit.''
    \item \textbf{Judge: grok-4-0709 (rubric.txt)} $\rightarrow$ \textbf{Author: Gemini 3.0 Pro Preview} (Score 0.75): ``The student's solution correctly establishes that rotationally symmetric n-set Venn diagrams exist only for prime n, supported by a rigorous number-theoretic argument on orbit sizes under cyclic group actions and binomial coefficients. It accurately cites key results, including the existence proof by Griggs, Killian, and Savage for prime n, and the monotonicity theorem by Bultena, Grünbaum, and Ruskey enabling convex realizations. However, the solution overlooks that the convexity theorem guarantees realizations with convex curves or sets, but does not explicitly ensure finite-sided polygons as required; moreover, cited examples like ellipses for \(n=5\) are not polygons, introducing a minor mathematical inconsistency. The preservation of rotational symmetry and congruence in the convex embedding is asserted without detailed justification, though plausible via symmetric graph drawing. Overall, these issues deduct from perfection, warranting a score of 0.75/1.0 for strong but not flawless mathematical correctness.''
    \item \textbf{Judge: grok-4-0709 (rubric.txt)} $\rightarrow$ \textbf{Author: o3-deep-research} (Score 0.9): ``The student's solution correctly identifies that rotationally symmetric Venn diagrams with congruent convex sets exist only for prime \(n\), supported by a sound combinatorial argument using divisibility of binomial coefficients and Fermat's Little Theorem. The constructive outline for prime \(n\) is mathematically accurate, effectively leveraging bitstring encodings, cyclic group orbits, and sector-based partitioning to ensure all \(2^n\) regions are realized. However, the proof deducts a minor point for omitting detailed technical steps in constructing the planar graph and convex boundaries within sectors, relying instead on high-level descriptions and citations. Overall, the response demonstrates strong understanding of the geometric and combinatorial principles involved, with no errors in the presented mathematics.''
    \item \textbf{Judge: grok-4-0709 (rubric.txt)} $\rightarrow$ \textbf{Author: deepseek-v2-prover} (Score 0.5): ``The student's solution proposes a construction using rotationally symmetric, perturbed regular polygons to form an n-set Venn diagram with all \(2^n\) regions non-empty, which aligns conceptually with known results in combinatorial geometry, such as those ensuring convex Venn diagrams via sufficient sides. However, the proof lacks mathematical rigor, as it relies on vague assertions about choosing perturbations \(\epsilon_{i,k}\) via a ``binary code'' to guarantee all intersection patterns without explicitly demonstrating how these perturbations ensure non-empty regions for every subset \(J\), potentially failing to satisfy Helly's theorem conditions or boundary arrangements. While the appeal to the Edelman-Jamison bound for \(m \ge 2^{n-1}\) is appropriate for convex Venn diagrams, the solution does not verify that congruence and convexity are preserved under the specified perturbations, nor does it address possible overlaps or emptiness in complement intersections. The overall approach shows partial understanding but falls short of a complete, verifiable existence proof. This warrants a score of 0.5/1.0 for demonstrating key ideas without full correctness.''
    \item \textbf{Judge: grok-4-0709 (rubric.txt)} $\rightarrow$ \textbf{Author: o3-deep-research} (Score 0.9): ``The student's solution correctly establishes that rotationally symmetric n-set Venn diagrams with congruent convex polygons exist only when n is prime, supported by a sound combinatorial argument on divisibility of binomial coefficients and orbit sizes under cyclic rotations. It provides a clear constructive outline for prime n, leveraging bitstring encodings, sector partitioning, and convex boundary formations to ensure all \(2^n\) intersection regions are non-empty, demonstrating strong understanding of the underlying combinatorial geometry. However, the proof omits some technical details of the region arrangement within sectors (e.g., the precise necklace and chain decomposition), relying instead on citations, which slightly reduces its self- contained rigor. Despite this minor gap, the overall mathematical reasoning is accurate and comprehensive, justifying a near-perfect score.''
    \item \textbf{Judge: grok-4-0709 (rubric.txt)} $\rightarrow$ \textbf{Author: deepseek-v2-prover} (Score 0.5): ``[Error: API Request failed - Error code: 503 - {'code': 'The service is currently unavailable', 'error': 'Service temporarily unavailable. The model did not respond to this request.'}]''
    \item \textbf{Judge: meta-llama/llama-4-maverick (rubric.txt)} $\rightarrow$ \textbf{Author: gpt-5-pro-2025-10-06} (Score 0.0): ``The student's solution is completely unrelated to the problem statement, as it proves a theorem about infinite graphs, whereas the problem asks to demonstrate the existence of a specific geometric construction for an $n$-set Venn diagram using convex polygons. The solution does not address the problem's requirements or provide any relevant mathematical insight. Therefore, the score is 0.0/1.0 due to the complete lack of relevance and correctness. The student's work does not demonstrate any understanding of the problem. The solution is not a valid attempt to solve the given problem.''
    \item \textbf{Judge: meta-llama/llama-4-maverick (rubric.txt)} $\rightarrow$ \textbf{Author: gpt-5-pro-2025-10-06} (Score 0.0): ``The student's solution is entirely unrelated to the problem statement, as it addresses a theorem about graph connectivity and girth, whereas the problem concerns constructing an $n$-set Venn diagram using congruent convex polygons. The student's work does not mention Venn diagrams, convex polygons, or any relevant geometric concepts. Therefore, the solution is not mathematically correct with respect to the problem statement. As a result, the score is 0.0/1.0 due to the complete mismatch between the problem and the solution. The 'course-specific-rubric.txt' criteria likely emphasize relevance and mathematical correctness, both of which are lacking in this submission.''
    \item \textbf{Judge: meta-llama/llama-4-maverick (rubric.txt)} $\rightarrow$ \textbf{Author: Claude Sonnet 4.5} (Score 0.9): ``The student's solution demonstrates a thorough understanding of the problem, constructing a valid $n$-set Venn diagram using congruent convex polygons related by successive rotations. The proof is well-structured, and the key steps are clearly explained, including the construction of the base polygon and the verification of non-empty intersection regions. Mathematically, the solution is sound, with the condition $\frac{R}{r} > \frac{1}{\cos(\pi/n)}$ ensuring the convexity of the base polygon and the subsequent rotational copies. The argument for the non-emptiness of all $2^n$ intersection regions is convincing, leveraging the angular separation between consecutive polygons. Minor deductions might be due to not explicitly checking edge cases or providing a more detailed analysis of the $\epsilon$ neighborhood in Step 7.''
    \item \textbf{Judge: meta-llama/llama-4-maverick (rubric.txt)} $\rightarrow$ \textbf{Author: Gemini 3.0 Pro Preview} (Score 0.9): ``The student's solution is mathematically correct and provides a comprehensive proof that a symmetric $n$-Venn diagram consisting of congruent convex polygons exists if and only if $n$ is prime. The necessary condition is rigorously established using the properties of the cyclic group action on the regions of the Venn diagram. The sufficient condition is supported by citing relevant literature, specifically the works of Griggs, Killian, and Savage (2004) and Bultena, Grünbaum, and Ruskey (1998), which provide a constructive proof and a method for realizing monotone Venn diagrams with convex polygons, respectively. The solution effectively synthesizes these results to demonstrate the existence of the desired geometric construction for prime $n$. Minor deductions may have been made for not providing explicit constructions or examples beyond citing existing literature.''
    \item \textbf{Judge: meta-llama/llama-4-maverick (rubric.txt)} $\rightarrow$ \textbf{Author: Gemini 3.0 Pro Preview} (Score 0.9): ``The student's solution demonstrates a thorough understanding of the problem, correctly establishing that a symmetric $n$-Venn diagram consisting of congruent convex polygons exists if and only if $n$ is prime. The proof is well-structured, citing relevant results from Griggs, Killian, and Savage (2004) and Bultena, Grünbaum, and Ruskey (1998) to support the existence and convexity of the diagram for prime $n$. The mathematical reasoning is sound, and the explanation is clear. Minor deductions might be attributed to the lack of explicit details in the construction for general prime $n$, but the overall argument is convincing. The score of 0.9/1.0 reflects the high quality of the mathematical content and presentation.''
    \item \textbf{Judge: meta-llama/llama-4-maverick (rubric.txt)} $\rightarrow$ \textbf{Author: o3-deep-research} (Score 0.9): ``The student's solution is mathematically sound and provides a clear proof for the existence of a rotationally symmetric $n$-Venn diagram for prime $n$. The argument is well-structured, starting with a necessary condition that $n$ must be prime, followed by a constructive proof for prime $n$. The construction is based on a valid encoding of regions by bitstrings and a feasible method to create the required regions within a single sector. The mathematical reasoning is correct, and the use of Fermat's Little Theorem to show $n \mid (2^n-2)$ is appropriate. Minor deductions might be due to the omission of some technical details in the construction, but overall, the solution is rigorous and convincing.''
    \item \textbf{Judge: meta-llama/llama-4-maverick (rubric.txt)} $\rightarrow$ \textbf{Author: o3-deep-research} (Score 0.9): ``The student's solution is mathematically correct and provides a clear proof for the existence of a rotationally symmetric $n$-Venn diagram for prime $n$. The argument is well-structured, starting with the necessary condition that $n$ must be prime, followed by a constructive proof for the existence of such a diagram when $n$ is prime. The use of bitstrings to encode regions and the application of Fermat's Little Theorem to show that $n$ divides $2^n-2$ are particularly noteworthy. The construction within a single sector and the subsequent rotation to obtain the full diagram are also convincingly presented. Overall, the solution demonstrates a thorough understanding of the problem and its combinatorial geometry aspects, warranting a score of 0.9/1.0 for its mathematical correctness and clarity.''
    \item \textbf{Judge: meta-llama/llama-4-maverick (rubric.txt)} $\rightarrow$ \textbf{Author: deepseek-v2-prover} (Score 0.9): ``The student's solution is awarded 0.9/1.0 due to its thorough and mathematically sound construction of $n$ congruent convex polygons arranged rotationally symmetrically around a common center, satisfying the requirements of an $n$-set Venn diagram. The use of angularly perturbed regular polygons and a well-structured perturbation argument ensures that all $2^n$ possible intersection regions are non-empty. The solution effectively applies combinatorial geometry principles and maintains convexity and congruence throughout the construction. However, minor deductions may be attributed to the lack of explicit proof for the existence of suitable $\epsilon_{i,k}$ values or a more detailed explanation of the binary coding used to select these perturbations. Overall, the solution demonstrates a strong grasp of the problem and provides a clear, rigorous construction.''
    \item \textbf{Judge: meta-llama/llama-4-maverick (rubric.txt)} $\rightarrow$ \textbf{Author: deepseek-v2-prover} (Score 0.9): ``The student's solution is awarded 0.9/1.0 due to its rigorous construction of $n$ congruent convex polygons arranged rotationally symmetrically, satisfying the requirements of an $n$-set Venn diagram. The mathematical justification is sound, leveraging angular perturbations and rotational symmetry to ensure all $2^n$ intersection regions are non-empty while maintaining convexity and congruence. The use of the Edelman-Jamison bound to specify the number of vertices ($m \geq 2^{n-1}$) is appropriate and demonstrates a strong understanding of convex Venn diagrams. However, the score is not perfect because the proof could benefit from more explicit details on how the perturbations $\epsilon_{i,k}$ are chosen to guarantee all required intersection patterns. Overall, the solution effectively balances geometric construction with combinatorial requirements.''
    \item \textbf{Judge: qwen/qwen-max (rubric.txt)} $\rightarrow$ \textbf{Author: gpt-5-pro-2025-10-06} (Score 0.0): ``The provided solution does not address the problem statement, which asks for a demonstration of the existence of a geometric construction of $n$-set Venn diagrams using congruent and symmetrically placed convex polygons. Instead, the student's solution presents a theorem and proof related to the existence of a spanning subgraph with specific properties in an infinitely connected graph, which is entirely unrelated to the Venn diagram construction. The solution fails to provide any geometric construction or proof relevant to the Venn diagram requirements, thus it is not mathematically correct in the context of the given problem.''
    \item \textbf{Judge: qwen/qwen-max (rubric.txt)} $\rightarrow$ \textbf{Author: gpt-5-pro-2025-10-06} (Score 0.0): ``The provided solution does not address the problem statement, which asks for a demonstration of the existence of a geometric construction for an $n$-set Venn diagram with specific properties. Instead, the student's solution presents a theorem and proof related to graph theory, specifically about the existence of a spanning subgraph with certain connectivity and girth properties. This solution is entirely unrelated to the problem, which involves planar sets, convex polygons, and Venn diagrams. Thus, the solution fails to meet the criteria for mathematical correctness and relevance, leading to a score of 0.0/1.0.''
    \item \textbf{Judge: qwen/qwen-max (rubric.txt)} $\rightarrow$ \textbf{Author: Claude Sonnet 4.5} (Score 0.75): ``The solution is mathematically sound and provides a clear, constructive proof of the existence of the required Venn diagram. The student demonstrates a good understanding of the geometric properties and the necessary conditions for the polygons. However, the proof could benefit from more rigorous justification for the specific choices of \(R\) and $r$, particularly in ensuring that the intersection regions are non-empty. Additionally, the explanation of the angular decomposition and the construction of points in each region, while intuitive, could be more formally detailed. These minor gaps in the argumentation lead to a score of 0.75/1.0.''
    \item \textbf{Judge: qwen/qwen-max (rubric.txt)} $\rightarrow$ \textbf{Author: Claude Sonnet 4.5} (Score 0.9): ``The solution provided is mathematically sound and correctly demonstrates the existence of an $n$-set Venn diagram using congruent, symmetrically placed convex polygons. The student carefully constructs the base polygon and its rotational copies, and rigorously verifies the convexity and non-emptiness of all $2^n$ intersection regions. The proof is detailed and logically structured, with clear steps and justifications. The only minor aspect that could be improved is a more explicit verification of the specific conditions for \(R\) and $r$ to ensure the non-emptiness of the intersection regions, which is why the score is 0.9/1.0.''
    \item \textbf{Judge: qwen/qwen-max (rubric.txt)} $\rightarrow$ \textbf{Author: Gemini 3.0 Pro Preview} (Score 0.9): ``The solution is mathematically sound and provides a thorough and rigorous proof of the existence of symmetric $n$-set Venn diagrams for prime $n$. The student correctly identifies and justifies the necessary and sufficient conditions, and references key results and theorems appropriately. The examples provided for specific values of $n$ (3, 5, 7) further illustrate the construction. The deduction of 0.1 is due to a minor lack of explicit detail in the step-by-step construction process, which could be more clearly articulated for full clarity.''
    \item \textbf{Judge: qwen/qwen-max (rubric.txt)} $\rightarrow$ \textbf{Author: o3-deep-research} (Score 0.9): ``The student's solution is mathematically sound and provides a thorough, detailed proof that an $n$-set Venn diagram with the required properties exists if and only if $n$ is a prime number. The proof includes a clear explanation of the necessary condition for the existence of such a diagram and a constructive approach for the case when $n$ is prime. The student correctly uses combinatorial arguments and references to known results, and the steps for constructing the Venn diagram are logically sound. The deduction of 0.1 is due to minor omissions in the technical details of the construction, which, while not crucial, could be more explicitly stated for completeness.''
    \item \textbf{Judge: qwen/qwen-max (rubric.txt)} $\rightarrow$ \textbf{Author: deepseek-v2-prover} (Score 0.9): ``The solution is mathematically sound, providing a clear and creative construction for the $n$-set Venn diagram using angularly perturbed regular polygons. The student correctly applies concepts of rotational symmetry, convexity, and small perturbations, along with a rigorous proof that all $2^n$ intersection regions are non-empty. The choice of parameters is justified, and the solution adheres to the problem's constraints. The deduction of 0.1 reflects minor room for improvement, such as a more detailed discussion on the exact method for choosing the perturbations $\epsilon_{i,k}$ to ensure all intersection patterns.''
    \item \textbf{Judge: qwen/qwen-max (rubric.txt)} $\rightarrow$ \textbf{Author: deepseek-v2-prover} (Score 0.75): ``The solution is mathematically sound and creatively constructs the required Venn diagram using angular perturbations and rotational symmetry, which is both novel and rigorous. The student clearly demonstrates the existence of all $2^n$ non-empty intersection regions and provides a convincing argument for maintaining convexity and congruence. However, the justification for the specific choice of parameters, particularly the value of $m$, could be more detailed and explicitly linked to the Edelman-Jamison bound, which slightly affects the completeness of the argument. The overall approach and execution are strong, but these minor gaps in the explanation result in a score of 0.75/1.0.''
\end{itemize}

\subsection*{Problem Domain: Discrete Math (Problem 32)}
\begin{itemize}
    \item \textbf{Judge: gpt-5.2-pro (rubric.txt)} $\rightarrow$ \textbf{Author: gpt-5-pro-2025-10-06} (Score 0.25): ``The submission contains no mathematical work: it restates the problem and then includes an error message (“No text content found”), with no proof of the floor identity and no derivation of a closed-form for \(A_n\). Since both required components are entirely missing, there is no basis to award credit for correctness or reasoning. The only minimal credit (0.25) is for recognizing/including the correct problem statement and intended targets, but no executable solution is provided. Hence the score is 0.25/1.0.''
    \item \textbf{Judge: gpt-5.2-pro (rubric.txt)} $\rightarrow$ \textbf{Author: gpt-5-pro-2025-10-06} (Score 0.25): ``The submission contains no mathematical work toward either proving the stated floor identity or deriving the closed-form for the recurrence; it only restates the problem and then shows an “Error: No text content found” message. Since neither the key inequality/floor manipulations nor the inductive use of the identity to obtain \(A_n\) are attempted, there is no basis to award credit for correctness or reasoning. A small amount of minimal credit (0.25/1.0) is justified only for including the relevant definitions and correctly transcribing the goal statements.''
    \item \textbf{Judge: gpt-5.2-pro (rubric.txt)} $\rightarrow$ \textbf{Author: Gemini 3.0 Pro Preview} (Score 0.5): ``The identity is not proved rigorously: after reducing to verifying \(Y^2\le 2X(X+1)<(Y+1)^2\), the key “Lower Bound” step contains unjustified claims (“for \(n\ge 1\), \(2X(X+1)-Y^2>0\)”) and loose/incorrect bounding, so the inequality is not actually established. The argument also only checks one special form (odd \(j\)) and then asserts the even \(j\) case “follows similarly” without providing the needed details, leaving a gap for half-integer coefficients. In the second part, the Beatty-sequence partition is correctly identified, but the subsequent selection rules for \(k,J\) are not fully justified and the final closed form is conditional and depends on unverifiable membership tests, not a clean closed-form expression derived from the recurrence. Overall there is substantial correct structure and some correct observations, but the central proof and the derivation of a closed form are incomplete, warranting partial credit (0.5/1.0).''
    \item \textbf{Judge: gpt-5.2-pro (rubric.txt)} $\rightarrow$ \textbf{Author: Gemini 3.0 Pro Preview} (Score 0.5): ``The proof of the key identity is not mathematically complete: after reducing to showing \(Y=\lfloor\sqrt{2X(X+1)}\rfloor\), the argument relies on a parity split and then asserts “the even case follows similarly” without supplying the needed inequalities, and the “lower bound” step contains handwaving (“clearly larger,” “for \(n\ge1\) … >0”) rather than a rigorous bound ensuring \(Y^2\le 2X(X+1)\). While the upper-bound inequality is handled correctly using \(m_1^2\le 2n^2\), the lower-bound is the harder direction and is not properly established. In Part 2, the Beatty-sequence idea is promising, but the solution assumes without adequate justification that one can always choose \(J\in\{0,1\}\) and propagate the representation through all iterates, and the final case distinction for \(k,J\) is not clearly derived/verified. Thus there is substantial correct structure and some correct calculations, but key logical gaps prevent full credit, warranting 0.5/1.0.''
    \item \textbf{Judge: gpt-5.2-pro (rubric.txt)} $\rightarrow$ \textbf{Author: o3-deep-research} (Score 0.5): ``[Error: OpenAI Responses API failed - Connection error.]''
    \item \textbf{Judge: gpt-5.2-pro (rubric.txt)} $\rightarrow$ \textbf{Author: o3-deep-research} (Score 0.5): ``The proof of the floor identity is not mathematically justified: after deriving bounds like \(\sqrt2 X<\sqrt{2X(X+1)}<\sqrt2(X+1)\) and \(\sqrt2 X\le k(\sqrt2^{j+1}+\sqrt2^j)<\sqrt2(X+1)\), the argument that “distance \(<2\) implies the same floor” is false in general (two numbers less than 2 apart can still lie on different sides of an integer), and no rigorous step shows both quantities lie in the same unit interval.  For the closed form, the solution crucially assumes (without proof) that every integer \(a\) can be written as \(\lfloor k(\sqrt2^j+\sqrt2^{j-1})\rfloor\), so the inductive parameterization does not actually yield an explicit formula in terms of the given \(a\). Moreover, the final “closed form” depends on unspecified \(\ell,\delta\) and parity cases rather than a determinate expression derived from \(a\).  Because there is some correct initial manipulation with floor inequalities but the key logical steps are missing, the work earns only partial credit (0.5/1.0).''
    \item \textbf{Judge: gpt-5.2-pro (rubric.txt)} $\rightarrow$ \textbf{Author: Claude Sonnet 4.5} (Score 0.25): ``The proof of the key identity is incomplete: the lemma only derives the weak bound \(|\sqrt2 Y-\sqrt{2X(X+1)}|<\sqrt2\) and then appeals to “numerical verification” for the needed \(<1\) estimate, which is not a valid mathematical argument and does not establish equality of the floors. Moreover, the treatment ignores edge cases (e.g., \(X=0\), which can occur for small \(k,j\)) where the lemma’s assumptions \(X\ge1\) fail. Because the identity is not rigorously proven, the subsequent closed form for \(A_n\) is not justified (it depends entirely on iterating that identity). Some algebraic steps (rewriting \(\sqrt2 Y\)) are correct, and the idea of choosing \(k_0\) via the interval \(k_0\in[a(\sqrt2-1),(a+1)(\sqrt2-1))\) is reasonable, but without the identity the solution does not meet the problem’s requirements. Hence only minimal partial credit (0.25/1.0) is warranted.''
    \item \textbf{Judge: gpt-5.2-pro (rubric.txt)} $\rightarrow$ \textbf{Author: Claude Sonnet 4.5} (Score 0.25): ``The core identity is not actually proved: the key lemma only establishes a bound \(<\sqrt2\) and then replaces the needed \(<1\) estimate with an unsupported “numerical verification,” so the step \(\lfloor\sqrt{2X(X+1)}\rfloor=\lfloor \sqrt2\,Y\rfloor\) is unjustified. Moreover, the lemma is stated for \(X\ge 1\) without handling \(X=0\), and the argument never rigorously rules out the possibility that \(\sqrt2\,Y\) lies within \(<1\) of an integer boundary (which is exactly where floor equality can fail). The closed form for \(A_n\) is derived entirely from iterating this unproven identity, so it is not established. There is some correct algebraic setup (\(k(\sqrt2^{j+1}+\sqrt2^j)=\sqrt2\,Y\) and the interval argument for \(k_0=\lceil a(\sqrt2-1)\rceil\)), but the essential correctness hinges on the missing rigorous inequality, warranting only minimal partial credit.''
    \item \textbf{Judge: gpt-5.2-pro (rubric.txt)} $\rightarrow$ \textbf{Author: deepseek-v2-prover} (Score 0.25): ``The proof of the key identity is not actually established: after bounding \(X\) by \(k(\alpha+\beta)-1<X\le k(\alpha+\beta)\), the solution switches to heuristic “\(\approx\)” reasoning and asserts the crucial inequality \(k(\sqrt2^{j+1}+\sqrt2^j)-1<\sqrt{2X(X+1)}\le k(\sqrt2^{j+1}+\sqrt2^j)\) without deriving it (the bounds in Step 4 are not carried through to such a floor-determining estimate). Since the identity is unproven, the claimed induction for the recurrence is unsupported. Moreover, the proposed closed form is incorrect/underdetermined: choosing \(k=\left\lfloor \frac{a}{1+1/\sqrt2}\right\rfloor\) does not guarantee \(a=\left\lfloor k(1+1/\sqrt2)\right\rfloor\), so the base case generally fails and \(k\) is not correctly characterized. The work contains the right general idea (matching the recurrence to a scaling-by-\(\sqrt2\) floor relation) but lacks the necessary rigorous inequalities and a correct determination of \(k\), so only minimal partial credit is warranted.''
    \item \textbf{Judge: gpt-5.2-pro (rubric.txt)} $\rightarrow$ \textbf{Author: deepseek-v2-prover} (Score 0.25): ``The identity is not actually proved: the solution replaces the needed floor- equality with informal “\(\approx\)” reasoning and an unverified claim that \(k(\sqrt2^{j+1}+\sqrt2^j)-1<\sqrt{2X(X+1)}\le k(\sqrt2^{j+1}+\sqrt2^j)\), without deriving these inequalities from the bounds on \(X\). In particular, Step 6 (“follows from continuity”) is insufficient for a floor argument, since one must control the error to be \(<1\) explicitly and check both sides. The closed form for \(A_n\) is also incorrect/incomplete: the choice of \(k\) is not uniquely determined by \(A_0=a\), the proposed \(k=\left\lfloor \frac{a}{1+1/\sqrt2}\right\rfloor\) generally does not satisfy \(\left\lfloor k(1+1/\sqrt2)\right\rfloor=a\), and the induction base case is therefore not established. Minimal credit is warranted only for identifying the intended functional form and the general induction strategy, but the crucial inequalities and parameter determination are missing, so the work earns 0.25/1.0.''
    \item \textbf{Judge: claude-opus-4-5-20251101 (rubric.txt)} $\rightarrow$ \textbf{Author: gpt-5-pro-2025-10-06} (Score 0.0): ``The student solution contains no mathematical content whatsoever—only the problem statement is restated followed by an error message indicating no response was generated. There is no attempt to establish the required identity relating floor functions involving \(\sqrt{2}\), nor any derivation of the closed-form solution for the sequence \(\{A_n\}\). The solution fails to address either part of the problem: proving the mathematical identity or using it to find the closed- form expression. Since no mathematical work or reasoning is presented, the solution earns zero points according to the grading rubric.''
    \item \textbf{Judge: claude-opus-4-5-20251101 (rubric.txt)} $\rightarrow$ \textbf{Author: gpt-5-pro-2025-10-06} (Score 0.0): ``The student submission contains no mathematical content whatsoever—only an error message stating ``No text content found in response for gpt-5-pro-2025-10-06.'' There is no attempt to establish the required identity relating $\lfloor\sqrt{2X(X+1)}\rfloor$ to $\lfloor k(\sqrt{2}^{j+1}+\sqrt{2}^j) \rfloor$. No derivation of the closed-form solution for the sequence $\{A_n\}$ is provided. The submission fails to address any part of the problem, demonstrating zero mathematical work or reasoning. A score of 0.0/1.0 is fully justified as no credit can be awarded for an empty solution.''
    \item \textbf{Judge: claude-opus-4-5-20251101 (rubric.txt)} $\rightarrow$ \textbf{Author: Gemini 3.0 Pro Preview} (Score 0.9): ``The student correctly establishes the key identity by analyzing the structure of \(\theta = k\alpha_j\) and verifying both the lower and upper bounds of the inequality \(Y^2 \leq 2X(X+1) < (Y+1)^2\). The application of Beatty's theorem to show that sequences generated by \(\alpha_0\) and \(\alpha_1\) partition the positive integers is mathematically sound, with the correct verification that \(1/P + 1/Q = 1\). The closed-form solution correctly identifies that \(A_n = \lfloor k\alpha_{J+n}\rfloor\) where \(k\) and \(J\) are determined by which Beatty sequence contains the initial value a. However, the proof of the lower bound inequality contains some imprecise reasoning with loose bounds that, while ultimately reaching the correct conclusion, lacks the rigor of a complete formal proof. The final case-based formula correctly captures the solution structure, though the presentation could be more streamlined.''
    \item \textbf{Judge: claude-opus-4-5-20251101 (rubric.txt)} $\rightarrow$ \textbf{Author: Gemini 3.0 Pro Preview} (Score 0.9): ``The student correctly establishes the key identity by analyzing the structure of \(\theta = k\alpha_j\) and verifying both the lower and upper bounds of the inequality \(Y^2 \leq 2X(X+1) < (Y+1)^2\). The application of Beatty's theorem to show that sequences generated by \(\alpha_0\) and \(\alpha_1\) partition the positive integers is mathematically sound, with the correct verification that \(1/P + 1/Q = 1\). The closed-form solution correctly identifies that \(A_n = \lfloor k\alpha_{J+n}\rfloor\) where \(k\) and \(J\) are determined by which Beatty sequence contains the initial value a. However, the solution has a minor issue in the explicit formula presentation: the condition checking uses \(\lambda = \sqrt{2}-1\) but the formulas for k in the \(J=0\) case should more carefully account for the relationship between the Beatty sequence index and the floor function. The overall mathematical framework and reasoning are correct, justifying the high score with only a small deduction for presentation clarity in the final explicit formula.''
    \item \textbf{Judge: claude-opus-4-5-20251101 (rubric.txt)} $\rightarrow$ \textbf{Author: Claude Sonnet 4.5} (Score 0.5): ``The student correctly identifies the key relationship between $Y = k\sqrt{2}^{j-1}(\sqrt{2}+1)$ and shows that $k(\sqrt{2}^{j+1}+\sqrt{2}^j) = \sqrt{2}Y$, properly setting up the proof structure. However, the lemma's proof is incomplete: the bound obtained is $\sqrt{2}$, not less than 1, and the student acknowledges this gap by appealing to ``numerical verification'' rather than providing a rigorous argument. For the closed-form derivation, the student correctly identifies $k_0 = \lceil a(\sqrt{2}-1) \rceil$ and applies the identity iteratively, but the justification that the interval $[a(\sqrt{2}-1), (a+1)(\sqrt{2}-1))$ always contains exactly one integer is incomplete since the interval length $\sqrt{2}-1 \approx 0.414$ means it could contain zero integers. The final formula is essentially correct in structure, but the incomplete proof of the main identity and the gap in showing the ceiling function is well-defined undermine the mathematical rigor required for full credit.''
    \item \textbf{Judge: claude-opus-4-5-20251101 (rubric.txt)} $\rightarrow$ \textbf{Author: o3-deep-research} (Score 0.75): ``The student correctly proves the main identity by establishing that \(\sqrt{2X(X+1)}\) and \(\sqrt{2} \cdot \alpha\) lie in the same unit interval, though the argument could be more rigorous in showing they share the same floor. The inductive approach for deriving the closed form is mathematically sound, correctly showing that if \(A_n\) has the form \(\lfloor k(\sqrt{2}^j + \sqrt{2}^{j-1})\rfloor\), then \(A_{n+1}\) has the form \(\lfloor k(\sqrt{2}^{j+1} + \sqrt{2}^j)\rfloor\). However, the student's claim that every positive integer admits a representation of the form \(\lfloor k(\sqrt{2}^j + \sqrt{2}^{j-1})\rfloor\) is asserted without proof, which is a significant gap since this is essential for the base case. The final closed-form expression involving \(\ell\) and \(\delta\) is correctly derived from the recursive structure, though the determination of these parameters from the initial value \(a\) is left somewhat implicit. The solution demonstrates strong understanding of the problem structure but lacks complete rigor in establishing the representability claim.''
    \item \textbf{Judge: claude-opus-4-5-20251101 (rubric.txt)} $\rightarrow$ \textbf{Author: o3-deep-research} (Score 0.75): ``The student correctly proves the main identity by establishing that \(\sqrt{2X(X+1)}\) and \(\sqrt{2} \cdot \alpha\) lie in the same unit interval, though the argument could be more rigorous in showing they have the same floor. The inductive approach for deriving the closed form is mathematically sound, correctly showing that if \(A_n\) has the form \(\lfloor k(\sqrt{2}^j + \sqrt{2}^{j-1})\rfloor\), then \(A_{n+1}\) has the form \(\lfloor k(\sqrt{2}^{j+1} + \sqrt{2}^j)\rfloor\). However, the student's claim that every positive integer admits a representation of the form \(\lfloor k(\sqrt{2}^j + \sqrt{2}^{j-1})\rfloor\) is asserted without proof, which is a significant gap since this is essential for the base case. The final closed-form expression involving parameters \(\ell\) and \(\delta\) is correctly derived from the recursive structure, though the determination of these parameters from the initial value \(a\) is not explicitly computed. The solution demonstrates strong understanding of the problem structure but lacks complete rigor in establishing the representability claim.''
    \item \textbf{Judge: claude-opus-4-5-20251101 (rubric.txt)} $\rightarrow$ \textbf{Author: deepseek-v2-prover} (Score 0.5): ``The student correctly identifies the structure of the problem and attempts to prove the identity by relating \(X\) to the expression involving powers of \(\sqrt{2}\). However, the proof of the identity in Part 1 lacks rigor—Step 6 claims bounds without proper justification, merely asserting that the result follows from ``continuity of the square root function'' without establishing the precise inequalities needed. For Part 2, the student correctly identifies the form of the closed-form solution but makes an error in determining \(k\): the correct relationship should be that \(k\) is determined such that the initial condition is satisfied exactly, not through an additional floor function on \(k\) itself. The formula \(k = \lfloor a/(1 + 1/\sqrt{2})\rfloor\) is problematic because it doesn't guarantee \(A_0 = a\) for all positive integers \(a\). The overall approach demonstrates understanding of the problem structure but contains significant gaps in mathematical rigor and correctness in the final answer.''
    \item \textbf{Judge: claude-opus-4-5-20251101 (rubric.txt)} $\rightarrow$ \textbf{Author: Claude Sonnet 4.5} (Score 0.5): ``The student correctly identifies the key relationship between $Y = k\sqrt{2}^{j-1}(\sqrt{2}+1)$ and shows that $k(\sqrt{2}^{j+1}+\sqrt{2}^j) = \sqrt{2}Y$, which is essential for the proof. However, the proof of the main identity is incomplete: the lemma only establishes that the difference is bounded by $\sqrt{2}$, not by 1, and the student acknowledges this gap by appealing to ``numerical verification'' rather than providing a rigorous argument. For the closed-form solution, the student correctly derives $A_n = \lfloor \lceil a(\sqrt{2}-1) \rceil \cdot \sqrt{2}^n(\sqrt{2}+1) \rfloor$, but the justification relies on the incompletely proven identity. The approach and structure are mathematically sound, but the critical bound needed to ensure floor functions agree is not rigorously established, leaving a significant gap in the proof.''
    \item \textbf{Judge: claude-opus-4-5-20251101 (rubric.txt)} $\rightarrow$ \textbf{Author: deepseek-v2-prover} (Score 0.5): ``The student correctly identifies the structure of the problem and attempts to prove the identity by setting up appropriate bounds on \(X\). However, the proof of the identity in Part 1 lacks rigor—Step 6 claims bounds without proper justification, essentially asserting the result rather than proving it through careful analysis of the fractional parts and error terms. For Part 2, the student correctly identifies the form of the closed-form solution but makes an error in determining \(k\): the correct relationship should be that \(k\) is chosen so that the initial condition is satisfied exactly, which requires more careful analysis than simply taking a floor. The inductive argument is sketched but not rigorously completed, and the formula for \(k\) is incorrect since it doesn't properly account for how \(a\) relates to the expression involving \(\sqrt{2}\). The solution demonstrates understanding of the problem's structure but fails to provide complete, rigorous proofs for either part.''
    \item \textbf{Judge: gemini-3-pro-preview (rubric.txt)} $\rightarrow$ \textbf{Author: gpt-5-pro-2025-10-06} (Score 0.0): ``The student submission contains only the problem statement followed by an error message indicating that no response was generated. As there is no mathematical work, proof, or derivation provided for either the identity or the closed-form solution, no credit can be awarded. The solution is entirely missing.''
    \item \textbf{Judge: gemini-3-pro-preview (rubric.txt)} $\rightarrow$ \textbf{Author: gpt-5-pro-2025-10-06} (Score 0.0): ``The student submission contains only the problem statement followed by an error message indicating that no solution text was generated. As there is no mathematical work, reasoning, or final answer provided to evaluate, no credit can be awarded. The submission fails to address any part of the prompt, including the proof of the identity or the derivation of the closed-form solution for the sequence. Therefore, a score of 0 is appropriate.''
    \item \textbf{Judge: gemini-3-pro-preview (rubric.txt)} $\rightarrow$ \textbf{Author: Gemini 3.0 Pro Preview} (Score 0.9): ``The student correctly establishes the required identity using a rigorous analysis of the inequalities involved. They also correctly identify the connection to Beatty sequences to determine the initial parameters $k$ and $J$ based on $A_0 = a$. However, the final closed-form solution contains a small error in the condition checking: the check $\lfloor (a+1)\lambda \rfloor \neq \lfloor a\lambda \rfloor$ is used to identify if $a$ is in the spectrum of $Q$ (where $a = \lfloor kQ \rfloor$), but the standard Beatty inversion formula to find $k$ from $a$ is simply $k = \text{round}(a/Q)$ or similar, and the specific logic used to derive $k$ from $a$ is slightly convoluted and potentially imprecise compared to the standard inversion $k = \lfloor (a+1)/Q \rfloor$. More importantly, the final case distinction logic is slightly flawed; if $a = \lfloor k \alpha_1 \rfloor$, then $k = \lfloor (a+1)/\alpha_1 \rfloor$, but the student writes $k = \lfloor (a+1)\lambda \rfloor$ where \(\lambda = \sqrt{2}-1\)''
    \item \textbf{Judge: gemini-3-pro-preview (rubric.txt)} $\rightarrow$ \textbf{Author: Gemini 3.0 Pro Preview} (Score 0.9): ``The student correctly identifies the core structure of the problem using Beatty sequences and provides a strong heuristic argument for the identity. However, the proof of the identity relies on an approximation ($m_1 \approx 1.414n$) and a ``precise check'' claim without fully rigorous algebraic verification for all cases, specifically regarding the tight bounds of the fractional parts involved. Additionally, the final closed-form solution, while structurally correct in its use of cases, is slightly unwieldy compared to the more standard unified form $A_n = \lfloor (a + 1/2) \sqrt{2}^n \rfloor$ or similar derived forms often expected in this context. The deduction of the specific $k$ and $J$ values is sound, but the minor gap in the rigorous inequality proof warrants the small deduction.''
    \item \textbf{Judge: gemini-3-pro-preview (rubric.txt)} $\rightarrow$ \textbf{Author: o3-deep-research} (Score 0.5): ``The student correctly proves the initial identity using a sandwich bound argument, demonstrating that $\sqrt{2X(X+1)}$ and $k(\sqrt{2}^{j+1}+\sqrt{2}^j)$ share the same floor. However, the derivation of the closed-form solution for $A_n$ is incomplete and partially incorrect. While the student identifies the pattern of exponents, the proposed formula involving $\ell$ and $\delta$ is vague and does not provide an explicit expression for $A_n$ solely in terms of $n$ and $a$ (specifically, the relationship $A_n = \lfloor a \cdot 2^n + \frac{a}{\sqrt{2}} 2^n \rfloor$ or similar simple exponential forms is missed or obscured by the case analysis). Therefore, full credit is awarded for the identity proof, but no credit is given for the closed form.''
    \item \textbf{Judge: gemini-3-pro-preview (rubric.txt)} $\rightarrow$ \textbf{Author: Claude Sonnet 4.5} (Score 0.5): ``The student correctly identifies the structure of the closed-form solution in the second part, deriving the appropriate constant $k_0 = \lceil a(\sqrt{2}-1) \rceil$. However, the proof of the initial identity is incomplete and mathematically unsound; the lemma claims $|\sqrt{2}Y - \sqrt{2X(X+1)}| < \sqrt{2}$ and then simply asserts without proof that a ``more careful analysis'' would show the difference is less than 1, which is the critical step required to equate the floor functions. Furthermore, showing the difference is less than 1 is necessary but not sufficient to prove $\lfloor A \rfloor = \lfloor B \rfloor$; one must show that no integer lies between $A$ and $B$. Therefore, while the application to the sequence is correct, the foundational identity remains unproven.''
    \item \textbf{Judge: gemini-3-pro-preview (rubric.txt)} $\rightarrow$ \textbf{Author: o3-deep-research} (Score 0.5): ``The student correctly proves the initial identity using a sandwich bound argument, demonstrating that $\sqrt{2X(X+1)}$ and $k(\sqrt{2}^{j+1}+\sqrt{2}^j)$ share the same floor. However, the derivation of the closed form for $A_n$ is incomplete and partially incorrect; while the student identifies that the sequence follows a pattern related to powers of $\sqrt{2}$, the proposed formula involving alternating parities and a constant $\ell$ is vague and does not constitute a true closed-form solution in terms of $n$ and $a$. Specifically, the solution fails to explicitly solve for the parameters $k_0$ and $j_0$ in terms of the initial condition $A_0=a$, leaving the final answer dependent on undetermined constants rather than the input $a$. Therefore, full credit is awarded for the identity proof, but no credit is given for the closed form.''
    \item \textbf{Judge: gemini-3-pro-preview (rubric.txt)} $\rightarrow$ \textbf{Author: Claude Sonnet 4.5} (Score 0.5): ``The student correctly identifies the structure of the closed-form solution in the second part, deriving the appropriate constant $k_0 = \lceil a(\sqrt{2}-1) \rceil$. However, the proof of the initial identity is incomplete and mathematically flawed; the lemma claims the difference is less than 1 based on a loose bound ($\sqrt{2}$) and then simply asserts without proof that ``numerical verification shows'' the bound holds, failing to rigorously establish the equality of the floor functions. Consequently, while the application to the sequence is correct, the foundational identity remains unproven.''
    \item \textbf{Judge: gemini-3-pro-preview (rubric.txt)} $\rightarrow$ \textbf{Author: deepseek-v2-prover} (Score 0.5): ``The student receives a score of 0.5/1.0 because while they correctly identified the structure of the closed-form solution, the derivation of the constant $k$ is incorrect and the proof of the identity is non-rigorous. In Part 1, the student merely asserts the necessary inequalities ("we can show that...") without actually performing the algebraic manipulation required to prove the floor function identity. Furthermore, in Part 2, the formula for $k$ is wrong; the correct constant is simply $a$ itself (specifically, $A_n = \lfloor a \cdot 2^n + a \cdot 2^{n-1} \dots \rfloor$ is not quite right, usually $A_n = \lfloor C \tau^n \rfloor$ or similar, but actually the problem implies $A_n$ follows the pattern starting from $j=0$ with $k=a$, so $A_n = \lfloor a (\sqrt{2}^n + \sqrt{2}^{n-1}) \rfloor$ is wrong because at $n=0$ this gives $\lfloor a(1 + 1/\sqrt{2}) \rfloor \neq a$). The student failed to check if their proposed''
    \item \textbf{Judge: gemini-3-pro-preview (rubric.txt)} $\rightarrow$ \textbf{Author: deepseek-v2-prover} (Score 0.5): ``The student correctly identifies the structure of the problem but fails to provide a rigorous proof for the identity in Part 1, relying instead on unproven approximations and hand-waving (e.g., ``we can show that'' without showing it). Furthermore, the derived closed-form solution in Part 2 is incorrect; the constant $k$ is not simply $\lfloor a / (1 + 1/\sqrt{2}) \rfloor$, but rather $a$ itself serves as the coefficient in a slightly different form, specifically $A_n = \lfloor a \tau^n + c \rfloor$ or similar, depending on the specific expansion of the recurrence relation. The student's proposed form $A_n = \lfloor k (\dots) \rfloor$ with a calculated $k$ does not generally satisfy $A_0=a$ for all integers $a$ due to floor function rounding errors. Consequently, while the intuition connecting the identity to the recurrence is present, the lack of a valid proof and the incorrect final formula warrant the partial score.''
    \item \textbf{Judge: deepseek-chat (rubric.txt)} $\rightarrow$ \textbf{Author: gpt-5-pro-2025-10-06} (Score 0.0): ``The solution is completely empty, containing only the restated problem and an irrelevant error message. It fails to address any part of the problem: it provides no proof of the required identity and no derivation of a closed-form solution for the sequence. Consequently, it earns zero points for mathematical correctness and completeness.''
    \item \textbf{Judge: deepseek-chat (rubric.txt)} $\rightarrow$ \textbf{Author: gpt-5-pro-2025-10-06} (Score 0.0): ``The solution is completely empty, containing only the restated problem and an irrelevant error message. It provides no mathematical reasoning, proof of the identity, or derivation of the closed-form sequence. Therefore, it fails to address any part of the problem statement and earns a score of zero for lacking any mathematical content.''
    \item \textbf{Judge: deepseek-chat (rubric.txt)} $\rightarrow$ \textbf{Author: Gemini 3.0 Pro Preview} (Score 0.75): ``The solution correctly establishes the identity and identifies the Beatty sequence structure, but the proof of the identity is incomplete and lacks rigorous handling of the even \(j\) case. The closed-form derivation is conceptually sound but contains minor errors in the final piecewise expression, specifically in the conditions and the nested floor functions. Overall, the mathematical reasoning is strong but not fully rigorous or polished.''
    \item \textbf{Judge: deepseek-chat (rubric.txt)} $\rightarrow$ \textbf{Author: Gemini 3.0 Pro Preview} (Score 0.75): ``The solution correctly identifies the Beatty sequence structure and uses it to derive a closed form, demonstrating strong conceptual understanding. However, the proof of the initial identity is incomplete and lacks rigorous handling of the floor function inequalities for all cases, particularly for even \(j\). The final formula is mathematically sound but could be more elegantly expressed. Therefore, a score of 0.75 reflects a mostly correct solution with a minor flaw in the proof's completeness.''
    \item \textbf{Judge: deepseek-chat (rubric.txt)} $\rightarrow$ \textbf{Author: o3-deep-research} (Score 0.75): ``The solution correctly proves the identity and uses it to derive a closed form, but it makes an unsubstantiated claim that every positive integer can be represented as \(\lfloor k(\sqrt{2}^j + \sqrt{2}^{j-1})\rfloor\). This existence claim is non-trivial and requires justification (e.g., via Beatty sequences or Pell equations), which is omitted. The inductive structure and final piecewise formula are otherwise sound and well-presented. The deduction from the identity to the recurrence is logically valid once the representation is assumed.''
    \item \textbf{Judge: deepseek-chat (rubric.txt)} $\rightarrow$ \textbf{Author: o3-deep-research} (Score 0.75): ``The solution correctly proves the identity and uses it to derive a closed form, but it contains a significant flaw: the claim that every positive integer can be represented as \(\lfloor k(\sqrt{2}^j + \sqrt{2}^{j-1})\rfloor\) is unsubstantiated and not generally true, undermining the induction's base case. The subsequent derivation of the piecewise formula depends on this unverified existence, making the final result incomplete. However, the algebraic manipulation for the identity and the inductive structure are otherwise sound. The score reflects a partially correct solution with a critical gap in reasoning.''
    \item \textbf{Judge: deepseek-chat (rubric.txt)} $\rightarrow$ \textbf{Author: Claude Sonnet 4.5} (Score 0.75): ``The solution correctly identifies the key structure and derives the closed form, but the proof of the lemma is incomplete: the bound obtained is \(\sqrt{2} > 1\), which is insufficient to guarantee the floors are equal. A tighter estimate using properties of \(\sqrt{2}\) is needed. The closed-form expression itself is correct and well-justified from the recurrence. Thus, the score reflects a largely correct solution with a significant gap in the main proof.''
    \item \textbf{Judge: deepseek-chat (rubric.txt)} $\rightarrow$ \textbf{Author: Claude Sonnet 4.5} (Score 0.5): ``The solution correctly identifies the key structure and proposes a plausible closed form, but the proof of the lemma is incomplete and insufficiently rigorous. Specifically, the bound \(|\sqrt{2}Y - \sqrt{2X(X+1)}| < \sqrt{2}\) does not guarantee the difference is less than 1, which is essential for the floor equality. A proper proof requires exploiting the specific algebraic form of \(Y\) to establish a tighter bound. Consequently, while the overall approach is insightful, the critical gap in the lemma's justification significantly undermines the mathematical correctness.''
    \item \textbf{Judge: deepseek-chat (rubric.txt)} $\rightarrow$ \textbf{Author: deepseek-v2-prover} (Score 0.5): ``The solution correctly identifies the structure of the closed form but fails to prove the key identity rigorously; the argument in Step 6 is merely a claim without algebraic justification. Moreover, the derivation of \(k\) is incorrect—the base case must hold exactly, but the proposed \(k\) does not guarantee \(A_0 = a\) for arbitrary positive integer \(a\). The inductive step is also incomplete because the identity requires \(X\) to be exactly of the form \(\lfloor k(\sqrt{2}^j + \sqrt{2}^{j-1}) \rfloor\), which is not proven to be preserved. Thus, the solution is conceptually flawed despite recognizing the pattern.''
    \item \textbf{Judge: deepseek-chat (rubric.txt)} $\rightarrow$ \textbf{Author: deepseek-v2-prover} (Score 0.5): ``The solution correctly identifies the structure of the closed form but fails to prove the key identity rigorously; the argument in Step 6 is merely a claim without algebraic justification. Moreover, the derivation of \(k\) is incorrect—the base case must hold exactly, not approximately, and the proposed \(k\) does not generally satisfy \(A_0 = a\). The inductive step is also incomplete without verifying the identity's applicability for the specific \(X = A_n\). Thus, while the overall form is plausible, the proof lacks necessary precision and contains critical errors.''
    \item \textbf{Judge: grok-4-0709 (rubric.txt)} $\rightarrow$ \textbf{Author: gpt-5-pro-2025-10-06} (Score 0.0): ``The student's submission merely restates the problem statement without providing any proof of the given identity or derivation of the closed-form solution for the sequence. Instead, it includes an error message ("Error: No text content found in response for gpt-5-pro-2025-10-06"), indicating a complete absence of original mathematical work. This lack of content fails to demonstrate any understanding or application of the required concepts, such as floor functions, irrational approximations, or recursive sequence analysis. Consequently, the solution exhibits no mathematical correctness and receives a score of 0.0/1.0 per the rubric's emphasis on accurate and complete problem- solving.''
    \item \textbf{Judge: grok-4-0709 (rubric.txt)} $\rightarrow$ \textbf{Author: gpt-5-pro-2025-10-06} (Score 0.0): ``The student's submission lacks any mathematical content, consisting only of a repetition of the problem statement followed by an error message indicating no response was generated. As a result, it fails to establish the required identity involving floor functions and irrational numbers, providing no proof or reasoning whatsoever. Furthermore, it does not derive a closed-form expression for the sequence \(A_n\), which is essential to solving the problem. This complete absence of mathematical engagement and correctness warrants a score of 0.0 out of 1.0 according to the rubric's emphasis on accurate proofs and derivations.''
    \item \textbf{Judge: grok-4-0709 (rubric.txt)} $\rightarrow$ \textbf{Author: Gemini 3.0 Pro Preview} (Score 0.75): ``The student's proof of the identity is mathematically incomplete, as the lower bound inequality is not rigorously verified—relying on approximations and loose bounds without precise computation of the difference \(2X(X+1) - Y^2\), which could fail for small values of \(k\) or \(j\). The upper bound is adequately shown to hold strictly due to the properties of the floor function and irrationality of \(\sqrt{2}\), but the overall proof lacks generality across all non-negative integers \(k\) and \(j\). In the derivation of the closed-form for \(A_n\), the application of Beatty's theorem is correct, properly partitioning positive integers into sequences generated by \(\alpha_0\) and \(\alpha_1\), leading to a valid case-based expression. However, the closed- form formula incorrectly uses \(\neq\) and \(=\) for the indicator condition without specifying the exact difference (it should check if the floor increases by 1), introducing a minor error in implementation. Overall, these issues warrant partial credit, resulting in a score of 0.75/1.0.''
    \item \textbf{Judge: grok-4-0709 (rubric.txt)} $\rightarrow$ \textbf{Author: Gemini 3.0 Pro Preview} (Score 0.75): ``The student's proof of the identity is mathematically sound in setting up the inequalities but lacks full rigor, as the lower bound verification relies on approximations and loose bounds without explicit computation of the difference to confirm non-negativity for all non-negative integers \(k\) and \(j\), leading to a partial deduction. The assumption that the even j case follows similarly is valid but should have been briefly demonstrated for completeness. The application of Beatty's theorem to derive the closed-form expression for \(A_n\) is correct and well-justified, accurately partitioning positive integers into two sequences and providing a conditional formula based on the indicator \(\delta_a\). However, the final closed-form could be refined for clarity, as the condition uses \(\neq\) instead of explicitly stating \(\delta_a = 1 \text{ or } 0\), though this does not introduce errors. Overall, the solution demonstrates strong understanding but falls short of perfect rigor in the identity proof, warranting a score of 0.75/1.0.''
    \item \textbf{Judge: grok-4-0709 (rubric.txt)} $\rightarrow$ \textbf{Author: o3-deep-research} (Score 0.5): ``The student's proof of the identity contains a critical flaw in arguing that the floor values are equal, as the bound on the difference between \(\sqrt{2X(X+1)}\) and \(\sqrt{2}\alpha\) being less than \(\sqrt{2} < 2\) does not guarantee they share the same integer part; counterexamples exist where quantities differing by less than 2 straddle an integer, leading to different floors. Despite this error, the student correctly identifies the recursive structure and applies induction to derive a closed-form expression for \(A_n\), assuming the identity holds, resulting in a piecewise formula that aligns with properties of sequences related to \(\sqrt{2}\) and Pell equations. However, the closed-form derivation is undermined by the unproven identity, rendering it mathematically incomplete. The attempt to express \(A_n\) in terms of powers of 2 and units like \(1 + \sqrt{2}\) shows good insight but lacks rigor in verifying the representation of every positive integer in the required form. Overall, these issues justify partial credit for the conceptual approach and inductive framework, but not full marks due to the incorrect proof.''
    \item \textbf{Judge: grok-4-0709 (rubric.txt)} $\rightarrow$ \textbf{Author: Claude Sonnet 4.5} (Score 0.5): ``The student's proof of the identity is incomplete, as the lemma claims a bound of less than 1 on the difference between \(\sqrt{2Y}\) and \(\sqrt{2X(X+1)}\), but the provided calculation only establishes a looser bound of \(\sqrt{2}\), with the tighter bound asserted via unshown numerical verification rather than rigorous analysis. This undermines the mathematical rigor required for the theorem. In deriving the closed-form expression for \(A_n\), the student arbitrarily sets \(j_0 = 1\) and identifies \(k_0\) as \(\lceil a(\sqrt{2} - 1)\rceil\), which works for some values of \(a\) but fails to generally represent arbitrary positive integers \(a\) in the required form, potentially leading to incorrect results for certain a. Despite these issues, the overall approach recognizes the iterative nature of the recurrence and attempts to link it to multiples of powers of \(\sqrt{2}\), capturing partial insight into the problem. Thus, the solution earns half credit for conceptual effort but lacks full correctness and completeness.''
    \item \textbf{Judge: grok-4-0709 (rubric.txt)} $\rightarrow$ \textbf{Author: Claude Sonnet 4.5} (Score 0.5): ``The student's proof of the identity is incomplete, as the lemma only establishes that the absolute difference is less than \(\sqrt{2} \approx 1.414\), not strictly less than 1 as required, and relies on unsubstantiated claims of further analysis and numerical verification without providing a rigorous bound. This partial progress on the theorem warrants some credit but not full marks. The derivation of the closed-form solution for \(A_n\) is mathematically incorrect, as it fixes \(j_0 = 1\) arbitrarily, which fails to represent all positive integers \(a\) (e.g., for \(a=1\), no integer \(k_0\) satisfies the condition, and the formula yields \(A_0 = 2 \neq 1\)). Consequently, the proposed expression does not generally hold and misapplies the proven relationship. Overall, these issues in correctness justify a score of 0.5/1.0.''
    \item \textbf{Judge: grok-4-0709 (rubric.txt)} $\rightarrow$ \textbf{Author: o3-deep-research} (Score 0.5): ``The student's proof of the identity contains mathematical flaws, such as incorrectly deriving strict inequalities after squaring and taking square roots, leading to an insufficient justification that \(\sqrt{2X(X+1)}\) and \(k(\sqrt{2}^{j+1} + \sqrt{2}^j)\) have the same floor despite their difference being less than \(\sqrt{2}\), which does not rigorously rule out straddling an integer. Additionally, the claim that every positive integer can be expressed as \(\lfloor k(\sqrt{2}^j + \sqrt{2}^{j-1})\rfloor\) for some non-negative integers \(k\) and \(j\) is unsubstantiated and appears false for certain values, undermining the inductive construction. The proposed closed-form expression for \(A_n\) is inconsistent across the solution, with mismatched exponents (e.g., \(2^{n/2}\) versus \(2^{\lfloor n/2 \rfloor}\)) and an unclear determination of parameters \(\ell\) and \(\delta\), rendering it not a valid closed form. Despite these issues, the student correctly identifies the recursive application of the identity and attempts a structured induction, earning partial credit. Overall, the solution demonstrates some understanding but lacks the precision required for full marks.''
    \item \textbf{Judge: grok-4-0709 (rubric.txt)} $\rightarrow$ \textbf{Author: deepseek-v2-prover} (Score 0.5): ``The student's proof of the identity is incomplete and lacks rigor, as it relies on vague approximations and claims that bounds follow from continuity without explicit calculations to confirm the floor function equality, resulting in an unconvincing argument. While the setup with α and β is appropriate, the transition to precise bounds on \(\sqrt{2X(X+1)}\) is not sufficiently justified, failing to address potential edge cases for arbitrary non-negative integers \(k\) and \(j\). For the closed-form solution, the proposed expression for \(A_n\) captures the recursive structure using the identity, but the determination of \(k\) as \(\lfloor a / (1 + 1/\sqrt{2})\rfloor\) is imprecise, as it may not always yield \(A_0 = a\) exactly due to the floor operation and irrational coefficients. Additionally, the induction claim is stated without a detailed base case verification or handling of how k ensures consistency across all n. Overall, the solution demonstrates partial understanding of the problem's essence but falls short of full mathematical correctness and completeness, warranting a score of 0.5/1.0.''
    \item \textbf{Judge: grok-4-0709 (rubric.txt)} $\rightarrow$ \textbf{Author: deepseek-v2-prover} (Score 0.5): ``The student's proof of the identity is mathematically incomplete, as it relies on vague approximations and claims about bounds without explicitly deriving the necessary inequalities to show that the floor functions match precisely, leading to an insufficiently rigorous argument. While the approximation \(\sqrt{2X(X+1)} \approx \sqrt{2} \cdot X\) is intuitively correct, the solution fails to quantify the error terms or handle the floor function's behavior under perturbations, which is essential for correctness. In deriving the closed-form for \(A_n\), the proposed expression captures the recursive structure but incorrectly determines \(k\) as \(\left\lfloor \frac{a}{1 + \frac{1}{\sqrt{2}}} \right\rfloor\), which does not guarantee \(A_0 = a\) due to the floor operation potentially undershooting the exact value. Furthermore, the induction step assumes the form holds without addressing possible discrepancies from repeated floor operations over iterations. Overall, these issues warrant a partial score, recognizing the conceptual insight but penalizing the lack of precision and verification.''
    \item \textbf{Judge: meta-llama/llama-4-maverick (rubric.txt)} $\rightarrow$ \textbf{Author: gpt-5-pro-2025-10-06} (Score 0.0): ``The student's solution is devoid of any mathematical content or reasoning, merely restating the problem statement without providing any attempt to prove the given identity or derive a closed-form solution for the sequence $\{A_n\}_{n \ge 0}$. As a result, there is no mathematical work to assess for correctness. The absence of any substantive mathematical argument or calculation directly leads to a score of 0.0/1.0. The 'rubric.txt' criteria likely emphasize the presence and accuracy of mathematical reasoning, which is entirely lacking here. Therefore, the score accurately reflects the lack of effort or content in the student's submission.''
    \item \textbf{Judge: meta-llama/llama-4-maverick (rubric.txt)} $\rightarrow$ \textbf{Author: gpt-5-pro-2025-10-06} (Score 0.0): ``The student solution is deemed to be entirely incorrect or incomplete as it lacks any substantial mathematical content or reasoning to address the given problem. The provided LaTeX document is empty in terms of actual solution steps, containing only the problem statement and a note indicating ``Error: No text content found in response.'' As a result, there is no mathematical derivation or proof to assess the claimed identity or to derive a closed-form solution for the sequence $\{A_n\}_{n \ge 0}$. Therefore, the score of 0.0/1.0 is justified due to the absence of any meaningful attempt to solve the problem. The solution does not demonstrate any understanding or application of the required mathematical concepts.''
    \item \textbf{Judge: meta-llama/llama-4-maverick (rubric.txt)} $\rightarrow$ \textbf{Author: o3-deep-research} (Score 0.9): ``The student's solution is mathematically correct and provides a clear, step-by- step derivation of the required identity and the closed-form expression for the sequence $\{A_n\}$. The proof of the identity $\lfloor\sqrt{2X(X+1)}\rfloor = \lfloor k(\sqrt{2}^{j+1}+\sqrt{2}^j) \rfloor$ is thorough and well-explained. The subsequent derivation of the closed-form solution for $A_n$ is also correct, using strong induction and a clear analysis of the parity of the exponent $j_0+n$. The final expression for $A_n$ is accurate and well- presented. Overall, the solution demonstrates a strong understanding of the mathematical concepts involved.''
    \item \textbf{Judge: meta-llama/llama-4-maverick (rubric.txt)} $\rightarrow$ \textbf{Author: Claude Sonnet 4.5} (Score 0.9): ``The student's solution is mathematically correct, as it accurately proves the given identity and derives a closed-form expression for the sequence $\{A_n\}_{n \ge 0}$. The proof of the identity is thorough, and the application of the identity to derive the closed-form solution is valid. However, the student's argument for bounding $|\sqrt{2}Y - \sqrt{2X(X+1)}|$ could be more rigorous, as it relies on a ``more careful analysis'' without explicit details. Nonetheless, the overall mathematical correctness and coherence justify a high score. The minor deduction likely stems from this slight lack of rigor in the bounding argument.''
    \item \textbf{Judge: meta-llama/llama-4-maverick (rubric.txt)} $\rightarrow$ \textbf{Author: Claude Sonnet 4.5} (Score 0.9): ``The student's solution is mathematically correct and effectively proves the given identity. The proof of the identity $\lfloor\sqrt{2X(X+1)}\rfloor = \lfloor k(\sqrt{2}^{j+1}+\sqrt{2}^j) \rfloor$ is rigorous, and the subsequent derivation of the closed-form solution for $A_n$ is well-explained and accurate. However, the student's proof of the lemma could be improved by providing a more detailed explanation for the claim that $|\sqrt{2}Y - \sqrt{2X(X+1)}| < 1$ for all admissible $k$ and $j$, as it relies on a ``more careful analysis'' without showing the details. Overall, the solution demonstrates a strong understanding of the problem and the required mathematical techniques. The minor deduction is likely due to the lack of detail in the lemma's proof.''
    \item \textbf{Judge: meta-llama/llama-4-maverick (rubric.txt)} $\rightarrow$ \textbf{Author: deepseek-v2-prover} (Score 0.9): ``The student's solution is mathematically correct and effectively proves the given identity. The derivation of the closed-form solution for the sequence $\{A_n\}$ is also accurate and well-supported by the previously proven identity. However, the solution could be improved by providing a more rigorous justification for the choice of $k$ in the closed-form expression. The student's work demonstrates a clear understanding of the problem and the required mathematical techniques. Overall, the minor issues with the solution are related to the level of rigor rather than mathematical correctness.''
    \item \textbf{Judge: meta-llama/llama-4-maverick (rubric.txt)} $\rightarrow$ \textbf{Author: deepseek-v2-prover} (Score 0.9): ``The solution is mostly mathematically correct, as it accurately proves the given identity and derives a closed-form solution for the sequence $\{A_n\}$. The proof is well-structured and clearly explained, with a logical flow of steps. However, some minor details, such as explicitly showing that $\sqrt{2X(X+1)}$ is within 1 of $k(\sqrt{2}^{j+1} + \sqrt{2}^j)$, are glossed over, which slightly affects the rigor. Nonetheless, the overall mathematical correctness and insight into the problem are strong. The score of 0.9/1.0 reflects the high quality of the solution with minor room for improvement in rigor.''
    \item \textbf{Judge: qwen/qwen-max (rubric.txt)} $\rightarrow$ \textbf{Author: gpt-5-pro-2025-10-06} (Score 0.0): ``The provided solution does not contain any attempt to prove the given mathematical identity or to derive a closed-form solution for the sequence. The response repeats the problem statement without any additional work or reasoning, which is insufficient to meet the criteria for mathematical correctness and completeness as outlined in the rubric. Therefore, the score of 0.0/1.0 is justified.''
    \item \textbf{Judge: qwen/qwen-max (rubric.txt)} $\rightarrow$ \textbf{Author: gpt-5-pro-2025-10-06} (Score 0.0): ``The student's solution is incomplete, as it reiterates the problem statement without providing any proof or derivation. There is no attempt to establish the required mathematical identity or to derive the closed-form solution for the sequence. The absence of any mathematical work or reasoning precludes the possibility of earning points based on the 'course-specific-rubric.txt' criteria, which likely emphasizes the importance of demonstrating understanding through correct and complete solutions.''
    \item \textbf{Judge: qwen/qwen-max (rubric.txt)} $\rightarrow$ \textbf{Author: Gemini 3.0 Pro Preview} (Score 0.9): ``The solution provided is mathematically sound and demonstrates a thorough understanding of the problem. The student correctly establishes the given identity and uses it to derive the closed-form solution for the sequence \( A_n \). The steps are logically structured, and the use of Beatty sequences to partition the positive integers is a sophisticated and appropriate approach. The final closed-form formula is correct and well-justified. The score of 0.9/1.0 reflects the high quality of the work, with a minor deduction for minor presentation and clarity issues.''
    \item \textbf{Judge: qwen/qwen-max (rubric.txt)} $\rightarrow$ \textbf{Author: Gemini 3.0 Pro Preview} (Score 0.9): ``The solution provided is mathematically thorough and correctly establishes the given identity. The student successfully derives the closed-form solution for the sequence \( A_n \) using the established identity and Beatty's Theorem, demonstrating a strong understanding of the problem. The presentation is clear, and the steps are logically sound. The minor deduction in the score is due to a slight lack of conciseness in the final closed-form expression, which could be streamlined for clarity. Overall, the solution is nearly perfect, warranting a high score of 0.9/1.0.''
    \item \textbf{Judge: qwen/qwen-max (rubric.txt)} $\rightarrow$ \textbf{Author: o3-deep-research} (Score 0.9): ``The solution is mathematically sound and provides a thorough, step-by-step proof of the given identity, along with a detailed derivation of the closed- form solution for the sequence. The student demonstrates a strong understanding of the problem and effectively uses inequalities and properties of the floor function to establish the required results. The closed-form solution is derived through a well-structured inductive argument, and the final boxed formula is correct and clearly articulated. The minor deduction in points (0.9/1.0) is due to a slight over-complication in the explanation, which could confuse some readers, but this does not detract from the overall correctness and completeness of the solution.''
    \item \textbf{Judge: qwen/qwen-max (rubric.txt)} $\rightarrow$ \textbf{Author: Claude Sonnet 4.5} (Score 0.9): ``The solution provided is mathematically sound and the proof is well-structured. The student correctly establishes the required identity and uses it to derive the closed-form solution for the sequence. The proof includes a detailed and rigorous analysis, including a lemma that carefully bounds the difference, ensuring the correctness of the floor function application. The final closed- form expression for \( A_n \) is derived logically and is correct. The score of 0.9/1.0 reflects a minor deduction for a few small gaps in the justification of the bounds, but overall, the solution is exemplary.''
    \item \textbf{Judge: qwen/qwen-max (rubric.txt)} $\rightarrow$ \textbf{Author: Claude Sonnet 4.5} (Score 0.9): ``The solution provided is mathematically sound and rigorously proves the given identity. The student correctly derives the closed-form solution for the sequence \( \{A_n\}_{n \ge 0} \) using the established identity and appropriately handles the initial condition and recurrence. The proof and the derivation are clear and well-structured, with all necessary steps and justifications included. The slight deduction in score is due to a minor lack of numerical verification in the lemma, which, while not affecting the overall correctness, could have been more explicitly detailed.''
    \item \textbf{Judge: qwen/qwen-max (rubric.txt)} $\rightarrow$ \textbf{Author: deepseek-v2-prover} (Score 0.75): ``The solution provided is largely correct and well-structured, demonstrating a clear understanding of the problem and the necessary steps to prove the identity and derive the closed-form solution. The student correctly identifies and uses the given identity, and the inductive step for the sequence is logically sound. However, the selection of \( k \) for the initial term \( A_0 \) is not fully justified; it is stated without a rigorous proof that \( k = \left\lfloor \frac{a}{1 + \frac{1}{\sqrt{2}}} \right\rfloor \) is the appropriate choice. This slight gap in the argument and the lack of a detailed check for the base case lead to a score of 0.75/1.0.''
    \item \textbf{Judge: qwen/qwen-max (rubric.txt)} $\rightarrow$ \textbf{Author: deepseek-v2-prover} (Score 0.5): ``The solution provides a clear and logical approach to proving the identity and deriving the closed-form solution for the sequence. However, the proof of the identity lacks rigorous justification for the approximation and the bounds used, particularly in Step 5, where the approximation is assumed without a detailed error analysis. Additionally, the determination of \( k \) in the closed-form solution is not fully justified, as the choice of \( k \) based on the initial condition \( A_0 = a \) is not clearly shown to be consistent for all \( n \). These gaps in the proof and the derivation impact the mathematical correctness, leading to a score of 0.5/1.0.''
\end{itemize}

\subsection*{Problem Domain: Graph Theory (Problem 14)}
\begin{itemize}
    \item \textbf{Judge: gpt-5.2-pro (course-specific-rubric.txt)} $\rightarrow$ \textbf{Author: gpt-5-pro-2025-10-06} (Score 0.75): ``[Error: OpenAI Responses API failed - Connection error.]''
    \item \textbf{Judge: gpt-5.2-pro (course-specific-rubric.txt)} $\rightarrow$ \textbf{Author: Claude Sonnet 4.5} (Score 0.9): ``The solution correctly constructs the dual by placing one vertex in each face and, for every edge \(e\) of \(G\), drawing a corresponding edge \(e^*\) between the incident faces so that \(e^*\) crosses only \(e\) once, which is the standard definition of a plane dual for a connected plane multigraph. It appropriately addresses multigraph features, allowing \(f_1=f_2\) (loops in the dual) and noting the two “sides” of an edge (counting multiplicity), so the edge-to-dual-edge correspondence is well defined. The main gap is that the planarity/no-crossing argument for the embedding of all dual edges is given only heuristically (“thin tube”/case discussion) and does not fully justify global non-intersection near shared faces/vertices in a rigorous way. Additionally, it asserts finiteness of faces “by Euler’s formula” without stating needed hypotheses, though this is not essential to the existence construction. Overall the proof is essentially correct but slightly under- justified in the embedding argument, warranting a small deduction to 0.9/1.0.''
    \item \textbf{Judge: gpt-5.2-pro (course-specific-rubric.txt)} $\rightarrow$ \textbf{Author: Gemini 3.0 Pro Preview} (Score 0.75): ``[Error: OpenAI Responses API failed - Connection error.]''
    \item \textbf{Judge: gpt-5.2-pro (course-specific-rubric.txt)} $\rightarrow$ \textbf{Author: Claude Sonnet 4.5} (Score 0.9): ``The solution correctly gives the standard construction of the plane dual: one dual vertex per face and one dual edge crossing each primal edge exactly once between the incident faces, allowing loops in the dual when an edge is incident to the same face on both sides. It also addresses the key well-definedness issue that every embedded edge has exactly two incident face-sides (counting multiplicity), so the correspondence \(e \leftrightarrow e^*\) is unambiguous for a plane multigraph. The main remaining gap is that the “no crossings” argument for drawing all dual edges simultaneously is only sketched (via “thin tubes” and local reasoning) and does not fully justify global noninterference near shared face-vertices, though the intended topological argument is standard and essentially correct. A small imprecision is the claim that finiteness of faces follows from Euler’s formula without explicitly assuming \(G\) is finite (often implicit in “multigraph” in such courses). Overall the proof is mathematically correct with minor rigor/assumption issues, justifying 0.9/1.0.''
    \item \textbf{Judge: gpt-5.2-pro (course-specific-rubric.txt)} $\rightarrow$ \textbf{Author: o3-deep-research} (Score 0.75): ``The solution gives the standard construction of the plane dual by placing a vertex in each face and drawing one dual edge crossing each primal edge exactly once, correctly allowing loops and parallel edges as required for multigraphs. It also correctly identifies the incidence condition (“faces share an edge” \(\leftrightarrow\) “dual vertices adjacent”) and handles bridges by producing loops in the dual. However, the key topological step—showing one can always route each dual edge within the incident face(s) so that dual edges do not cross each other—relies on informal homotopy/deformation language and the claim that dual edges “cannot cross except by also crossing some original edge,” which is not fully justified. A more careful argument (e.g., using small neighborhoods of edges/faces or Jordan curve separation) is needed to guarantee a crossing-free embedding of all dual edges simultaneously. Hence the proof is largely correct but missing some rigor in the planarity/noncrossing justification, warranting 0.75/1.0.''
    \item \textbf{Judge: gpt-5.2-pro (course-specific-rubric.txt)} $\rightarrow$ \textbf{Author: Gemini 3.0 Pro Preview} (Score 0.9): ``[Error: OpenAI Responses API failed - Connection error.]''
    \item \textbf{Judge: gpt-5.2-pro (course-specific-rubric.txt)} $\rightarrow$ \textbf{Author: o3-deep-research} (Score 0.9): ``[Error: OpenAI Responses API failed - Connection error.]''
    \item \textbf{Judge: gpt-5.2-pro (course-specific-rubric.txt)} $\rightarrow$ \textbf{Author: deepseek-v2-prover} (Score 0.25): ``The solution does not address the asked claim (“every connected plane multigraph admits a plane dual”) and instead proves a different theorem about bipartite plane graphs and even face boundaries. It also introduces unjustified extra hypotheses (e.g., “since \(G\) is 2-connected”) that are not given and are false in general for connected plane multigraphs, so key steps (like “the boundary of every face is a cycle” and using an ear decomposition) do not follow. No construction of the dual graph (vertices as faces, dual edges crossing primal edges, handling loops/multiple edges) is provided, nor is planarity of the dual embedding argued. Thus only minimal credit is awarded for containing generally correct graph-theoretic reasoning unrelated to the required duality result, justifying 0.25/1.0.''
    \item \textbf{Judge: gpt-5.2-pro (course-specific-rubric.txt)} $\rightarrow$ \textbf{Author: deepseek-v2-prover} (Score 0.25): ``The solution does not address the stated problem (“prove that a connected plane multigraph admits a plane dual”) and instead proves a different claim about bipartiteness and even face boundaries, so it is largely irrelevant to the prompt. Even within that off-topic argument, key hypotheses are introduced without justification: the work repeatedly assumes \(G\) is 2-connected (needed for “the boundary of a face is a cycle” and for an ear decomposition), but the problem only assumes \(G\) is connected, and for multigraphs face boundaries need not be simple cycles. As a result, several steps are incorrect or unjustified in the stated generality (e.g., face boundaries can be closed walks with repeated vertices/edges). The minimal credit (0.25/1.0) reflects some correct planar-graph reasoning about bipartite graphs/even cycles, but essentially no valid progress toward constructing/justifying the existence of the plane dual.''
    \item \textbf{Judge: claude-opus-4-5-20251101 (rubric.txt)} $\rightarrow$ \textbf{Author: Gemini 3.0 Pro Preview} (Score 0.9): ``The solution correctly constructs the geometric dual by placing vertices in face interiors and creating dual edges that cross primal edges exactly once. The proof properly handles the topological details, including the use of Jordan arcs and the homeomorphism of faces to open disks. The verification that dual edges only intersect at their endpoints is handled through the radial construction within each face. The treatment of bridges (where both incident faces are the same) is correctly noted, allowing for loops in the dual. A minor deduction is warranted because the proof could more explicitly verify that the resulting graph is connected (which follows from the original graph being plane) and could more rigorously justify the existence of non-crossing paths in the general case.''
    \item \textbf{Judge: claude-opus-4-5-20251101 (rubric.txt)} $\rightarrow$ \textbf{Author: Gemini 3.0 Pro Preview} (Score 0.9): ``The solution provides a rigorous and mathematically correct construction of the geometric dual graph for a connected plane multigraph. The proof properly identifies faces using the topological complement, correctly establishes the bijection between faces of G and vertices of G*, and carefully constructs dual edges through Jordan arcs that cross primal edges exactly once. The verification that dual edges only intersect at their endpoints is handled through the radial construction within each face, which is topologically sound. The treatment of the bridge case (where an edge borders the same face on both sides, creating a loop in the dual) is correctly noted. The minor deduction reflects that while the proof is essentially complete, the argument for why the constructed arcs can always be chosen to avoid each other could be slightly more formally justified using standard topological results about path connectivity in open sets.''
    \item \textbf{Judge: claude-opus-4-5-20251101 (rubric.txt)} $\rightarrow$ \textbf{Author: Claude Sonnet 4.5} (Score 0.9): ``The student provides a well-structured proof that correctly constructs the plane dual by placing vertices in face interiors and drawing dual edges crossing primal edges. The construction is mathematically sound, and the argument for well-definedness correctly identifies that each edge is incident to exactly two faces (counting multiplicity). The planarity argument in the lemma addresses the key cases but lacks complete rigor—the claim that edges can be drawn without crossings relies on intuitive geometric arguments rather than a formal proof using the combinatorial structure of the embedding. The verification of dual properties is correct, though the correspondence between faces of G* and vertices of G is stated without proof. The solution demonstrates strong understanding but falls slightly short of complete rigor in establishing the non-crossing property of the dual embedding.''
    \item \textbf{Judge: claude-opus-4-5-20251101 (rubric.txt)} $\rightarrow$ \textbf{Author: Claude Sonnet 4.5} (Score 0.9): ``The student provides a well-structured proof that correctly constructs the plane dual by placing vertices in face interiors and drawing dual edges crossing primal edges. The construction is mathematically sound, and the argument for well-definedness correctly identifies that each edge is incident to exactly two faces (counting multiplicity). The planarity argument in the lemma addresses the key cases but lacks complete rigor—the claim that edges can be drawn without crossings relies on intuitive geometric arguments rather than a formal proof using the combinatorial structure of the embedding. The verification of dual properties is correct, though the correspondence between faces of G* and vertices of G is stated without proof. The solution demonstrates strong understanding but the planarity argument would benefit from more rigorous justification.''
    \item \textbf{Judge: claude-opus-4-5-20251101 (rubric.txt)} $\rightarrow$ \textbf{Author: deepseek-v2-prover} (Score 0.0): ``The student's solution addresses a completely different problem than what was asked. The problem requires proving that a connected plane multigraph admits a plane dual, which involves constructing the dual graph by placing vertices in faces and connecting them through edges. Instead, the student proved a theorem about bipartite graphs and even cycle face boundaries, which is unrelated to the existence of plane duals. The solution demonstrates no understanding of what a plane dual is or how to construct one. Since the student failed to address the actual problem statement, no credit can be awarded.''
    \item \textbf{Judge: claude-opus-4-5-20251101 (rubric.txt)} $\rightarrow$ \textbf{Author: o3-deep-research} (Score 0.9): ``The student provides a complete and mathematically correct construction of the plane dual graph. The construction properly handles both cases: edges separating distinct faces and edges (bridges) with the same face on both sides, correctly producing loops in the dual. The argument for planarity of the dual embedding is sound, demonstrating that dual edges can be drawn without crossings. The verification that the construction satisfies the dual graph definition is thorough, including the bijection between faces/vertices and edges. A minor deduction is warranted because the claim that bridges have ``the same face on both sides'' conflates bridges with cut-edges in the planar embedding context, though the construction itself remains valid.''
    \item \textbf{Judge: claude-opus-4-5-20251101 (rubric.txt)} $\rightarrow$ \textbf{Author: deepseek-v2-prover} (Score 0.0): ``The student's solution addresses a completely different problem than what was asked. The problem requires proving that a connected plane multigraph admits a plane dual, which involves constructing the dual graph by placing vertices in faces and connecting them through edges. Instead, the student proved a theorem about bipartite graphs and even cycle face boundaries, which is unrelated to the existence of plane duals. The solution demonstrates no understanding of what a plane dual is or how to construct one. Since the student failed to address the actual problem statement, no credit can be awarded.''
    \item \textbf{Judge: gemini-3-pro-preview (rubric.txt)} $\rightarrow$ \textbf{Author: o3-deep-research} (Score 0.9): ``The student provides a clear and standard construction of the geometric dual graph $G^*$, correctly defining dual vertices and edges. However, the solution contains a minor error in terminology regarding bridges: an edge incident to the same face on both sides is a cut-edge (bridge) only if it is not part of a cycle, but in a plane graph, an edge incident to the same face on both sides is actually an isthmus (bridge) *only if* the graph is a tree or similar structure, whereas typically a bridge separates two *different* components (and thus usually incident to the *same* face is correct for a bridge in a plane graph, but the student's phrasing ``bridge whose 'both sides' lie in face f'' is slightly imprecise as bridges usually merge two faces into one if removed, or rather, are incident to the same face). More importantly, the student claims that if $f_1=f_2$, the edge is a bridge; actually, an edge incident to the same face on both sides is a bridge (cut-edge), but the student's description of the dual loop construction is correct regardless of the terminology. The deduction of 0.1 is justified because the argument for why dual edges do not cross each other is somewhat hand-wavy ("homotopically deform them"), lacking a rigorous topological justification for why the specific curves chosen won't intersect within a face.''
    \item \textbf{Judge: gemini-3-pro-preview (rubric.txt)} $\rightarrow$ \textbf{Author: deepseek-v2-prover} (Score 0.0): ``The student's solution receives a score of 0.0 because it addresses a completely different problem than the one stated. The problem asks to prove that a connected plane multigraph admits a plane dual, but the student provides a proof characterizing bipartite plane graphs based on face boundary lengths. As the submission does not attempt to construct a dual graph or prove its existence, it fails to address the prompt entirely.''
    \item \textbf{Judge: gemini-3-pro-preview (rubric.txt)} $\rightarrow$ \textbf{Author: deepseek-v2-prover} (Score 0.0): ``The student's solution receives a score of 0.0 because it addresses a completely different problem than the one stated. The problem asks to prove that a connected plane multigraph admits a plane dual, but the student provides a proof characterizing bipartite plane graphs in terms of face boundary lengths. Consequently, the submission contains no relevant work toward the actual prompt.''
    \item \textbf{Judge: deepseek-chat (rubric.txt)} $\rightarrow$ \textbf{Author: Gemini 3.0 Pro Preview} (Score 0.9): ``The solution correctly constructs the geometric dual and demonstrates a clear understanding of the topological embedding, earning high marks. However, it lacks explicit justification for why the dual edges can be drawn without intersections beyond the described radial arrangement, which is a subtle but important detail for full rigor. The handling of bridges (cut-edges) is noted but not fully elaborated in the embedding argument. Overall, the proof is essentially correct but omits some technical justifications, warranting a minor deduction.''
    \item \textbf{Judge: deepseek-chat (rubric.txt)} $\rightarrow$ \textbf{Author: Claude Sonnet 4.5} (Score 0.75): ``The solution correctly constructs the dual graph and addresses planarity, but it lacks rigorous justification for handling multigraphs and edge cases like bridges. The proof of planarity is intuitive but not fully formal, particularly regarding the non-crossing embedding. Overall, the core ideas are correct but some details are insufficiently rigorous.''
    \item \textbf{Judge: deepseek-chat (rubric.txt)} $\rightarrow$ \textbf{Author: Claude Sonnet 4.5} (Score 0.9): ``The solution correctly constructs the dual graph and addresses planarity, but it lacks rigorous justification for the existence of faces in a multigraph embedding and the handling of bridges/loops. The proof sketch for avoiding edge crossings is intuitive but not fully formal. Overall, it demonstrates strong conceptual understanding with minor gaps in formal detail, justifying a 0.9/1.0.''
    \item \textbf{Judge: deepseek-chat (rubric.txt)} $\rightarrow$ \textbf{Author: o3-deep-research} (Score 0.9): ``The solution correctly constructs the dual graph and demonstrates its planarity, satisfying the core requirements. However, it lacks a rigorous justification for why the dual edges can be drawn without crossing each other, relying on informal deformation arguments rather than a precise topological proof. This minor omission in mathematical rigor prevents a perfect score.''
    \item \textbf{Judge: deepseek-chat (rubric.txt)} $\rightarrow$ \textbf{Author: deepseek-v2-prover} (Score 0.0): ``The student's solution is completely irrelevant to the problem statement. The problem asks to prove that a connected plane multigraph admits a plane dual, but the student instead proves a characterization of bipartite plane graphs via even face cycles. This is a different theorem entirely, demonstrating a fundamental misunderstanding of the task. Consequently, the solution earns 0 points for failing to address the given proposition.''
    \item \textbf{Judge: deepseek-chat (rubric.txt)} $\rightarrow$ \textbf{Author: deepseek-v2-prover} (Score 0.0): ``The student's solution is completely irrelevant to the problem statement. The problem asks to prove that a connected plane multigraph admits a plane dual, which is a fundamental theorem in topological graph theory. Instead, the student proved an unrelated characterization of bipartite plane graphs. This demonstrates a fundamental failure to address the given problem, resulting in a score of 0.''
    \item \textbf{Judge: grok-4-0709 (rubric.txt)} $\rightarrow$ \textbf{Author: Gemini 3.0 Pro Preview} (Score 0.9): ``The student's proof correctly constructs the vertex set of the dual graph by placing a vertex in each face of the primal graph and establishes a bijection with the faces. The edge construction via Jordan arcs ensures proper incidence and intersection properties with the primal graph in the general case where edges separate distinct faces. However, the proof lacks precision in handling bridges, where an edge is incident to the same face on both sides, as it does not explicitly require the two arcs γ\_1 and γ\_2 to be distinct and disjoint except at endpoints to form a proper simple loop in the dual without self- intersections. Despite this minor omission, the argument for planarity through radially disjoint segments within each face is sound and effectively demonstrates that the dual is a plane multigraph. Overall, the mathematical correctness is high, justifying a score of 0.9/1.0 with a small deduction for the incomplete treatment of bridges.''
    \item \textbf{Judge: grok-4-0709 (rubric.txt)} $\rightarrow$ \textbf{Author: Claude Sonnet 4.5} (Score 0.9): ``The student's proof correctly constructs the plane dual \(G^*\) for a connected plane multigraph \(G\) by placing vertices in faces and drawing dual edges that cross original edges exactly once, while appropriately handling cases where an edge bounds the same face on both sides. The well-definedness and dual properties are accurately verified, including the bijection between edges and the correspondence between vertices of \(G^*\) and faces of \(G\). However, the lemma proving that \(G^*\) can be embedded without crossings relies on somewhat informal arguments, particularly in Case 2, where the assurance of non- intersection is intuitive rather than rigorously topological. This minor lack of formality in ensuring the planarity of \(G^*\) justifies a small deduction from a perfect score. Overall, the solution demonstrates strong mathematical understanding and is nearly flawless.''
    \item \textbf{Judge: grok-4-0709 (rubric.txt)} $\rightarrow$ \textbf{Author: Claude Sonnet 4.5} (Score 0.9): ``The student's proof correctly constructs the dual graph \(G^*\) by placing vertices in the faces of \(G\) and drawing dual edges that cross exactly one original edge, while appropriately handling multigraph aspects such as potential loops in the dual when an edge bounds the same face on both sides. The argument for well-definedness is sound, invoking the Jordan curve theorem to ensure each edge of \(G\) borders exactly two face-sides. However, the proof of the lemma on planarity is somewhat informal, particularly in Case 2, where the assurance that dual edges can be drawn without crossings relies on intuitive descriptions rather than a fully rigorous topological justification. Despite this minor lack of precision, the overall verification of dual properties and the bijection between edges and faces are accurate and complete. This results in a near-perfect score, with a small deduction for the incomplete rigor in embedding the dual without crossings.''
    \item \textbf{Judge: grok-4-0709 (rubric.txt)} $\rightarrow$ \textbf{Author: deepseek-v2-prover} (Score 0.0): ``The student's solution addresses an entirely different theorem—proving that a plane graph is bipartite if and only if every face boundary is a\(n\) even cycle—rather than the assigned problem of showing that a connected plane multigraph admits a plane dual. This mismatch renders the submission mathematically incorrect for the given task, as it fails to engage with the concepts of dual graphs, embeddings, or multigraph properties required to prove the existence of a plane dual. While the provided proof is logically structured and correct for the unintended theorem, it does not attempt to construct or verify the dual graph's embedding in the plane. No partial credit is warranted under the rubric, as the core objective of demonstrating dual existence via face-vertex correspondences or similar methods is absent. Consequently, the score of 0.0/1.0 reflects the complete irrelevance to the problem statement.''
    \item \textbf{Judge: grok-4-0709 (rubric.txt)} $\rightarrow$ \textbf{Author: deepseek-v2-prover} (Score 0.0): ``The student's solution addresses a theorem about the bipartiteness of a planar graph being equivalent to all face boundaries being even cycles, which is unrelated to the given problem of proving that a connected plane multigraph admits a plane dual. This mismatch demonstrates a fundamental misunderstanding of the problem statement, as the provided proof does not engage with the concept of plane duality or the conditions under which a multigraph has a dual embedding in the plane. While the submitted proof is mathematically coherent for the incorrect theorem it targets, it fails to provide any relevant argument or evidence for the existence of a plane dual in the context of connected plane multigraphs. Consequently, the solution receives a score of 0.0/1.0 for lacking mathematical correctness with respect to the assigned problem.''
    \item \textbf{Judge: grok-4-0709 (rubric.txt)} $\rightarrow$ \textbf{Author: Gemini 3.0 Pro Preview} (Score 0.9): ``The student's proof correctly constructs the vertex set of the dual graph by placing a vertex in each face of the primal graph and establishes a bijection with edges, while ensuring the dual embedding avoids improper intersections through radial path selection within faces. However, the handling of bridges in the primal graph is mathematically imprecise, as the union of two Jordan arcs from the same dual vertex to the midpoint of the bridge may not form a simple loop unless the arcs are explicitly chosen to be disjoint except at endpoints. This minor flaw slightly undermines the rigor of the construction for that case, though it does not invalidate the overall existence argument. The proof otherwise verifies planarity and duality properties effectively, reflecting strong mathematical correctness. Thus, the solution merits a score of 0.9 out of 1.0.''
    \item \textbf{Judge: meta-llama/llama-4-maverick (rubric.txt)} $\rightarrow$ \textbf{Author: deepseek-v2-prover} (Score 0.0): ``The student's solution does not address the problem statement, which asks to prove that a connected plane multigraph $G$ admits a plane dual. Instead, the student proves that a graph $G$ is bipartite if and only if the boundary of every face of $G$ is a\(n\) even cycle, a completely different statement. The proof provided is mathematically correct but irrelevant to the problem at hand. Therefore, the score is 0.0/1.0 due to the failure to address the assigned problem. The solution demonstrates a good understanding of graph theory concepts but lacks relevance to the task.''
    \item \textbf{Judge: meta-llama/llama-4-maverick (rubric.txt)} $\rightarrow$ \textbf{Author: deepseek-v2-prover} (Score 0.0): ``The student's solution does not address the problem statement, which asks to prove that a connected plane multigraph $G$ admits a plane dual. Instead, the student proves that a graph $G$ is bipartite if and only if the boundary of every face of $G$ is a\(n\) even cycle, a statement unrelated to the problem. The proof provided is mathematically correct but irrelevant to the task. Therefore, the score is 0.0 due to the failure to address the problem statement. The student's work demonstrates a misunderstanding of the assigned task.''
    \item \textbf{Judge: qwen/qwen-max (rubric.txt)} $\rightarrow$ \textbf{Author: deepseek-v2-prover} (Score 0.0): ``The student's solution does not address the problem of proving that a connected plane multigraph $G$ admits a plane dual. Instead, the solution focuses on proving that a graph is bipartite if and only if the boundary of every face is a\(n\) even cycle, which is a different and unrelated problem. The provided proof, while mathematically correct for the statement it addresses, is not relevant to the original problem. Therefore, the solution does not meet the criteria for the problem, leading to a score of 0.0/1.0.''
    \item \textbf{Judge: qwen/qwen-max (rubric.txt)} $\rightarrow$ \textbf{Author: deepseek-v2-prover} (Score 0.0): ``The student's solution does not address the problem statement, which requires proving that a connected plane multigraph $G$ admits a plane dual. Instead, the student provides a proof for a different statement: that a 2-connected plane graph is bipartite if and only if the boundary of every face is a\(n\) even cycle. This proof, while mathematically correct for the statement it addresses, is irrelevant to the original problem. Therefore, the solution does not meet the criteria for the problem and cannot be awarded any points.''
\end{itemize}

\subsection*{Problem Domain: Number Theory (Problem 9)}
\begin{itemize}
    \item \textbf{Judge: gpt-5.2-pro (rubric.txt)} $\rightarrow$ \textbf{Author: o3-deep-research} (Score 0.5): ``The argument incorrectly uses (and even states) the Prime Number Theorem-type claim “\(p_n\sim n\log n\)” to conclude \(\log p_n/\log n\to 1\), which is essentially the asymptotic behavior the problem is about and is not justified from the stated assumptions. In addition, the steps translating inequalities for \(p_n\) into conclusions about \(\limsup_{x\to\infty}\frac{\pi(x)\log x}{x}\) or \(\liminf_{x\to\infty}\frac{\pi(x)\log x}{x}\) are logically flawed: having the ratio \(<1\) (or \(>1\)) along a subsequence \(x=x_n\) only implies \(\liminf\le 1/(1+\varepsilon)\) (or \(\limsup\ge 1/(1-\varepsilon)\)), not the claimed global bounds on \(\limsup\) or \(\liminf\). The equalities relating \(\limsup\) over all \(x\) to \(\limsup\) over \(x=p_n\), and then to \(1/\liminf a_n\), are asserted without proof and are not valid as written. Because the intended inequalities are plausible but the proof relies on unjustified and incorrect limit manipulations, only partial credit is warranted.''
    \item \textbf{Judge: gpt-5.2-pro (rubric.txt)} $\rightarrow$ \textbf{Author: deepseek-v2-prover} (Score 0.25): ``The submitted solution does not address the stated problem about the asymptotic behavior of \(a_n=\frac{p_n}{n\log n}\) and proving \(\liminf a_n\le 1\) and \(\limsup a_n\ge 1\). Instead, it proves an unrelated Dirichlet-convolution identity \((\mu*d)(n)=1\), which has no demonstrated connection to bounds on \(p_n\) or to prime-counting estimates. As a result, essentially none of the required reasoning (e.g., relating \(p_n\) to \(\pi(x)\) and using standard estimates such as Chebyshev/PNT-style comparisons) appears. A small amount of credit is awarded only for presenting correct mathematics in itself (the convolution/Dirichlet series argument), but it does not solve the given question. Hence the work merits only minimal partial credit: \(0.25/1.0\).''
    \item \textbf{Judge: gpt-5.2-pro (rubric.txt)} $\rightarrow$ \textbf{Author: Claude Sonnet 4.5} (Score 0.9): ``The solution correctly invokes the Prime Number Theorem to conclude \(p_n\sim n\log n\), which indeed implies \(\liminf \frac{p_n}{n\log n}\le 1\) and \(\limsup \frac{p_n}{n\log n}\ge 1\) (in fact the full limit equals 1). The overall logic for both inequalities is sound and would be essentially complete if it simply deduced them from \(a_n\to 1\). However, the write-up cites a “refined form” \(p_n=n\log n+n\log\log n+o(n\log\log n)\) without justification, and there are some presentation/notation glitches (stray text fragments before the proofs) that obscure the argument. Additionally, the second proof’s “infinitely many values exceed \(1-2\varepsilon\)” is weaker than necessary given the preceding bound actually holds for all sufficiently large \(n\). These issues are minor relative to the core correctness, hence a small deduction to 0.9/1.0.''
    \item \textbf{Judge: gpt-5.2-pro (rubric.txt)} $\rightarrow$ \textbf{Author: Gemini 3.0 Pro Preview} (Score 0.5): ``The solution’s core contradiction argument using \(\log(\lfloor x\rfloor!)=\sum_{n\le x}\psi(x/n)\) and Stirling’s formula is essentially the right route and, if fully justified, would yield \(\limsup_{x\to\infty}\psi(x)/x\ge 1\) and \(\liminf_{x\to\infty}\psi(x)/x\le 1\), hence the desired inequalities. However, it asserts without proof a “standard result” that \(\psi(x)/x\) and \(\pi(x)\log x/x\) have the same liminf/limsup; this is nontrivial and closely tied to prime number theorem–level information, so the reduction from \(\psi\) to \(p_n/(n\log n)\) is not adequately established. Additionally, the step \(\frac{p_n}{n\log n}\sim \frac{x}{\pi(x)\log x}\) relies on \(\log\pi(x)\sim \log x\), which itself needs quantitative bounds on \(\pi(x)\) that are not provided. Because a substantial gap remains in connecting the proved bounds on \(\psi\) to the specific sequence \(p_n/(n\log n)\), partial credit (0.5/1.0) is appropriate.''
    \item \textbf{Judge: claude-opus-4-5-20251101 (rubric.txt)} $\rightarrow$ \textbf{Author: o3-deep-research} (Score 0.5): ``The student's proof by contradiction approach is mathematically sound in structure, correctly assuming the negation of each inequality and deriving contradictions. However, there is a critical circular reasoning issue: the student uses ``$p_n \sim n\ln n$'' (which is equivalent to the Prime Number Theorem) to establish that $\frac{\ln p_n}{\ln n} \to 1$, but this asymptotic equivalence is essentially what the problem asks to prove bounds toward. The algebraic manipulations connecting $\pi(x)$ bounds to the sequence $a_n$ are correctly executed, and the logical structure of both contradiction arguments is valid. The student correctly identifies the relationship between the prime counting function and the sequence bounds. The score of 0.5 reflects that while the proof structure and many calculations are correct, the reliance on the asymptotic result $p_n \sim n\ln n$ undermines the independence of the argument from the Prime Number Theorem.''
    \item \textbf{Judge: claude-opus-4-5-20251101 (rubric.txt)} $\rightarrow$ \textbf{Author: Claude Sonnet 4.5} (Score 0.75): ``The student correctly invokes the Prime Number Theorem and its asymptotic expansion $p_n = n\log n + n\log\log n + o(n\log\log n)$ to establish both inequalities. For Part 1, the proof correctly shows that upper bounds on $p_n$ lead to $\liminf \leq 1$, though the logical connection between bounding the sequence above and concluding about the liminf could be more precise. For Part 2, there is a subtle error: the argument establishes a lower bound that holds for all sufficiently large $n$, which actually proves $\liminf \geq 1$, not directly $\limsup \geq 1$ (though the conclusion is still correct since $\limsup \geq \liminf$). The final remark correctly notes that the PNT implies the limit equals 1, making both inequalities immediate consequences. The deduction loses points for the imprecise reasoning in connecting bounds to the appropriate limit concepts.''
    \item \textbf{Judge: claude-opus-4-5-20251101 (rubric.txt)} $\rightarrow$ \textbf{Author: Gemini 3.0 Pro Preview} (Score 0.9): ``The student demonstrates strong understanding of the relationship between prime counting functions and the Chebyshev function \(\psi(x)\), correctly establishing the inverse relationship between lim inf/sup of \(p_n/(n \log n)\) and lim sup/inf of \(\psi(x)/x\). The proof strategy using contradiction and the analysis of log(n!) via Legendre's formula is mathematically sound. However, there are minor technical issues: the claim that \(\log \pi(x) \sim \log x\) requires more careful justification, and the transition between the asymptotic equivalence of \(\psi(x)/x\) and \(\pi(x)\log(x)/x\) could be more rigorously established. The contradiction arguments in Parts A and B are correctly structured and lead to valid conclusions. Overall, the solution captures the essential ideas and arrives at correct results, with only minor gaps in rigor preventing a perfect score.''
    \item \textbf{Judge: claude-opus-4-5-20251101 (rubric.txt)} $\rightarrow$ \textbf{Author: deepseek-v2-prover} (Score 0.0): ``The student's solution is completely unrelated to the assigned problem. The problem asks to prove two statements about the asymptotic behavior of the sequence \(a_n = \frac{p_n}{n \log n}\) involving the prime number theorem, specifically that $\liminf_{n\to\infty} a_n \leq 1$ and $\limsup_{n\to\infty} a_n \geq 1$. Instead, the student proves an entirely different identity involving the Möbius function and divisor function: $\sum_{d|n} \mu(d)d(n/d) = 1$. While the student's proof of this unrelated identity appears mathematically correct, it does not address any aspect of the assigned problem about prime number asymptotics. Therefore, the score of 0.0/1.0 is justified as no credit can be awarded for solving a different problem.''
    \item \textbf{Judge: gemini-3-pro-preview (rubric.txt)} $\rightarrow$ \textbf{Author: deepseek-v2-prover} (Score 0.0): ``The student's solution addresses a completely different problem than the one stated. The problem asks for a proof regarding the asymptotic behavior of the $n$-th prime number $p_n$ (specifically limits involving $p_n / (n \log n)$), whereas the student provided a proof of the number-theoretic identity $\sum_{d|n} \mu(d)d(n/d) = 1$. Since the submission is unrelated to the assigned task, it receives no credit.''
    \item \textbf{Judge: deepseek-chat (rubric.txt)} $\rightarrow$ \textbf{Author: o3-deep-research} (Score 0.75): ``The solution correctly establishes the desired inequalities using contradiction arguments based on the prime counting function, demonstrating a solid understanding of the underlying asymptotic relationships. However, it implicitly assumes the existence of the limit of \(\frac{\ln p_n}{\ln n}\) and the interchangeability of limits in the expressions involving \(\limsup\) and \(\liminf\), which requires more careful justification. This oversight in rigorous handling of limit operations slightly weakens the proof. Therefore, while the core idea is valid and well-structured, the mathematical precision is not fully complete, justifying the deduction of 0.25 points.''
    \item \textbf{Judge: deepseek-chat (rubric.txt)} $\rightarrow$ \textbf{Author: Gemini 3.0 Pro Preview} (Score 0.5): ``The solution incorrectly assumes the equivalence between the limits of \( \frac{p_n}{n \log n} \) and the reciprocals of the limits of \( \frac{\psi(x)}{x} \) without rigorous justification, particularly neglecting the asymptotic relationship between \( \log n \) and \( \log p_n \). While the core idea of using Chebyshev's function and factorial bounds is mathematically sound, the execution contains logical gaps in these asymptotic conversions. Therefore, the argument is incomplete and not fully rigorous, warranting a partial score.''
    \item \textbf{Judge: deepseek-chat (rubric.txt)} $\rightarrow$ \textbf{Author: deepseek-v2-prover} (Score 0.0): ``The student's solution is completely irrelevant to the problem statement, which asks for asymptotic bounds on the ratio \( p_n/(n \log n) \). Instead, the student proves an unrelated identity involving the Möbius and divisor functions. This demonstrates a fundamental failure to address the given problem, resulting in a score of 0 for mathematical correctness.''
    \item \textbf{Judge: grok-4-0709 (rubric.txt)} $\rightarrow$ \textbf{Author: o3-deep-research} (Score 0.5): ``The student's solution correctly identifies the need for proof by contradiction and appropriately uses the prime-counting function π(x) to relate to the asymptotic behavior of \(p_n\), but it confuses the implications for limit inferior and limit superior. Specifically, in part (i), deriving that the ratio π(x) ln x / x falls below 1/(1+ε) for infinitely many x implies a bound on the lim inf, not the lim sup as claimed, leading to an incorrect contradiction setup. Similarly, in part (ii), showing the ratio exceeds 1/(1-ε) for infinitely many x bounds the lim sup, not the lim inf, flipping the intended inequalities. The subsequent relations to \(a_n\) and the overall conclusion are undermined by these errors, though the approach demonstrates partial understanding of the Prime Number Theorem's implications without fully invoking it. This warrants partial credit for the conceptual framework but deducts for the critical mathematical inaccuracies in handling limits.''
    \item \textbf{Judge: grok-4-0709 (rubric.txt)} $\rightarrow$ \textbf{Author: deepseek-v2-prover} (Score 0.0): ``The student's solution addresses an unrelated identity involving the Möbius function and divisor function, rather than the assigned problem on the asymptotic behavior of primes via the sequence \(a_n = \frac{p_n}{n \log n}\). While the provided proof is mathematically correct for the Möbius-divisor identity, using Dirichlet series and convolutions appropriately, it fails to engage with the prime number theorem or any relevant estimates for \(p_n\). No attempt is made to establish the required lim inf \(\leq 1\) or lim sup \(\geq 1\), rendering the submission irrelevant to the problem. This complete mismatch in content justifies the score of 0.0/1.0, as the rubric emphasizes solving the specific problem posed with accurate mathematical reasoning.''
    \item \textbf{Judge: grok-4-0709 (rubric.txt)} $\rightarrow$ \textbf{Author: Claude Sonnet 4.5} (Score 0.75): ``The student's solution correctly invokes the Prime Number Theorem and recognizes that the limit of \(a_n\) is 1, from which the required lim inf and lim sup inequalities follow, but it incorrectly swaps the proofs for the two parts. Specifically, the first proof establishes an upper bound on \(a_n\), which supports \(\limsup a_n \leq 1\) rather than the intended \(\liminf a_n \leq 1\), while the second proof provides a lower bound supporting \(\liminf a_n \geq 1\) instead of \(\limsup a_n \geq 1\). Additionally, the asymptotic expansion used is a stronger refinement of the PNT, but the problem likely expects a derivation directly from the basic PNT form \(\pi(x) \sim x / \log x\) without assuming such refinements. The concluding remark about the limit being 1 is accurate and redeems some credit, but the logical errors in assigning bounds to lim inf and lim sup prevent a full score. Overall, the solution demonstrates good understanding of the topic but falls short in precise execution.''
    \item \textbf{Judge: meta-llama/llama-4-maverick (rubric.txt)} $\rightarrow$ \textbf{Author: deepseek-v2-prover} (Score 0.0): ``The student's solution does not address the problem statement regarding the asymptotic behavior of the sequence \(a_n = \frac{p_n}{n \log n}\) and instead proves an unrelated identity involving the M\"obius function and divisor function. The problem requires proving statements about the limit inferior and limit superior of \(a_n\), but the student's work is focused on a different mathematical concept. As a result, the solution is entirely irrelevant to the problem posed. Therefore, the score is 0.0 due to the complete mismatch between the problem and the solution provided. The student's work is mathematically correct but unrelated to the task.''
    \item \textbf{Judge: qwen/qwen-max (rubric.txt)} $\rightarrow$ \textbf{Author: Claude Sonnet 4.5} (Score 0.9): ``The student's solution is mathematically sound and provides a clear, rigorous proof for both the limit inferior and limit superior of the sequence \(a_n = \frac{p_n}{n \log n}\). The use of the Prime Number Theorem and its refined form is appropriate, and the asymptotic analysis is carried out correctly. The student demonstrates a good understanding of the asymptotic behavior and the implications of the bounds. The deduction of 0.1 is due to a minor issue in the strictness of the inequalities, where the student could have more explicitly justified the transition from the asymptotic bounds to the limit inferior and superior, but this does not significantly impact the overall correctness.''
    \item \textbf{Judge: qwen/qwen-max (rubric.txt)} $\rightarrow$ \textbf{Author: Gemini 3.0 Pro Preview} (Score 0.75): ``The solution provided is mathematically sound and correctly applies the principles of analytic number theory, including the use of the Chebyshev function and Stirling's approximation. The proof by contradiction for both the limit inferior and limit superior is logically structured and effectively demonstrates the required bounds. However, the explanation could be more concise, and some steps, such as the detailed splitting of the sum in the proof, could be more clearly articulated. The score of 0.75 reflects the overall correctness and completeness of the solution, with a minor deduction for clarity and conciseness.''
    \item \textbf{Judge: qwen/qwen-max (rubric.txt)} $\rightarrow$ \textbf{Author: deepseek-v2-prover} (Score 0.0): ``The student's solution is entirely unrelated to the given problem. The problem asks for a proof regarding the asymptotic behavior of the sequence \(a_n = \frac{p_n}{n \log n}\) and the limit inferior and limit superior of this sequence. Instead, the student provided a detailed proof of an identity involving the Möbius function and the divisor function, which, while mathematically correct, does not address the problem's requirements. Hence, the solution does not meet the criteria and is scored 0.0/1.0.''
\end{itemize}

\end{document}